%% file: main.tex
\documentclass[Afour,sageh,times]{sagej}
\usepackage{moreverb,url}
\usepackage[colorlinks=false,bookmarksopen,bookmarksnumbered,citecolor=black,urlcolor=black]{hyperref}
\usepackage{array}
\usepackage{textcomp}
\usepackage{stfloats}
\usepackage{url}
\usepackage{verbatim}
\usepackage[utf8]{inputenc}
\usepackage[english]{babel}
\usepackage{setspace}
\usepackage{multirow}
\usepackage[table,xcdraw]{xcolor}
\usepackage{epsfig}
\usepackage{graphics}
\usepackage{graphicx}
\graphicspath{{fig/}}
\usepackage[mathscr]{euscript} 
\usepackage{multicol}
\usepackage{multirow}
\usepackage{algorithm}
\usepackage{algpseudocode}
\usepackage{amsthm} 
\usepackage{amsmath, amsfonts}
\usepackage{amssymb}
\usepackage[percent]{overpic}
\usepackage{siunitx}
\usepackage{soul}
\usepackage{sidecap}
\usepackage{caption}
\usepackage{subfigure}
\usepackage{float}
\usepackage{tensor}
\usepackage{adjustbox}
\usepackage{booktabs}
\usepackage{enumerate}
\usepackage[normalem]{ulem}
\usepackage{makecell}

\usepackage{mathtools}
\mathtoolsset{showonlyrefs}

\newlength\figureHeight 
\newlength\figureWidth

\newcommand\BibTeX{{\rmfamily B\kern-.05em \textsc{i\kern-.025em b}\kern-.08em
T\kern-.1667em\lower.7ex\hbox{E}\kern-.125emX}}

\def\volumeyear{2025}

\theoremstyle{plain}
\newtheorem{theorem}{Theorem}
\newtheorem{remark}{Remark}
\newtheorem{lemma}{Lemma}
\newtheorem{proposition}{Proposition}

\newtheorem{corollary}{Corollary}
\newtheorem*{corollary*}{Corollary}

\newtheorem*{example*}{Toy Example}
\newtheorem*{examplestop*}{Toy Example.}
\newtheorem{property}{Property}
\newtheorem{assumption}{Assumption}
\newtheorem{problem}{Problem}
\input{boldfacemath.tex}

\newcommand{\review}[1]{{\color{red}#1}}
\newcommand{\reviewr}[1]{{\color{red}\sout{#1}}}

\begin{document}

\runninghead{Pustina et al.}

\title{Recursive Model-agnostic Inverse Dynamics of Serial Soft-Rigid Robots}

\author{Pietro Pustina\affilnum{1, 2}, Cosimo Della Santina\affilnum{2, 3} and Alessandro De Luca\affilnum{1}}

\affiliation{\affilnum{1} Department of Computer, Control and Management Engineering (DIAG), Sapienza University of Rome, Rome, Italy.\\
\affilnum{2} Department of Cognitive Robotics, Delft University of Technology, Delft, The Netherlands.\\
\affilnum{3}Institute of Robotics and Mechatronics, German Aerospace Center (DLR), Oberpfaffenhofen, Germany.
}

\corrauth{Pietro Pustina, pietro.pustina@uniroma1.it}

\begin{abstract}
Robotics is shifting from rigid, articulated systems to more sophisticated and heterogeneous mechanical structures. Soft robots, for example, have continuously deformable elements capable of large deformations. The flourishing of control techniques developed for this class of systems is fueling the need of efficient procedures for evaluating their inverse dynamics (ID), which is challenging due to the complex and mixed nature of these systems. As of today, no single ID algorithm can describe the behavior of generic (combinations of) models of soft robots. We address this challenge for generic series-like interconnections of possibly soft structures that may require heterogeneous modeling techniques. Our proposed algorithm requires as input a purely geometric description (forward-kinematics-like) of the mapping from configuration space to deformation space. With this information only, the complete equations of motion can be given an exact recursive structure which is essentially independent from (or `agnostic' to) the underlying reduced-order kinematics. We achieve this by exploiting Kane's method to manipulate the equations of motion, showing then their recursive structure. The resulting ID algorithms have optimal computational complexity within the proposed setting, i.e., linear in the number of distinct modules. Further, a variation of the algorithm is introduced that can evaluate the generalized mass matrix without increasing computation costs. We showcase the method applicability to robot models involving a mixture of rigid and soft elements, described via possibly heterogeneous reduced order models (ROMs), such as Volumetric FEM, Cosserat strain-based, and volume-preserving deformation primitives. None of these systems can be handled using existing ID techniques. 
\end{abstract}

\maketitle

\section{Introduction}
In recent years, robotic systems have undergone a remarkable transformation, evolving from assemblies of rigid links and joints to articulated mechanical architectures built from complex modules. These include serial interconnections of parallel mechanisms~\citep{lee2009development}, flexible link and joint manipulators~\citep{deluca2016robots}, robots made of meta-materials~\citep{hu2020origami, bevilacqua2022bio}, continuum and soft arms\citep{rus2015design, russo2023continuum, yin2021modeling, hoeijmakers2022investigation, liu2023large}, bio-hybrid robots~\citep{sun2020biohybrid, wang2021endoscopy, lin2022emerging}, and rigid robots manipulating deformable objects~\citep{zhu2022challenges,tiburzio2024model}. 
In this work, we specifically focus on one such class of systems: soft robots. Unlike rigid robots, where motion originates from the joints, soft robots leverage their compliant and highly deformable bodies to move~\citep{della2020soft}. 

However, the advantages of the soft body come with substantial challenges in modeling and control~\citep{dellasantina2023survey} due to the infinite dimensionality of the dynamics. The control of infinite-dimensional mechanical systems is still an emerging field of research~\citep{han2021distributed, zheng2022pde, chang2023energy}.
%
%
%
Reduced order models (ROMs) have been thoroughly investigated as an alternative to PDE descriptions \citep{qin2024modeling, renda2018discrete, dinev2018fepr, grazioso2019geometrically, boyer2020dynamics}. These range from volumetric FEM \citep{dubied2022sim, menager2023direct, ferrentino2023finite}, possibly combined with model order reduction techniques, to strain-based Cosserat descriptions \citep{renda2020geometric, boyer2020dynamics, caasenbrood2023control}, to other techniques specifically tailored to soft robots - like quasi-rigid models \citep{sfakiotakis2014multi} or volume-preserving primitives \citep{xu2023model}. Despite the vast heterogeneity, all these techniques follow the principles of Lagrangian mechanics and are described by ordinary differential equations of the form
\begin{equation}\label{eq:finite dimensional equations}
    \Mm(\qv)\ddqv + \cv(\qv, \dqv) + \gv(\qv) + \sv(\qv, \dqv) = \nuv(\qv, \uv),
\end{equation}
where symbols are defined in Table~\ref{tab:notation}. 
The Inverse Dynamics (ID) problem 
refers to the challenge of computing 
the left-hand side of~\eqref{eq:finite dimensional equations}. The ID problem plays a pivotal role in various applications that are crucial for the advancement of autonomous and intelligent robotic systems, including real-time control~\citep{buondonno2016efficient, boyer2020dynamics, liu2023hierarchical}, system identification~\cite{gaz2019dynamic}, trajectory generation~\citep{longhini2023edo} and optimization~\citep{saunders2010modeling, ferrolho2021inverse}, and mechanical design~\citep{chen2020design, pinskier2022bioinspiration}. For example, in case of full actuation, it is a common control approach to feedback linearize the system via direct application of the ID law: $\vect{\nu} = \Mm(\qv)\alphav + \cv(\qv, \dqv) + \gv(\qv) + \sv(\qv, \dqv)$, where $\alphav$ is a control acceleration.
%
In all these contexts, the ID must be executed in real-time or faster. Thus, computational efficiency becomes essential, especially when the dynamics has many DoFs\footnote{It is noteworthy that a procedure for the ID problem can also be used to solve - efficiently although not optimally - the forward dynamics (FD) problem, namely the problem of computing $\ddqv$ from $\qv$, $\dqv$ and $\uv$, and to implement implicit integration schemes~\citep{boyer2024implicit}. For example, this is an essential tool for data-hungry learning techniques trained in simulation \citep{li2021deep, li2022towards, li2024continual}.}. In this context, recursive formulations of the ID have been imposed themselves as a way of solving the ID problem with minimal computational time \citep{featherstone2014rigid}. \review{Indeed, recursive ID algorithms scale linearly with the number of bodies, whereas non-recursive algorithms scale quadratically. As a result, the former require less resources and offer faster computation times, particularly in systems with many bodies.}

Substantial research has been done on recursive ID for non-conventional robotic structures, as discussed in Sec. \ref{sec:SoA}, with some focused attention already being devoted to continuum soft robots \citep{ godage2019center, boyer2020dynamics, jensen2022tractable}. However, all existing \review{recursive} methods have the disadvantage of being specifically tailored to a single ROM technique, thus severely constraining the range of systems that can be dealt with.

This work proposes a more holistic recursive formulation of soft-rigid systems. Being agnostic to the underlying reduced order kinematic modeling technique, it can deal with a vast range of systems and is readily applicable to novel kinds of structures or models.\reviewr{, including existing methods as sub-cases.} A key enabling idea of our approach is to treat kinematics as an input to the procedure rather than a predefined information. In particular, the robot is seen as an assembly of kinematic modules (joints or deformable bodies) whose relative motion is parameterized by a generic kinematics-like function of material and configuration variables. Beside assuming that the kinematics belongs to a finite-dimensional space, the only hypothesis is that the contact area between two adjacent modules does not deform. The proposed derivations build on the Kane method~\citep{kane1983use, banerjee2022flexible}, \review{which is an equivalent form of the virtual work principle. For the first time, we demonstrate that the Kane equations can be manipulated to obtain a recursive formulation of the dynamics, independent of the soft body kinematics. This result has two implications. First, both existing and new kinematic models can be seamlessly integrated into our framework. Second, it enables the efficient computation of the ID.} The resulting equations of motion (EoM) can be viewed as a recursive form of the Euler-Lagrange (EL) equations and a generalization to the seminal results of~\citep{hollerbach1980recursive, book1984recursive}. These works derived a recursive EL equations with linear complexity for serial rigid~\citep{hollerbach1980recursive} and flexible link robots~\citep{book1984recursive}. Analogous to the Newton-Euler approach for rigid systems, our \reviewr{proposed} method has cost linear in the number of bodies or, equivalently, in the number of DoFs. \review{However, as we shall see, treating the kinematics as a black-box function limits recursivity to the system bodies, preventing its extension to the material domain --- unlike ID algorithms designed for specific kinematic models.}

The main contributions of the paper are summarized below.
\begin{enumerate}
    \item We derive a model-agnostic recursive formulation of the EoM for generic (i.e., agnostic to discretization technique) serial soft-rigid robots \review{whose bodies are rigidly connected to the joints}.
    \item By exploiting this recursive representation of the dynamics, we introduce two novel algorithms for computing the generalized active and actuation forces. In both cases, we prove that the computational complexity grows linearly with the number of DoFs.
    \item We demonstrate how the algorithms solving the ID problem can be minimally modified to evaluate the mass matrix without affecting computational complexity. Additionally, we propose embedding such computation within the ID algorithms, as it provides greater flexibility for control purposes.
\end{enumerate}

The findings are supported by numerical results on multiple robotic systems - which cannot be handled using current recursive methods. We present for the first time a dynamic simulation with the control of a trimmed helicoid robot using the locally volume preserving (LVP) primitives method~\citep{xu2023model} \review{for 3D continua}. In a second simulation, we consider a hybrid arm consisting of a rigid manipulator, a slender soft body modeled using the geometric variable strain (GVS) approach~\citep{boyer2020dynamics}, and a soft gripper with LVP kinematics. We show then how the proposed recursive equations scale with the number of bodies compared to the EL approach, which is the only framework that  currently can describe the class of robotic systems considered in this paper. 

\begin{table}[t]\caption{\small \review{Nomenclature}}
\centering
\begin{tabular}
{p{0.35\linewidth} p{0.55\linewidth} }
\toprule
\toprule
Symbol & Description\\
\toprule
$\R^{n}$ & Euclidean space of dimension $n$\\
$\R^{n \times m}$ & Space of $n \times m$ matrices over $\R$\\
$\calB_{i}$ &Body $i$ of the system\\
$\calJ_{i}$ &Joint $i$ of the system\\
$\{ S_{0} \}$ & Inertial reference frame\\
$\{ S_{i} \}$ & Reference frame of $\calB_{i}$\\
$\{ S_{\calJ_{i}} \}$ & Reference frame of $\calJ_{i}$\\
$V_{i} \subset \R^{3}$ & Undeformed volume of $\calB_{i}$\\
$\xv_{i} \in V_{i}$ & Material coordinates of $\calB_{i}$
\\
$\qv, \dqv, \ddqv \in \R^{n}$ & Configuration variables and their time derivatives\\
$\uv \in \R^{m}$ & Control variables\\
$\yv \in \R^{m}$ & Actuation coordinates\\
$\zv \in \R^{n-m}$ & Completing set of coordinates \\
$\fv_{\calB_{i}}(\xv_{i}, \qv_{i})$ & Reduced-order kinematics of $\calB_{i}$\\
$(\nv_{1_{i}} \,\, \nv_{2_{i}} \,\, \nv_{3_{i}})$ & Orthonormal basis for $\{ S_{i} \}$\\
$\Mm(\qv) \in \R^{n \times n}$ & Generalized mass matrix\\
$\cv(\qv, \dqv) \in \R^{n}$ & Generalized Coriolis and centrifugal force\\
$\gv(\qv)$ & Generalized gravitational force\\
$\sv(\qv, \dqv)$ & Generalized visco-elastic force\\
$\nuv(\qv, \uv)$ & Generalized actuation force\\
$\Qm^{*}(\qv, \dqv, \ddqv)$ & Generalized inertial force\\
$\Qm(\qv, \dqv, \uv)$ & Generalized active force\\
$\mat{\mathbb{I}}_{n}$ & Identity matrix of dimension $n$\\
$\zerov_{n \times m}$ & $n \times m$ matrix of zeros\\
$\Sm_{i}$ & $i$-th column of $\Sm \in \R^{n \times m}$\\
$S_{ij}$ & Element $(i, j)$ of $\Sm$\\
$\Am \otimes \Bm$ & Kronecker product\\
$\mathrm{vec}(\Am)$ & Column-wise vectorization of a matrix\\
$\tilde{\rv} = \small \begin{carray}{ccc}
    0 \hspace{-6pt} & \hspace{-6pt} -r_{3} \hspace{-6pt} & \hspace{-6pt} r_{2}\\
     r_{3} \hspace{-6pt} & \hspace{-6pt} 0 \hspace{-6pt} & \hspace{-6pt} -r_{1}\\
    -r_{2} \hspace{-6pt} & \hspace{-6pt} r_{1} \hspace{-6pt} & \hspace{-6pt} 0
\end{carray}$ & Skew symmetric matrix from $\rv$\\
$\av \times \bv = \Tilde{\av}\bv$ & Cross product\\
$\jac{\fv}{\xv} \in \R^{l \times h} $ & Jacobian of $\fv \in \R^{l}$ at $\xv \in \R^{h}$\\
$\parv{\fv}{\xv} = \rb{\jac{\fv}{\xv}}^{T}$ & Transposed Jacobian of $\fv$ at $\xv$\\
$\text{div}_{\xv} \fv$ & Divergence of $\fv$ at $\xv$\\
\bottomrule
\bottomrule
\end{tabular}
\label{tab:notation}
\end{table}

\begin{table}[t]\caption{\small \review{Main abbreviations}}
\centering
\begin{tabular}
{p{0.35\linewidth} p{0.55\linewidth} }
\toprule
\toprule
CC & Constant Curvature\\
EL & Euler-Lagrange\\
EoM & Equations of Motion\\
GVS & Geometric Variable Strain\\
ID & Inverse Dynamics\\
IID & Inertial Inverse Dynamics\\
LVP & Locally Volume Preserving\\
MID & Mass Inverse Dynamics\\
MIID & Mass Inertial Inverse Dynamics\\
PCC & Piecewise Constant Curvature\\
PCS & Piecewise Constant Strain\\
ROM & Reduced-Order Model\\
\bottomrule
\bottomrule
\end{tabular}
\label{tab:acronyms}
\end{table}

\subsection{State of the art on ID problem}\label{sec:SoA}

Numerous studies have explored the computation of the ID problem for slender flexible link robots, i.e., under the small deformations assumption, using different approaches. These include the EL method~\citep{book1984recursive, chen2001dynamic, zhang2009recursive, my2019efficient}, the generalized Newton-Euler (NE) equations~\citep{shabana1990dynamics, boyer1998efficient, khalil2017general}, the Gibb-Appell equations~\citep{korayem2014systematic}, and the Kane equations~\citep{singh1985dynamics, banerjee1990dynamics, amirouche1993explicit}. All the mentioned works share three main assumptions: (i) small deflections, (ii) slender bodies, and (iii) the possibility to approximate the deformable motion, \review{i.e., changes in shape and size,} as a \review{linear} function of the configuration variables \review{--- a set of independent parameters that uniquely define the body relative position}. Unfortunately, assumptions (i) and (iii) are seldom verified by ROMs of soft robotic systems. Consequently, new ID algorithms have been developed to relax the above assumptions to some extent. 

Under a rigid-body discretization and the piecewise constant curvature kinematic model,~\cite{rone2013continuum} presents a procedure based on the Kane method to evaluate the ID of slender continuum manipulators. Later, under the same kinematic assumptions,~\cite{godage2019center} follows a Lagrangian approach to compute the dynamic matrices with a $O(n^2)$ cost. Although not explicitly stated, this procedure can also be used to evaluate the ID. In~\citep{jensen2022tractable}, the authors apply the recursive Newton-Euler algorithm for rigid body systems to a lumped mass model of a six-DoFs slender soft arm with piecewise constant curvature. Under the Cosserat rod hypothesis,~\cite{renda2018discrete} proposes an ID procedure for robots with piecewise constant strain (PCS). By leveraging the rigidity of the robot cross sections, the authors derive a recursive ID algorithm that generalizes the Newton-Euler procedure for rigid arms. This algorithm has computational complexity linear in the number of bodies, making it optimal. The method was extended in~\cite{renda2018geometric} to consider hybrid systems consisting of PCS soft and rigid bodies. Later,~\cite{boyer2020dynamics} relaxed the PCS hypothesis in favor of a modal Ritz reduction of the strain. Recently,~\cite{anup2024reduced} introduced a state- and time-dependent basis of the strain. These \review{recursive methods} are limited to bodies satisfying the Cosserat rod assumption, i.e., thin bodies (hypothesis (ii)) with rigid cross sections, and, consequently, are inapplicable to kinematic models of non-slender soft robotic systems~\citep{sharp2023data, xu2023model}. \review{Nonetheless, it is important to note that ID algorithms for discrete Cosserat rods enable a more efficient computation compared to our method. Specifically, the degree of recursivity applies not only between the bodies of the system --- as in our approach --- but also within the soft material domain. This advantage arises because the kinematics of Cosserat rods can be leveraged to obtain recursivity also within the soft material domain and, consequently, improve the speed of the algorithm. }

\review{Finally,~\cite{sadati2021tmtdyn} introduced the TMT method for soft bodies. This approach enables the derivation of~\eqref{eq:finite dimensional equations} in vector form, reducing the number of required steps compared to the EL method. While it facilitates a more efficient implementation of the dynamics, it still maintains the quadratic computational cost of a non-recursive formulation. Moreover, although the TMT method has been successfully applied to compute the ID of various reduced-order kinematic models for soft robots, its extension to 2D and 3D geometries assumes that these bodies can be modeled as wire meshes, where edges correspond to 1D Euler-Bernoulli beams and connections are point masses. Consequently, its applicability to the generic kinematic models considered in this work remains unexplored.}

\review{
To summarize, the main benefits of our approach are as follows.
\begin{itemize}
    \item 
    It is possible to evaluate the ID for robots with arbitrarily deformable bodies, such as rigid arms with soft grippers and non-slender soft manipulators. Table~\ref{tab:comparison} provides a non-exhaustive list of models supported by our framework compared to other existing solutions.
    \item In view of the recursive formulation, the computation of the ID scales linearly with number of bodies.
    \item 
    Since the method is agnostic to the kinematics, new models can be easily conceptualized and applied in control tasks. 
\end{itemize}

On the other hand, the following limitations exist.
\begin{itemize}
    \item 
    It is not possible to exploit recursivity within the soft material domain. As a result, for certain subclasses of kinematic models, more efficient algorithms may be available.
    \item The derivations rely on the assumption that the soft bodies kinematics is purely geometrical. 
    \item 
    The introduction of an abstraction layer on the kinematics could make the model more difficult to understand and interpret. 
\end{itemize}

}


\begin{table*}[t]
\scriptsize	
\caption{\small \review{Comparison of kinematic models for deformable bodies supported by our and other existing frameworks. A cross indicates whether a particular model is supported.}}
\centering
\begin{tabular}{lccccccc }
\toprule
\toprule
Kinematic model & Proposed framework & \makecell{SOFA\\\cite{coevoet2017software}} & \makecell{TMTDyn\\\cite{sadati2021tmtdyn}} & \makecell{SoroSim\\\cite{mathew2022sorosim}} & \makecell{Sorotoki\\\cite{caasenbrood2024sorotoki}} \\
\toprule
\makecell[cl]{Piecewise constant curvature\\\cite{webster2010design}} & $\times$ & & $\times$ & $\times$ & $\times$\\
\makecell[cl]{Geometric variable strain\\\cite{boyer2020dynamics}} & $\times$ & & $\times$ & $\times$ & $\times$\\
\makecell[cl]{Ritz--Galerkin on position\\\cite{sadati2017control}} & $\times$ & $\times$ & $\times$ & & $\times$\\
\makecell[cl]{Locally volume preserving primitives\\\cite{xu2023model}} & $\times$ & & &\\
\makecell[cl]{Learning-based kinematics\\\cite{sharp2023data}} & $\times$ & & & & & &\\
\bottomrule
\bottomrule
\end{tabular}
\label{tab:comparison}
\end{table*}

\subsection{Notation}
We denote vectors and matrices with bold letters. Arguments of the functions are omitted when clear from the context. 
Table~\ref{tab:notation} presents the notation adopted in the paper. \review{Additionally, Table~\ref{tab:acronyms} summarizes the main abbreviations used in the text.}

\subsection{Code}

We also provide in a GitHub repository an expandable object-oriented~\href{https://github.com/piepustina/Jelly}{MATLAB library} with full C/C++ code generation functionality implementing all the algorithms presented in the following.

\subsection{Supplementary material}
In this paper, we show numerical results on recently introduced kinematic models of soft robots. However, the proposed strategy can be seamlessly applied to more conventional robots, such as Cosserat rods. The interested reader can find supplementary material detailing the application of the method to other systems at the link~\href{http://arxiv.org/abs/2402.07037}{Supplementary Material}.

\subsection{Structure of the paper}

The rest of the paper is organized as follows. In Sec.~\ref{sec:problem statement}, we formally introduce the goal of the paper. In Sec.~\ref{section:kinematics}, we discuss the proposed \textit{kinematics-as-an-input} setting and establish the configuration space that characterizes the considered class of systems. Herein, we introduce the direct and the first- and second-order differential kinematics. Sec.~\ref{section:kane equations} presents the Kane equations for the system. Their recursive form is obtained in Sec.~\ref{section:recursive formulation}, offering immediate utility in assessing generalized active forces with cost linear in the number of bodies. We also demonstrate how this recursive structure can be exploited to evaluate the mass matrix without affecting computational complexity.
In Sec.~\ref{section:actuation inverse dynamics}, we expand the active forces to derive a recursive procedure for evaluating only the actuation forces in the ID problem. Sec.~\ref{section:simulations} presents simulations results, and Sec.~\ref{section:conclusions} draws the main conclusions of the paper and includes future work.

\begin{figure}[t!]
    \centering
\includegraphics[width=0.65\columnwidth]{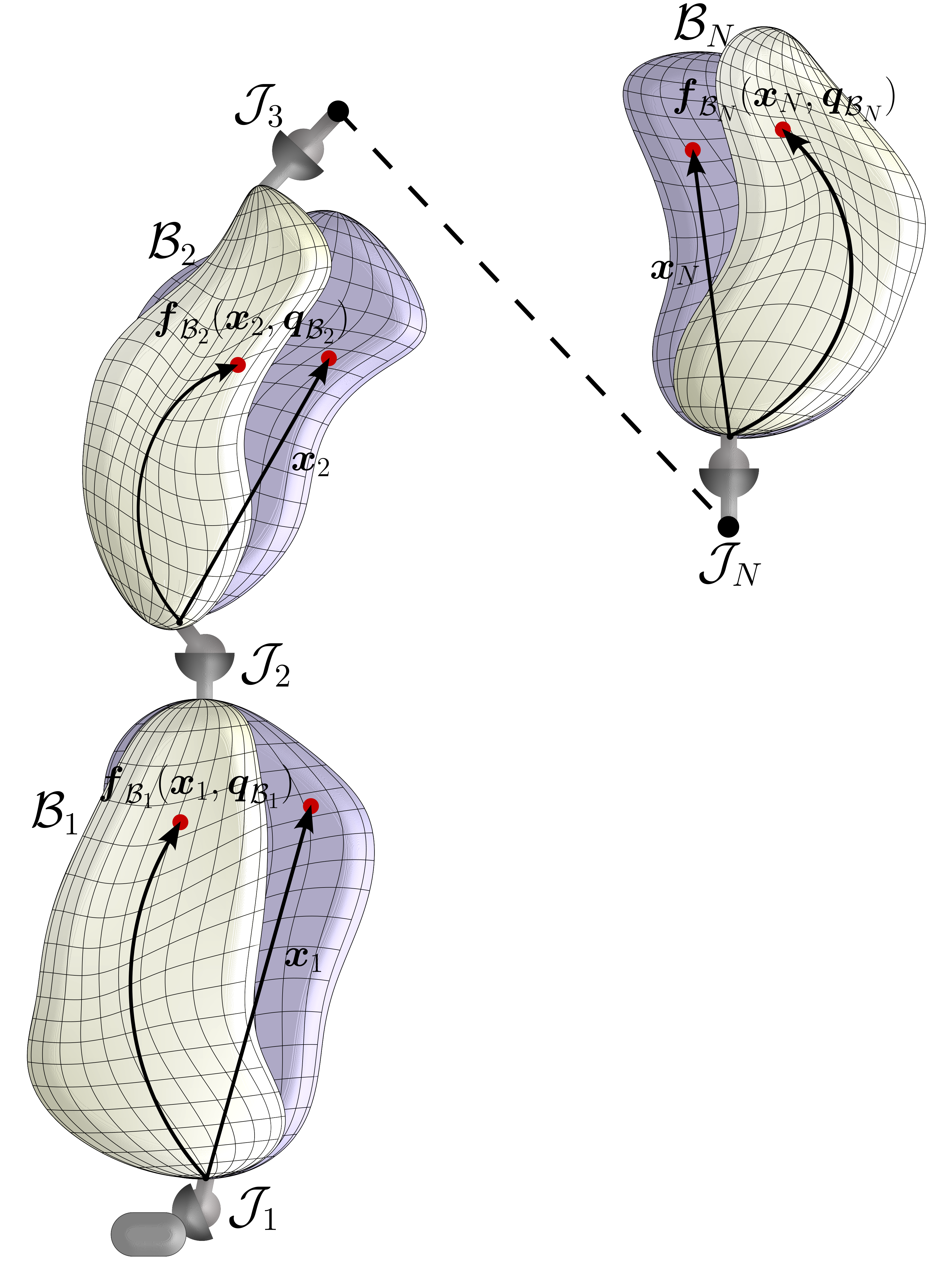}
    \caption{\small Schematic representation of the considered class of soft robotic systems, conceptualized as a sequence of $N$ deformable bodies $\calB_{i}$ and joints $\calJ_{i}$. The purple and pale brown volumes illustrate the robot bodies in their stress-free and deformed configuration, respectively. The individual kinematics of each body $\calB_{i}$ is modeled as a generic function $\fv_{\calB_{i}}(\xv_{i}, \qv_{\calB_{i}}) \in \R^{3}$, where $\xv_{i} \in \R^{3}$ denotes the (relative) position of $\calB_{i}$ points in the stress-free configuration and $\qv_{\calB_{i}} \in \R^{n_{\calB_{i}}}$ is a configuration vector parametrizing the body motions.
    }
    \label{fig:system sketch}
\end{figure}

\section{Problem statement}\label{sec:problem statement}
\review{In this section, we define the ID problem for the class of robots under consideration and introduce the two hypotheses under which it is addressed: (i) the kinematics of deformable bodies can be described geometrically using a finite number of configuration variables, and (ii) the bodies are connected through rigid joints.}

Consider a holonomically constrained~\review{\footnote{\color{red} A robot is holonomically constrained if it is subject to constraints that depend only on position (and possibly time), but not on velocity.}} serial chain with fixed base constituted by $N$ moving bodies $\calB_{i}$, each connected to its predecessor by a joint $\calJ_{i}$, for $i=1,\dots,N$. \review{From now on, if not explicitly stated, the subscript $(\cdot)_{i}$ refers to a quantity of body or joint $i$.}

Assume that the following conditions hold, as illustrated in Fig.~\ref{fig:system sketch}.
\begin{assumption}\label{assumption:finite dimensional kinematics}
     The kinematics of each body $\calB_{i}$ belongs to a finite-dimensional \review{configuration} space.
\end{assumption}
Mathematically, Assumption~\ref{assumption:finite dimensional kinematics} requires that the position $\pv_{\calB_{i}} \in \R^{3}$ of each body point relative to the distal end of its joint takes the form
\begin{equation}\label{eq:functional expression body kinematics}
    \pv_{\calB_{i}} = \fv_{\calB_{i}}(\xv_{i}, \qv_{\calB_{i}}),
\end{equation}
where $\xv_{i} \in V_{i} \subset \R^{3}$ collects the Cartesian material coordinates of $\calB_{i}$ in the reference (stress-free) configuration $V_{i}$, $\qv_{\calB_{i}} \in \R^{n_{\calB_{i}}}$ parameterizes the body motions, and $\fv_{\calB_{i}}$ is a generic function, possibly highly nonlinear in both arguments. 
\review{
The previous equation implicitly requires that the kinematics can be described purely geometrically. In other words, the dependence of $\pv_{\calB_{i}}$ on external forces, such as those of actuation $\uv$, is not modeled explicitly. Instead, this dependence arises through the dynamic influence of $\uv$ on $\qv_{\calB_{i}}$, as expressed in~\eqref{eq:finite dimensional equations}. This simplification, which is widely adopted in the reduced-order modeling of soft robots, \cite{boyer2020dynamics, della2020model, sadati2021tmtdyn, xu2023model, caasenbrood2024sorotoki}, can be motivated as follows. The instantaneous effects of $\uv$ on the kinematics can be reasonably captured by introducing additional configuration variables, which are then influenced by $\uv$ through the EoM.
Examples of this approach can be found in Simulation~2 of the paper and Simulation~4 in the supplementary material. 
}
\begin{assumption}\label{assumption:contact area}
    For each body, the contact areas at the joints do not deform.  
\end{assumption}
This assumption reasonably holds for the majority of soft robotic systems since a joint typically introduces some degree of rigidity at the contact areas. As we shall see, Assumption~\ref{assumption:contact area} is not required from the perspective of the dynamics equations and does not effect the recursion, but it is necessary for describing the robot kinematics in presence of non-fixed joints. As a matter of fact, when the bodies are linked through fixed joints, it can be removed. Relaxation of the above assumption will be considered in one of the simulations. 
%
\begin{remark}
We do not assume any specific structure for~\eqref{eq:functional expression body kinematics} and consider it an input to the ID procedure, similar to the approach of~\cite{featherstone2014rigid} for rigid robots. This allows us to decouple the kinematic assumption typically made to solve the ID problem for specific sub-classes of soft robots from the algorithm itself. For the purposes of this paper, the reader should consider $\fv_{\calB_{i}}$ as a black-box function, which could be obtained in various ways~\citep{sharp2023data, xu2023model}. This structure also encompasses popular kinematic functions used in soft robot modeling\review{~\citep{sadati2020reduced}}, such as reduced-order FEM~\citep{menager2023direct} and discrete Cosserat models~\citep{renda2020geometric, dellasantina2023survey}.
\end{remark}
Assumption~\ref{assumption:finite dimensional kinematics} implies that the robot dynamics follows the principles of Lagrangian mechanics, thus taking the form of~\eqref{eq:finite dimensional equations} and leading to the following problems.
%
%
\begin{problem}[\bf Inertial Inverse Dynamics Problem]\label{problem:IDP}
Under Assumption~\ref{assumption:finite dimensional kinematics} and~\ref{assumption:contact area}, find an algorithm $\mathrm{IID}(\qv, \dqv, \ddqv)$ with $O(N)$ complexity such that
\begin{equation*}
    \mathrm{IID}(\qv, \dqv, \ddqv) = \Mm(\qv)\ddqv + \cv(\qv, \dqv).
\end{equation*}
\end{problem}
\begin{problem}[\bf Inverse Dynamics Problem]\label{problem:AIDP}
Under the same hypotheses of Problem~\ref{problem:IDP}, find an algorithm $\mathrm{ID}(\qv, \dqv, \ddqv)$ with $O(N)$ complexity such that
\begin{equation*}
    \mathrm{ID}(\qv, \dqv, \ddqv) = \Mm(\qv)\ddqv + \cv(\qv, \dqv) + \gv(\qv) + \sv(\qv, \dqv).
\end{equation*}
\end{problem}
It is worth remarking that we consider the output of the ID as the generalized actuation force $\nuv(\qv, \uv)$, rather than the control variables $\uv$, as done for rigid fully-actuated robots. This choice is motivated by two  considerations. First, the expression of $\nuv$ depends on the type of actuation and how it is modeled. Consequently, deriving a general method to compute $\uv$ from $\nuv$ is not straightforward. As shown in Sec.~\ref{section:simulations}, when $\nuv(\qv, \uv) = \Am(\qv)\uv$ and $\Am(\qv)$ satisfies certain integrability conditions~\citep{pustina2024input}, a change of coordinates allows $\uv$ to directly influence the dynamics, making it possible to compute $\uv$ directly through the ID algorithm. However, this is a relatively specific case and does not universally apply. Second, the actuation forces commonly used in soft robotics are often subject to constraints~\citep{tonkens2021soft, bruder2021data}. For example, a tendon can only generate a pulling force, which must be taken into account when computing $\uv$ from $\nuv$~\citep{dellasantina2019constraints}. 

Another important aspect to discuss is the relationship between the inputs to the $\mathrm{IID}$ and $\mathrm{ID}$ procedures, namely $\qv$, $\dqv$ and $\ddqv$, the generalized actuation force $\nuv(\qv, \uv)$, and the control variables $\uv$. Even in finite-dimensional models of soft robotic systems, the number of control variables is typically less than the number of generalized coordinates, i.e., $m < n$. This implies that $\uv$ can only directly control the evolution of a subset of the configuration variables $\qv$ or, in general, a vector of output variables $\yv$ of dimension $m$ (assuming all control inputs are independent). Consequently, $n-m$ configuration variables do not evolve freely. As a result, if the ID algorithm in a given state $(\qv,\dqv)$ is supplied with an arbitrary acceleration value $\ddqv$, there is no guarantee that a command $\uv$ exists such that~\eqref{eq:finite dimensional equations} holds. In other words, at any given instant $t$, there may not exist a $\uv$ such that $\nuv(\qv(t), \uv) = \mathrm{ID}(\qv(t), \dqv(t), \ddqv(t))$. This issue can potentially be resolved if there exists a set of generalized coordinates $\thetav = \rb{\yv^{T} \,\, \zv^{T} }^{T}$ such that $\yv \in \R^{m}$ -- referred to hereinafter as \textit{actuation coordinates} -- can be directly controlled through $\uv$, and $\zv \in \R^{n-m}$ is the complement to $\yv$ of coordinates not directly actuated by $\uv$. In this scenario, the unactuated coordinates $\zv$ satisfy a set of second-order differential constraints that takes the form
$
    \ddzv = -\Mm_{\zv \zv}^{-1}\rb{ \Mm_{\zv \yv}\ddyv + \cv_{\zv} + \gv_{\zv} + \sv_{\zv}},
$
with suitable sub-blocks of the dynamic terms in~(\ref{eq:finite dimensional equations})  rewritten in the $\thetav$ coordinates.
Thus, for any generic state $(\yv, \zv, \dyv, \dzv)$ and arbitrary $\ddyv$, selecting $\ddzv$ according to the previous equation ensures that the output $\nuv(\thetav, \uv)$ of the ID algorithm can be achieved by an appropriate choice of $\uv$. The existence of such actuation coordinates in generic soft robotic systems remains an open question and has only been explored so far when the dynamics is \review{linear} in the input~\citep{pustina2024input}. 

Note also that in addition to the ID problem (Problem~\ref{problem:AIDP}), we introduce another problem to be solved, namely the Inertial Inverse Dynamics (IID) problem (Problem~\ref{problem:IDP}), for three reasons. First,  active forces on the robot may also differ from those considered in~\eqref{eq:finite dimensional equations} --- despite $\gv(\qv) + \sv(\qv,\dqv)$ is the most common case, or some of them may be absent, such as when the robot is moving on a horizontal plane. This approach allows us to initially focus on computing the inertial forces in $\mathrm{IID}$, and only later distinguish the actuation forces (the unknowns) from the active forces. Second, by addressing Problem~\ref{problem:IDP}, we can concentrate on the inertial terms alone, which are the most challenging to compute. Third, the Kane equations yield naturally a distinction between inertial and active forces, driving us naturally to consider Problem~\ref{problem:IDP} as well. Although we differentiate between these two problems, their solutions will be developed almost in parallel.
\begin{remark}
    Because we are considering robots with a serial topology, the ID problem cannot be solved in less than $O(N)$ steps~\citep{featherstone2014rigid}. Additionally, since the number of DoFs $n$ is a multiple of the number of bodies $N$, in terms of computational complexity one has $O(N) = O(n)$.
\end{remark}

\review{
\begin{examplestop*}
    To facilitate the understanding of our algorithms, throughout the paper we present the computation of the main terms required by our procedures for a planar soft body modeled under the Constant Curvature (CC) assumption~\citep{della2020model}. This body is connected to its predecessor via a fixed joint and has a cylindrical shape with unitary radius and length. We choose a CC body for three main reasons. First, the CC kinematics is a well-known and widely used model in soft robotics. Second, despite its relative simplicity, the CC model exhibits significant nonlinearities, making the solution of the ID problem non trivial, even when the system has only a few bodies. Finally, Simulation~3 is a natural continuation of this example. 
    
    For the ease of reading, the short notation $\mathrm{c}_{x}$ and $\mathrm{s}_{x}$ stands for $\cos(x)$ and $\sin(x)$, respectively. Moreover, to avoid unnecessary use of subscripts, in this example only, we use the notation $\qv_{\calB_{i}} = q$. 

    For the CC model, the reduced-order kinematics~\eqref{eq:functional expression body kinematics} reads
    \begin{equation}\label{eq:PCC:kinematics}
        \pv_{\calB_{i}} = \begin{pmatrix} 
        \displaystyle \frac{\mathrm{c}_{q x_3} - 1}{q} + x_1 \mathrm{c}_{q x_3} \\[4pt] 
        x_2 \\[3pt] 
        \displaystyle \frac{\mathrm{s}_{q x_3} (q x_1 + 1)}{q} 
        \end{pmatrix},
    \end{equation}
    where $x_{1}$ and $x_{2}$ parametrize a circle of unitary radius, namely the robot cross section, and $x_{3} \in [0, 1]$, i.e., $x_{3}$ represents the distance from the base. 
\end{examplestop*}
}

\section{Kinematics}\label{section:kinematics}
As a preliminary step towards solving Problem~\ref{problem:IDP}, this section characterizes the forward kinematics for a robot whose bodies kinematics take the generic form~\eqref{eq:functional expression body kinematics}. We also briefly introduce the first and second-order differential kinematics, which will be used later in the solution of the ID. 

\begin{figure}
    \centering
    \includegraphics[width=0.7\columnwidth]{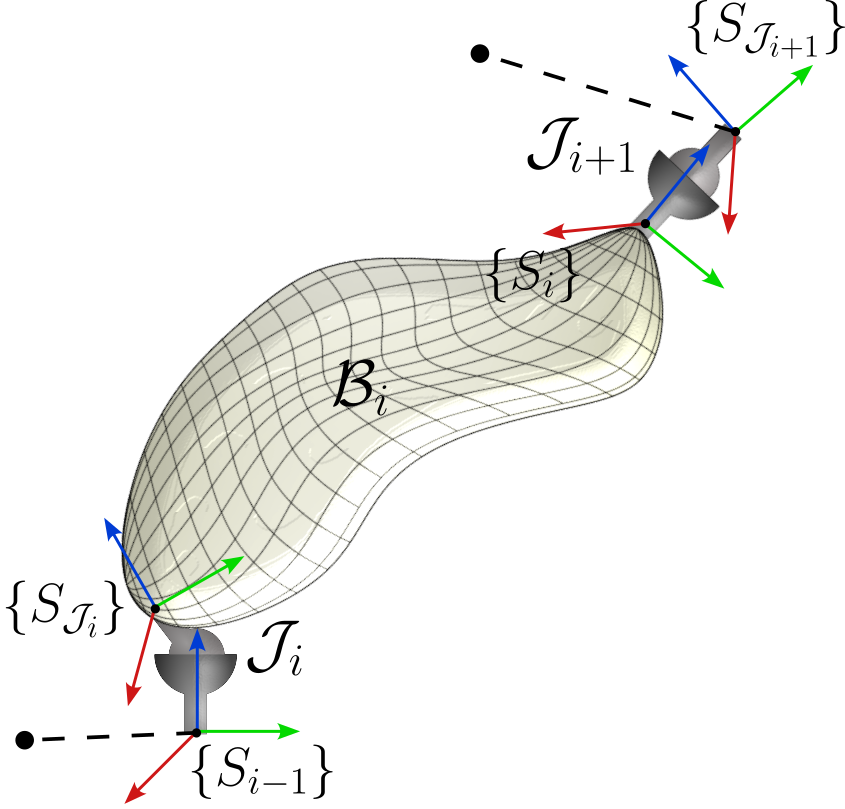}
    \caption{\small For each body $\calB_{i}$, we introduce two reference frames $\{ S_{i} \}$ and $\{ S_{\calJ_{i}} \}$. The frame $\{ S_{i} \}$ is attached at the connection point between $\calB_{i}$ and its successor joint $\calJ_{i+1}$, and accounts for body deformability. On the other hand, $\{ S_{\calJ_{i}} \}$ is attached at the distal end of $\calJ_{i}$, which coincides with the connection point between $\calJ_{i}$ and $\calB_{i}$, and is used to model the relative motion between $\calB_{i-1}$ and $\calB_{i}$ due to the joint. {\color{red} The reference frames are oriented as the contact areas between the corresponding body and joint.}}
    \label{fig:kinematics_frames}
\end{figure}

A complete description of the kinematics requires introducing two types of reference frames: a reference frame $\{S_{i}\}$ at the pivot point between $\calB_{i}$ and $\calJ_{i+1}$ with the same orientation of their rigid contact area, and another one $\{ S_{\calJ_{i}} \}$ at the distal end of $ \calJ_{i}$ with the same orientation of the area between $\calJ_{i}$ and $\calB_{i}$, as illustrated in Fig.~\ref{fig:kinematics_frames}. Note that other choices are possible, but still a pair of reference frames is needed for each body as explained below. An inertial reference frame $\{ S_{0} \}$ is also attached to the robot base, where the dynamics will be formulated. 

The frame $\{ S_{i} \}$ is used in conjunction with $\{ S_{\calJ_{i}} \}$ to track the relative motion between adjacent bodies as a consequence of the joint presence. 
Indeed, the relative position and orientation between $\calJ_{i}$ and $\calJ_{i+1}$ changes not only because of the joint motion but also because of $\calB_{i}$ deformability, thus requiring a reference frame ($\{ S_{i}\}$) attached to $\calB_{i}$ as well. In other words, at the kinematic level only, $\calB_{i}$ can be seen as a distributed joint that requires as such its own reference frame. This is also the reason why Assumption~\ref{assumption:contact area} is taken. As we shall see, this hypothesis allows us to describe the orientation of $\{ S_{i}\}$ starting from $\fv_{\calB_{i}}$. On the other hand, if the robot contains some bodies connected by fixed joints, one can adopt three possible strategies: (i) remove Assumption~\ref{assumption:contact area} for the bodies connected by fixed joints and describe the kinematics using only the pairs $\{ S_{i} \}$ and $\{ S_{\calJ_{i}} \}$ associated to bodies connected by non-fixed joints; (ii) retain Assumption~\ref{assumption:contact area} and introduce a single reference frame $\{ S_{i} \}$ for bodies connected by fixed joints; or (iii) maintain all reference frames and let $\{ S_{i} \} = \{ S_{\calJ_{i + 1}} \}$. In the following, to consider the more general case, i.e., the case of a robot with moving joints, we adopt the third strategy. Note also that because of the same above reason, for the last body $\calB_{N}$, one does not need in principle to introduce $\{ S_{N} \}$ but only $\{ S_{\calJ_{N}} \}$. We now move to the computation of the homogeneous transformation matrices describing the relative motion between the above frames.

\begin{figure}
    \centering
    \includegraphics[width=1\columnwidth]{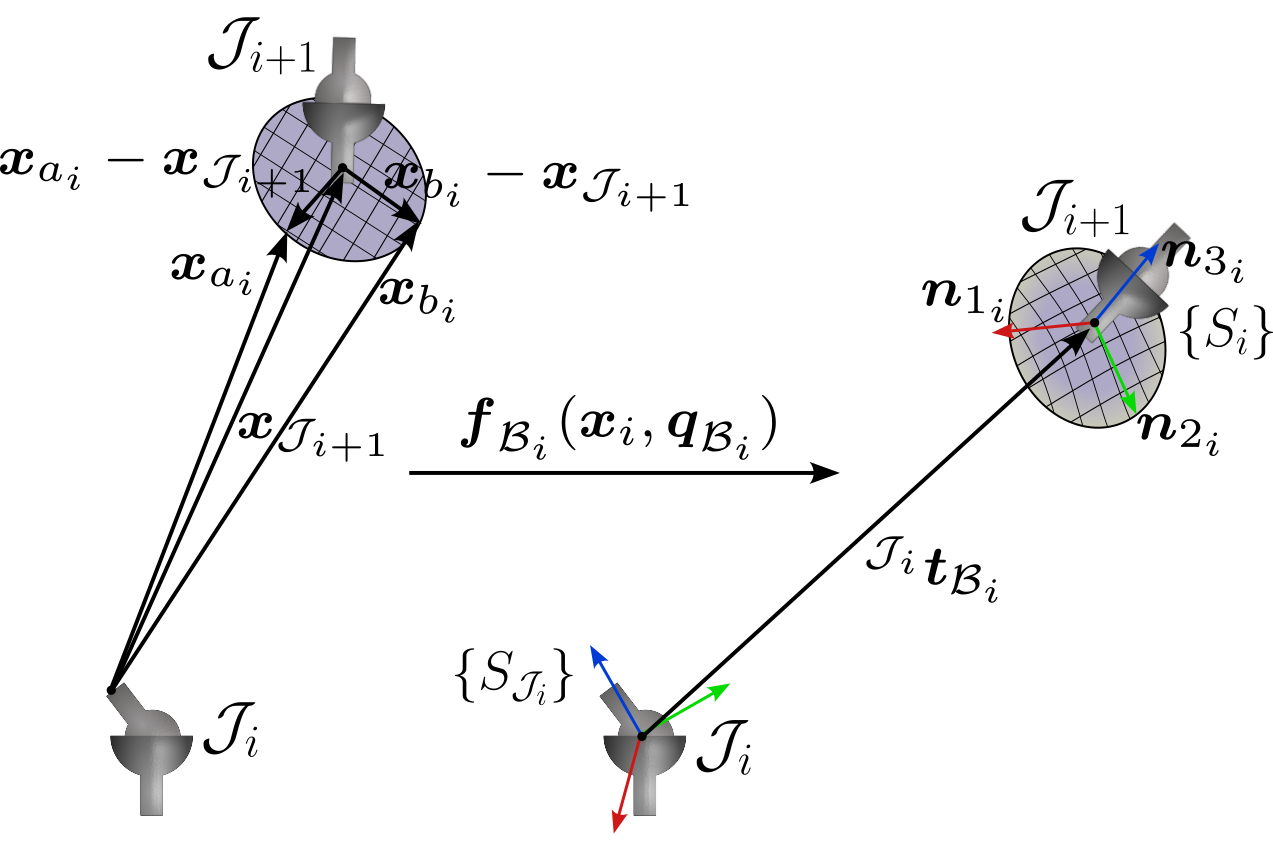}
    \caption{\small \color{red} Graphical representation of how the transformation matrix from $\{ S_{i} \}$ to $\{S_{\calJ_{i}}\}$ can be constructed using the reduced-order kinematics, under the hypothesis that the bodies are rigidly connected to the joints.}
    \label{fig:frames_steps}
\end{figure}

Let $\xv_{\calJ_{i+1}}$ be the pivot point of joint $\calJ_{i+1}$ at body $\calB_{i}$, and $\xv_{a_{i}}$ and $\xv_{b_{i}}$ two points of $\calB_{i}$ belonging to the contact area with $\calJ_{i+1}$ such that 
\begin{equation}\label{eq:ortogonality condition}
(\xv_{a_{i}} - \xv_{\calJ_{i+1}})^{T}(\xv_{b_{i}} - \xv_{\calJ_{i+1}}) = 0.
\end{equation}
By leveraging Assumption~\ref{assumption:contact area}, it is possible to construct the homogeneous transformation matrix from $\{ S_{i} \}$ to $\{ S_{\calJ_{i}} \}$ as
\begin{equation}\label{eq:body to joint transformation}\small
    \Transm[\calB_i][\calJ_i][(\qv_{\calB_{i}})] = \begin{carray}{cccc}
     \nv_{1_{i}} & \nv_{2_{i}} & \nv_{3_{i}} & {}^{\calJ_{i}}\tv_{\calB_{i}} \\
    0 & 0 & 0 & 1
    \end{carray},
\end{equation}
where
\begin{equation}\label{eq:body to joint transformation:n}
    \begin{split}
    \nv_{1_{i}}(\qv_{\calB_{i}}) &= \frac{\fv_{\calB_{i}}(\xv_{a_{i}}, \qv_{\calB_{i}}) - \fv_{\calB_{i}}(\xv_{\calJ_{i+1}}, \qv_{\calB_{i}})}{\norm{\fv_{\calB_{i}}(\xv_{a_{i}}, \qv_{\calB_{i}}) - \fv_{\calB_{i}}(\xv_{\calJ_{i+1}}, \qv_{\calB_{i}})}},\\
    \nv_{2_{i}}(\qv_{\calB_{i}}) &= \frac{\fv_{\calB_{i}}(\xv_{b_{i}}, \qv_{\calB_{i}}) - \fv_{\calB_{i}}(\xv_{\calJ_{i+1}}, \qv_{\calB_{i}})}{\norm{\fv_{\calB_{i}}(\xv_{b_{i}}, \qv_{\calB_{i}}) - \fv_{\calB_{i}}(\xv_{\calJ_{i+1}}, \qv_{\calB_{i}})}} ,\\
    \nv_{3_{i}}(\qv_{\calB_{i}}) &= \nv_{1_{i}} \times \nv_{2_{i}},
\end{split}
\end{equation}
and
\begin{equation}\label{eq:body to joint transformation:t}
    {}^{\calJ_{i}}\tv_{\calB_{i}}(\qv_{\calB_{i}}) = \fv_{\calB_{i}}(\xv_{\calJ_{i+1}}, \qv_{\calB_{i}}).
\end{equation}
The above transformation contains the information about the position and orientation of the rigid contact area between $\calB_{i}$ and $\calJ_{i+1}$ as seen from $\calJ_{i}$. \review{Fig.~\ref{fig:frames_steps} illustrates how the elements of~\eqref{eq:body to joint transformation} are built starting from the body kinematics~\eqref{eq:functional expression body kinematics} and the three points $\xv_{a_{i}}, \xv_{b_{i}}$ and $\xv_{\calJ_{i+1}}$.} Note that, without loss of generality, we choose $\nv_{3_{i}}$ as the unit normal vector to the body-joint contact area. 

It is also worth observing that~\eqref{eq:body to joint transformation} is a function of $\qv_{\calB_{i}}$ only because it encodes the information due to deformability. When the body is rigid,~\eqref{eq:functional expression body kinematics} reduces to the identity, i.e.,~$\pv_{\calB_{i}} = \xv_{i}$, and~\eqref{eq:body to joint transformation}--\eqref{eq:body to joint transformation:t} is indeed constant.
Because of Assumption~\ref{assumption:contact area}, the orthogonality between $\xv_{a_{i}}-\xv_{\calJ_{i+1}}$ and $\xv_{b_{i}}-\xv_{\calJ_{i+1}}$ implies that between the unitary norm vectors $\nv_{1_{i}}$ and $\nv_{2_{i}}$. Consequently, the triad $\{\nv_{1_{i}} \,\, \nv_{2_{i}} \,\, \nv_{3_{i}}\}$ forms an orthonormal basis. Moreover,~\eqref{eq:body to joint transformation:n} is always well defined since the denominators are never zero and, from Assumption~\ref{assumption:contact area}, it holds 
\begin{equation*}
    \norm{\fv_{\calB_{i}}(\xv_{a_{i}}, \qv_{\calB_{i}}) - \fv_{\calB_{i}}(\xv_{\calJ_{i+1}}, \qv_{\calB_{i}})} = \norm{\xv_{a_{i}} - \xv_{\calJ_{i+1}}},
\end{equation*}
and
\begin{equation*}
    \norm{\fv_{\calB_{i}}(\xv_{b_{i}}, \qv_{\calB_{i}}) - \fv_{\calB_{i}}(\xv_{\calJ_{i+1}}, \qv_{\calB_{i}})} = \norm{\xv_{b_{i}} - \xv_{\calJ_{i+1}}},
\end{equation*}
which are constants.
\begin{remark}
    In some modeling approaches, such as the \review{geometric variable strain} or lumped mass models, the above transformation matrix is readily available since each point of the body is associated with a transformation matrix. However, this is not always the case as in the \review{locally volume preserving} method or reduced-order FEM.
\end{remark}

The relative motion between $\calB_{i-1}$ and $\calB_{i}$ due to $\calJ_{i}$ has not been considered yet. This effect on the kinematics is modeled by the joint homogeneous transformation 
\begin{equation}\label{eq:joint to body transformation}\small
    \Transm[\calJ_i][\calB_{i-1}][\rb{\qv_{\calJ_i}}] = \begin{carray}{cc}
    \Rotm[\calJ_i][\calB_{i-1}][\rb{\qv_{\calJ_i}}] & {}^{\calB_{i-1}}\tv_{\calJ_i}\rb{\qv_{\calJ_i}}\\
    \zerov_{1 \times 3} & 1
    \end{carray},
\end{equation}
where $\qv_{\calJ_i} \in \R^{n_{\calJ_{i}}}$ is the configuration vector of the joint and $n_{\calJ_{i}}$ its number of DoFs. Note that $\Transm[\calJ_i][\calB_{i-1}][\rb{\qv_{\calJ_i}}]$ is specified only by the joint type, as for rigid bodied systems~\citep{featherstone2014rigid}. 

For the sake of convenience, we define the vectors
\begin{equation*}
\qv_i = \begin{carray}{c} \qv_{\calJ_i}\\\qv_{\calB_i} \end{carray} \in \R^{n_i},
\end{equation*}
with $n_i = n_{\calB_i} + n_{\calJ_i}$, and
\begin{equation*}
\qv = \begin{carray}{c}
    \qv_1\\ \vdots\\\qv_{N}
\end{carray}\in \R^{n},
\end{equation*}
with $n = \sum_{i = 1}^{N}n_{i}$, which group the generalized coordinates of each body and of the entire system, respectively. 

\review{
\begin{example*}[(Continuation)]
    We can now proceed with the computation of $\,{}^{\calJ_{i}}\Tm_{\calB_{i}}$. To this end, we take the tip along the backbone as pivot point, i.e., $\xv_{\calJ_{i+1}} = \rb{0 \,\, 0 \,\, 1}^{T}$. Moreover, the choice $\xv_{a_{i}} = \rb{1 \,\, 0 \,\, 1}^{T}$ and $\xv_{b_{i}} = \rb{0 \,\, 1 \,\, 1}^{T}$ guarantees the orthogonality condition~\eqref{eq:ortogonality condition}. 

    Applying~\eqref{eq:body to joint transformation}-\eqref{eq:body to joint transformation:t} to the CC kinematics yields
    \begin{equation*}
        {}^{\calJ_{i}}T_{\calB_{i}} = \begin{pmatrix} 
            \mathrm{c}_{q x_3} & 0 & -\mathrm{s}_{q x_3} & \displaystyle \frac{\mathrm{c}_{q x_3} - 1}{q} \\[3pt] 
            0 & 1 & 0 & 0 \\[2pt] 
            \mathrm{s}_{q x_3} & 0 & \mathrm{c}_{q x_3} & \displaystyle \frac{\mathrm{s}_{q x_3}}{q} \\[3pt] 
            0 & 0 & 0 & 1 
            \end{pmatrix},
    \end{equation*}
    which corresponds to the well-known transformation matrix from the tip to the base of a CC segment~\citep{della2019control}.
    Since the body is assumed to be connected to a fixed joint, we have
    \begin{equation*}
        {}^{\calB_{i-1}}T_{\calJ_{i}} = \mat{\mathbb{I}}_{4}.
    \end{equation*}
\end{example*}
}

\review{In the rest of the section, we present the robot forward and differential kinematics. Although the derivation introduces little new information, it is included as a necessary starting point for formulating the dynamics.}
\review{We use the superscript ${}^{i}(\cdot)$ to indicate that a vector is expressed in the reference frame $\{S_{i}\}$. For ease of presentation, the superscript is omitted when the quantity is expressed in $\{S_{0}\}$.}
%
\subsection*{Forward kinematics}

%
Using~\eqref{eq:functional expression body kinematics},~\eqref{eq:body to joint transformation}--\eqref{eq:body to joint transformation:t} and~\eqref{eq:joint to body transformation}, we can compute the forward kinematics for all robot points. To this end, define the relative transformation from $\{ S_{i} \}$ to $\{ S_{i-1} \}$
\begin{equation}\label{eq:body to body transformation}
\begin{split}
    \Transm[i][i-1](\qv_{i}) &= \Transm[\calJ_i][\calB_{i-1}][\rb{\qv_{\calJ_i}}] \Transm[\calB_i][\calJ_i][\rb{\qv_{\calB_i}}]\\
    &=\begin{carray}{cc}
            \Rotm[i][i-1] & {}^{i-1}\tv_{i}\\
            \zerov_{1 \times 3} & 1
        \end{carray},
\end{split}
\end{equation}
and, by concatenation, that from $\{ S_{i} \}$ to $\{ S_{0} \}$
\begin{equation}\label{eq:body to base transformation}
\begin{split}
    \Transm[i](\qv_{1}, \dots, \qv_{i}) &= \Transm[1][0](\qv_1)\Transm[2][1](\qv_2) \dots \Transm[i][i-1](\qv_i)\\
    &= \begin{carray}{cc}
        \Rotm[i] & \tv_{i}\\
        \zerov_{1 \times 3} & 1
    \end{carray}.
\end{split}
\end{equation}
Denote, respectively, with ${}^{i}\pv_i$ and $^{i}\pv_{\comi} \in \R^{3}$ the position of a point of $\calB_i$ relative to $\calJ_{i}$ and of $\calB_{i}$ center of mass, both expressed in $\{S_{i}\}$. These can be computed from~\eqref{eq:functional expression body kinematics} and~\eqref{eq:body to joint transformation} as
\begin{equation*}
\begin{split}
    {}^{i}\pv_i(\xv_{i}, \qv_{\calB_{i}}) &=\Transm[\calB_i][\calJ_i][{}^{-1}] \begin{carray}{c}
        \fv_{\calB_{i}}\\1
    \end{carray},
\end{split}
\end{equation*}
and
\begin{equation*}
	^{i}\pv_{\comi}(\qv_{\calB_{i}}) = \frac{1}{m_{i}}\, \int_{V_{i}} {}^{i}\pv_{i}(\xv_{i}, \qv_{\calB_{i}}) \rho_{i}(\xv_{i}) \drm V,
\end{equation*}
where $m_{i} = \int_{V_{i}} \rho_{i}(\xv_{i})\drm V$ is the body mass and $\rho_{i}(\xv_i)$ its mass density. Since ${}^{i}\pv_{i}$ and ${}^{i}\pv_{\comi}$ are elements of $\R^{3}$, there always exists a unique vector ${}^{i}\rv_{i} \in \R^{3}$ such that
\begin{equation}
	\label{eq:i_p_i_com}
	{}^{i}\pv_{i} = {}^{i}\rv_{i} + {}^{i}\pv_{\comi},
\end{equation}
where $^{i}\rv_{i}$ satisfies the notable property
\begin{equation}\label{eq:r integral}
    \int_{V_{i}} {}^{i}\rv_{i} \rho_{i} \drm V = \int_{V_{i}} {}^{i}\pv_{i} \rho_{i} \drm V - {}^{i}\pv_{\comi} \int_{V_{i}} \rho_{i} \drm V = \zerov_{3}.
\end{equation}
%
\begin{remark}
    In the flexible link case, ${}^{i}\rv_{i}$ is typically approximated following a modal-Ritz approach, i.e.,
    \begin{equation}\label{eq:r_i flexible}
        {}^{i}\rv_{i}(\qv_{\calB_{i}}, \xv_{i}) = \Phim_{\rv_{i}}(\xv_{i})\qv_{\calB_{i}}. 
    \end{equation}
    In this this work, we never consider such hypothesis. In fact, the function $\fv_{\calB_{i}}$ defines a generic and possibly highly nonlinear functional dependence of ${}^{i}\rv_{i}$ on $\qv_{\calB_{i}}$ and $\xv_{i}$. 
\end{remark}

\review{
\begin{example*}[(Continuation)]
    Recalling that ${}^{\calJ_{i}}T_{\calB_{i}} = \mat{\mathbb{I}}_{4}$ the transformation from $\calB_{i}$ to $\calB_{i-1}$ is given by
    \begin{equation}\label{eq:PCC:body to body}
        {}^{i-1}T_{i} = {}^{\calB_{i-1}}T_{\calJ_{i}} {}^{\calJ_{i}}T_{\calB_{i}} = {}^{\calJ_{i}}T_{\calB_{i}}.
    \end{equation}
    Moreover, after some computations, the position vectors ${}^{i}\pv_{i}$, ${}^{i}\pv_{\comi}$ and ${}^{i}\rv_{i}$ in the body frame are
    \begin{equation*}
        {}^{i}\pv_{i} = 
        \left(\!\begin{array}{c} 
        \displaystyle \frac{\mathrm{c}_{q} - 1}{q} + x_{1} \mathrm{c}_{q} \\ 
        x_{2} \\ 
        \displaystyle -\frac{\mathrm{s}_{q} (q x_{1} + 1)}{q} 
        \end{array}\!\right),
    \quad
        {}^{i}\pv_{\comi} = 
        \begin{pmatrix} 
            \displaystyle -\frac{q - \mathrm{s}_{q}}{q^2} \\[3pt] 
            0 \\[2pt] 
            \displaystyle \frac{\mathrm{c}_{q} - 1}{q^2} 
        \end{pmatrix},
    \end{equation*}
    and
    \begin{equation*}
        {}^{i}\rv_{i} = 
        \left(\!\begin{array}{c} 
        \displaystyle \frac{q \mathrm{c}_{q} - \mathrm{s}_{q} + q^2 x_{1} \mathrm{c}_{q}}{q^2} \\ 
        x_{2} \\ 
        \displaystyle -\frac{\mathrm{c}_{q} - 1 + q \mathrm{s}_{q} (q x_{1} + 1)}{q^2}
        \end{array}\!\right).
    \end{equation*}
\end{example*}
}

We can now use~\eqref{eq:body to base transformation} and~\eqref{eq:i_p_i_com} to describe the forward kinematics, which \review{reads}
\begin{equation}\label{eq:point position}
	\pv_{i} = \pv_{\comi} + \Rotm[i] {}^{i}\rv_{i},
\end{equation}
with
\begin{equation}\label{eq:com position}
    \pv_{\comi} = \tv_{i} + \Rotm[i] {}^{i}\pv_{\comi}.
\end{equation}

%
\subsection*{First- and second-order forward differential kinematics}
\review{As we will see later in Sec.~\ref{section:kane equations}, the equations of motion depend on the first- and second-order time derivatives of $\pv_{i}$, which we briefly introduce below. To keep the presentation concise, we provide only a minimal overview, as the derivation of the differential kinematics follows standard arguments. For a detailed derivation, we refer the interested reader to Appendix~\ref{appendix:differential kinematics}.

From~\eqref{eq:point position} and~\eqref{eq:com position}, it follows that $\dpv_{i}$ depends on the time derivatives of $\tv_{i}, {}^{i}\pv_{\comi}, {}^{i}\rv_{i}$, namely
\begin{equation*}
    \vv_{i} = \dt{\tv_{i}}, \;\; {}^{i}\dpv_{\comi} = \dt{{}^{i}\pv_{\comi}}, \;\; {}^{i}\drv_{i} = \dt{{}^{i}\rv_{i}}, 
\end{equation*}
along with the angular velocity of $\{S_{i}\}$, denoted as $\omegav_{i}$ in the following. 

Similarly, the acceleration $\ddpv_{i}$ can be expressed using the terms
\begin{equation*}
    \av_{i} = \dt{\vv_{i}}, \;\; {}^{i}\ddpv_{\comi} = \dt{{}^{i}\dpv_{\comi}}, \;\; {}^{i}\ddrv_{i} = \dt{{}^{i}\drv_{i}}, \
\end{equation*}
and
\begin{equation*}
    \domegav_{i} = \dt{\omegav_{i}}. 
\end{equation*}
From the expression of $\dpv_{i}$ and $\ddpv_{i}$ (see Appendix~\ref{appendix:differential kinematics}), the following lemma holds.}
\begin{lemma}\label{lemma:velocities}
    Given $\qv, \dqv$ and $\ddqv$, the first- and second-order differential kinematics expressed in the body frame, i.e., ${}^{i}\vv_{i}, {}^{i}\omegav_{i}, {}^{i}\vv_{\comi}, {}^{i}\av_{i}, {}^{i}\domegav_{i}$ and $\av_{\comi}$, can be computed recursively forward in space from $\calB_{1}$ to $\calB_{N}$ by initializing ${}^{0}\vv_{0} = \zerov_{3}~[\si{\meter \per \second}], {}^{0}\omegav_{0} = \zerov_{3}~[\si{\radian \per \second}], {}^{0}\av_{0} = \zerov_{3}~[\si{\meter \per \square\second}]$ and ${}^{0}\domegav_{0}= \zerov_{3}~[\si{\radian \per \square\second}]$. In addition, the computational complexity for such evaluation is $O(N)$.
\end{lemma}
\begin{proof}
    The result follows immediately by observing the recursive structure of~\eqref{eq:v_i local}--\eqref{eq:v_comi local} and~\eqref{eq:^{i}av_i}--\eqref{eq:^{i}acom_i}.
\end{proof}
%
\review{
\begin{example*}[(Continuation)]
    Since the expression for the differential kinematics is quite complex even in this elementary example, we omit its presentation. However, the reader can verify that velocities and accelerations in the body frame can be readily obtained from~${}^{i-1}\Tm_{i}$, and the time derivatives of ${}^{i}\pv_{\comi}$ and ${}^{i}\rv_{i}$. 
\end{example*}
}

\section{Kane Equations}\label{section:kane equations}
Here, we briefly present the Kane equations for the class of robotic systems introduced in Sec.~\ref{sec:problem statement}. In a nutshell, the Kane method projects the weak form of the dynamics in the configuration space by exploiting the system holonomic nature, \review{specifically the fact that all constraints depend only on time and configuration variables, but never on their time derivatives}. While the content of this section is not new per se, we believe it can be of interest to the soft robotics community, which is mostly familiar with the EL approach. We also demonstrate that the Kane and EL approach are indeed equivalent. However, as we shall see later in Sec.~\ref{section:recursive formulation}, the Kane equations can be manipulated to obtain a recursive form of the dynamics that can be directly used to solve Problems~\ref{problem:IDP} and~\ref{problem:AIDP}.
According to the weak form of the EoM for a continuum~\citep{lacarbonara2013nonlinear}, for every body of the robot, one has
\begin{equation}\label{eq:integral form d'alembert}
    \int_{V_{i}} \delta \pv_{i}^{T} \rb{ \fv_{\mathrm{int}_{i}} + \fv_{\mathrm{ext}_{i}} - \dt{\rb{\dpv_{i} \rho_{i}}} } \drm V = 0.
\end{equation}
Here, $\fv_{\mathrm{int}_{i}}$ is the vector modeling internal forces per unit volume, such as the mechanical stress and actuation forces (when the system is internally actuated). 
The vector $\fv_{\mathrm{ext}_{i}}$ represents the resultant external force per unit volume, including, for example, gravity. Furthermore, $\dt{\rb{\dpv_{i} \rho_{i}}}$ is the time derivative of the linear momentum per unit volume. 

According to the Kane method, in the computations, it suffices to consider among $\fv_{\mathrm{int}_{i}}$ and $\fv_{\mathrm{ext}_{i}}$ the only forces that perform work on the body in the course of a virtual displacement, i.e., the forces for which $\int_{V_{i}}\delta \pv_{i}^{T} \fv_{\mathrm{int}_{i}} \drm V \neq 0$ and $\int_{V_{i}}\delta \pv_{i}^{T} \fv_{\mathrm{ext}_{i}} \drm V \neq 0$. 
Thanks to the holonomic nature of the system, the reaction forces at the contact area between two bodies are non-working~\citep{whittaker1964treatise}.
As a result, they have no effect in the above balance equation. In other words, the internal reaction forces between adjacent bodies, either rigidly connected or in relative motion due to a joint, can be completely neglected in~\eqref{eq:integral form d'alembert}. \review{While this equation already describes the dynamics of the body, it cannot be used in this form to compute the dynamics because it contains the virtual displacement $\delta \pv_{i}$.}

Recalling that $\pv_i$ depends on $\qv$, it follows that $\delta \pv_i = \rb{\parv{\pv_{i}}{\qv}}^{T} \delta \qv$. Substituting this into the above equation and rearranging the terms yields
\begin{equation*}
     \int_{V_{i}} \delta \qv^{T}\parv{\pv_{i}}{\qv} \rb{\drm\fv_{i} - \dt{\rb{\dpv_{i} \rho_{i}}} \drm V } = 0,
\end{equation*}
where we have defined for compactness the net force $\drm\fv_{i} = \rb{\fv_{\mathrm{int}_{i}} + \fv_{\mathrm{ext}_{i}}}\drm V$. 
Since the system is holonomic, each generalized coordinate can experience a virtual displacement independently of the others, which implies 
\begin{equation*}
    \int_{V_{i}} \parv{\pv_{i}}{\qv} \rb{\drm\fv_{i} - \dt{\rb{\dpv_{i} \rho_{i}}}\drm V } = \zerov_{n}.
\end{equation*}
Summing the contributions for all body gives the reduced-order dynamics
\begin{equation*}
    \begin{split}    
    &\sum_{j = 1}^{N} \int_{V_{j}} \parv{\pv_{j}}{\qv} \rb{\drm\fv_{j} - \dt{\rb{\dpv_{j} \rho_{j}}}\drm V } = \zerov_{n},
    \end{split}
\end{equation*}
and the Kane equations for the system
\begin{gather}
	\label{eq:Q_k_deformable}
	\Qm = \sum_{j = 1}^{N} \int_{V_{j}} \parv{\dpv_{j}}{\dqv} \drm\fv_{j},\\
	\label{eq:Q^*_k_deformable}
	\Qm^{*} = -\sum_{j = 1}^{N} \int_{V_{j}} \parv{\dpv_{j}}{\dqv} \dt{\rb{\dpv_{j} \rho_{j}}}\drm V,\\
	\label{eq:kane_deformable_system}
	\Qm + \Qm^{*} = \zerov_{n},
\end{gather}
where the identity $\parv{\pv_{j}}{\qv} = \parv{\dpv_{j}}{\dqv}$ has been used.
The terms $\Qm$ and $\Qm^{*}$ are called, respectively, the generalized active and inertia force \review{because}
\begin{equation*}
    \Qm = -\gv(\qv) - \sv(\qv, \dqv) + \nuv(\qv, \uv),
\end{equation*}
and
\begin{equation*}
    \Qm^{*} = -\Mm(\qv)\ddqv - \cv(\qv, \dqv).
\end{equation*}
\review{In a nutshell, $\Qm$ models all external forces, while $\Qm^{*}$ the inertia of the system.}
\review{In this paper,} we restrict the analysis to systems for which the variation of the mass density is negligible, i.e., $\Dot{\rho}_{i} = 0$, $i \in \{ 1, \dots, N \}$. 
This way, we can simplify the second term in the integrand of~\eqref{eq:Q^*_k_deformable} to 
\begin{equation}\label{eq:time derivative linear momentum}
    \dt{\rb{\dpv_{i} \rho_{i}}} \drm V = \ddpv_i \rho_i \drm V = \ddpv_i \drm m_{i},
\end{equation}
where we defined the infinitesimal mass $\drm m_{i} = \rho_{i} \drm V$. 
From~\eqref{eq:r integral} and the Reynolds transport theorem~\review{\footnote{\color{red} (\cite{marsden2016manifolds})~Let $\fv(\xv, t) : U \subseteq \R^{n} \times \R \rightarrow V \subseteq \R^{3}$ be a $C^{1}$ mapping with $U$ constant in time. Then
    \begin{equation*}
        \frac{\drm }{\drm t} \int_{U}\fv(\xv, t) \drm u = \int_{U} \frac{\drm }{\drm t} \fv(\xv, t) \drm u.
    \end{equation*}}}, this also implies
\begin{equation}\label{eq:r derivatives integral:1}
\int_{V_i} {}^{i}\drv_i \rho_i \drm V = \dt{}\rb{\int_{V_i} {}^{i}\rv_i \rho_i \drm V} = \zerov_{3},
\end{equation}
and
\begin{equation}\label{eq:r derivatives integral:2}
\int_{V_i} {}^{i}\ddrv_i \rho_i \drm V = \dt{}\rb{\int_{V_i} {}^{i}\drv_i \rho_i \drm V} = \zerov_{3}.
\end{equation}
%
%
\begin{proposition}\label{proposition:equivalence}
Equations~\eqref{eq:Q_k_deformable}--\eqref{eq:time derivative linear momentum} are equivalent to~\eqref{eq:finite dimensional equations}, i.e., 
\begin{equation}\label{eq:equivalance Q* EL-Kane}
    -\Qm^{*}(\qv, \dqv, \ddqv) = \Mm(\qv)\ddqv + \cv(\qv, \dqv) = \Qm(\qv, \dqv, \uv).
\end{equation}
\end{proposition}
\begin{proof}
    See Appendix~\ref{appendix:EL equivalence}.
\end{proof}
%
%
For the sake of the following derivations, it proves advantageous to expand $\Qm$ and $\Qm^{*}$ into the rows associated to the single bodies, namely to consider the EoM~\eqref{eq:Q_k_deformable}--\eqref{eq:kane_deformable_system} in the equivalent form
\begin{gather}
	\label{eq:Q_j,l_deformable}
	\Qm_{i} = \sum_{j = 1}^{N} \int_{V_{j}} \parv{\dpv_{j}}{\dqv_{i}} \drm\fv_{j},\\
	\label{eq:Q^*_j,l_deformable}
	\Qm^{*}_{i} = -\sum_{j = 1}^{N} \int_{V_{j}}\parv{\dpv_{j}}{\dqv_{i}} \ddpv_{j}\drm m_{j},\\
    \label{eq:Kane}
	\Qm_{i} + \Qm^{*}_{i} = \zerov_{n_{i}},
\end{gather}
with $i \in \{1, \dots, N\}$.
\section{Recursive Formulation of the Equations of Motion}\label{section:recursive formulation}
This section presents the main result of the paper, i.e., a recursive formulation of~\eqref{eq:Q_j,l_deformable}--\eqref{eq:Kane}, which immediately yields a  procedure for the solution of the ID problem. We provide a pseudo-code for implementing the algorithm, encompassing all necessary terms. We also show how this procedure can be modified to evaluate the mass matrix.

To this end, rewrite the left-hand side of~\eqref{eq:Kane} as
\begin{equation*}
    \Qm_{i} + \Qm^{*}_{i} = \sum_{j = 1}^{N} \int_{V_{j}} \parv{\dpv_{j}}{\dqv_{i}} \rb{  \drm\fv_{j} - \ddpv_{j}\drm m_j}.
\end{equation*}
For $j < i$, one has $\parv{\dpv_j}{\dqv_{i}} = \zerov_{n_{i} \times 3}$ because $\pv_j$ depends only on $\qv_{1}, \dots, \qv_{j-1}$ and $\qv_{j}$. Thus, the lower bound ($j=1$) of the above summation can be replaced with the index associated to $\calB_{i}$ ($j = i$), leading to a more concise form
\begin{equation*}
    \Qm_{i} + \Qm^{*}_{i} = \sum_{j = i}^{N} \int_{V_{j}} \parv{\dpv_{j}}{\dqv_{i}} \rb{ \drm\fv_{j} - \ddpv_{j}\drm m_j}.
\end{equation*}
By leveraging~\eqref{eq:v_p_i}, the previous equation can be rewritten as
\begin{equation*}
    \begin{split}
        \Qm_{i} + \Qm^{*}_{i} &= \sum_{j = i}^{N} \parv{\vv_{\com_{j}}}{\dqv_{i}} \int_{V_j} \drm \fv_j - \ddpv_j \drm m_j\\
        &\quad+ \parv{\omegav_{j}}{\dqv_{i}} \int_{V_j} \rb{ \Rotm[j] {}^{j}\rv_j } \times \rb{\drm \fv_j - \ddpv_j \drm m_j }\\
        &\quad+\int_{V_j}\parv{{}^{j}\drv_j}{\dqv_{i}}\Rotm[j]^{T} \rb{\drm \fv_j - \ddpv_j \drm m_j }.
    \end{split} 
\end{equation*}
Given that ${}^{i}\rv_i$ depends only on the configuration variables of the corresponding body $\calB_i$, it holds $\parv{{}^{j}\drv_j}{\dqv_{i}} = \zerov_{n_{i} \times 3}; \,\, j \neq i$, which simplifies the above expression to
\begin{equation}\label{eq:Kane with momenta}
    \begin{split}
        \Qm_{i} + \Qm^{*}_{i} &= \int_{V_i}\parv{{}^{i}\drv_i}{\dqv_{i}}\Rotm[i]^{T} \rb{\drm \fv_i - \ddpv_i \drm m_i }\\
        &\quad+ \sum_{j = i}^{N} \parv{\vv_{\com_{j}}}{\dqv_{i}} \int_{V_j} \drm \fv_j - \ddpv_j \drm m_j\\
        &\quad+ \parv{\omegav_{j}}{\dqv_{i}} \int_{V_j} \drm \tauv_j -\rb{ \Rotm[j] {}^{j}\rv_j } \times \ddpv_j \drm m_j,
    \end{split} 
\end{equation}
being $\drm \tauv_j = \rb{\Rotm[j]{}^{j}\rv_j} \times \drm \fv_j$. 
\review{At this stage, it is useful to summarize what we have achieved so far. In essence, we have simplified the Kane equations in two key ways. First, we have demonstrated that the reduced-order dynamics of the $i$-th body does not directly depend on its predecessors, as indicated by the summation index $j$ starting from $i$. Second, we have shown that the dynamics relies on three distinct types of terms. The first term captures effects local to the body. The other two account for the balance of forces and momenta with respect to the center of mass, and propagate also to the previous bodies through the summation. We now proceed with a further manipulation of the equations that will allow to obtain a recursive form.}

By exploiting the invariance of the scalar product under rotations, it is possible to express~\eqref{eq:Kane with momenta} in the body frame $\{ S_{i} \}$ as
\begin{equation}\label{eq:kane body frame}
    \begin{split}
        \Qm_{i} + \Qm^{*}_{i} &= \int_{V_i}\parv{{}^{i}\drv_i}{\dqv_{i}} \rb{ \drm {}^{i}\fv_i - {}^{i}\ddpv_i \drm m_i }\\
        &\quad+ \sum_{j = i}^{N} \parv{{}^{j}\vv_{\com_{j}}}{\dqv_{i}} \int_{V_j}\drm {}^{j}\fv_j - {}^{j}\ddpv_j \drm m_j\\
        &\quad+ \parv{{}^{j}\omegav_{j}}{\dqv_{i}} \int_{V_j} \drm {}^{j}\tauv_j -{}^{j}\rv_j \times {}^{j}\ddpv_j \drm m_j,
    \end{split}
\end{equation}
where $\drm {}^{i}\fv_i = \Rotm[i]^{T}\drm \fv_i$ and $\drm {}^{i}\tauv_i = \Rotm[i]^{T}\drm \tauv_i$ denote the force and torque in $\{ S_{i} \}$, respectively. 
For the sake of readability, we introduce the following definitions
\begin{equation}\label{eq:calFv_i 1}
    {}^{i}\calFv_{i} = \int_{V_i}\drm {}^{i}\fv_i, \,\, {}^{i}\calFv_{i}^{*} = -\int_{V_i}{}^{i}\ddpv_i \drm m_i, \,\, {}^{i}\calTv_{i} = \int_{V_i} \drm {}^{i}\tauv_i,
\end{equation}
and
\begin{equation}\label{eq:calTv_i 1:2}
    {}^{i}\calTv_{i}^{*} = -\int_{V_i}{}^{i}\rv_i \times {}^{i}\ddpv_i \drm m_i.
\end{equation}
Henceforth, the terms active force and torque of $\calB_i$ are used to denote ${}^{i}\calFv_{i}$ and ${}^{i}\calTv_{i}$, respectively. Similarly, ${}^{i}\calFv_{i}^{*}$ and ${}^{i}\calTv_{i}^{*}$ are its inertial force and torque. \review{Examining the previous equations, the two active terms can be interpreted as the net external force and torque averaged over the entire body. Analogously, the inertial terms represent the time derivatives of the linear and angular momentum.}

Note also that the dimension of all the above vectors is three, independently of the number of generalized coordinates.

Substituting~\eqref{eq:calFv_i 1} into~\eqref{eq:kane body frame} leads to the compact form
\begin{equation}\label{eq:Kane compact}
    \begin{split}
        \Qm_{i} + \Qm^{*}_{i} &= \int_{V_i}\parv{{}^{i}\drv_i}{\dqv_{i}} \rb{ \drm {}^{i}\fv_i - {}^{i}\ddpv_i \drm m_i }\\
        &\quad+\sum_{j = i}^{N} \parv{{}^{j}\vv_{\com_{j}}}{\dqv_{i}} \rb{{}^{j}\calFv_{j}
        + {}^{j}\calFv_{j}^{*}}\\
        &\quad\quad+ \parv{{}^{j}\omegav_{j}}{\dqv_{i}} \rb{ {}^{j}\calTv_{j} + {}^{j}\calTv_{j}^{*} }.
    \end{split}
\end{equation}
The previous expression can be further simplified recalling~\eqref{eq:v_comi local}, which implies that, for $j = i$,
\begin{equation*}
\begin{split}
    \parv{{}^{j}\vv_{\com_{j}}}{\dqv_{i}} &= \parv{{}^{j}\vv_{j}}{\dqv_{i}} + \parv{{}^{j}\omegav_{j}}{\dqv_{i}}  {}^{j}\Skew{\pv}_{\com_{j}} \\&\quad+ \parv{{}^{j}\dpv_{\com_{j}}}{\dqv_{i}},
\end{split}
\end{equation*}
and, for $j > i$,
\begin{equation*}
    \begin{split}
        \parv{{}^{j}\vv_{\com_{j}}}{\dqv_{i}} &= \parv{{}^{j}\vv_{j}}{\dqv_{i}} + \parv{{}^{j}\omegav_{j}}{\dqv_{i}}  {}^{j}\Skew{\pv}_{\com_{j}}.
    \end{split}
\end{equation*}
Indeed, the position of the center of mass in the body frame, and consequently also its time derivative, depends only on the configuration variables of the body.
The substitution of the above identities into~\eqref{eq:Kane compact} gives
\begin{equation}\label{eq:Kane final}
     \Qm_{i} + \Qm^{*}_{i} = {}^{i}\piv_{{i}} + {}^{i}\piv_{i}^{*} + \calMv_{i},
\end{equation}
with 
\begin{align}
    \label{eq:pi expression final}
    \begin{split}    
    {}^{i}\piv_{i} &= \int_{V_{i}} \parv{\rb{{}^{i}\drv_{i} + {}^{i}\dpv_{\com_{i}}}}{\dqv_{i}} \drm {}^{i}\fv_i\\
    &= \int_{V_{i}} \parv{{}^{i}\dpv_{i}}{\dqv_{i}} \drm {}^{i}\fv_i,
    \end{split}\\
    \label{eq:pi* expression final}
    \begin{split}    
    {}^{i}\piv_{i}^{*} &= -\int_{V_{i}} \parv{\rb{{}^{i}\drv_{i} + {}^{i}\dpv_{\com_{i}}}}{\dqv_{i}} {}^{i}\ddpv_i \drm m_i\\
    &= -\int_{V_{i}} \parv{{}^{i}\dpv_{i}}{\dqv_{i}} {}^{i}\ddpv_i \drm m_i,
    \end{split}
\end{align}
and
\begin{equation}\label{eq: M_{jl} definition}
\begin{split}
    \calMv_{i} &= \sum_{j = i}^{N}\parv{{}^{j}\vv_{j}}{\dqv_{i}} \rb{{}^{j}\calFv_{j} + {}^{j}\calFv_{j}^{*} } + \parv{{}^{j}\omegav_{j}}{\dqv_{i}}\\
    &\quad \rb{ {}^{j}\calTv_{j} +{}^{j}\calTv_{j}^{*} + {}^{j}\pv_{\com_{j}}\times \rb{{}^{i}\calFv_{j} + {}^{j}\calFv_{j}^{*}}}.
\end{split}
\end{equation}
It is easy to observe that ${}^{i}\piv_{i}$ and ${}^{i}\piv_{i}^{*}$
contain terms that depend only on $\calB_{i}$. As such, these terms can be computed without any recursion. In contrast, $\calMv_{i}$ involves the forces and torques of $\calB_{i}$ as well as those of all subsequent bodies in the chain. \review{We thus need a way to compute recursively the latter.} The following theorem formalizes the main contribution of this work, namely that $\calMv_{i}$ can indeed be expressed in a recursive form. 
\begin{theorem}\label{theorem:recurisive M}
    Given $\qv, \dqv$ and $\ddqv$, $\calMv_{i}; i \in \{N, \dots, 1\},$ admits the following backward recursive expression
    \begin{equation}\label{eq:M_{jl} recursive expression}
    \begin{split}
        \calMv_{i} &= \parv{{}^{i}\vv_{i}}{\dqv_{i}} \rb{{}^{i}\Fv_{i} + {}^{i}\Fv_{i}^{*}}\\
        &\quad+ \parv{{}^{i}\omegav_{i}}{\dqv_{i}} \rb{{}^{i}\Tv_{i} + {}^{i}\Tv_{i}^{*}},
    \end{split}
\end{equation}
where
\begin{equation}\label{eq:M_{jl} recursive expression update}
\begin{split}
    {}^{i}\Fv_{i} + {}^{i}\Fv_{i}^{*} &= {}^{i}\calFv_{i} + {}^{i}\calFv_{i}^{*}\\
    &\quad+ \Rotm[i+1][i] \rb{{}^{i+1}\Fv_{i+1} + {}^{i+1}\Fv_{i+1}^{*}},\\
    {}^{i}\Tv_{i} + {}^{i}\Tv_{i}^{*} &= {}^{i}\calTv_{i} + {}^{i}\calTv_{i}^{*} + {}^{i}\pv_{\com_{i}}\times \rb{{}^{i}\calFv_{i} + {}^{i}\calFv_{i}^{*} }\\
    &\quad+ \Rotm[i+1][i] \rb{{}^{i+1}\Tv_{i+1} + {}^{i+1}\Tv_{i+1}^{*}} \\
    &\quad+ {}^{i}\tv_{i+1} \times \Rotm[i+1][i] \rb{{}^{i+1}\Fv_{i+1} + {}^{i+1}\Fv_{i+1}^{*}},
\end{split}
\end{equation}
with ${}^{{N+1}}\Fv_{N+1} = {}^{{N+1}}\Fv_{N+1}^{*} = \zerov_{3}~[\si{\newton}]$ and $ {}^{N+1}\Tv_{N+1} = {}^{N+1}\Tv_{N+1}^{*} = \zerov_{3}~[\si{\newton\meter}]$. 
\end{theorem}
\begin{proof}
    See Appendix~\ref{appendix:recurisve relations details}. 
\end{proof}
In~\eqref{eq:M_{jl} recursive expression update}, the terms ${}^{i}\Fv_{i}$ and ${}^{i}\Tv_{i}$ account for the total effects of the linear and angular active forces on $\calB_{i}$ due to $\calB_{i}$ itself and all its successor bodies. Similarly, ${}^{i}\Fv_{i}^{*}$ and ${}^{i}\Tv_{i}^{*}$ are the corresponding inertial forces. It is worth remarking that the active and inertial terms can be computed alone by setting the dual terms to zero. 
\review{For example, we can obtain ${}^{i}\Fv_{i}^{*}$ setting ${}^{i}\calFv_{i} = {}^{i+1}\Fm_{i+1} = \zerov_{3}$ in~\eqref{eq:M_{jl} recursive expression update}.}
Additionally, note that $\parv{{}^{i}\vv_{i}}{\dqv_{i}}$ and $\parv{{}^{i}\omegav_{i}}{\dqv_{i}}$ project the system dynamics in the direction of $\qv_{i}$. These projectors can be obtained seamlessly from the kinematic model because they depend only on quantities of the body.
Indeed, ~\eqref{eq:v_i local} and~\eqref{eq:omega_i local} imply
\begin{equation}\label{eq:partial velocities:1}
    \begin{split}
        \parv{{}^{i}\vv_{i}}{\dqv_{i}} &= \parv{{}^{i-1}\vv_{i-1, i}}{\dqv_{i}}\Rotm[i][i-1](\qv_{i}),
    \end{split}
\end{equation}
and
\begin{equation}\label{eq:partial velocities:2}
        \begin{split}
        \parv{{}^{i}\omegav_{i}}{\dqv_{i}} &= \parv{{}^{i-1}\omegav_{i-1, i}}{\dqv_{i}}\Rotm[i][i-1](\qv_{i}).
    \end{split}
\end{equation}

\review{
\begin{example*}[(Continuation)]
    Using the kinematics derived in the previous section, the projectors for a CC body are given by
    \begin{equation*}
        \parv{{}^{i}\vv_{i}}{\dqv_{i}} = \left(\!\begin{array}{c} 
            \displaystyle \frac{\mathrm{c}_q - 1}{q^2} \\ 
            0 \\ 
            \displaystyle \frac{q - \mathrm{s}_q}{q^2} 
            \end{array}\!\right),
    \end{equation*}
    and
    \begin{equation*}
        \parv{{}^{i}\omegav_{i}}{\dqv_{i}} = \left(\!\begin{array}{c} 0\\ -1\\ 0 \end{array}\!\right).
    \end{equation*}
\end{example*}
}

%
\begin{remark}
    Equations~\eqref{eq:Kane final}--\eqref{eq:partial velocities:2} represent a recursive form of the dynamics, parameterized by ${}^{i}\piv_{i}, {}^{i}\piv_{i}^{*}, {}^{i}\Fv_{i},{}^{i}\Fv_{i}^{*},{}^{i}\Tv_{i}$ and ${}^{i}\Tv_{i}^{*}$. These equations hold for any soft robotic system satisfying Assumption~\ref{assumption:finite dimensional kinematics}. 
\end{remark}
Observe also that $\calMv_{i}$ satisfies an important property that is useful for solving the ID problem. 
\begin{property}\label{property:M}
    The operator $\calMv_{i}$ is linear in all its arguments.  
\end{property}
\begin{proof}
    The property follows from~\eqref{eq: M_{jl} definition} (or, equivalently, \eqref{eq:M_{jl} recursive expression update}) since the matrix and cross products are linear operators.
\end{proof}
The above result is quite useful for our derivation since it allows isolating the contributions of the generalized active forces. In turn, this implies that $\Qm_{i}$ and $\Qm_{i}^{*}$ can be computed by nullifying the inertial and active terms, respectively, i.e.,
\begin{equation}\label{eq:Qm_{j} M expression}
    \Qm_{i} = {}^{i}\piv_{i} + \calMv_{i}( {}^{i}\Fv_{i},\zerov_{3}, {}^{i}\Tv_{i}, \zerov_{3}) = -\Qm_{i}^{*},
\end{equation}
and
\begin{equation}\label{eq:Q_{j}^{*} M expression}
    \Qm_{i}^{*} = {}^{i}\piv_{i}^{*} + \calMv_{i}(\zerov_{3}, {}^{i}\Fv_{i}^{*}, \zerov_{3}, {}^{i}\Tv_{i}^{*}) = -\Qm_{i},
\end{equation}
where the last identity in~\eqref{eq:Qm_{j} M expression} and~\eqref{eq:Q_{j}^{*} M expression} arises from $\Qm_{i} + \Qm_{i}^{*} = \zerov_{n_{i}}$. 
\review{Using the last equation and the recursive form of $\calMv_{i}$ in~\eqref{eq:M_{jl} recursive expression}, the inertial inverse dynamics can be computed as
\begin{equation}\label{eq:Qm_{j} M expression star}
    \Qm_{i} =-{}^{i}\piv_{i}^{*} -\parv{{}^{i}\vv_{i}}{\dqv_{i}} {}^{i}\Fv_{i}^{*} - \parv{{}^{i}\omegav_{i}}{\dqv_{i}} {}^{i}\Tv_{i}^{*}.
\end{equation}
This expression represents the core of our ID algorithms.
}
\subsection{Evaluation of the inertial force and torque}
%
In this section, we address the evaluation of ${}^{i}\calFv_{i}^{*}$, ${}^{i}\calTv_{i}^{*}$ and ${}^{i}\piv_{i}^{*}$. Indeed, if these terms are known,~\eqref{eq:Qm_{j} M expression star} and Theorem~\ref{theorem:recurisive M} provide a way to recursively calculate $\Qm_{i}$, thereby solving Problem~\ref{problem:IDP}. In other words, thanks to Theorem~\ref{theorem:recurisive M}, we can immediately solve the IID problem once the expression of the body inertial terms is available. 

Using~\eqref{eq:point acceleration}, ${}^{i}\calFv_{i}^{*}$ takes the form
\begin{equation}\label{eq:inertia force}
	\begin{split}
		{}^{i}\calFv_{i}^{*} = - {}^{i}\av_{\comi}m_{i},
	\end{split}
\end{equation}
where~\eqref{eq:r integral} and~\eqref{eq:r derivatives integral:1}--~\eqref{eq:r derivatives integral:2} have been employed. 

A similar computation for ${}^{i}\calTv_{i}^{*}$ leads to
\begin{equation}\label{eq:integral r times ddotp}
	\begin{split}
		{}^{i}\calTv_{i}^{*} = -\int_{V_{i}} \dt{} \left( {}^{i}\rv_{i} \times \Rotm[i]^{T} \dt{} \rb{\Rotm[i] {}^{i}\rv_{i}} \right) \drm m_{i}.
	\end{split}
\end{equation}
Note that the integrand of the right-hand side represents the time derivative of the angular momentum about the center of mass.
Now, using the identity 
\begin{equation*}
\dt{} \rb{ \Rotm[i]{}^{i}\rv_i} = \omegav_{i} \times \Rotm[i]{}^{i}\rv_{i} + \Rotm[i] {}^{i}\drv_{i},    
\end{equation*}
after some computations,~\eqref{eq:integral r times ddotp} becomes
\begin{equation}\label{eq:int r x ddpi}
	\begin{split}
		{}^{i}\calTv_{i}^{*} 
		&= -{}^{i}\IIm_{i} {}^{i}\domegav_{i} - {}^{i}\omegav_{i} \times {}^{i}\IIm_{i}{}^{i}\omegav_{i} - {}^{i}\dot{\IIm}_{i} \omegav_{i}- {}^{i}\omegav_{i}\\
		&\quad \times \int_{V_{i}} {}^{i}\rv_{i} \times {}^{i}\drv_{i} \drm m_{i} - \int_{V_{i}} {}^{i}\rv_{i} \times {}^{i}\ddrv_{i}\drm m_{i},
	\end{split}
\end{equation}
where we defined the body inertia
\begin{equation*}
    ^{i}\IIm_{i} = \int_{V_{i}} {}^{i}\Skew{\rv}_{i}^{T} {}^{i}\Skew{\rv}_{i}\drm m_{i},
\end{equation*}
and its time derivative
\begin{equation*}    
    {}^{i}\dot{\IIm}_{i} = \dt{{}^{i}\IIm_{i}} = \int_{V_{i}} {}^{i}\Skew{\drv}_{i}^{T}{}^{i}\Skew{\rv}_{i} + {}^{i}\Skew{\rv}_{i}^{T}{}^{i}\Skew{\drv}_{i} \drm m_{i}.
\end{equation*}
Indeed, recall that, when the body is deformable, ${}^{i}\rv_{i}$ is a function of time through $\qv_{i}(t)$. Note also that the first two terms in the right-hand side of~\eqref{eq:int r x ddpi} model the rigid motion, while the remaining arise because of deformability, as expected~\citep{stramigioli2024principal}. 

Now consider ${}^{i}\piv_{i}^{*}$. By substituting~\eqref{eq:ddpv_i local} into~\eqref{eq:pi* expression final} and performing some computations, one obtains
\begin{equation}\label{eq:int dri ddpi}
	\begin{split}
		{}^{i}\piv_{i}^{*} &= -\parv{}{\dqv_{i}} \left( \int_{V_{i}} {}^{i}\rv_{i} \times {}^{i}\drv_{i} \drm m_{i} \right) {}^{i}\domegav_{i}\\ 
        &\quad+ \frac{1}{2} \rb{\vect{\mathbb{I}}_{n_{i}} \otimes \rb{{}^{i}\omegav_{i}^{T} \rb{{}^{i}\omegav_{i}^{T} \otimes \vect{\mathbb{I}}_{3}}}} \\
        &\quad\quad\quad\mathrm{vec}\rb{\rb{ \parv{\mathrm{vec}({}^{i}\dot{\IIm}_{i})}{\qv_{i}}}^{T}}
        \\
	&\quad - 2 \left( \int_{V_{i}} {}^{i} \drv_{i} \times \parv{{}^{i}\drv_{i}}{\dqv_{i}} \drm m_{i} \right) {}^{i}\omegav_{i}\\
		&\quad - \int_{V_{i}} \parv{{}^{i}\drv_{i}}{\dqv_{i}} {}^{i}\ddrv_{i} \drm m_{i} + \parv{{}^{i}\dpv_{\com_{i}}}{\dqv_{i}} {}^{i}\calFv^{*}_{i},
	\end{split}
\end{equation}
which can be computed using the differential kinematics and $\calFv_{i}^{*}$. 

\begin{remark}
    All the inertial terms are functions of $\qv_{i}, \dqv_{i}$ and $\ddqv_{i}$ and their functional expression can be computed offline once the kinematic model of the body and its mass density are known. 
    When a closed-form expression for the above integrals is not available, numerical integration techniques, such as the Gaussian quadrature rule, must be used. Since the numerical approximation of integrals is a distinct research area and not the focus of this work, we refer the reader to~\cite{davis2007methods} and the references therein for an introduction to the topic. For the numerical results presented in the following, the integrals have been implemented through Gaussian quadrature rules with Legendre polynomials bases. 
\end{remark}
Combining all the above results, $\Qm$ can be computed -- and so Problem~\ref{problem:IDP} solved -- recursively using Algorithm~\ref{alg:Kane inverse recursive general}. Analogous to the rigid body case~\citep{featherstone2014rigid}, the procedure involves a forward and a backward step. In the forward step, $\qv$ and its time derivatives are used to compute velocities and accelerations, enabling the evaluation of ${}^{i}\calFv_{i}^{*}$, ${}^{i}\calTv_{i}^{*}$ and ${}^{i}\piv_{i}^{*}$. \review{At the same time, also the configuration space projectors $\parv{{}^{i}\vv_{i}}{\dqv_{i}}$ and $\parv{{}^{i}\omegav_{i}}{\dqv_{i}}$ are computed.} In the backward step, the generalized inertial force is computed by projecting the inertial terms in the configuration space. This is achieved using~\eqref{eq:Qm_{j} M expression star} and the recursive expression for ${}^{i}\Fv_{i}^{*}$ and ${}^{i}\Tv_{i}^{*}$ given by~\eqref{eq:M_{jl} recursive expression update}. The computational complexity grows linearly with the number of bodies, as formalized below. 
\begin{corollary}\label{corollary:computational cost M}
    The computational complexity of Algorithm~\ref{alg:Kane inverse recursive general} is $O(N)$. 
\end{corollary}
\begin{proof}
    From Lemma~\ref{lemma:velocities}, ~\eqref{eq:inertia force} and~\eqref{eq:int r x ddpi}--\eqref{eq:int dri ddpi}, the cost of the forward step in Algorithm~\ref{alg:Kane inverse recursive general} is $O(N)$. Similarly, given the recursive form of $\calMv_{i}$ in Theorem~\ref{theorem:recurisive M} the computational complexity of the backward step is $O(N)$. Recall indeed that $\calMv_{i}$ can be evaluated backward through $\Fv_{i}^{*}$ and $\Tv_{i}^{*}$, which are in turn computed from the differential kinematics, $\calFv_{i}^{*}$ and $\calTv_{i}^{*}$.
\end{proof}
\review{To better illustrate Algorithm~\ref{alg:Kane inverse recursive general} and its computational advantages over the EL approach, we present flow diagrams for both methods in Appendix~\ref{appendix:EL comparison flow}.}
\review{
\begin{example*}[(Continuation)]
    According to~\eqref{eq:inertia force} and~\eqref{eq:int r x ddpi}–\eqref{eq:int dri ddpi}, the inertial terms required by Algorithm~\ref{alg:Kane inverse recursive general} are derived from the kinematics and the body mass density. Assuming the latter has a unitary value, and after some simplifications, the inertial force and torque can be expressed as
    \begin{equation*}\small
        {}^{i}\calFv_{i}^{*} = 
        \begin{pmatrix} 
            \displaystyle \frac{\pi \left( \ddot{q} q^2 (1 - \mathrm{c}_{q}) + 2 q \mathrm{s}_{q} (\ddot{q} - {\dot{q}}^2) + 6 {\dot{q}}^2 \mathrm{s}_{q} - 4 {\dot{q}}^2 q \right)}{q^4} \\[3pt] 
            0 \\[3pt] 
            \displaystyle \frac{\pi \left( {\dot{q}}^2 (q^2 + 6 \mathrm{c}_{q} - 6) + q \mathrm{s}_{q} (2 {\dot{q}}^2 - \ddot{q} q) - 2 \ddot{q} q \mathrm{c}_{q} \right)}{q^4} 
        \end{pmatrix},
    \end{equation*}
    \begin{equation*}\small
        {}^{i}\calTv_{i}^{*} = 
        \left(\!\!\!\! \begin{array}{c} 
            0 \\ 
            \displaystyle 
            \frac{\pi}{2 q^5} 
            \begin{aligned}[t]
            & \ddot{q} q \left(6 - q^2 - 6 \mathrm{c}_{q} - 2 q \mathrm{s}_{q} \right)  \\
            & + {\dot{q}}^2 \left(-16 + 4 q^2 + 16 \mathrm{c}_{q} + 4 q \mathrm{s}_{q} \right)  
            \end{aligned} \\ 
            0 
        \end{array}\!\!\!\!\right).
    \end{equation*}
    Proceeding similarly for the local inertial forces yields
    \begin{equation*}\small
        \begin{split}
            {}^{i}\piv_{i}^{*} &= \frac{\pi}{6 q^6} \big( 18 \ddot{q} q^2 - \ddot{q} q^4 + 60 {\dot{q}}^2 \mathrm{s}_{q} - 24 \ddot{q} q \mathrm{s}_{q} \\
            &\quad\;\; + 6 \ddot{q} q^2 \mathrm{c}_{q} - 12 {\dot{q}}^2 q \mathrm{c}_{q} - 48 {\dot{q}}^2 q + 4 {\dot{q}}^2 q^3 \big).
        \end{split}
    \end{equation*}

    All the necessary terms needed to compute the IID for a robot containing CC bodies connected are now available. Indeed, the forward step of Algorithm~\ref{alg:Kane inverse recursive general} computes ${}^{i}\calFv_{i}^{*}$, ${}^{i}\calTv_{i}^{*}$ and ${}^{i}\piv_{i}^{*}$. These are then projected in the configuration space by~$\parv{{}^{i}\vv_{i}}{\dqv_{i}}$ and~$\parv{{}^{i}\omegav_{i}}{\dqv_{i}}$ during the backward step. 

    Note also that we have derived closed-form expressions for all terms. This implies that, for a robot with planar CC bodies, the ID can be computed analytically for an arbitrary number of bodies using Algorithm~\ref{alg:Kane inverse recursive general}.
\end{example*}
}

It is worth noting that, in the case where the system contains only rigid bodies, the IID for rigid robots are immediately recovered, as one might expect. 

\review{
In assessing the computational complexity of Algorithm~\ref{alg:Kane inverse recursive general}, we do not account for the cost of evaluating the intra-body terms, i.e., those that depend on the body kinematics~\eqref{eq:body to body transformation}. However, in practice, the computational resources required for their evaluation may be significant. For an analysis of how kinematics affects the computation of the ID, the interested reader is referred to Appendix~\ref{appendix:impact evaluation}.
}
\begin{remark}    
    Similar to ID algorithms for rigid robots~\citep{featherstone2014rigid}, Algorithm~\ref{alg:Kane inverse recursive general} is well-suited for parallelization. Indeed, the relative kinematic and dynamic terms, such as $ \parv{{}^{i}\vv_{i}}{\dqv_{i}}$ and ${}^{i}\piv_{i}^{*}$, can be evaluated in a parallel loop prior to executing the forward and backward steps of Algorithm~\ref{alg:Kane inverse recursive general}. Their computation \review{typically accounts for the majority of the processing time. This is primarily due to the need for repeated evaluations of highly nonlinear functions that depend on the kinematics~\eqref{eq:functional expression body kinematics}, which cannot be accelerated since the latter is treated as a black-box. Therefore, significant attention should be devoted to implementing these terms as efficiently as possible.}
\end{remark}
\subsection[O(N) computation of M]{$O(N)$ computation of $\Mm(\qv)$}\label{N computation of M}
We now discuss how Algorithm~\ref{alg:Kane inverse recursive general} can be modified to compute the generalized mass matrix $\Mm(\qv)$ alongside the generalized active force $\Qm$ in $O(N)$ steps. 

First, observe that
\begin{equation*}
\Mm(\qv)\ddqv = \text{IID}(\qv, \zerov_{n}, \ddqv),
\end{equation*}
which implies
\begin{equation}\label{eq:M ID}
\Mm(\qv) = \jac{\text{IID}(\qv, \zerov_{n}, \ddqv)}{\ddqv} = -\jac{{\piv^{*}}^{\zerov}}{\ddqv} - \jac{\calMv^{\zerov}}{\ddqv},
\end{equation}
where the superscript ${}^{\zerov}$ on $\piv^{*}$ and $\calMv$ indicates that these terms are computed with zero velocity. In other words, the evaluation of $\Mm(\qv)$ can be performed by setting the velocities to zero and differentiating both $\piv^{*}$ and $\calMv$ with respect to $\ddqv$. It is worth noting that the idea of exploiting the linearity of the dynamics in the accelerations to evaluate the mass matrix has already been proposed for rigid manipulators, both in configuration space~\citep{rodriguez1992spatial, pu1996parallel} and task space~\citep{lilly1990n}. However, to the best of our knowledge, this is the first time such a method is discussed for robots with deformable bodies. We also propose embedding the computation of $\Mm$ directly into the ID process because it has minimal impact on computation time while providing greater flexibility for control purposes. In Appendix~\ref{appendix:mass inverse dynamics}, we detail how $\Mm$ can be computed alongside $\Qm$ and present a new algorithm called Mass Inertial Inverse Dynamics (MIID) for this computation. 

Having a procedure that allows evaluating also $\Mm$ while solving the ID offers two benefits. First, it enables the computation of nonlinear control techniques for underactuated mechanical systems~\citep{spong1994partial}, such as continuum soft robots, and variable gain PD control~\citep{santibanez2001pd, della2020exciting, wotte2023discovering} with a single call to the ID. Second, it provides a means of solving the FD problem, i.e., given $\qv, \dqv$ and $\Qm(\qv, \dqv, \uv)$ compute $\ddqv$, with a significant speedup over the inertia-based algorithm (IBA)~\citep{featherstone2014rigid}. Computing forward dynamics using an ID procedure, whether the MIID or the IBA, is never optimal. Nonetheless, it is worth mentioning this class of approaches because they are almost effortless to implement and have been used in the simulations of this paper. 

Given the MIID, one can solve the FD in two steps 
\begin{algorithmic}[]
\State $\rb{ \cv, \Mm } = \textnormal{MIID}(\qv, \dqv, \zerov_{n})$,
\State $\ddqv = \Mm^{-1}(-\cv + \Qm)$.
\end{algorithmic}
Recall that, in the above algorithm, $\Qm$ is known because we are solving the FD problem. On the other hand, in the IBA, $\ddqv$ is computed in three stages
\begin{algorithmic}[]
\State $\cv = \text{IID}(\qv, \dqv, \zerov_{n})$,
\For{$i = 1 \rightarrow n$}
    \State $\Mm_{i} = \text{IID}(\qv, \dqv, (\boldsymbol{\mathbb{I}}_{n})_{i})$,
\EndFor
\State $\ddqv = \Mm^{-1}(-\cv + \Qm)$.
\end{algorithmic}
Because of the mass matrix inversion, the complexity for solving the FD with the MIID is the same as that of the IBA, namely $O(N^{3})$. However, in the IBA, the computation of $\Mm$ requires $n$ calls to the ID, resulting in a $O(N^{2})$ cost. Additionally, the kinematic and inertial terms must be re-evaluated at every call with an obvious waste of computational resources. When the body geometry is complex, the computation of these terms can easily become a bottleneck and exceed the time required for matrix inversion, especially if the total number of DoFs remains in the order of tens, as is customary in ROMs for control purposes.
\begin{algorithm}[t]
\caption{\small Inertial Inverse Dynamics: $\Qm \hspace{-2pt} = \hspace{-2pt} \text{IID}(\qv, \dqv, \ddqv$)}\label{alg:Kane inverse recursive general}
\begin{algorithmic}
\Require $\qv, \dqv, \ddqv$
\For{$i = 1 \rightarrow N$}\Comment{Forward step}
    \State Compute ${}^{i}\av_{i}$, ${}^{i}\av_{\com_{i}}$ and ${}^{i}\domegav_{i}$
    \State Compute ${}^{i}\calFv_{i}^{*}$, ${}^{i}\calTv_{i}^{*}$ and ${}^{i}\piv_{i}^{*}$
    \State \review{Compute $\parv{{}^{i}\vv_{i}}{\dqv_{i}}$ and $\parv{{}^{i}\omegav_{i}}{\dqv_{i}}$}
\EndFor
\For{$i = N \rightarrow 1$}\Comment{Backward step}
    \State \review{Compute ${}^{i}\Fv_{i}^{*}$ and ${}^{i}\Tv_{i}^{*}$}
    \State \review{Compute $\Qm_{i} = -{}^{i}\piv_{i}^{*} -\parv{{}^{i}\vv_{i}}{\dqv_{i}} {}^{i}\Fv_{i}^{*} - \parv{{}^{i}\omegav_{i}}{\dqv_{i}} {}^{i}\Tv_{i}^{*} $}
\EndFor
\end{algorithmic}
\end{algorithm}
\section{Inverse Dynamics}\label{section:actuation inverse dynamics}
After solving Problem~\ref{problem:IDP}, we now demonstrate how to solve the ID problem (Problem~\ref{problem:AIDP}). 
At this stage, the ID can be conceptually addressed by leveraging the IID algorithm and Property~\ref{property:M} to separate\review{, without affecting computational complexity,} the effects of the actuators from those of other active forces, such as the gravitational load and the interaction forces due to body deformability. 

Specifically, being elements of vector spaces, $\drm {}^{i}\fv_i$ and $\drm {}^{i}\tauv_{i}$ can be decomposed in the forces and torques acting on the body. This paper considers, other than actuation, two additional types of active forces, namely gravitational and \review{visco-elastic} forces. 
Thus, the active force and torque take the form
\begin{equation}\label{eq:active force components}
	\drm {}^{i}\fv_{i} = \drm{}^{i}\fv^{g}_{i} + \drm{}^{i}\fv^{s}_{i} + \drm{}^{i}\fv^{a}_{i},
\end{equation}
and
\begin{equation}\label{eq:active torque components}
	\drm {}^{i}\tauv_{i} = \drm{}^{i}\tauv^{s}_{i} + \drm{}^{i}\tauv^{a}_{i},
\end{equation}
where $\drm{}^{i}\fv^{g}_{i}$, $\drm{}^{i}\fv^{s}_{i}$ and $\drm{}^{i}\fv^{a}_{i}$ denote the force due to gravity, stress and actuation, respectively. The vectors $\drm{}^{i}\tauv^{s}_{i}$ and $\drm{}^{i}\tauv^{a}_{i}$ represent their rotational counterparts. Note that the effect of gravity appears only as a linear force, namely $\drm{}^{i}\tauv^{g}_{i} = \zerov_{3}$. Recalling the definition of $\Qm_{i}$ in~\eqref{eq:Q_j,l_deformable} and the subsequent derivations, the generalized active force is
\begin{equation}\label{eq:generalized active force components}
	\Qm_{i}(\qv, \dqv, \uv) = -\gv_{i}(\qv) -\sv_{i}(\qv, \dqv) + \vect{\nu}_{i}(\qv, \uv),
\end{equation}
where each term in the right-hand side of the equation is obtained by replacing $\drm\fv_{i}$ and $\drm \tauv_{i}$ with the corresponding force and torque as in~\eqref{eq:active force components} and~\eqref{eq:active torque components}. 
We can now compute the ID replacing~\eqref{eq:generalized active force components} into~\eqref{eq:Kane} as
\begin{equation*}
    \vect{\nu}_{i} = \gv_{i}(\qv) +\sv_{i}(\qv, \dqv) - \Qm_{i}^{*},
\end{equation*}
or, equivalently by using~\eqref{eq:Qm_{j} M expression}--\eqref{eq:Q_{j}^{*} M expression} and the linearity of $\calMv_{i}$ (Property~\ref{property:M}),
\begin{equation}\label{eq:Qm_{j}^{a} final}
\begin{split}
    \vect{\nu}_{i} &= -{}^{i}\piv_{i}^{*} - {}^{i}\piv_{i}^{g} - {}^{i}\piv_{i}^{s}\\
    &\quad- \calMv_{i}({}^{i}\Fv_{i}^{g} + {}^{i}\Fv_{i}^{s}, {}^{i}\Fv_{i}^{*}, {}^{i}\Tv_{i}^{s}, {}^{i}\Tv_{i}^{*}),
\end{split}
\end{equation}
with ${}^{i}\Fv_{i}^{\{g, s\}}$ and ${}^{i}\Tv_{i}^{s}$ obtained recursively by using ${}^{i}\calFv_{i}^{\{ g, s\}}$ and ${}^{i}\calTv_{i}^{s}$ in~\eqref{eq:M_{jl} recursive expression update}. The last two terms and ${}^{i}\piv_{i}^{\{ g, s \}}$ are obtained replacing the respective infinitesimal forces and torques into~\eqref{eq:calFv_i 1}--\eqref{eq:calTv_i 1:2} and~\eqref{eq:pi expression final}, respectively. We address their computation in the following subsections.

The pseudo-code for the computation of $\nuv_{i}$ is given in Algorithm~\ref{alg:Kane inverse recursive}, which is similar to that of the IID. The difference lies in accounting for the terms that correspond to active forces which are not those generated by actuation. Note that it is possible to embed the computation of the mass matrix also in Algorithm~\ref{alg:Kane inverse recursive} by using the same recursive formulas of the Jacobians presented in Appendix~\ref{appendix:mass inverse dynamics} for Algorithm~\ref{alg:Kane inverse recursive general} and leading to an algorithm analogous to the MIID, denoted later as Mass Inverse Dynamics (MID).
\begin{algorithm}[t]
\caption{\small Inverse Dynamics: $\nuv = \mathrm{ID}(\qv, \dqv, \ddqv$)}\label{alg:Kane inverse recursive}
\begin{algorithmic}
\Require $\qv, \dqv, \ddqv$
\For{$i = 1 \rightarrow N$}\Comment{Forward step}
    \State Compute ${}^{i}\av_{i}$, ${}^{i}\av_{\com_{i}}$ and ${}^{i}\domegav_{i}$
    \State Compute ${}^{i}\calFv_{i}^{*}$, ${}^{i}\calTv_{i}^{*}$, ${}^{i}\piv_{i}^{*}$, 
    ${}^{i}\calFv_{i}^{g}$, ${}^{i}\piv_{i}^{g}$, 
    ${}^{i}\calFv_{i}^{s}$, ${}^{i}\calTv_{i}^{s}$ and ${}^{i}\piv_{i}^{s}$
    \State \review{Compute $\parv{{}^{i}\vv_{i}}{\dqv_{i}}$ and $\parv{{}^{i}\omegav_{i}}{\dqv_{i}}$}
\EndFor
\For{$i = N \rightarrow 1$}\Comment{Backward step}
    \State \review{Compute ${}^{i}\Fv_{i}^{*}$, ${}^{i}\Tv_{i}^{*}$, ${}^{i}\Fv_{i}^{g}$, ${}^{i}\Fv_{i}^{s}$, ${}^{i}\Tv_{i}^{s}$}
    \State \review{Compute $\nuv_{i} = - \rb{{}^{i}\piv_{i}^{*} - {}^{i}\piv_{i}^{g} - {}^{i}\piv_{i}^{s} }\vspace{-0.2cm}$
    \begin{equation*}\quad\quad\,\,
        \begin{split}
            &- \parv{{}^{i}\vv_{i}}{\dqv_{i}} \rb{{}^{i}\Fv_{i}^{*} - {}^{i}\Fv_{i}^{g} - {}^{i}\Fv_{i}^{s}}\\
            &- \parv{{}^{i}\omegav_{i}}{\dqv_{i}} \rb{ {}^{i}\Tv_{i}^{*} - {}^{i}\Tv_{i}^{s} }
        \end{split}
    \end{equation*}}
\EndFor
\end{algorithmic}
\end{algorithm} 
\subsection{Gravitational force}
\review{In the following, we focus on computing the gravitational load, which is obtained by considering the gravitational force acting on an infinitesimal volume element of the body and by integrating over the entire body volume.}

\review{Specifically, } the gravitational active force performing work on $\calB_i$ is
\begin{equation}\label{eq:gravity force}
	{}^{i}\calFv_{i}^{g} = \int_{V_{i}} \Rotm[i]^{T}{}^{0}\gv \drm m_i = \Rotm[i]^{T} {}^{0}\gv m_{i} = {}^{i}\gv m_{i},
\end{equation}
being ${}^{0}\gv \in \R^{3}$ the gravity vector in $\{ S_0 \}$ and ${}^{i}\gv$ its representation in body coordinates. Similarly, we have
\begin{equation}\label{eq:gravity force:pi}
    {}^{i}\piv_{i}^{g} = \parv{{}^{i}\dpv_{\com_{i}}}{\dqv_{i}} {}^{i}\gv m_i.
\end{equation}
The computation of ${}^{i}\calFv_{i}^{g}$ can be performed sequentially from the first body to the last. However, by comparing~\eqref{eq:gravity force} with~\eqref{eq:inertia force} and~\eqref{eq:gravity force:pi} with~\eqref{eq:int dri ddpi}, it is possible to see that the effect of gravity can be incorporated into the calculations by setting the acceleration of the base to ${}^{0}\av_{0} = -{}^{0}\gv$.

\review{
\begin{examplestop*}[(Continuation)]
    Under the assumption that the robot base is rotated so that the gravitational force acts along the $Z$ direction, i.e., ${}^{0}\gv = \rb{0 \,\, 0\,\, g_{0}}^{T}$, the above terms become 
    \begin{equation*}
        {}^{i}\calFv_{i}^{g} = \left(\!\begin{array}{c} 
            \displaystyle -\pi g_{0} \mathrm{s}_{q} \\ 
            0 \\ 
            \displaystyle -\pi g_{0} \mathrm{c}_{q} 
            \end{array}\!\right),
    \end{equation*}
    and
    \begin{equation*}
        {}^{i}\piv_{i}^{g} = -\frac{\pi \,g_{0}\,\left(2\,\mathrm{c}_{q}+q\,\mathrm{s}_{q}-2\right)}{q^3}.
    \end{equation*}
\end{examplestop*}
}
\subsection{\review{Visco-elastic} force}\label{sec:visco_elastic_force}
\review{As final step in evaluating the ID, we assess visco-elastic generalized forces, which model the tendency of a continuum to return in its undeformed configuration.} The derivation of these force requires necessarily further hypotheses because a stress-strain model is required. In the following, we assume that $\sv_{i}(\qv, \dqv) = \kv_{i}(\qv) + \dv_{i}(\qv, \dqv)$, and denote the corresponding infinitesimal forces (torques) as $\drm{}^{i}\fv^{e}_{i}$ ($\drm{}^{i}\tauv^{e}_{i}$) and $\drm{}^{i}\fv^{d}_{i}$ ($\drm{}^{i}\tauv^{d}_{i}$), respectively. 
Furthermore, let
\begin{equation*}
    {}^{i}\Bm_{i} = \jac{\fv_{\calB_{i}}}{\xv_{i}} \rb{\jac{\fv_{\calB_{i}}}{\xv_{i}}}^{T},
\end{equation*}
the Green strain tensor.
Assuming an incompressible Neo-Hookean solid one has
\begin{equation*}
    \drm {}^{i}\fv_{i}^{e} = 2 C_{{i}} \text{div}_{\xv_{i}}{}^{i}\Bm_{i}  \drm V,
\end{equation*}
and
\begin{equation*}
     \drm {}^{i}\tauv_{i}^{e} = {}^{i}\rv_{i} \times \drm {}^{i}\fv_{i}^{e},
\end{equation*}
where $C_{{i}}$ is a material parameter. Note that the divergence operator projects the internal stress into the dynamic equations~\citep{lacarbonara2013nonlinear}.
Similarly, considering a Kelvin–Voigt model for the viscous forces leads to
\begin{equation*}
    \drm {}^{i}\fv_{i}^{d} = \eta_{i} C_{i} \text{div}_{\xv_{i}} {}^{i}\Dot\Bm_{i}  \drm V,
\end{equation*}
and
\begin{equation*}
     \drm {}^{i}\tauv_{i}^{d} = {}^{i}\rv_{i} \times \drm {}^{i}\fv_{i}^{d},
\end{equation*}
being $\eta_{i}$ the material damping factor.

Finally, ${}^{i}\calFv_{i}^{\{e, d\}}$, ${}^{i}\calTv_{i}^{\{e, d\}}$ and ${}^{i}\piv_{i}^{\{e, d\}}$ can be obtained by proper integration over the body domain using~\eqref{eq:calFv_i 1}--\eqref{eq:calTv_i 1:2} and~\eqref{eq:pi expression final}.

\review{
\begin{examplestop*}[(Continuation)]
    It can be shown that~\citep{armanini2023soft}, for the above stress-strain model, the visco-elastic force is linear and takes the form $\sv_{i} = \frac{\pi}{4}\rb{q - \dot{q}}$, assuming again unitary values for elastic and damping parameters. 
\end{examplestop*}
}

\begin{remark}
    Many different stress-strain models could have been considered. Indeed, from Property~\ref{property:M}, a different model of these forces will affect only the computation of $\sv_{i}$. 
\end{remark}
\section{Numerical Results}\label{section:simulations}
The main contribution of this work is a unified and model-agnostic procedure to evaluate the ID of soft robots ROMs for control purposes. The previous algorithms are validated through numerical simulations, which are illustrated below. Experimental validation is beyond the scope of this paper and will be considered in future work. Additionally, it is important to keep in mind that the derived ID procedures, by design, do not depend on the specific kinematic model used and produce dynamics equivalent to the EL equations, as proven in Appendix~\ref{appendix:EL equivalence}. Therefore, the accuracy of the dynamic models obtained with our algorithms is equivalent, up to numerical precision, to that of the EL approach, which primarily depends on the accuracy of the underlying kinematic model. \review{For the same reasons mentioned above, we decided not to validate our method using open-source datasets available in the literature, such as~\cite{grassmann2022dataset}~\footnote{\color{red} Note also that available datasets are more suitable to assess the accuracy of a novel kinematic model and/or a forward dynamics procedure, neither of which is the focus of this work.}.}

Below, we utilize the MID algorithm described in Sec.~\ref{N computation of M} to simulate -- despite this is not its intended use -- and control a soft continuum robot with 3D bodies and a hybrid rigid-soft manipulator. In both simulations, we use a kinematic model based on LVP primitives~\citep{xu2023model}. It is the first time such model is used for dynamic simulation and thus also this section presents a (small) contribution of the paper. Through these simulations we also show the benefit from a control perspective of including the computation of $\Mm$ in the solution of the ID problem. In a third simulation, we compare the computation time of the IID algorithm with that of the EL equations to show the scalability of the recursive EoM. 

The same ID algorithm is used in all the following simulations. Every simulation is implemented by providing a different robot model to the procedure. The simulations are implemented in MATLAB through \href{https://github.com/piepustina/Jelly}{Jelly}, an object-oriented library offering C/C++ code generation functionality both in the MATLAB and Simulink environment.
\subsection{Simulation 1. Trimmed helicoid continuum soft robot}
\begin{figure}[th]
    \centering
    \subfigure[]{
        \includegraphics[width=0.45\columnwidth]{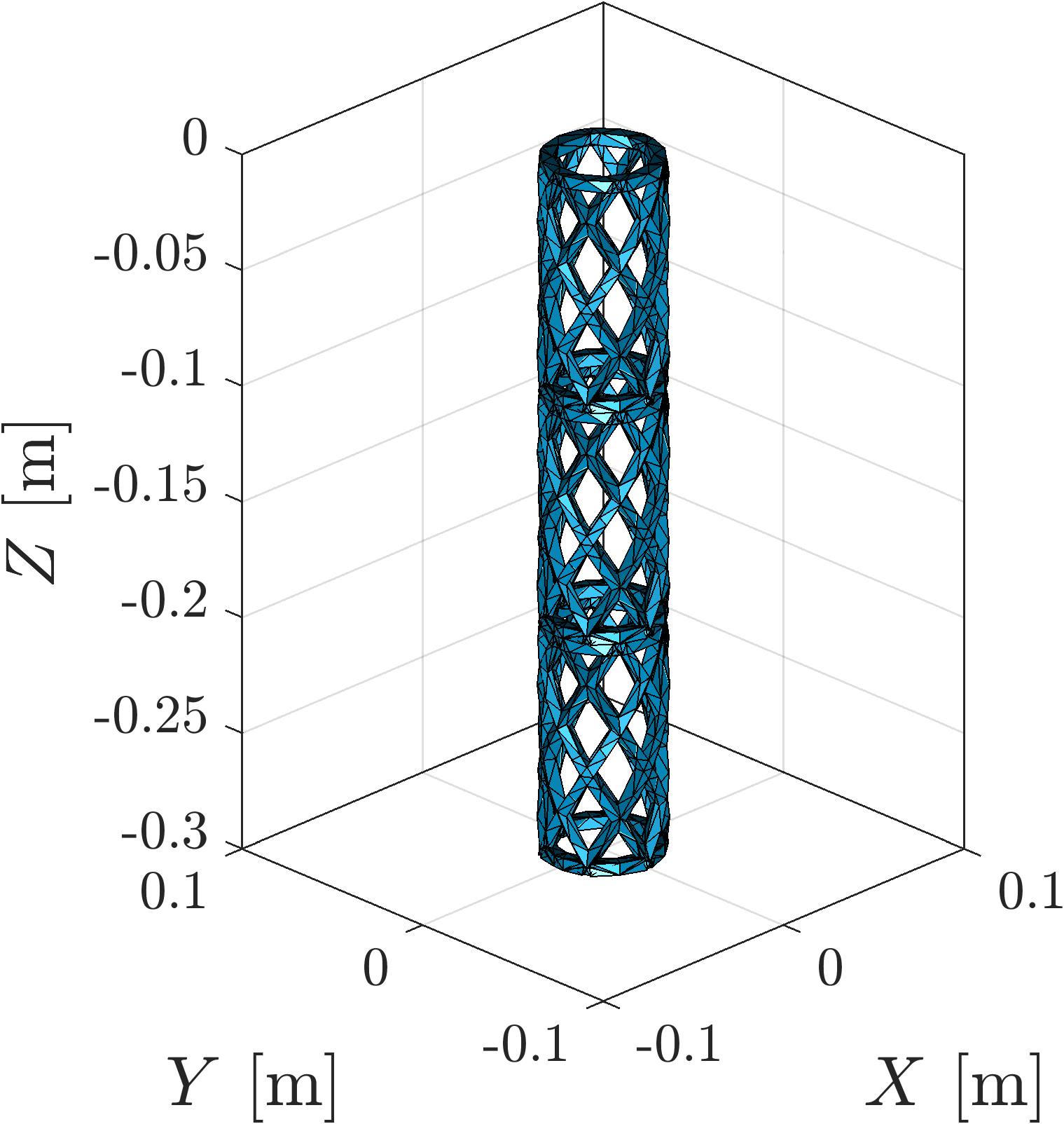}
    }
    \subfigure[]{
        \includegraphics[width=0.45\columnwidth]{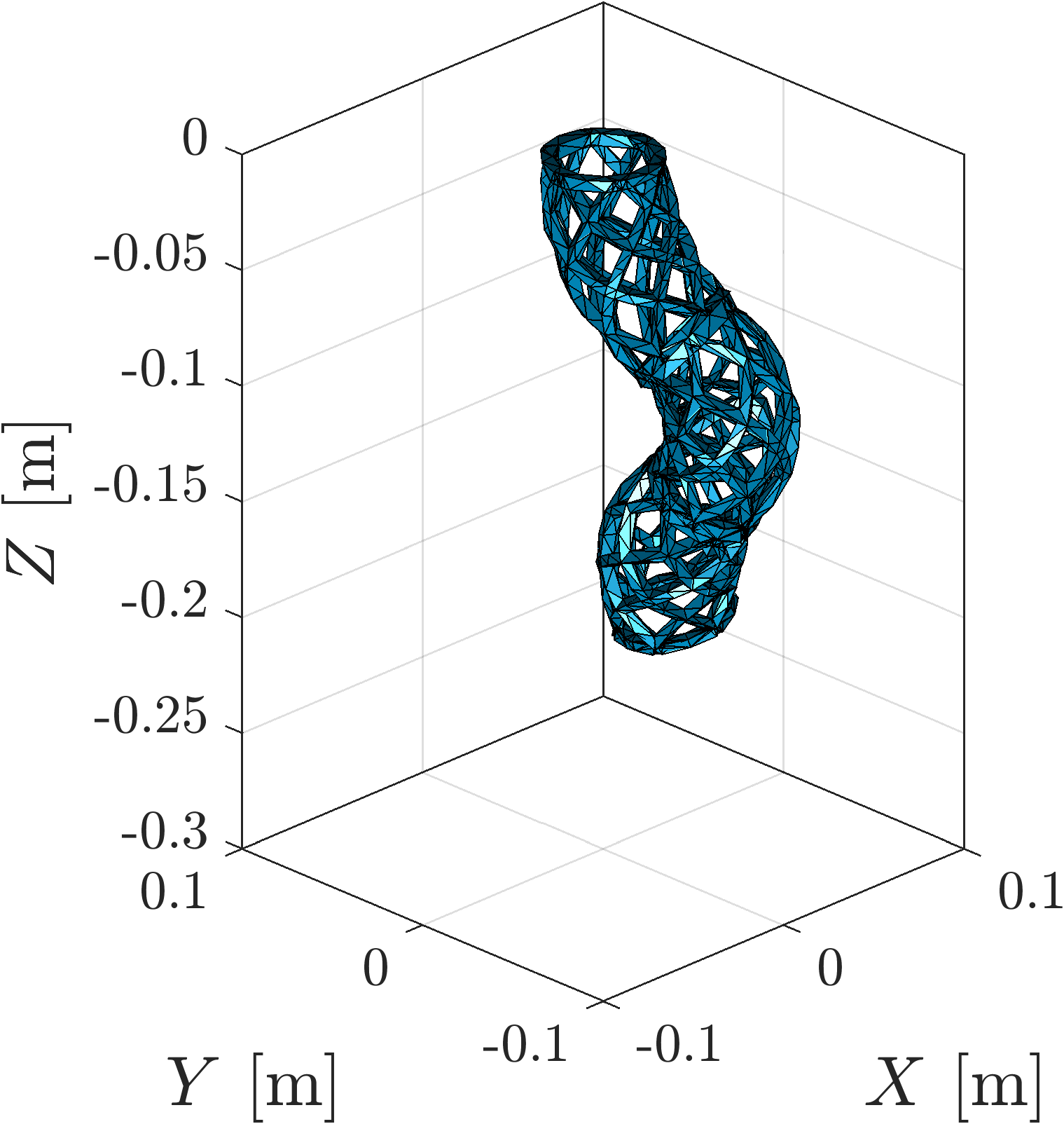}
    }
    \caption{\small Simulation 1. Trimmed helicoid continuum soft robot in (a) its stress free configuration and (b) a sample configuration. Note the radial deformation of the bodies in the deformed configuration.}
    \label{fig:simulation8:1:stress_free}
\end{figure}
We consider the task of controlling a trimmed helicoid continuum soft robot inspired by~\cite{guan2023trimmed} and illustrated in Fig.~\ref{fig:simulation8:1:stress_free}. These soft robotic systems do not satisfy the Cosserat rod assumption because they deform in all directions. As a consequence, other kinematic models should be considered to capture the system motion. The robot consists of three helical bodies of rest length $L_{i} = 10 \cdot 10^{-2}~[\si{\meter}]$ and radius $2 \cdot 10^{-2}~[\si{\meter}]$. The base is rotated so that the arm is aligned with the gravitational field in the straight configuration. Each body has uniform mass distribution $\rho_{i} = 960~[\si{\kilogram \per \cubic \meter}]$, elastic parameter $C_{i} = 0.178~[\si{\mega \pascal}]$ and damping coefficient $\eta_{i} = 0.05~[\si{\second}]$. 
The robot is actuated through nine straight tendons, running in triplets from one body to its successor and routed along their external perimeters. The kinematics is modeled using LVP primitives~\citep{xu2023model}, a new reduced-order FEM-like approach for modeling soft robots that has been simulated only in static conditions. The primitives are functions $\hv_{p}(\xv_{i}, \qv_{\calB_{i}}) \in \R^{3}; p \in \{ 1, \dots, P \}$ modeling elementary types of deformation and satisfying the property that $ \text{det}(\jac{\hv_{p}}{\xv}) = 1$ so that volume is locally preserved. Given a set of primitives, the body kinematics is obtained by function composition leading to
\begin{equation*}
    \fv_{\calB_{i}}(\xv_{i}, \qv_{\calB_{i}}) = \hv_{P}(\hv_{P-1}(\dots \hv_{1}(\xv_{i}, \qv_{\calB_{i}}), \qv_{\calB_{i}}), \qv_{\calB_{i}}),
\end{equation*}
which is a highly nonlinear function in both arguments. It is worth remarking that ID algorithms for flexible bodies or discrete Cosserat rods are not suitable for solving the ID problem for such kinematics. We consider six different types of LVP primitives, namely elongation, bending, shear and twist. 
\begin{table}[t!]
\caption{\small Modal expansion of the body primitives for Simulation 1. The body subscript is omitted for the sake of readability.}
\centering
\begin{tabular}{p{0.5\columnwidth}p{0.4\columnwidth}}
\hline
\hline
Deformation                   & Modal expansion              \\ \hline
Stretch and compression       & $(x_{3}/L)(1-x_{3}/L)\review{q_{1}}$       \\
Planar bending ($x_{1}$ axis) & $(q_{2} + x_{3}q_{3})/L$     \\
Planar bending ($x_{2}$ axis) & $(q_{4} + x_{3}q_{5})/L$     \\
Twist                         & $(x_{3}/L) q_{6}$            \\
Shear ($x_{1}$ axis)          & $(x_{3}/L) q_{7}$            \\
Shear ($x_{2}$ axis)          & $(x_{3}/L) q_{8}$            \\ \hline\hline
\end{tabular}
\label{tab:simulation1:modal_expansion}
\end{table}
Table~\ref{tab:simulation1:modal_expansion} summarizes, for each primitive, the modes functions used, which provide eight DoFs to each body. We refer the reader to~\cite{xu2023model} for the functional structures of the primitives and more details on this modeling approach. Given that the motion of the system is parameterized by a finite number of configuration variables, the dynamics takes the form of~\eqref{eq:finite dimensional equations}, namely
\begin{equation*}
    \Mm(\qv)\ddqv + \cv(\qv, \dqv) + \gv(\qv) + \sv(\qv, \dqv) = \vect{\nu}(\qv, \uv) = \Am(\qv)\uv,
\end{equation*}
where $\qv \in \R^{24}$, $\Am(\qv) \in \R^{24 \times 9}$ is the actuation matrix projecting the control inputs into the configuration space and $\uv \in \R^{9}$ denotes the vector of actuation inputs, i.e., the tendons tensions. The actuation matrix is computed using the principle of virtual work, following steps similar to those of~\cite{renda2022geometrically}, as
\begin{equation*}
    \delta \qv^{T} \vect{\nu}(\qv, \uv) = \delta \lv(\qv)^{T}\uv = \delta \qv^{T} \Am(\qv)\uv,
\end{equation*}
where $\lv(\qv)$ is the actuator length in the given configuration. Thus, the robot is highly underactuated with a $15$ degree of underactuation.
We use the stress-strain model described in Sec.~\ref{sec:visco_elastic_force}. More accurate and sophisticated actuation and stress-strain models could have been considered. Here, we choose a reasonable balance between fidelity and applicability for control purposes. 
The volume integrals appearing in the ID algorithms are computed numerically using meshes of $2749$ tetrahedrons for each body.

The control problem consists in following three circular trajectories assigned at the bodies tips. The commanded references in the robot base coordinates are (in [\si{\meter}])
\begin{equation*}\label{simulation1: reference}
    \tv_{1_{d}} = \begin{carray}{c}
        0.02~\cos\rb{\frac{2\pi}{5}t}\\
        0.02~\sin\rb{\frac{2\pi}{5}t}\\
        0.08
    \end{carray}, \,\, 
    \tv_{2_{d}} = \begin{carray}{c}
        0.025~\cos\rb{\frac{2\pi}{5}t}\\
        0.025~\sin\rb{\frac{2\pi}{5}t}\\
        0.15
    \end{carray}, 
\end{equation*}
and
\begin{equation*}
    \tv_{3_{d}} = \begin{carray}{c}
        0.030~\cos\rb{\frac{2\pi}{5}t}\\
        0.030~\sin\rb{\frac{2\pi}{5}t}\\
        0.24
    \end{carray}.
\end{equation*}
\begin{figure}
    \centering
    \subfigure[]{
        \includegraphics[width = 0.86\columnwidth]{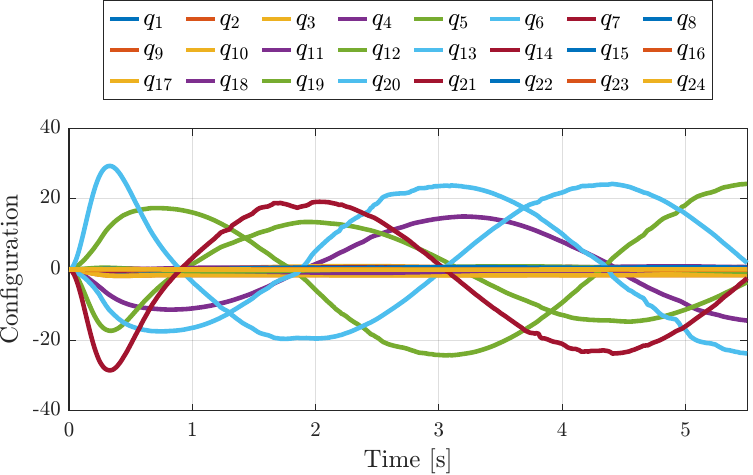}
    }
    \subfigure[]{
        \includegraphics[width = 0.86\columnwidth, trim={0cm 0cm 0cm 2cm}, clip]{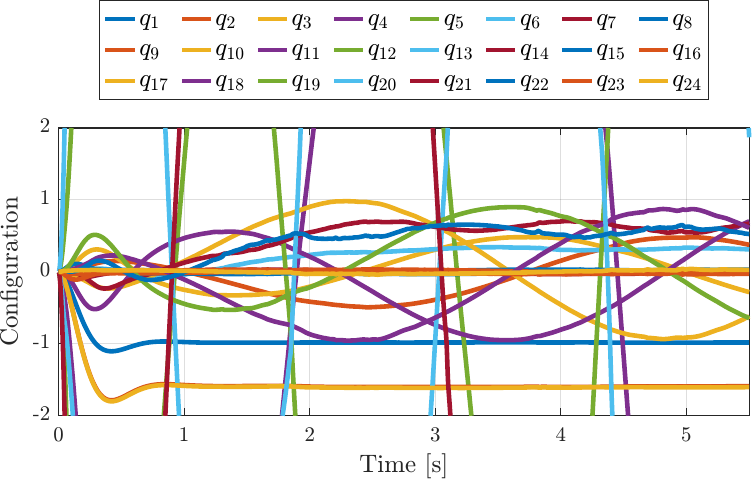}
    }
    \caption{\small Simulation 1. Time evolution of the configuration variables (in $[\si{\radian}]$ and $[\si{\meter}]$) with a zoomed view in panel (b).}
    \label{sim1:q}
\end{figure}
\begin{figure}
    \centering
    \subfigure[]{
        \includegraphics[width = 0.86\columnwidth]{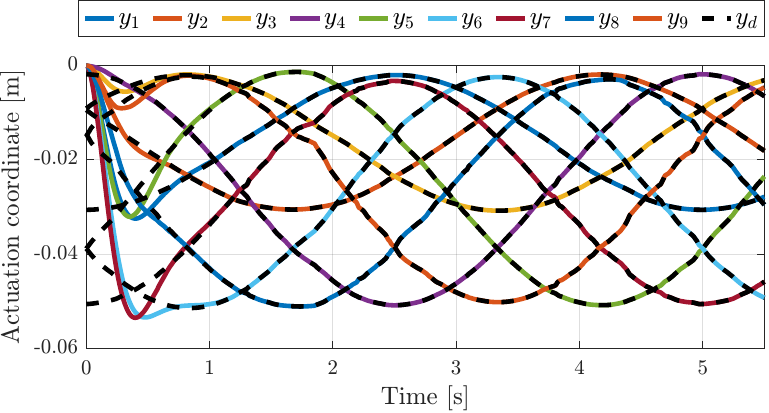}
    }
    \subfigure[]{
        \includegraphics[width = 0.86\columnwidth]{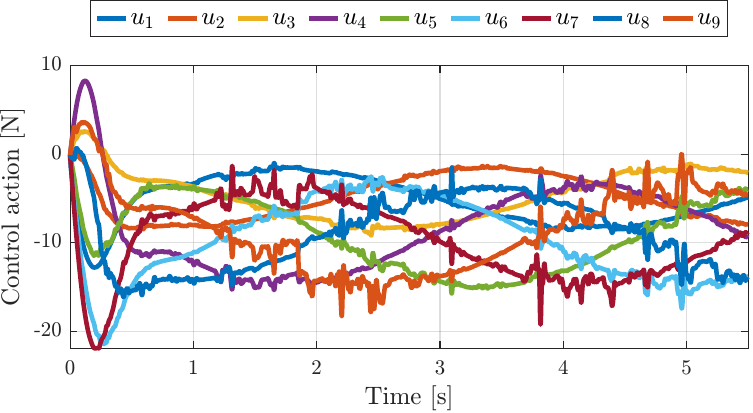}
    }
    \caption{\small Simulation 1. Time evolution of the (a) actuation coordinates and (b) control action.}
    \label{sim1:yu}
\end{figure}
\begin{figure*}[ht!]
    \centering
    \subfigure[{$t = 0~[\si{\second}]$}]{
        \includegraphics[width = 0.23\textwidth]{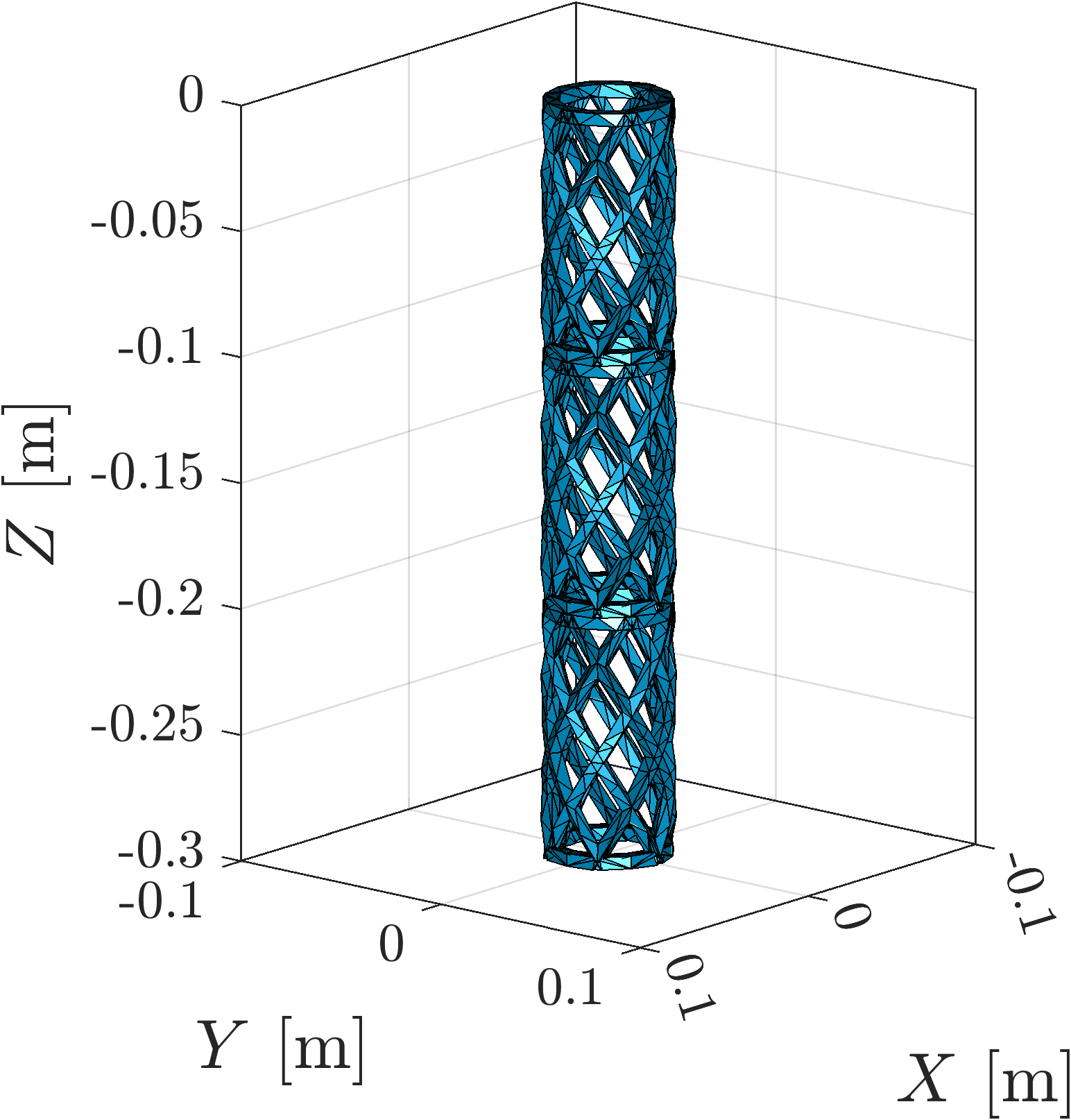}
    }
    \subfigure[{$t = 0.5~[\si{\second}]$}]{
        \includegraphics[width = 0.23\textwidth]{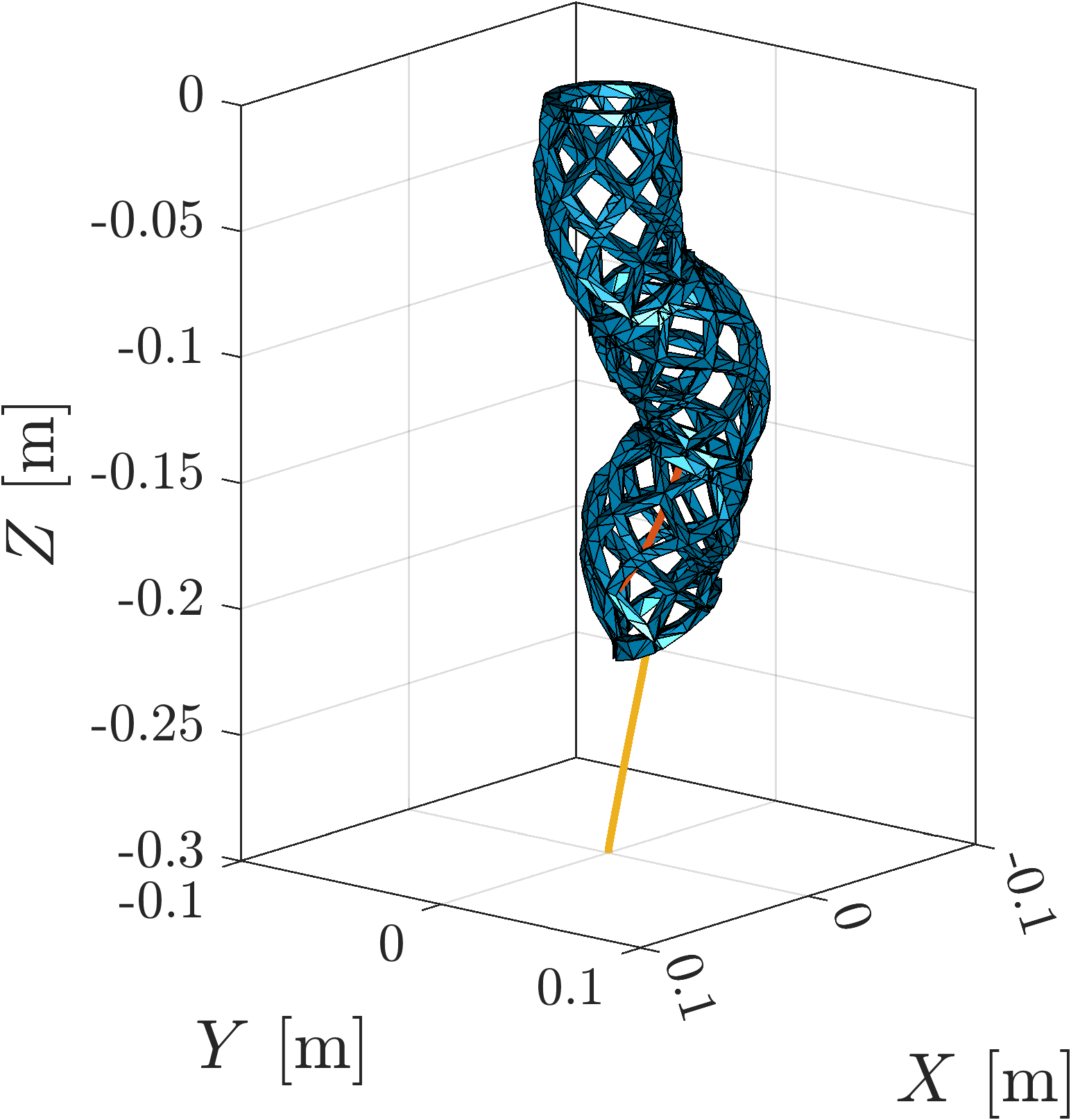}
    }
    \subfigure[{$t = 1~[\si{\second}]$}]{
        \includegraphics[width = 0.23\textwidth]{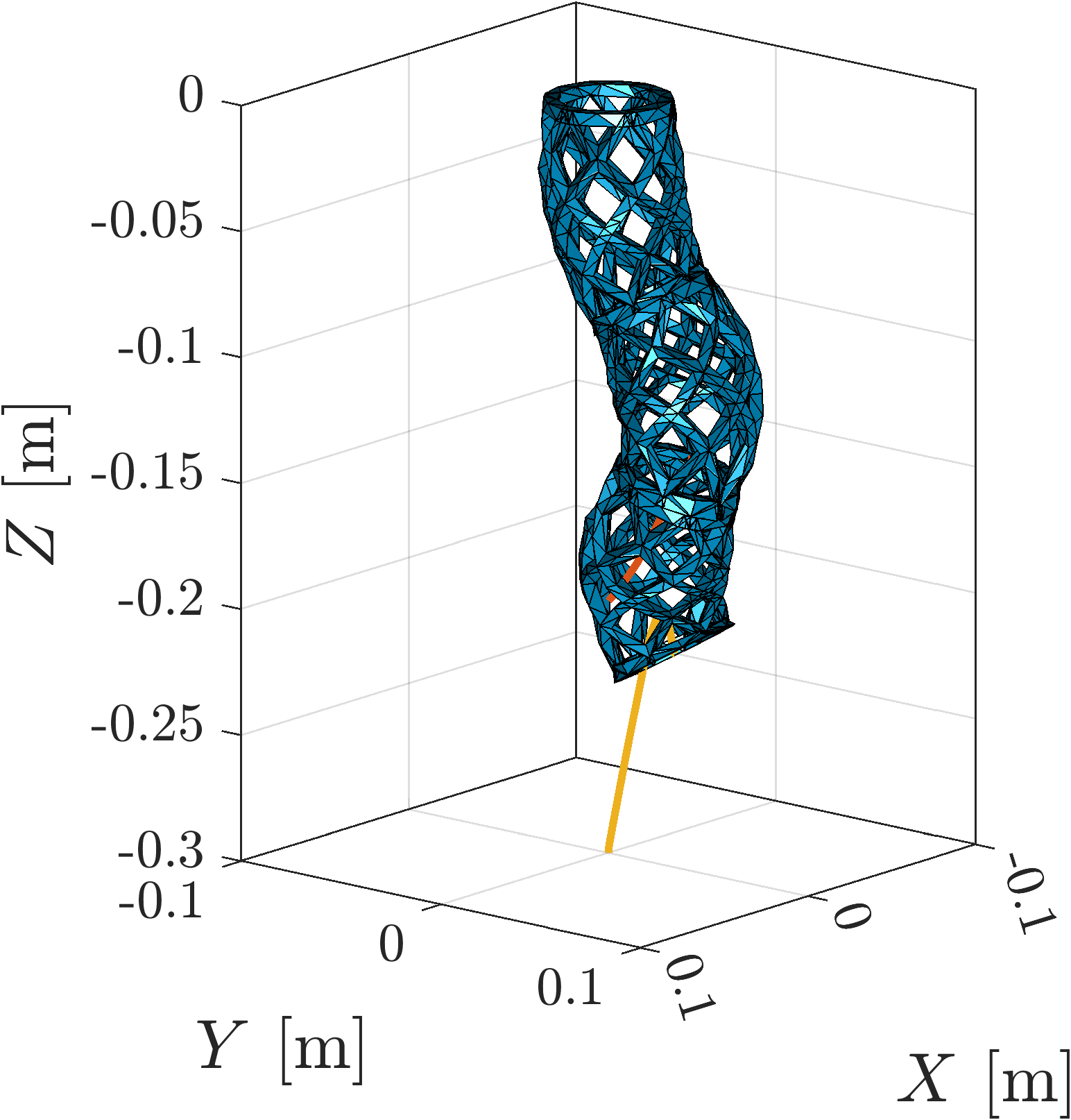}
    }
    \subfigure[{$t = 1.5~[\si{\second}]$}]{
        \includegraphics[width = 0.23\textwidth]{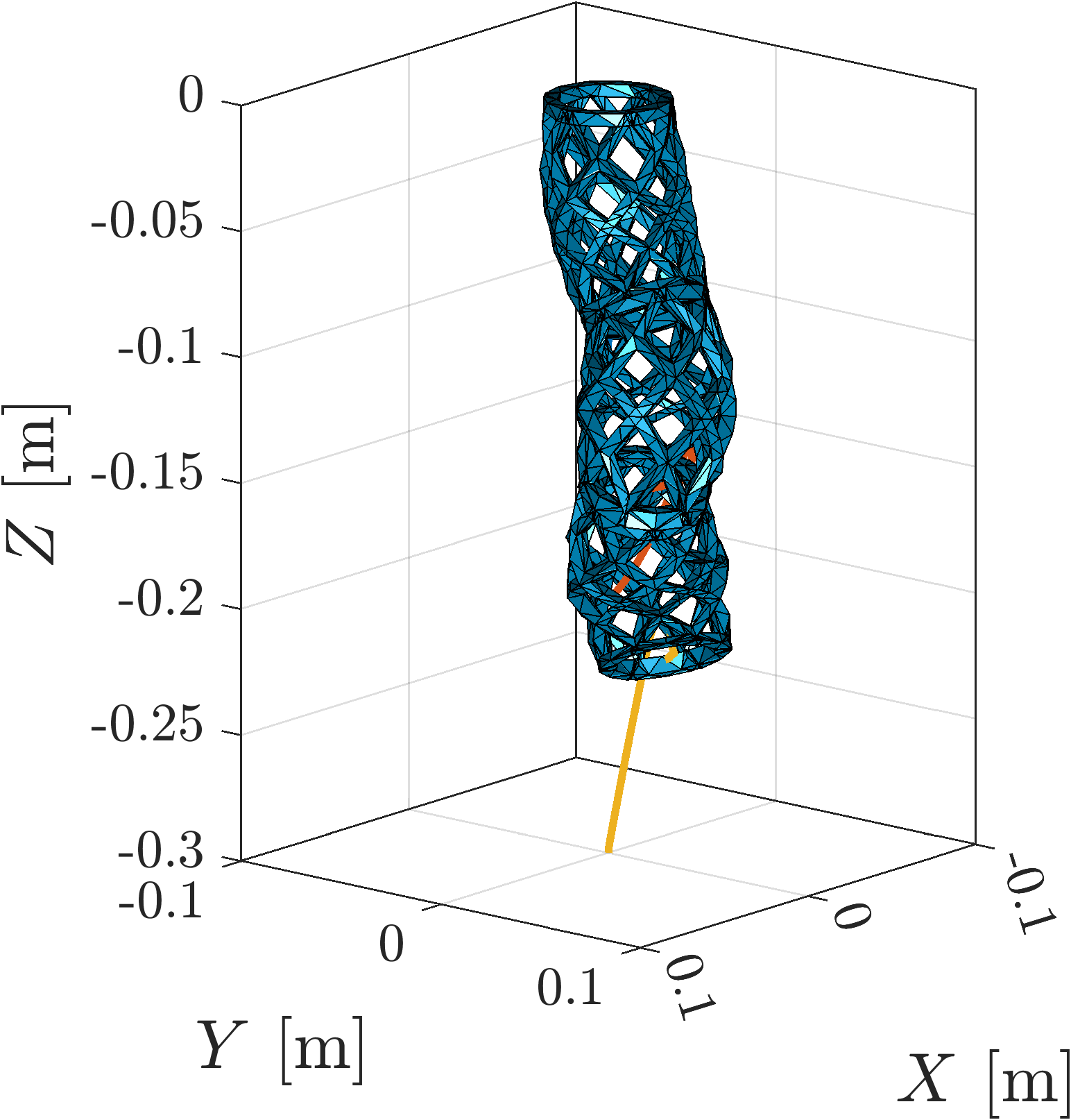}
    }
    \subfigure[{$t = 2~[\si{\second}]$}]{
        \includegraphics[width = 0.23\textwidth]{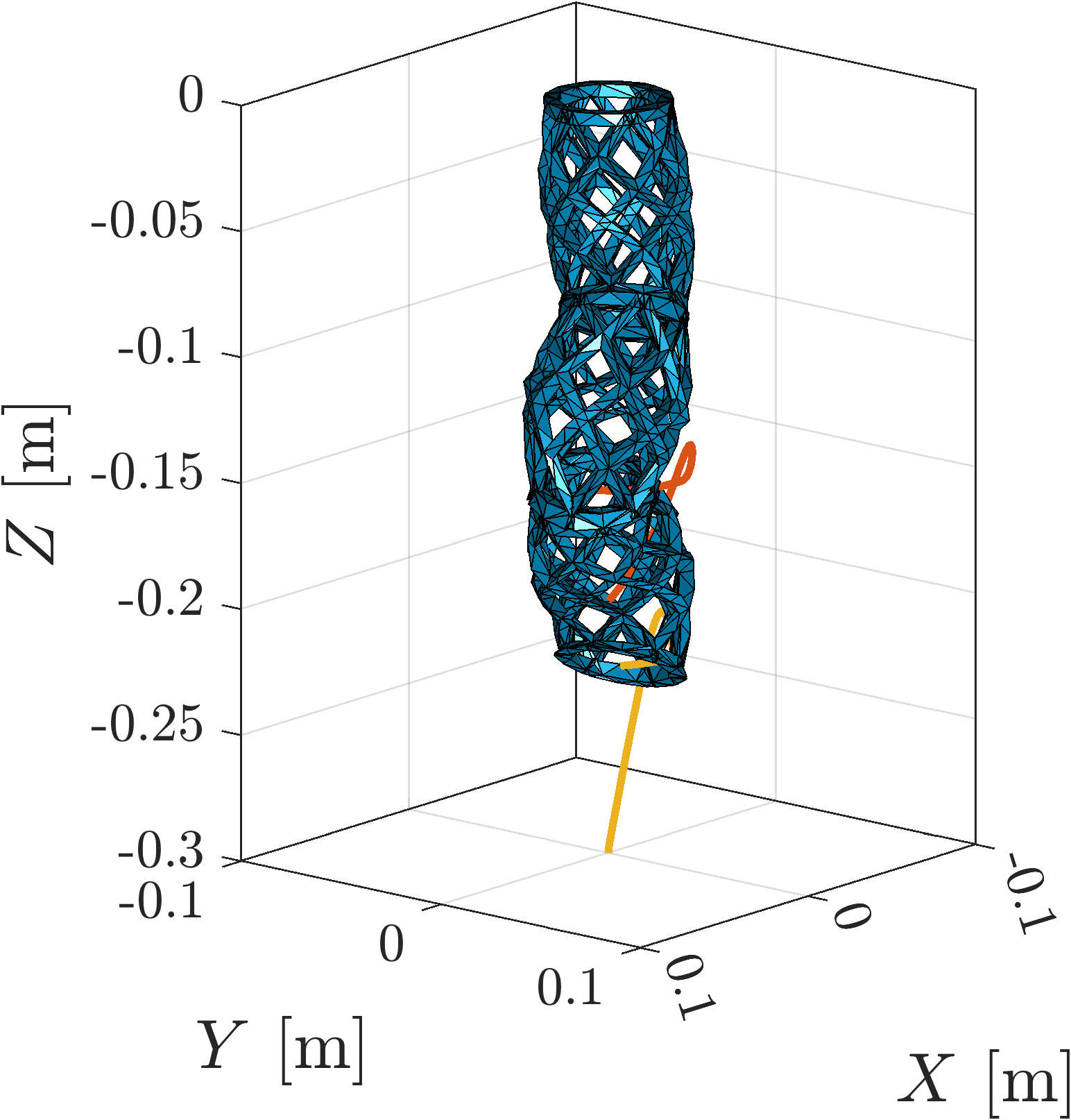}
    }
    \subfigure[{$t = 2.5~[\si{\second}]$}]{
        \includegraphics[width = 0.23\textwidth]{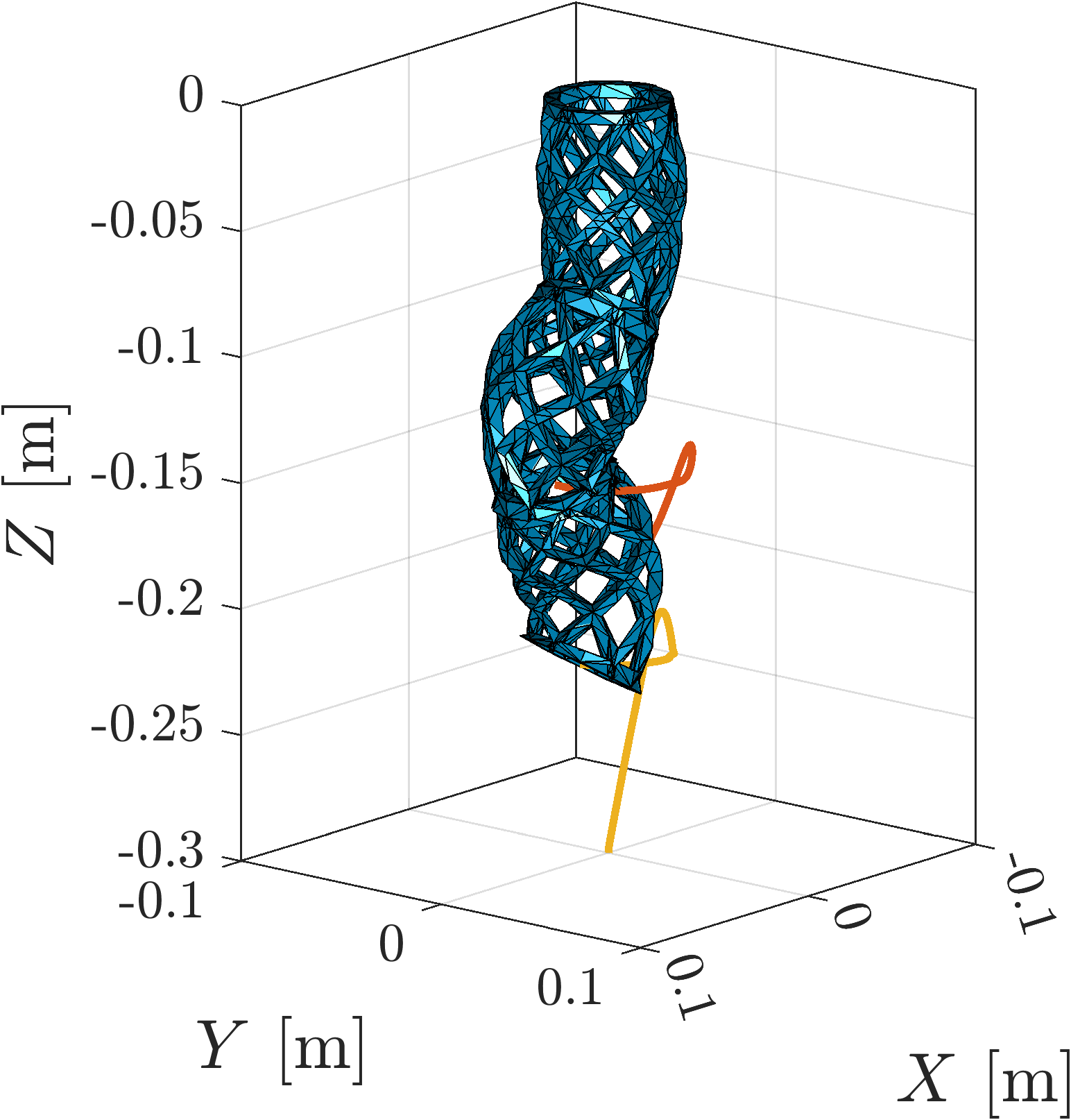}
    }
    \subfigure[{$t = 3~[\si{\second}]$}]{
        \includegraphics[width = 0.23\textwidth]{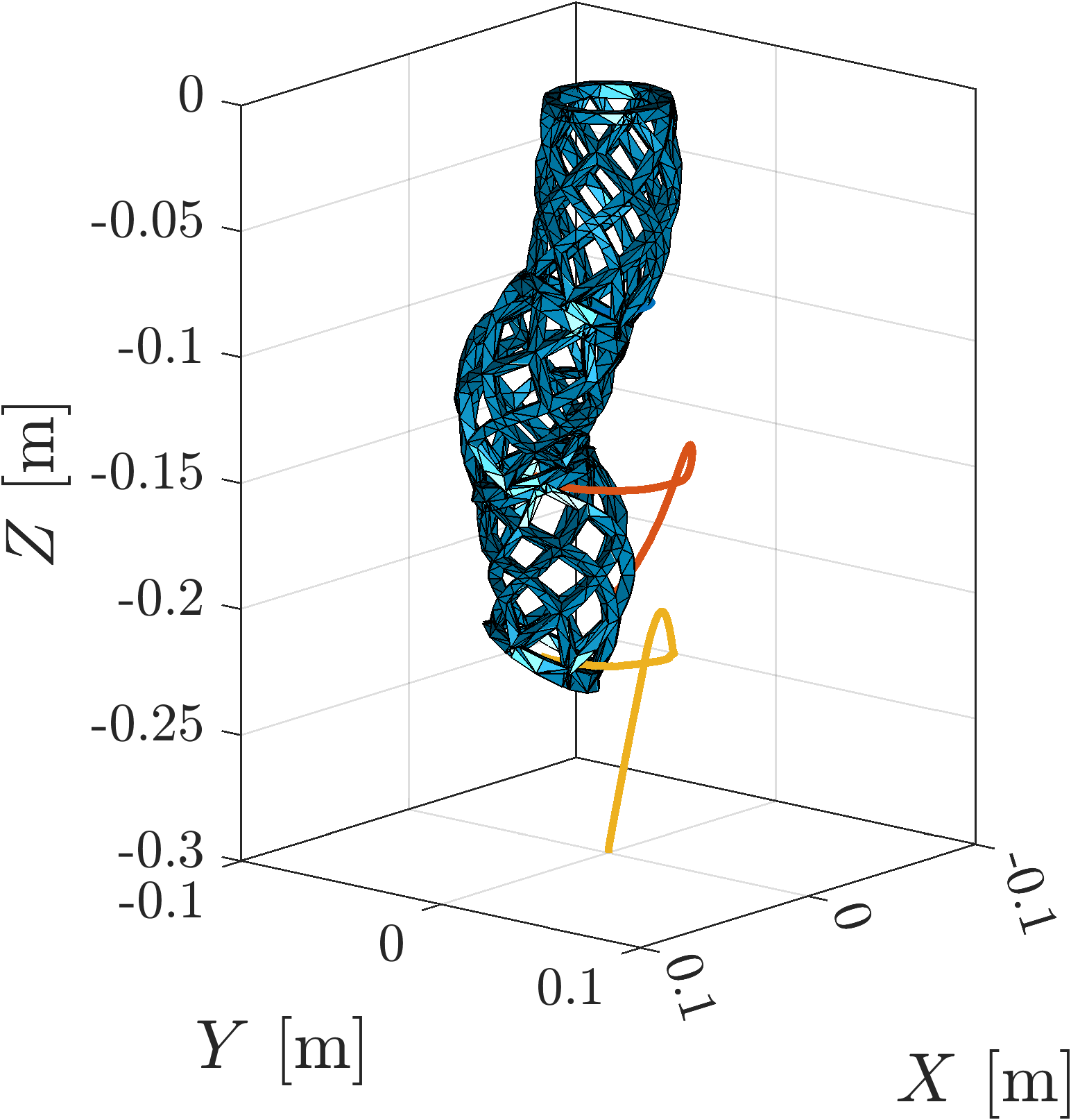}
    }
    \subfigure[{$t = 3.5~[\si{\second}]$}]{
        \includegraphics[width = 0.23\textwidth]{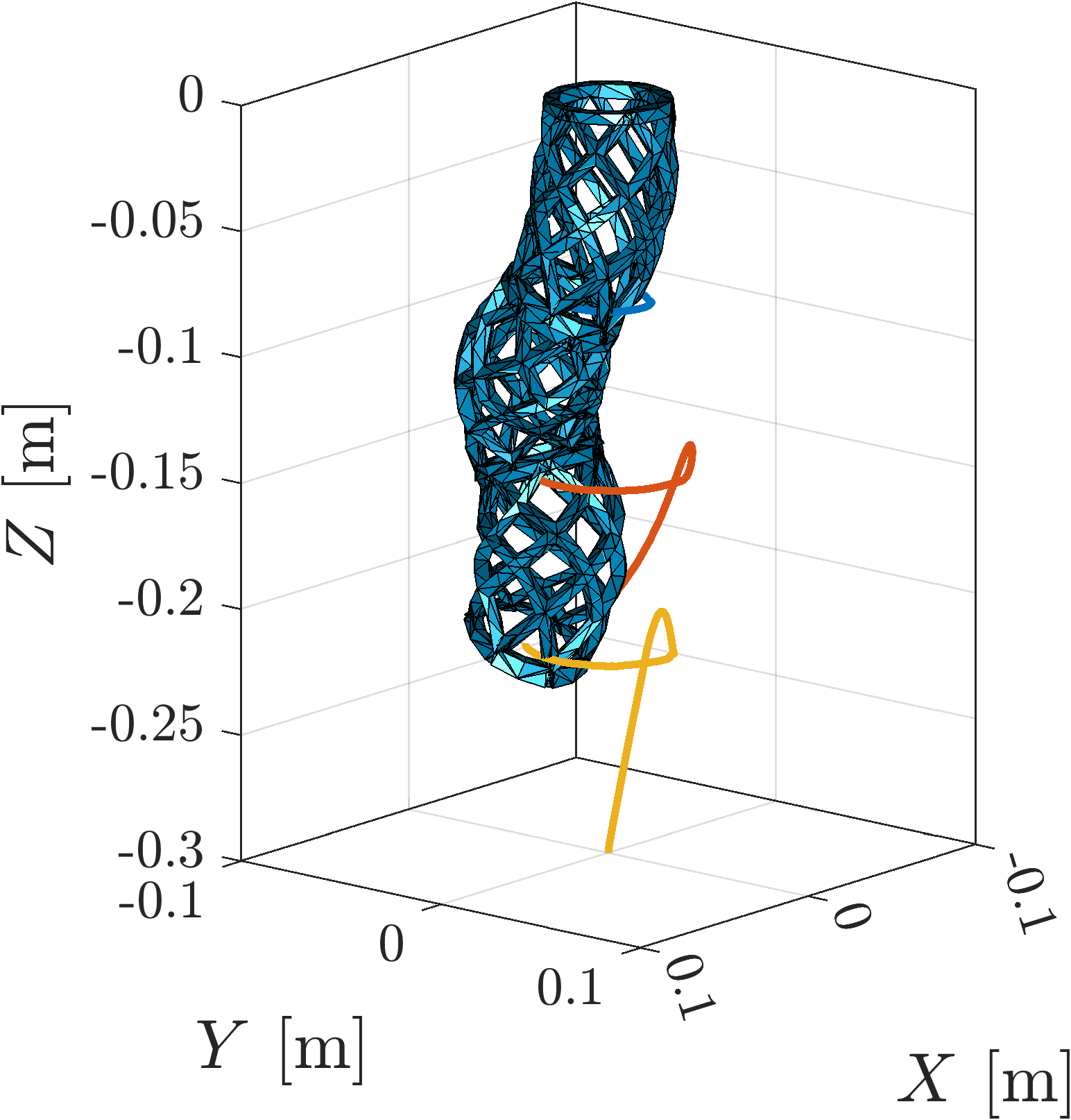}
    }
    \subfigure[{$t = 4~[\si{\second}]$}]{
        \includegraphics[width = 0.23\textwidth]{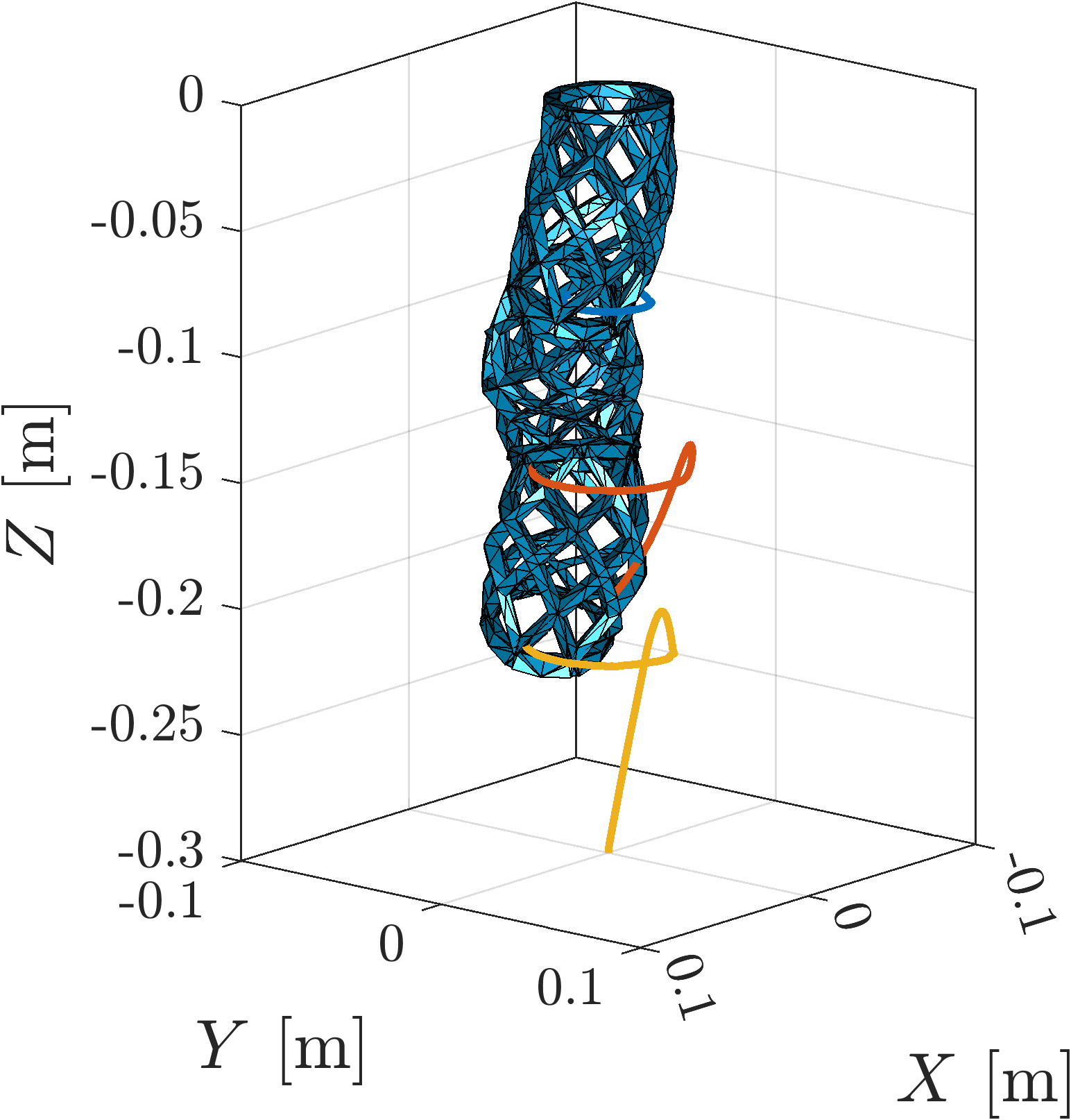}
    }
    \subfigure[{$t = 4.5~[\si{\second}]$}]{
        \includegraphics[width = 0.23\textwidth]{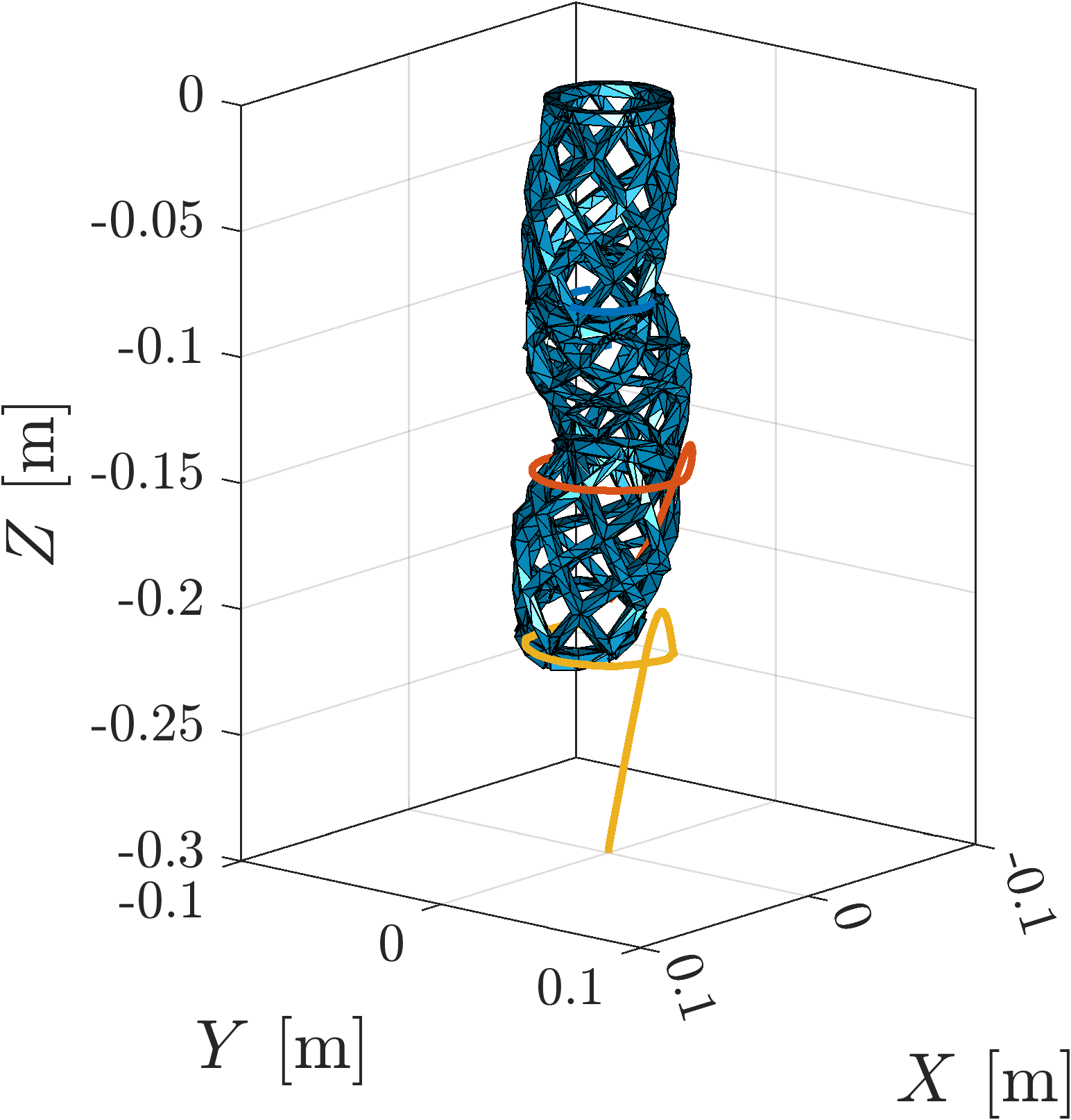}
    }
    \subfigure[{$t = 5~[\si{\second}]$}]{
        \includegraphics[width = 0.23\textwidth]{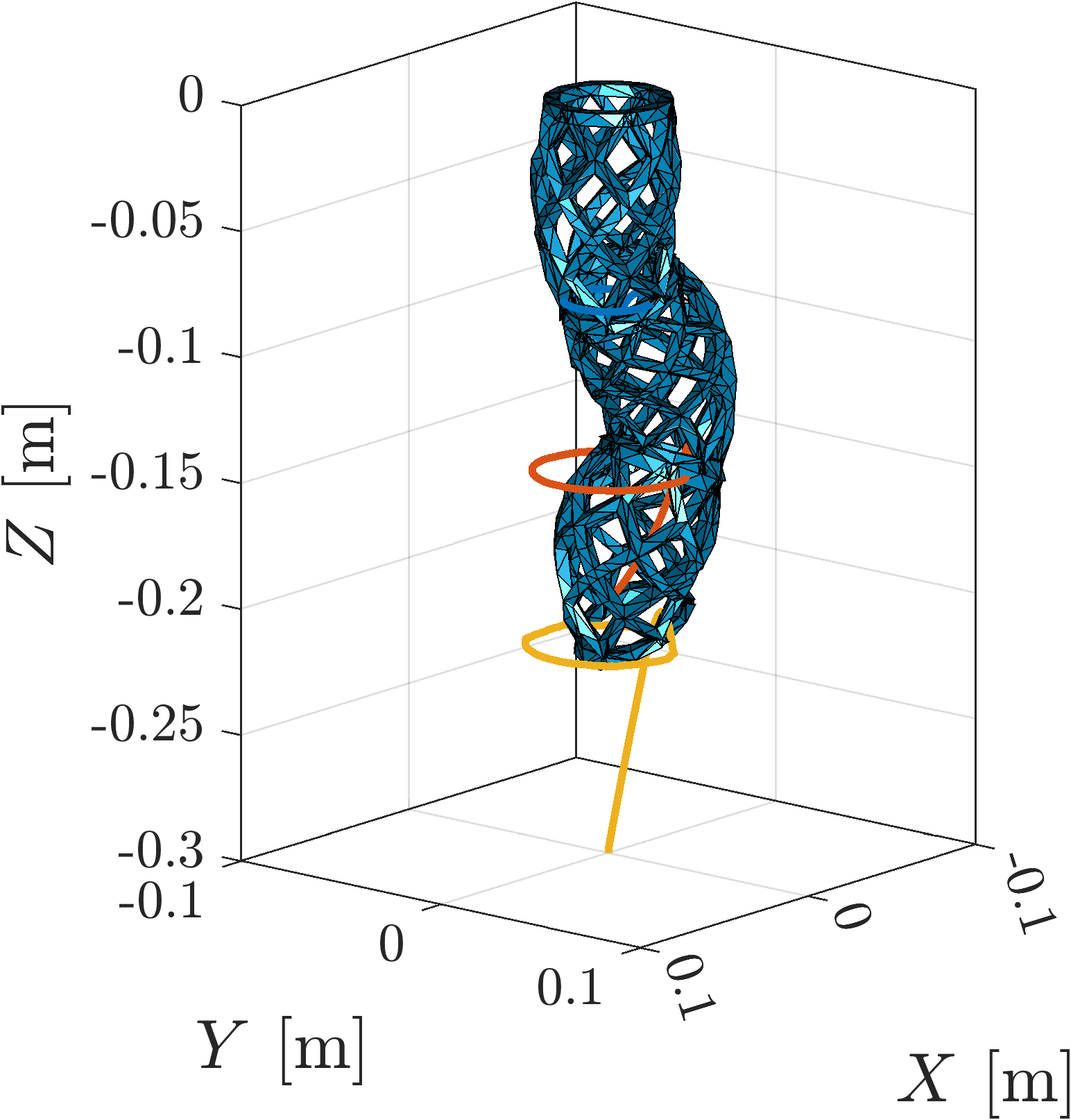}
    }
    \subfigure[{$t = 5.5~[\si{\second}]$}]{
        \includegraphics[width = 0.23\textwidth]{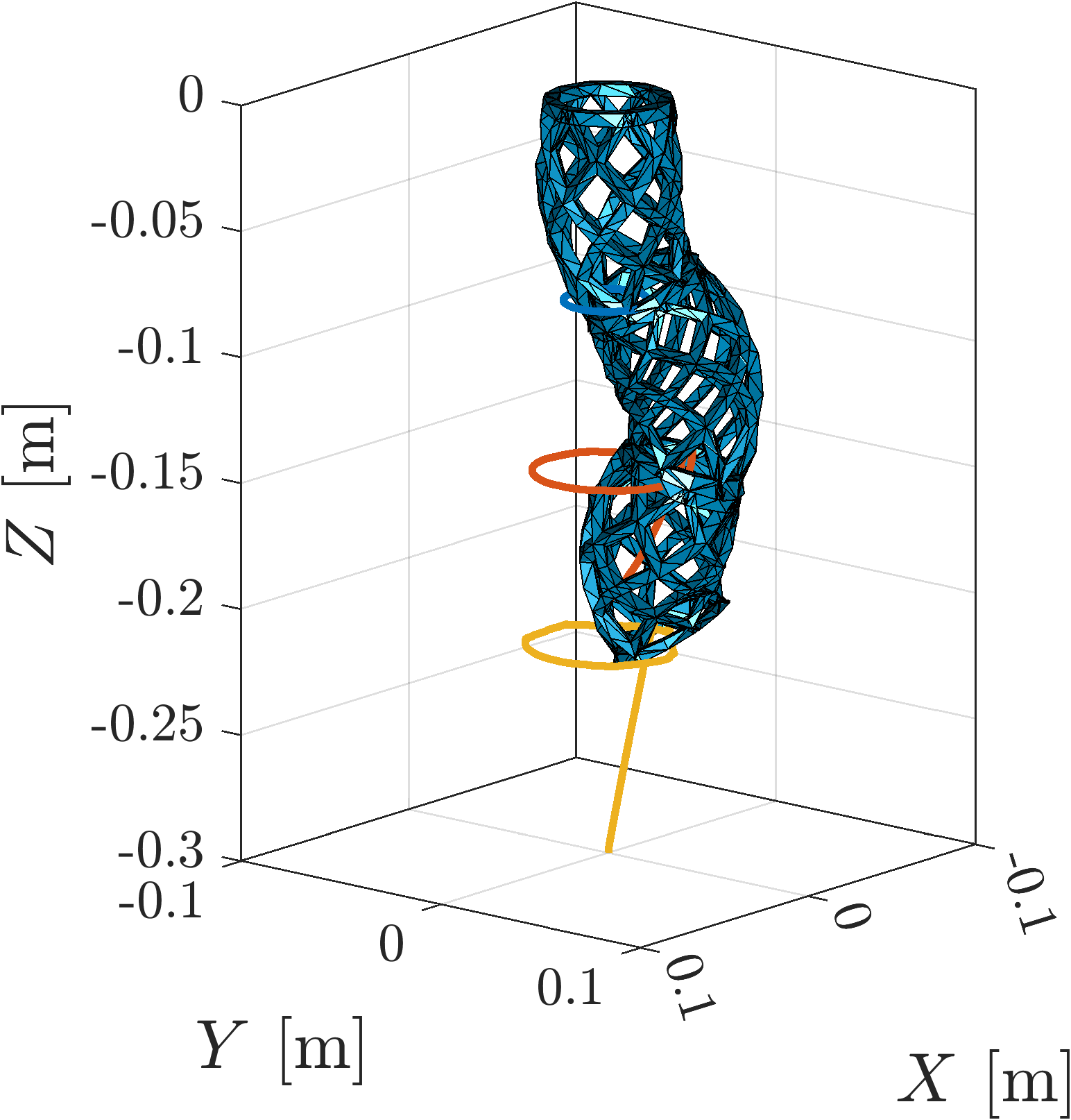}
    }
    \caption{\small Simulation 1. Snapshots of the robot motion during execution of the control task.}
    \label{fig:sim1:snapshot}
\end{figure*}
Given the highly underactuation, the derivation of a Cartesian controller with proven stability guarantees is not straightforward and not addressed yet in the control literature. We thus solve the control problem in two steps. First, the Cartesian trajectory $\tv_{d} = \rb{ \tv_{1_{d}}^{T} \,\, \tv_{2_{d}}^{T} \,\, \tv_{3_{d}}^{T}  }^{T}$ is sampled and used to solve a static problem, which provides the unique equilibrium compatible with the unactuated dynamics for the given reference. Remarkably, the ID procedure offers a mean of efficiently solving also the statics since, in a robot equilibrium, the following holds
\begin{equation}\label{eq:sim1:statics}
   \gv(\qv_{eq}) + \sv(\qv_{eq}, \zerov) + \Am(\qv_{eq})\uv_{eq}(\qv_{eq}) = \zerov_{n},
\end{equation}
where $\uv_{eq}(\qv_{eq})$ is the actuation force -- here obtained from a Cartesian regulator -- and $\qv_{eq}$ the unknown equilibrium. Specifically, the above equation can be solved numerically with a Newton-Raphson routine calling iteratively Algorithm~\ref{alg:Kane inverse recursive}. The equilibria, obtained by solving~\eqref{eq:sim1:statics} at the sampled time instants, are then interpolated into a twice differential trajectory and converted into a smooth reference $\yv_{d}(t)$ for the actuation coordinates (the tendons elongation inside the robot), which are then commanded through a partial feedback linearization (PFBL) on the collocated dynamics~\citep{pustina2024input}. Thanks to the stability of the zero dynamics~\citep{pustina2022feedback}, the controller guarantees proven global exponential stability and takes the form
\begin{equation*}
\left\{ 
\begin{split}
    \uv &= (\Mm^{\thetav}_{aa}-\Mm^{\thetav}_{au}\Mm{^{\thetav}_{uu}}^{-1}\Mm^{\thetav}_{ua})\av + \cv^{\thetav}_{a}\\
    &\quad\quad- \Mm^{\thetav}_{au}\Mm{^{\thetav}_{uu}}^{-1}\left(\cv^{\thetav}_{u} + \gv^{\thetav}_{u} + \sv^{\thetav}_{u} \right),\\
    \av &= \Ddot{\yv}_{d} + \Km_{D} ( \dot{\yv}_{d} - \dyv ) + \Km_{P} (\yv_{d} - \yv),
\end{split}
\right.
\end{equation*}
where the system dynamics -- expressed through the superscript $(\cdot)^{\thetav}$ in the decoupling coordinates $\thetav = (\yv^{T} \,\, \zv^{T})$ with $\zv \in \R^{n-m}$ a complement to $\yv$ -- has been partitioned in actuated and unactuated part, and $\Km_{P} = 10 \cdot \boldsymbol{\mathbb{I}}_{9}~[\si{\newton}]$ and $\Km_{D} = 1 \cdot \boldsymbol{\mathbb{I}}_{9}~[\si{\newton \second}]$ are the control gains. The dynamic terms needed for the computation of the control action can be obtained with a single call to the MID as $\textnormal{MID}(\qv(\thetav), \dqv(\thetav, \dthetav), \zerov_{n})$. Note the appearance of $\Mm$ in the above control law and so the utility of having a ID procedure that allows the computation of $\Mm$ as well. Figs.~\ref{sim1:q} and~\ref{sim1:yu} show the time evolution of the configuration variables, and of the actuation coordinates and of the control action, respectively. As expected, $\yv$ converges exponentially fast to $\yv_{d}$. In Fig.~\ref{fig:sim1:snapshot}, we illustrate snapshots of the robot motion in its workspace during the execution of the task.

\subsection{Simulation 2. Hybrid rigid-soft robotic arm}\label{sim:simulation 2}
\begin{figure}[th]
    \centering
    \subfigure[]{
        \includegraphics[width=0.9\columnwidth, trim={0cm 0cm 0cm 0cm},clip]{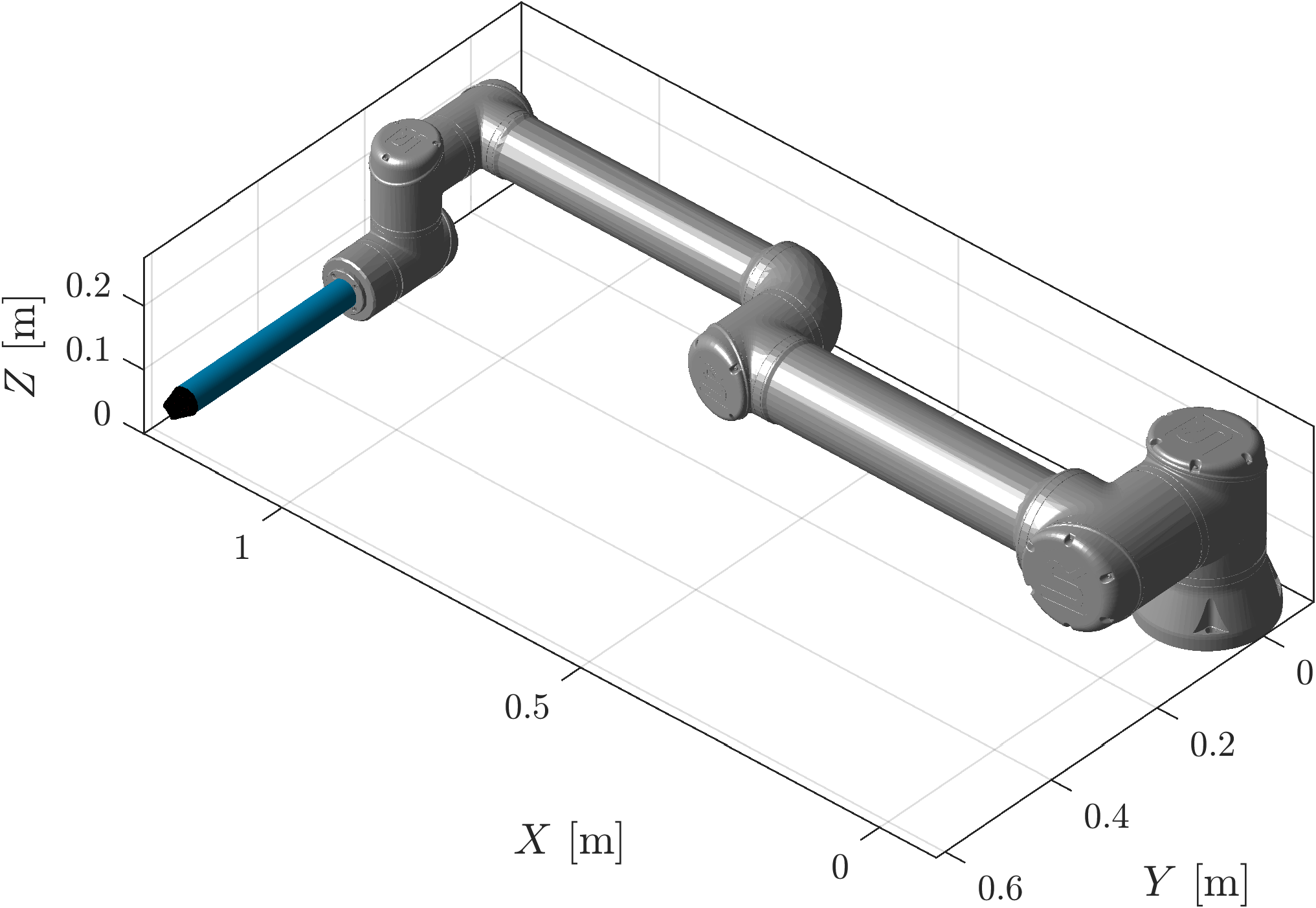}
    }\\
    \subfigure[]{
        \includegraphics[width=0.55\columnwidth, trim={0cm 0cm 0cm 0cm},clip]{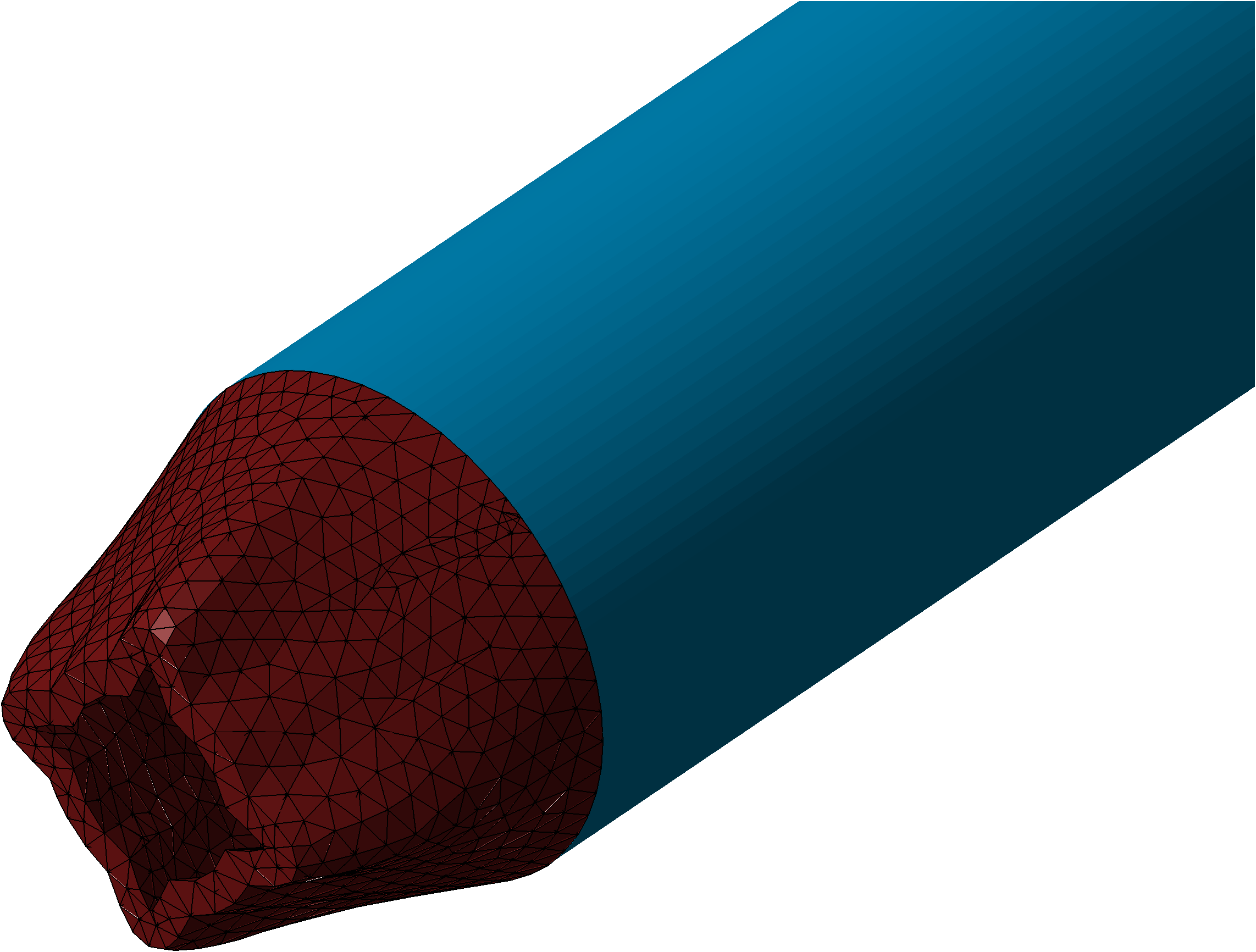}
    }
    \caption{\small Simulation 2. Hybrid rigid-soft robot consisting of a Universal Robot UR10e (in gray), a slender soft body modeled as a discrete Cosserat continuum (in blue) and a soft gripper modeled with LVP primitives (in red and zoomed in panel (b)).}
    \label{fig:simulation8:2:stress_free}
\end{figure}
In this simulation, we demonstrate how the proposed approach can handle robots with rigid and soft bodies, each possibly modeled using different kinematics. We consider a hybrid manipulator including a Universal Robot UR10e, a slender continuum soft arm modeled as a Cosserat rod with the GVS approach~\citep{boyer2020dynamics}, and a soft gripper modeled as a 3D continuum with LVP primitives. The setup\review{, which consists of $N = 8$ bodies,} is illustrated in Fig.~\ref{fig:simulation8:2:stress_free}. The gravitational force acts along the negative direction of the $z$-axis. The soft arm has radius $0.02~[\si{\meter}]$, rest length $0.3~[\si{\meter}]$, mass density $1070~[\si{\kilogram\per\cubic\meter}]$. The material elastic and damping coefficients are $0.17857~[\si{\mega\pascal}]$ and $0.1~[\si{\second}]$, respectively. Three tendons displaced by an angle of $120\si{\degree}$ and at a distance from the center backbone of $0.015~[\si{\meter}]$ actuate the Cosserat segment.  For the sake of simplicity, its backbone strain is modeled under the piecewise constant curvature hypothesis with elongation, thus resulting in three DoFs for this body. The effect of actuation is modeled as in~\cite{renda2022geometrically}. On the other hand, the soft gripper has mass density $960~[\si{\kilogram\per\cubic\meter}]$, elastic coefficient $22.2~[\si{\mega\pascal}]$ and damping factor $0.05~[\si{\second}]$. The gripper kinematics is modeled with two LVP primitives: an elongation primitive and a cavity primitive. The first accounts for the compression when the gripper closes and the second for the radial change due to inflation of the inner air chamber. The mode functions used by these primitives are reported in Table~\ref{tab:simulation2:modal_expansion}.
\begin{table}[t!]
\caption{\small Modal expansion of the body primitives for the soft gripper of Simulation 2.}
\centering
\begin{tabular}{p{0.5\columnwidth}p{0.4\columnwidth}}
\hline
\hline
Deformation                   & Modal expansion              \\ \hline
Stretch and compression       & $(x_{3}/L)^2\review{q_{10}}$       \\
Source       & $(x_{3}/L)^2\review{q_{11}}$       \\ \hline\hline
\end{tabular}
\label{tab:simulation2:modal_expansion}
\end{table}
Because the gripper is the last body in the chain, Assumption~\ref{assumption:contact area} can be relaxed to allow deformation of the top area. Thus, we allow for the mode functions of both primitives to be non zero at the body distal end, i.e., when $x_{3} = L$. The gripper is actuated by applying a pressure difference with respect to the atmospheric one. This input is projected into the configuration space using the top and lateral surface area of the pressurized chamber.

\begin{figure}
    \centering
    \subfigure[]{
        \includegraphics[width = 0.86\columnwidth]{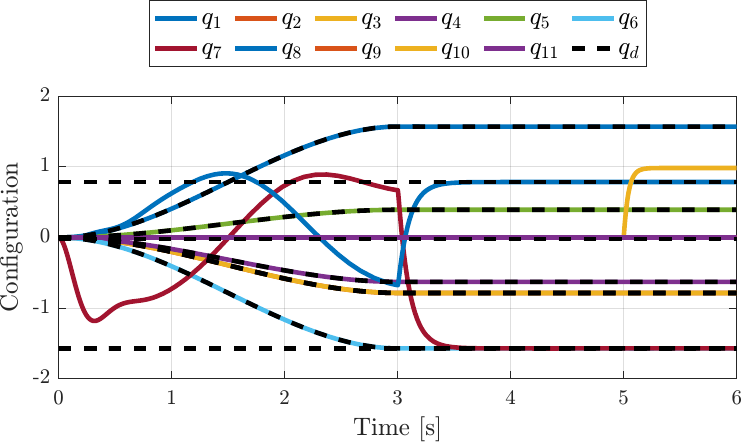}
    }\\
    \subfigure[]{
        \includegraphics[width = 0.94\columnwidth]{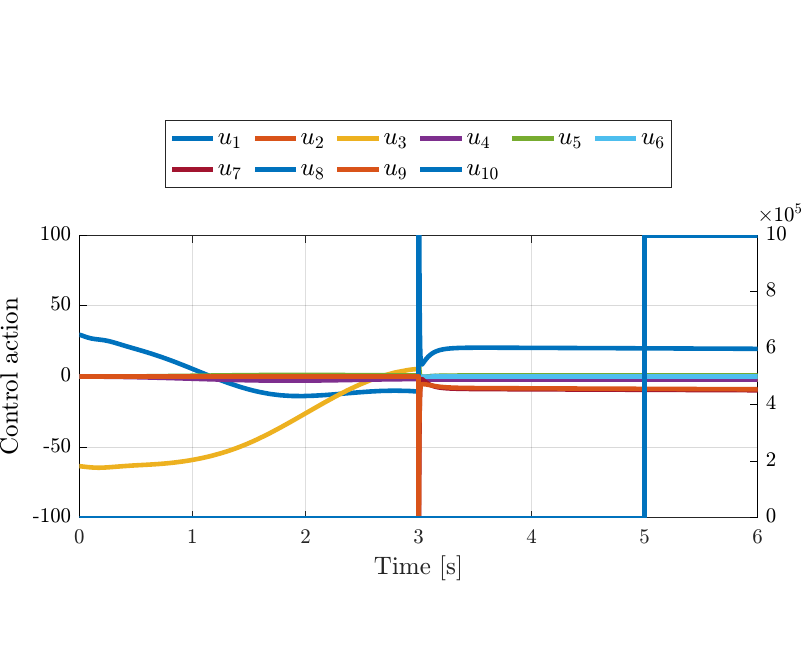}
    }
    \caption{\small Simulation 2. Time evolution of the (a) configuration variables (in $[\si{\radian}]$ and $[\si{\meter}]$) and (b) system input (in $[\si{\newton\meter}]$, $[\si{\newton}]$ and $[\si{\pascal}]$). In panel (a), the reference for the controlled configuration variables is illustrated with black dashed lines.}
    \label{fig:sim2:time evolutions}
\end{figure}

\begin{figure*}
    \centering
    \subfigure[{$t = 0~[\si{\second}]$}]{
        \includegraphics[width = 0.23\textwidth]{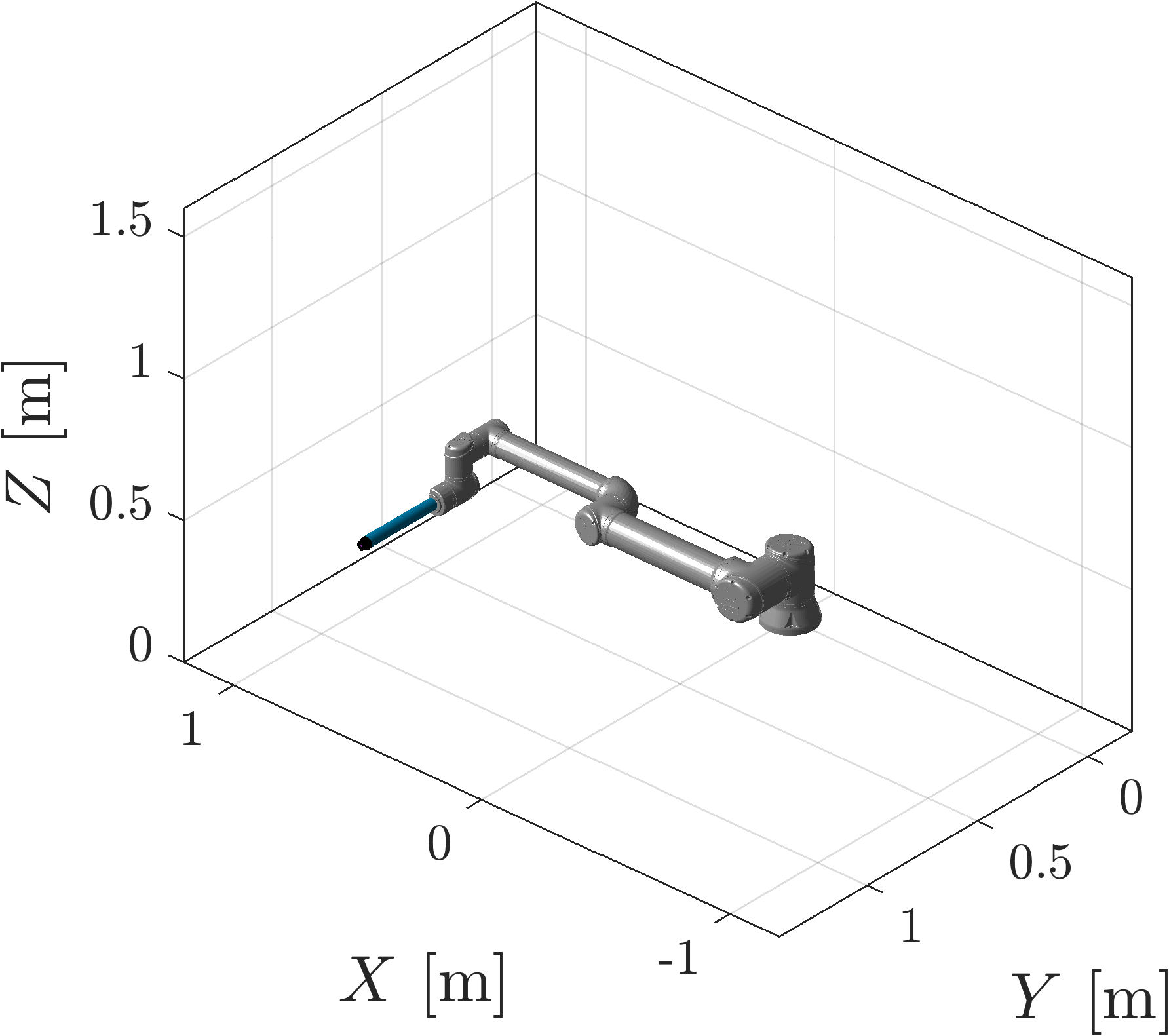}
    }
    \subfigure[{$t = 0.5~[\si{\second}]$}]{
        \includegraphics[width = 0.23\textwidth]{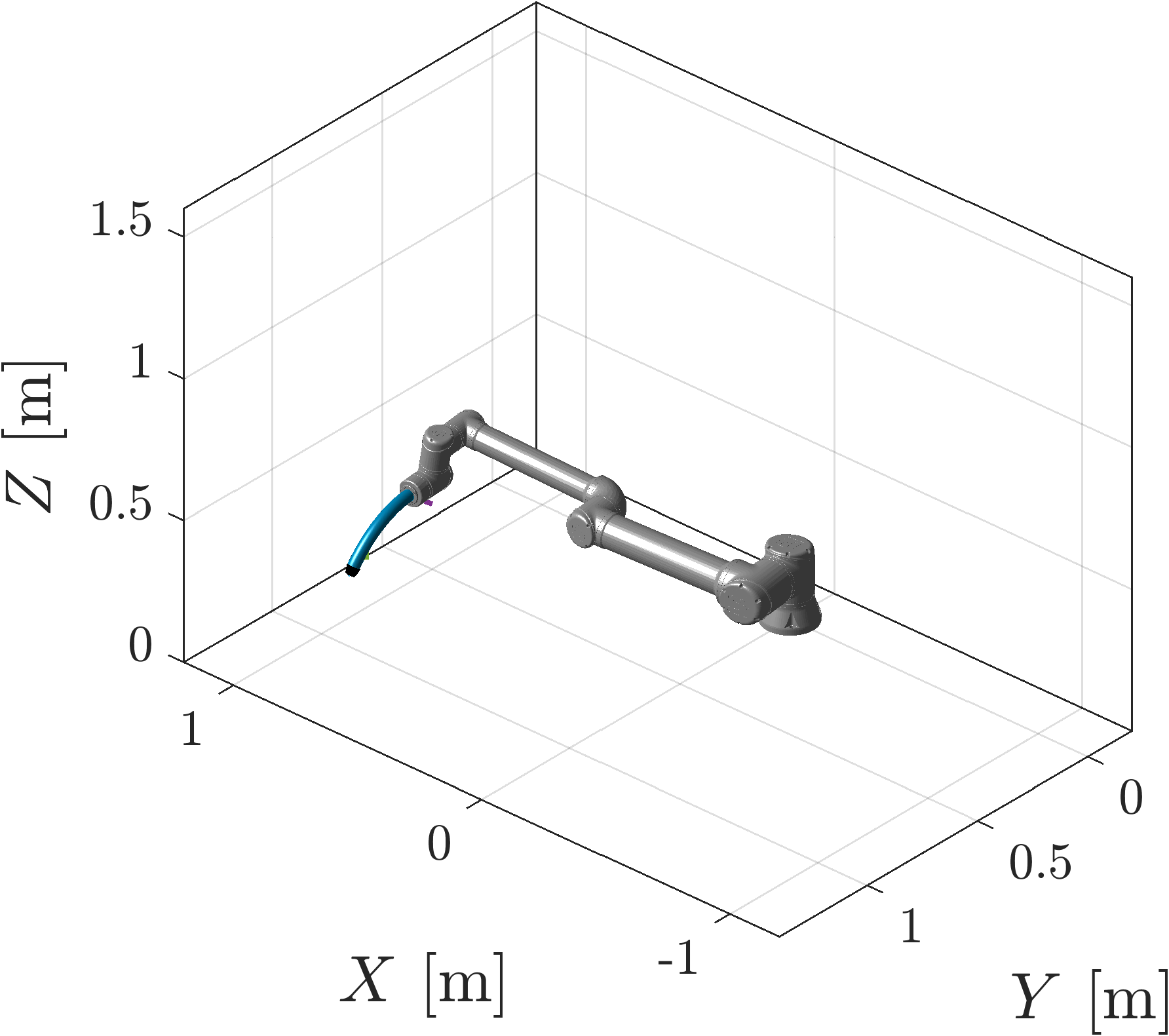}
    }
    \subfigure[{$t = 1~[\si{\second}]$}]{
        \includegraphics[width = 0.23\textwidth]{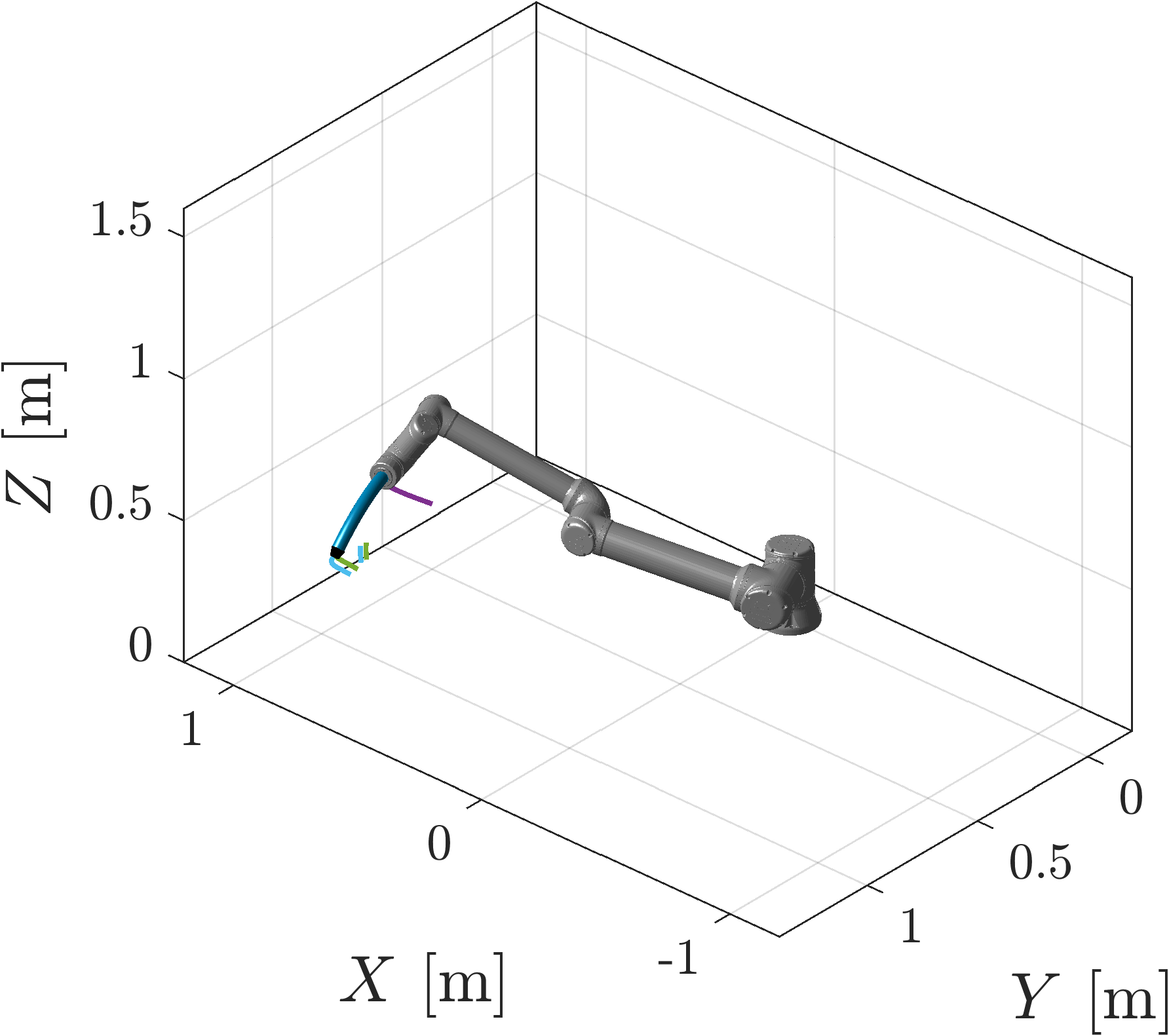}
    }
    \subfigure[{$t = 1.5~[\si{\second}]$}]{
        \includegraphics[width = 0.23\textwidth]{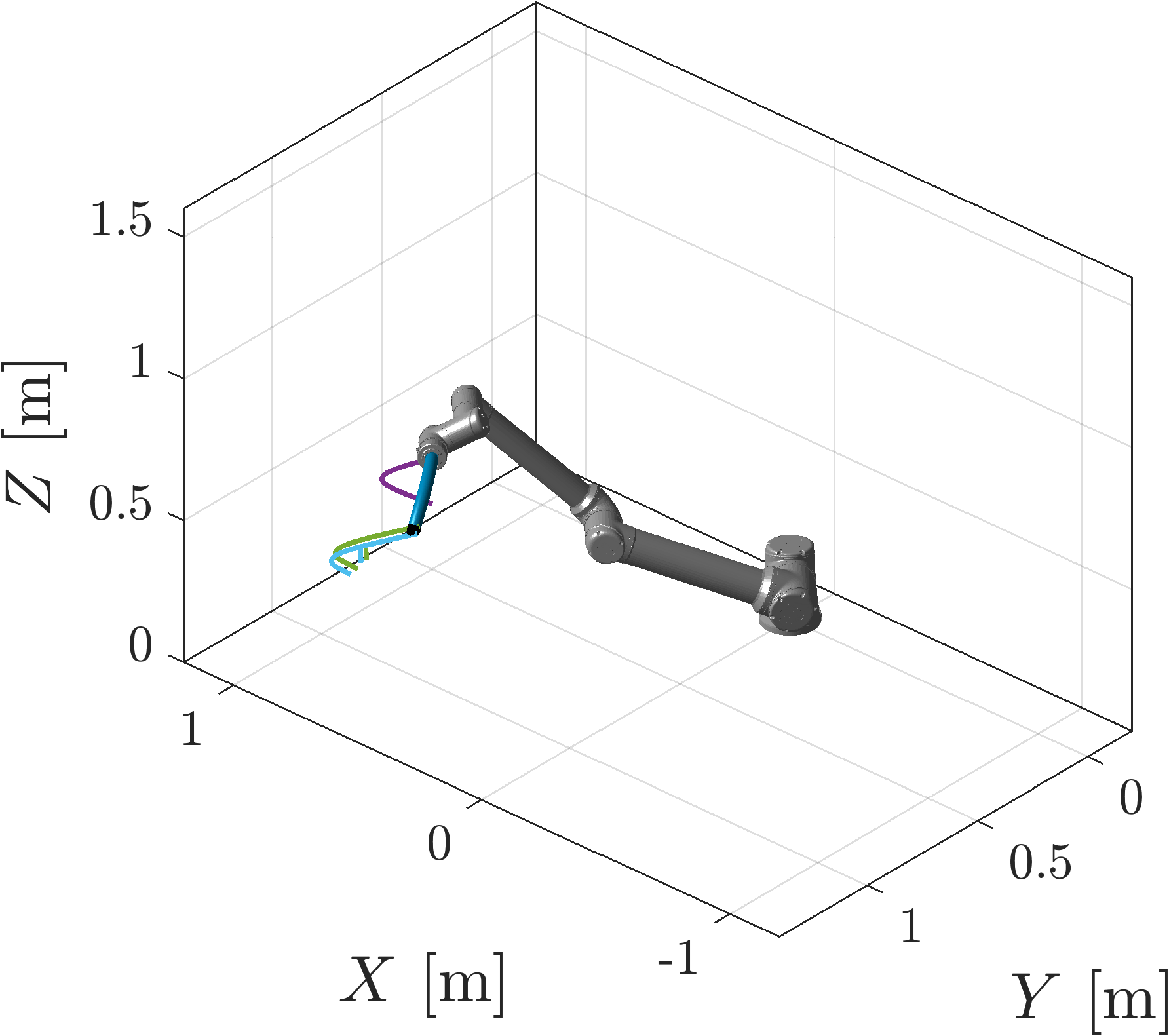}
    }
    \subfigure[{$t = 2~[\si{\second}]$}]{
        \includegraphics[width = 0.23\textwidth]{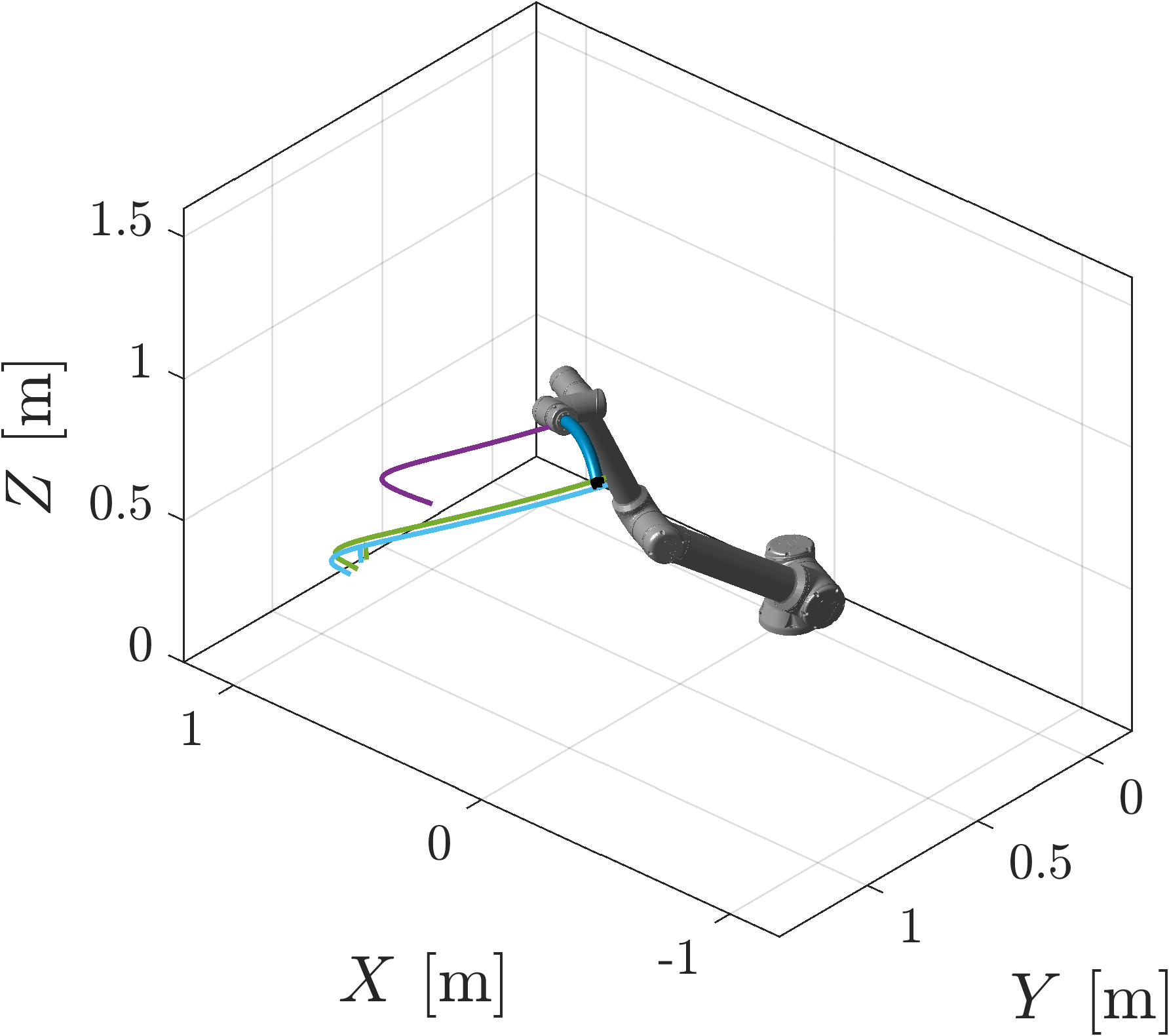}
    }
    \subfigure[{$t = 2.5~[\si{\second}]$}]{
        \includegraphics[width = 0.23\textwidth]{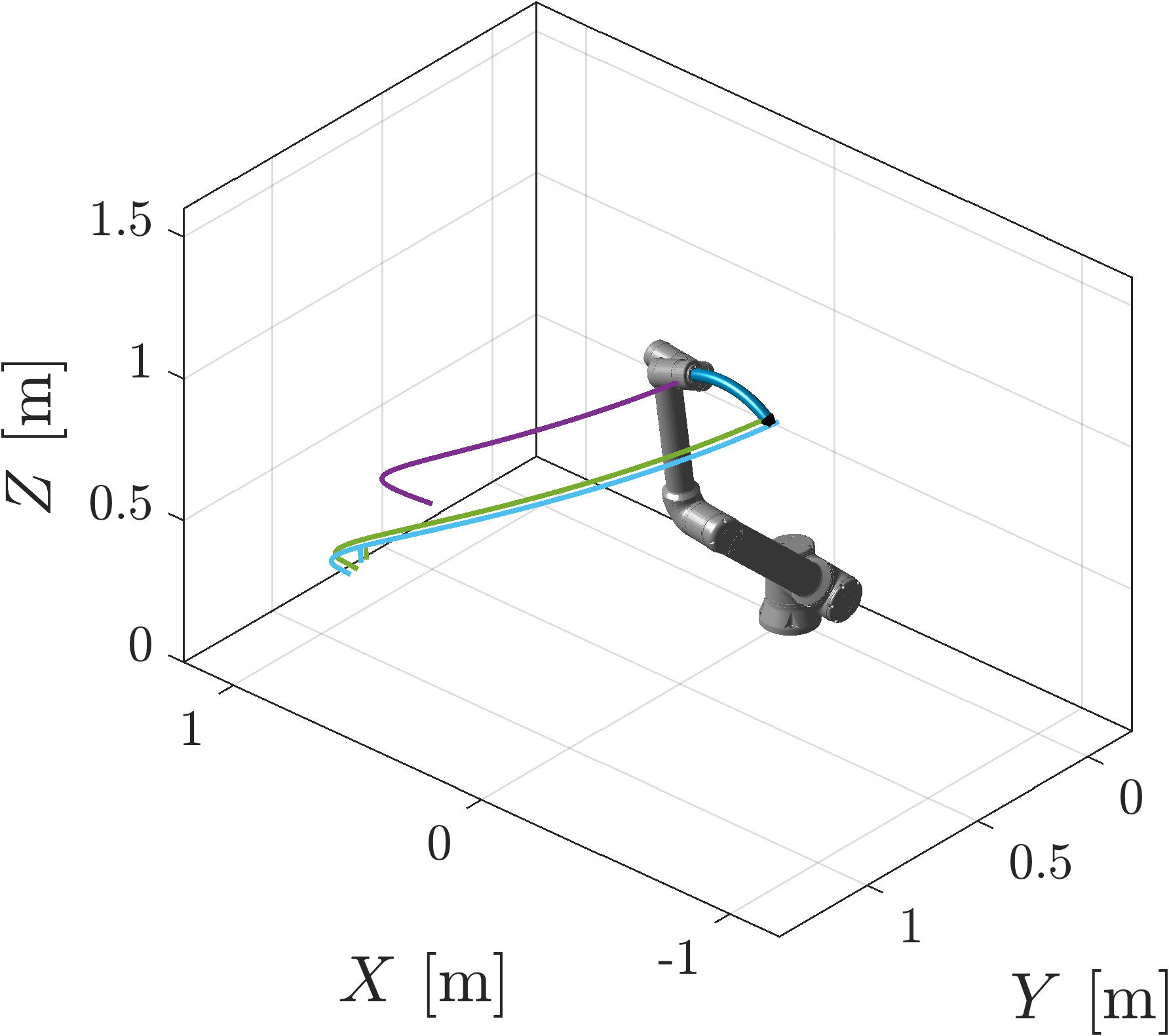}
    }
    \subfigure[{$t = 3.1~[\si{\second}]$}]{
        \includegraphics[width = 0.23\textwidth]{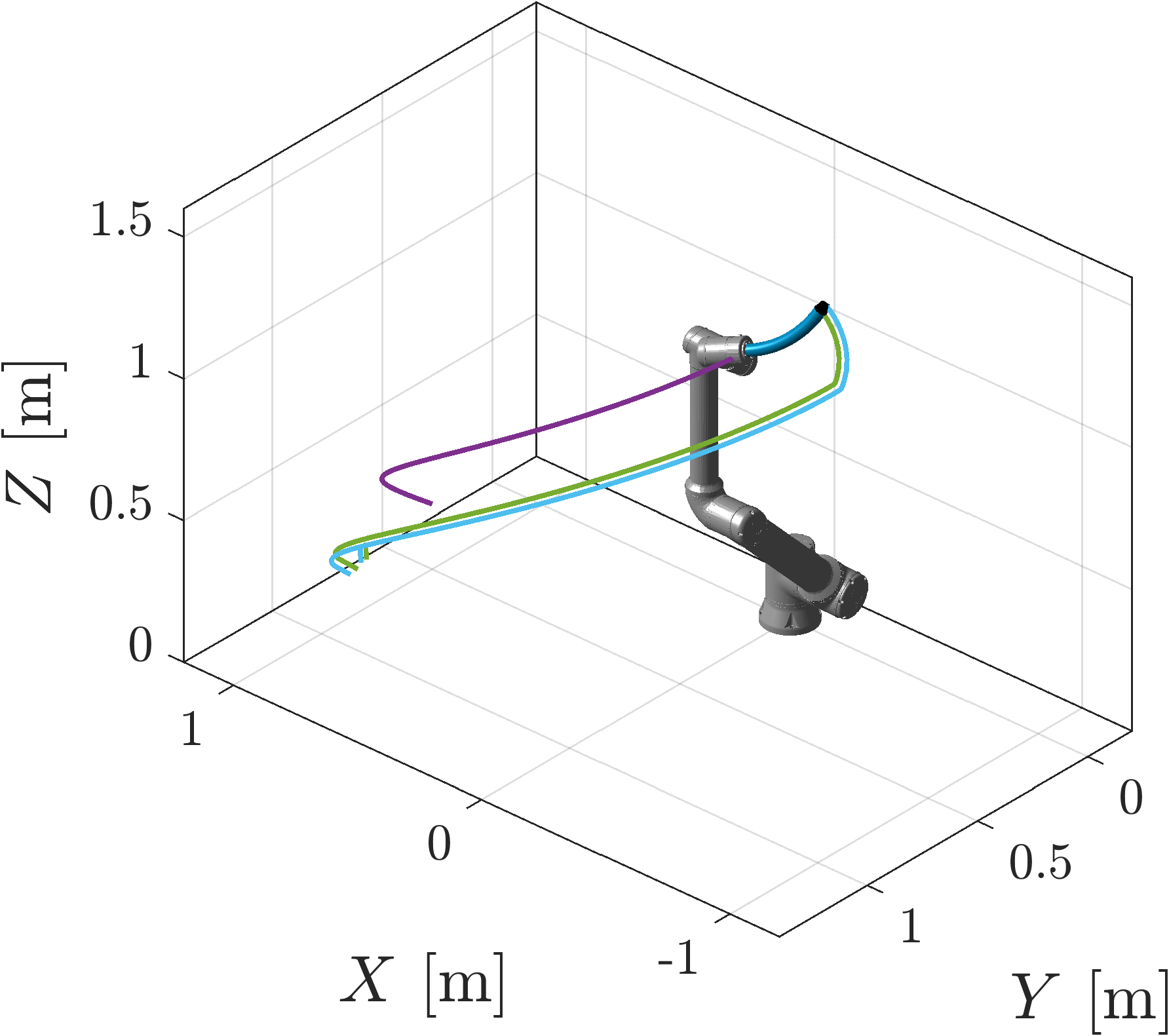}
    }
    \subfigure[{$t = 3.15~[\si{\second}]$}]{
        \includegraphics[width = 0.23\textwidth]{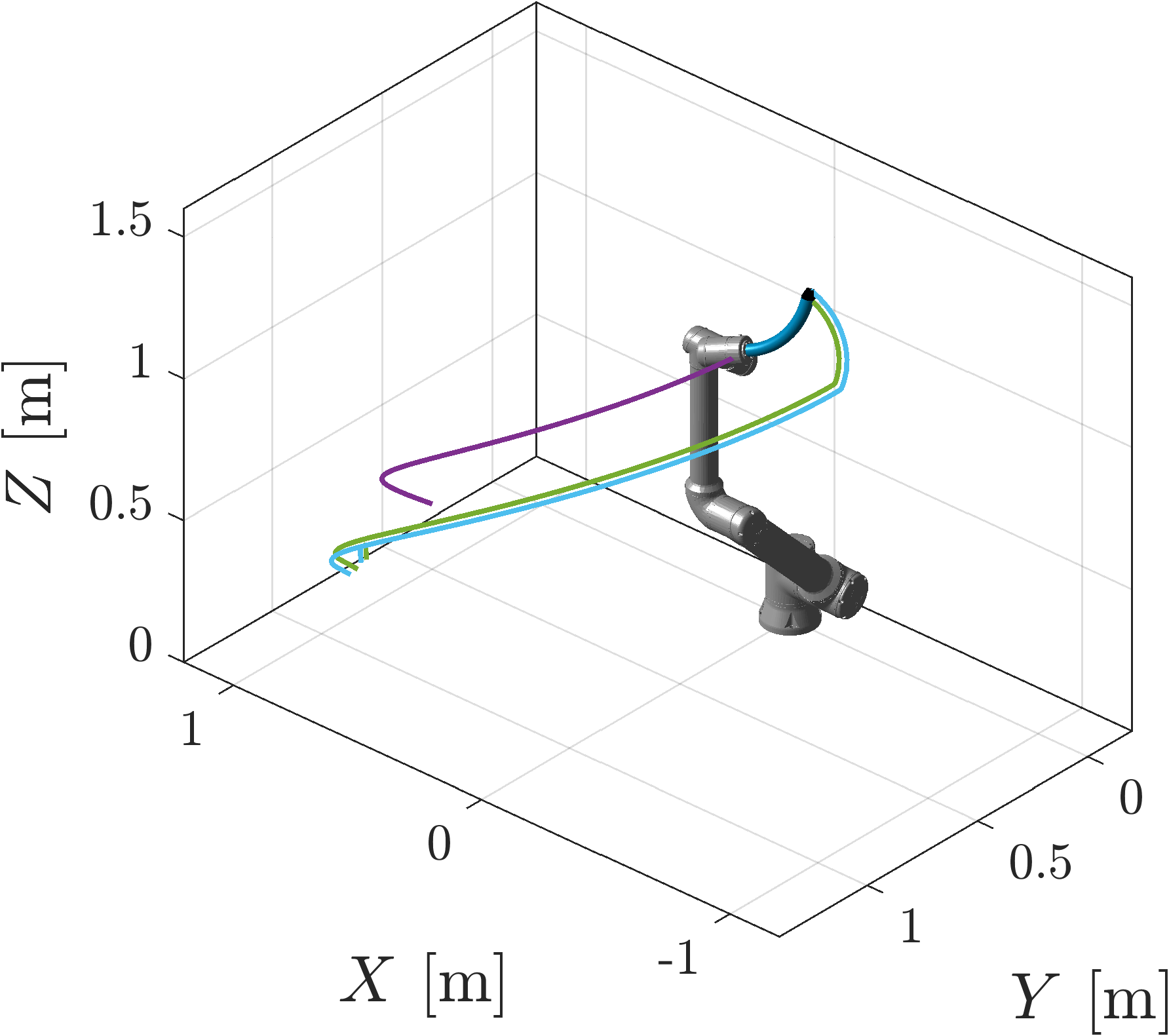}
    }
    \subfigure[{$t = 3.2~[\si{\second}]$}]{
        \includegraphics[width = 0.23\textwidth]{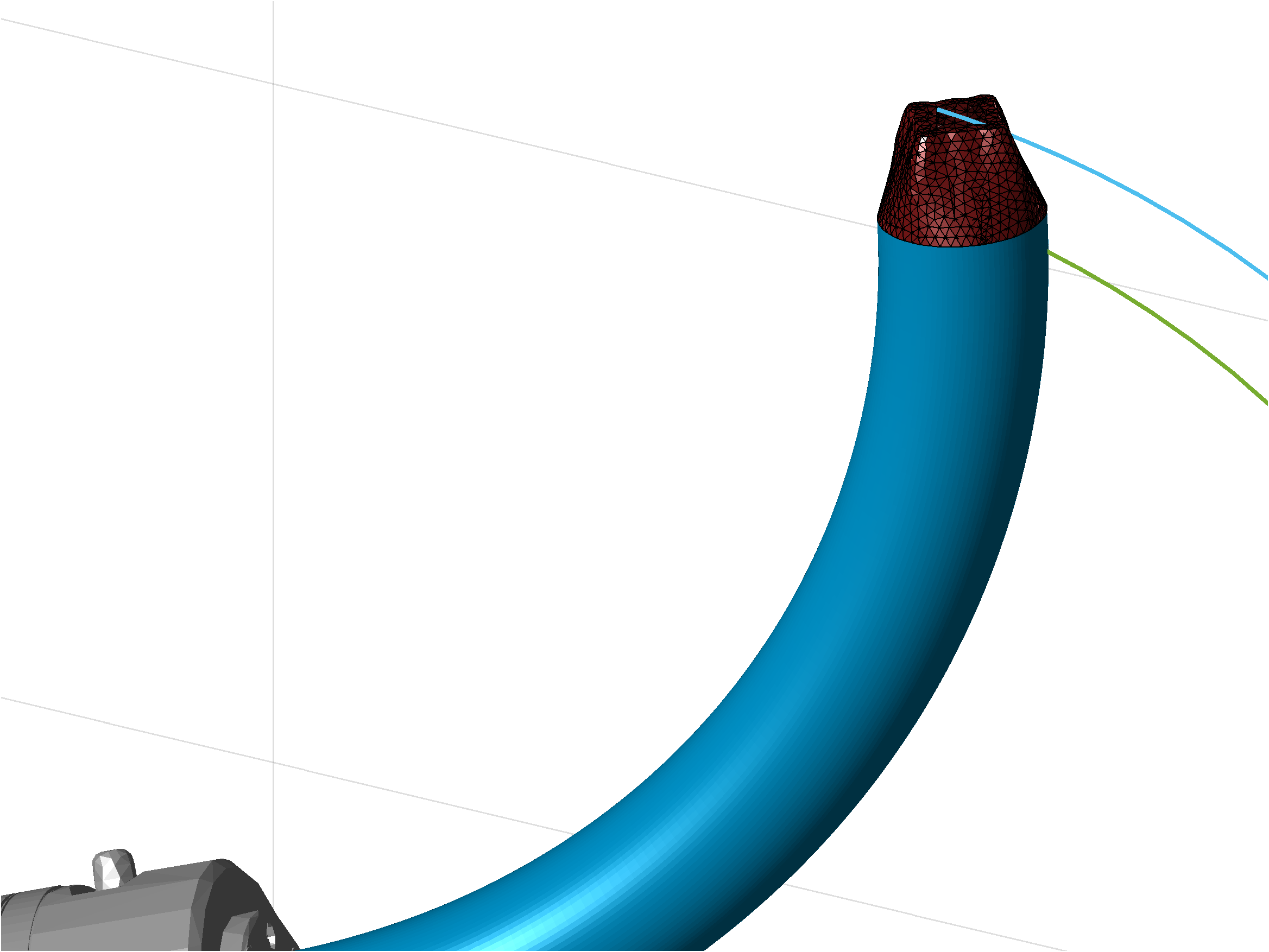}
    }
    \subfigure[{$t = 5~[\si{\second}]$}]{
        \includegraphics[width = 0.23\textwidth]{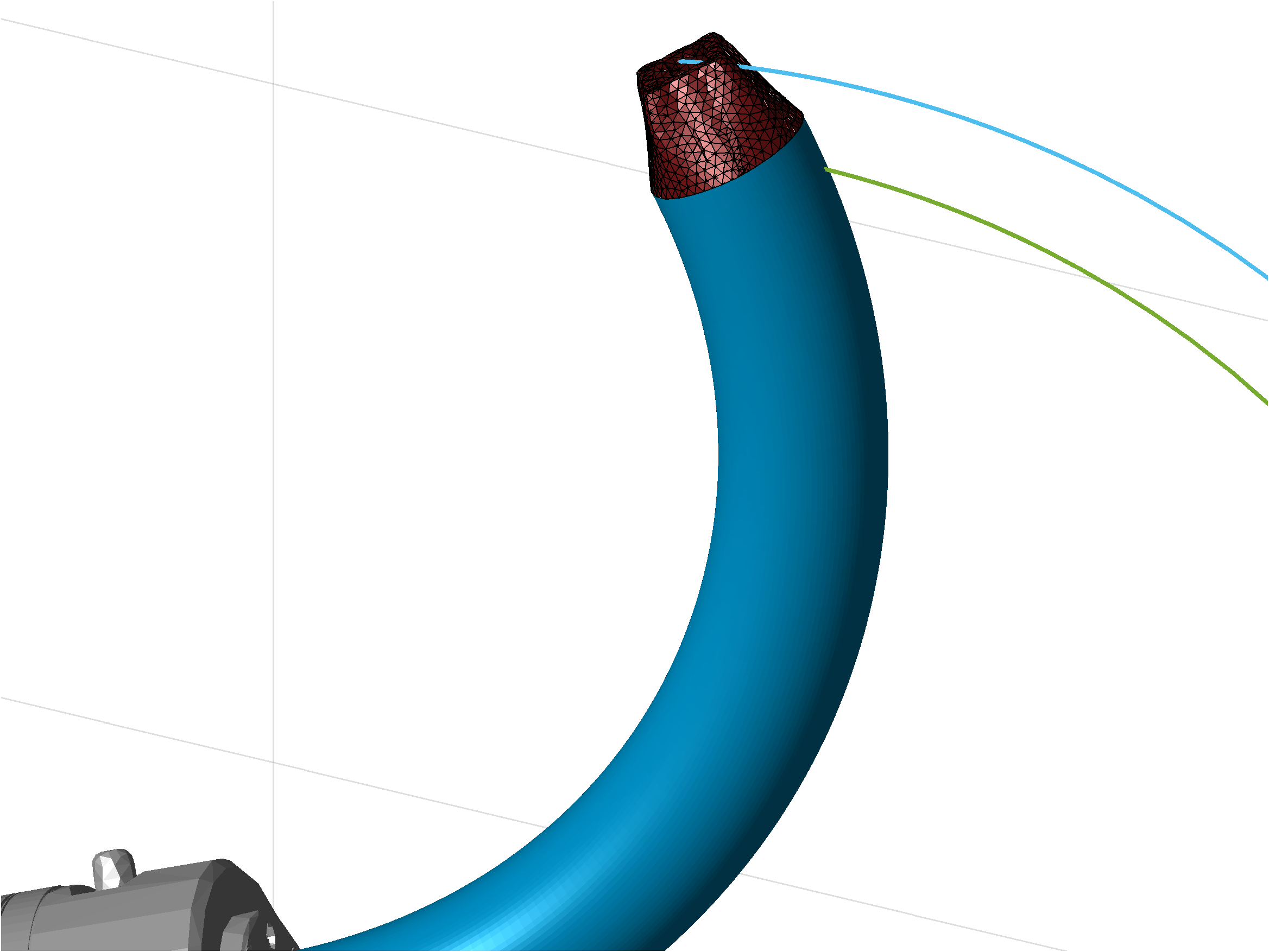}
    }
    \subfigure[{$t = 5.1~[\si{\second}]$}]{
        \includegraphics[width = 0.23\textwidth]{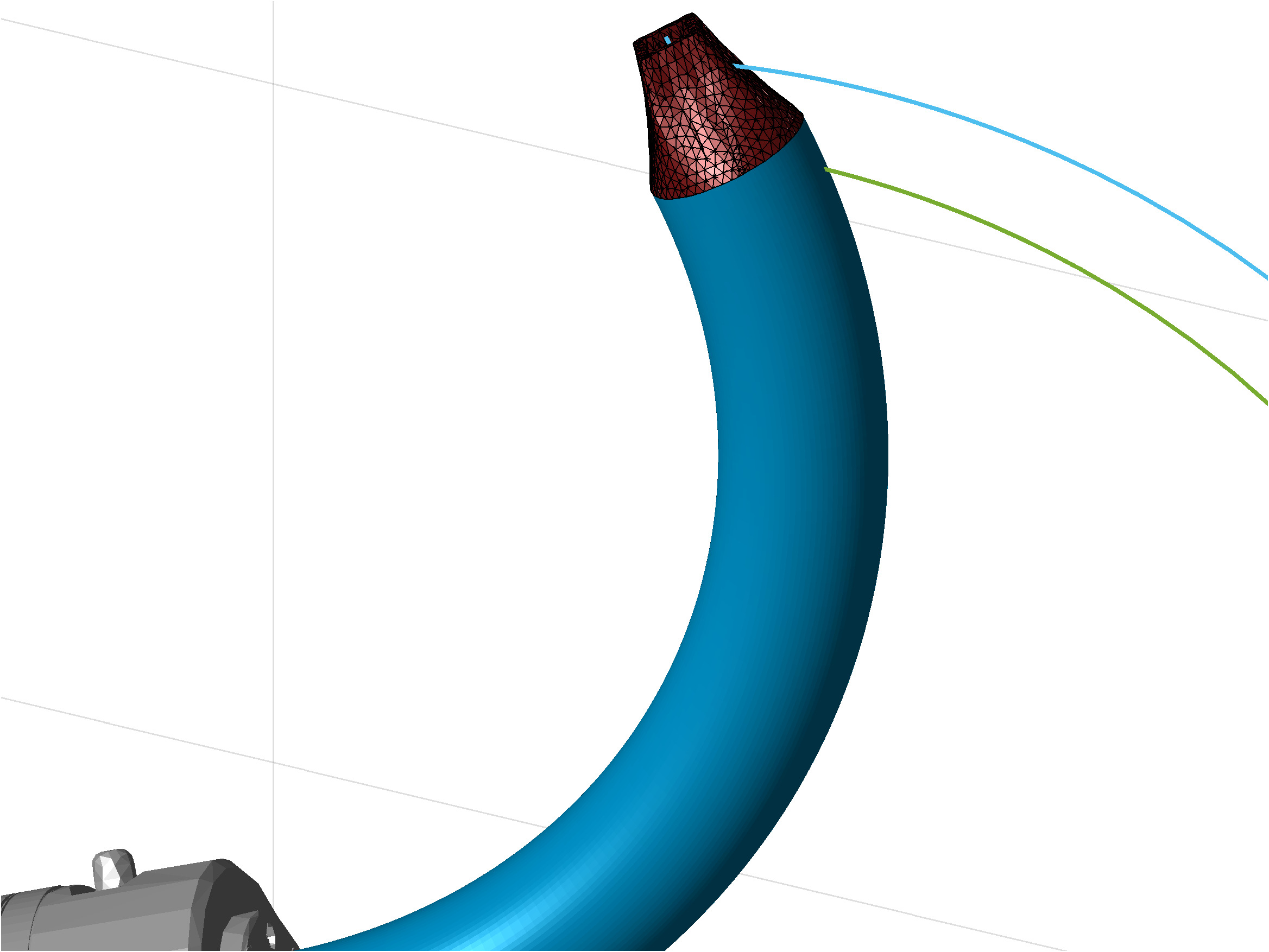}
    }
    \subfigure[{$t = 5.2~[\si{\second}]$}]{
        \includegraphics[width = 0.23\textwidth]{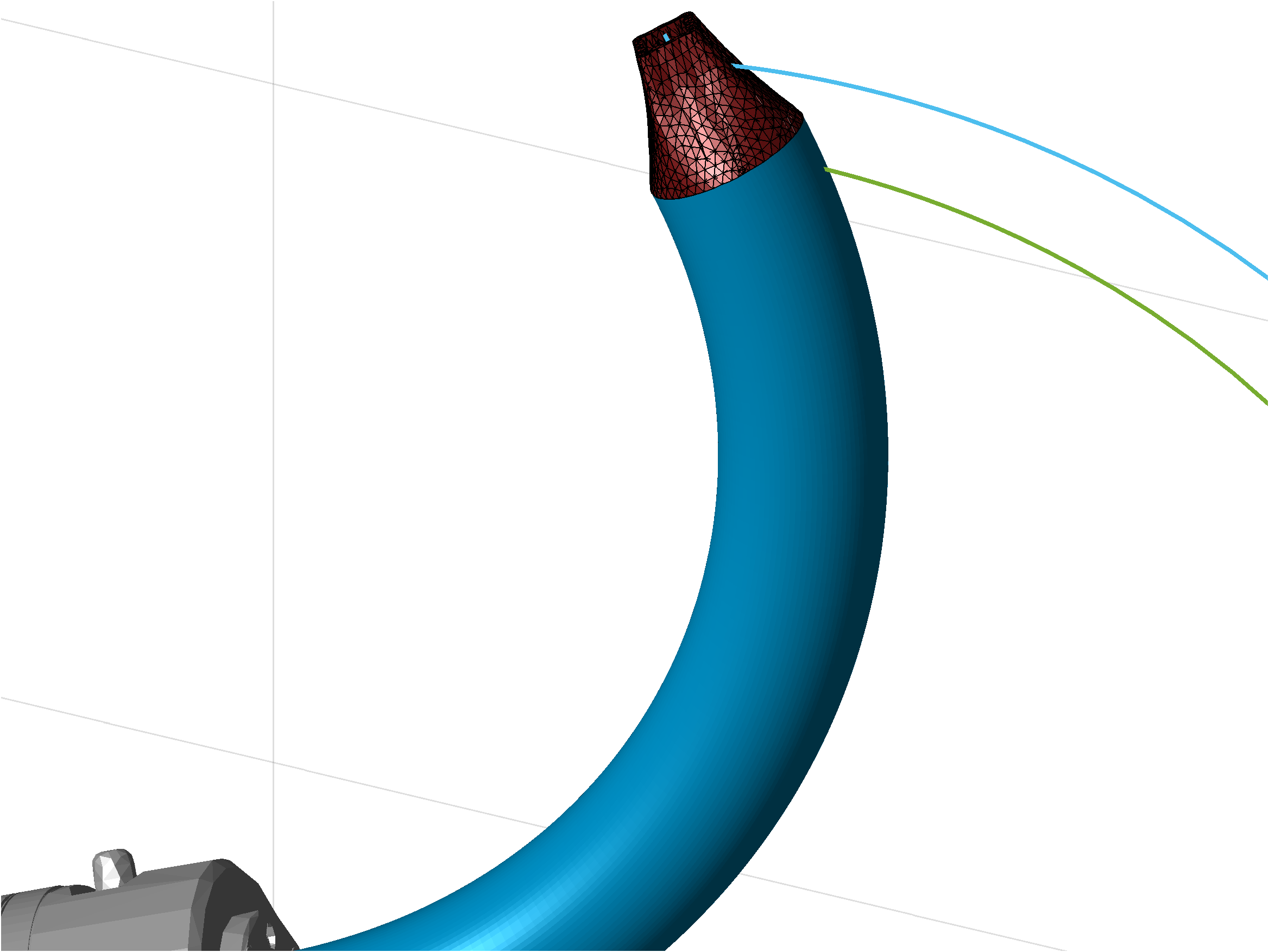}
    }
    \caption{\small Simulation 2. Snapshots of the robot motion during execution of the three control phases.}
    \label{fig:sim2:snapshot}
\end{figure*}

The control task involves using both rigid and deformable bodies to navigate the environment and grasp an object with the end-effector. This task is divided into three sub-tasks: (i) track a rest-to-rest cubic joint trajectory for the UR10e between
\begin{equation*}
    \qv_{i_{\mathrm{UR10e}}} = \begin{carray}{cccccc}
        0 & 0 & 0 & 0 & 0 & 0
    \end{carray}^T, 
\end{equation*}
and
\begin{equation*}
    \qv_{f_{\mathrm{UR10e}}} = \begin{carray}{cccccc}
        \pi/2 & -\pi/4 & -\pi/4 & -\pi/5 & \pi/8 & -\pi/2
    \end{carray}^T, 
\end{equation*}
while keeping the soft bodies passive, (ii) perform a swing-up motion with the soft segment to the final configuration
\begin{equation*}
    \qv_{f_{\mathrm{GVS}}} = \begin{carray}{ccc}
        -\pi/2 & \pi/4 & -0.02
    \end{carray}^T,
\end{equation*}
and (iii) close the gripper. Each task is executed sequentially within three time windows $t_{1} \in [0, 3)~[\si{\second}]$, $t_{2} \in [3, 5)~[\si{\second}]$ and $t_{3} \in [5, 6]~[\si{\second}]$. Sub-task (i) is solved with a PFBL similar to that of Simulation 1. However, in this case, the controller can be implemented directly into the configuration space since the actuation coordinates correspond to the joint angles. The control gains are chosen as $\Km_{P_{\mathrm{UR10e}}} = 100 \cdot \mathbb{I}_{6}~ [\si{\newton\meter}]$ and $\Km_{D_{\mathrm{UR10e}}} = 10\cdot \mathbb{I}_{6}~[\si{\newton\meter\second}]$. Sub-task (ii) is accomplished through a PD regulator with cancellation of gravitational and stress elastic forces, namely
\begin{equation*}
\begin{split}
    \uv_{\mathrm{GVS}} &= \Am_{\mathrm{GVS}}^{-1}\left[ \Km_{P_{\mathrm{GVS}}} \rb{\qv_{f_{\mathrm{GVS}}} - \qv_{\mathrm{GVS}}} - \Km_{D_{\mathrm{GVS}}}\dqv \right.\\
    &\quad\quad\quad\quad\quad\quad\left. + \gv_{\mathrm{GVS}}(\qv) + \sv_{\mathrm{GVS}}(\qv_{\mathrm{GVS}}, \zerov_{n_{\mathrm{GVS}}})   \right].
\end{split}
\end{equation*}
Also the model-based term of this controller can be implemented with a single call to the ID as $\mathrm{ID}(\qv, \zerov_{n}, \zerov_{n})$.
Finally, sub-task (iii) is solved by commanding a feedforward pressure of $2~[\si{\mega\pascal}]$. We do not use any feedback controller in this case since it would require the gripper state which cannot be easily measured through proprioceptive sensors. Fig.~\ref{fig:sim2:time evolutions} shows the time evolution of the (a) configuration variables and (b) control action (note the different scale in $[\si{\pascal}]$ for the gripper input). The two feedback controllers guarantee tracking of the desired references while keeping the closed loop system provably stable. During the execution of sub-task (i), namely when the soft bodies are not actuated, the system state and input remain bounded thanks to the stability of the zero dynamics. In Fig.~\ref{fig:sim2:snapshot}, it is possible to see some snapshots of the robot motion in its workspace together with the trajectories of the distal ends of the UR10e, soft arm and gripper. 

\subsection{Simulation 3}
In this last simulation, we compare the computation time needed to evaluate the inertial and centrifugal generalized forces, i.e., $\Mm(\qv)\ddqv$ and $\cv(\qv, \dqv)$, between our approach and the EL approach. The goal of this simulation is to demonstrate the scalability of a recursive formulation of the dynamics compared to an energetic one. We use models of planar Piecewise Constant Curvature (PCC) continuum soft robots because, within a small number of bodies, the EL equations can still be computed symbolically. \review{As shown through the toy example,} also the terms appearing in our algorithm can be computed exactly for such class of robots. This implies that the computation of the dynamics with the ID algorithm remains exact independent of the number of bodies. Note that we do not compare our method with specific implementations for discrete Cosserat rods~\citep{mathew2022sorosim, caasenbrood2024sorotoki} for two main reasons. First, the EL approach is the only method available for computing the ID for the class of soft robots considered in this paper. Second, a fair comparison would require using the same programming language and implementation by the same person, as performance can vary significantly due to these factors. 
In addition, we do not claim that our algorithms are more efficient than specific implementations for soft robot subclasses. On the other hand, as shown in the previous simulations, we can solve the ID problem not only for discrete Cosserat rods but also for more generic soft robot models.

By using Algorithm~\ref{alg:Kane inverse recursive general}, the inertial and centrifugal forces are computed with the following arguments
\begin{equation*}
    \Mm(\qv)\ddqv = \mathrm{IID}(\qv, \zerov_{n}, \ddqv),
\end{equation*}
and
\begin{equation*}
    \cv(\qv, \dqv) = \mathrm{IID}(\qv, \dqv, \zerov_{n}).
\end{equation*}
In contrast, the evaluation of the above terms using the EL formulation involves computing the Lagrangian symbolically with the moving frame algorithm. The inertial and centrifugal forces are then obtained through symbolic differentiation. These expressions are simplified and exported as MATLAB functions for numerical evaluation. 

\begin{figure*}[t]
    \centering
    \subfigure[{Necessary pre-computation}]{
    \includegraphics[width = 0.48\columnwidth, trim={0 0 0 1.0cm}, clip]{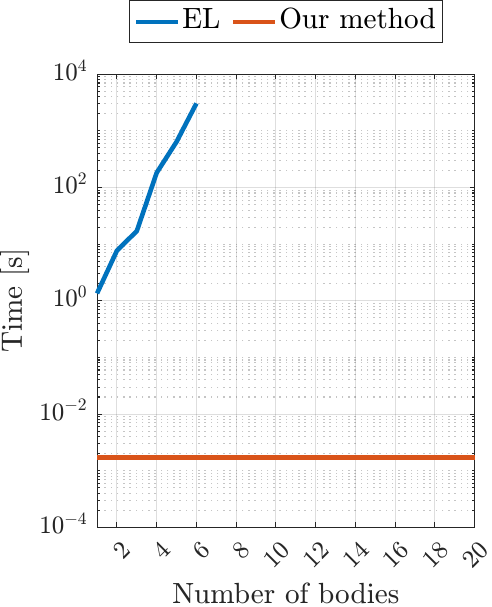}
    \label{simulation:planar comparison time model}
    }
    \subfigure[{Inertial force $\Mm(\qv)\ddqv$}]{
    \includegraphics[width = 0.48\columnwidth]{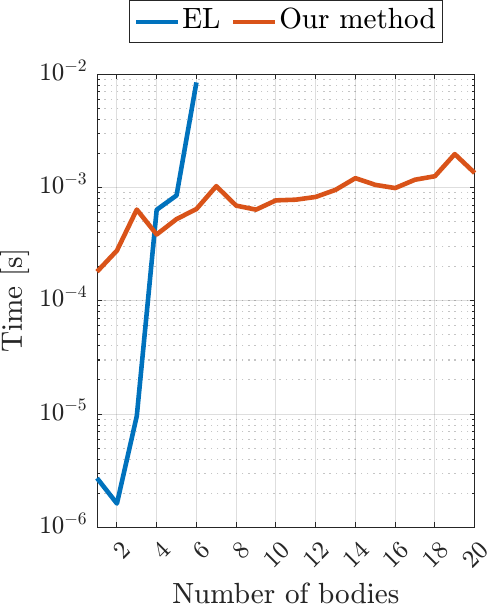}
    \label{simulation:planar comparison time mass}
    }
    \subfigure[{Centrifugal force $\cv(\qv, \dqv)$}]{
    \includegraphics[width = 0.48\columnwidth, trim={0 0 0 1cm}, clip]{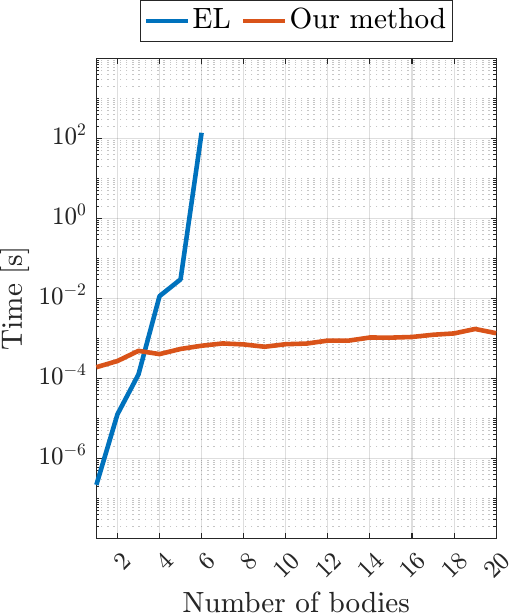}
    \label{simulation:planar comparison time apparent}
    }
    \subfigure[{ID relative difference norm}]{
    \includegraphics[width = 0.48\columnwidth, trim={0 0 0 0}, clip]{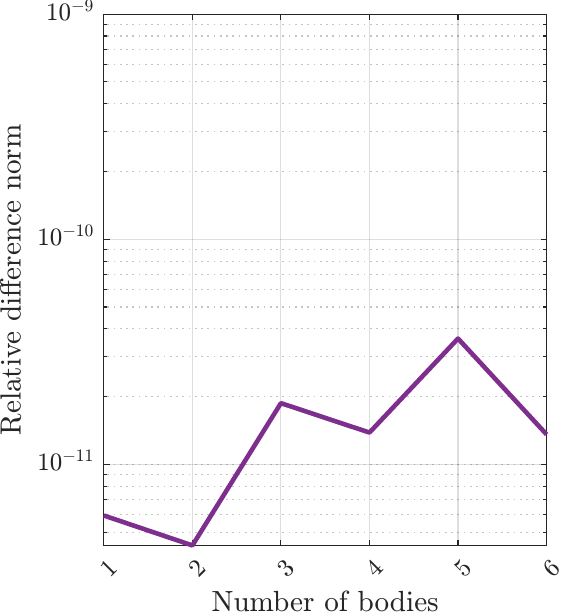}
    \label{simulation:planar comparison ID error}
    }
    \caption{\small Simulation 3. Comparison of ID using our method or using EL on planar soft robotic systems with increasing number of CC bodies. Time needed for (a) the computation of the model, and the evaluation of the (b) inertial and (c) centrifugal forces. Norm of the relative difference between the ID evaluated using the EL approach and our method (d). The EL approach is faster when few bodies are considered. However, its evaluation time increases quickly becoming soon impractical. This is because the functional expressions of $\Mm(\qv)\ddqv$ and $\cv(\qv, \dqv)$ become progressively more complex and cannot be evaluated recursively, thus requiring more resources both in terms of CPU time and memory.}
    \label{simulation:planar comparison time}
\end{figure*}

We compare the two approaches by measuring the time needed to build the model, and evaluate $\Mm(\qv)\ddqv$ and $\cv(\qv, \dqv)$ through multiple calls of the corresponding methods, subsequently calculating the median and standard deviation of these measurements. 
%
The bodies have radius $0.03~[\si{\meter}]$, rest length~$0.3~[\si{\meter}]$ and mass density~$1000~[\si{\kilogram\per\cubic\meter}]$.
Fig~\ref{simulation:planar comparison time model}--\ref{simulation:planar comparison time apparent} shows the simulation outcomes for unitary values of the configuration variables and their time derivatives. 
Using the EL equations, we could compute the model for only up to six bodies. After this number, the computation time exploded and it became impractical to obtain the dynamics. Instead, for our approach, the time to build the model remains below $\SI{4.52}{[\second]}$ up to twenty bodies (see Fig.~\ref{simulation:planar comparison time model}) as it is just the time needed to compute the functional expression of the integrals from the kinematics and allocate in memory the objects representing the bodies. When the number of bodies is small, the EL method is faster because it provides a closed-form expression of the generalized forces. However, the computation time increases exponentially (note the logarithmic scale) and quickly explodes. On the other hand, once the local body kinematics is updated, our algorithm computes the terms iteratively by performing algebraic operations, the most of which have dimension three. Indeed, recall that $\calMv_{i}$ depends on three-dimensional vectors. Furthermore, memory efficiency is enhanced by coding the functional expression of integrals and partial derivatives only once for all bodies. 
Fig.~\ref{simulation:planar comparison ID error} shows the norm of the difference in the evaluation of the ID between the two approaches. Specifically, for each model, we evaluate the ID using the EL method and our algorithm with hundred random samples of triplets $(\qv, \dqv, \ddqv)$ ranging in the intervals $[\qv_{\min}, \qv_{\max}] = [-\pi,\pi]$~[\si{\radian}], $[\dqv_{\min},\dqv_{\max}] = [-10, 10 ]~[\si{\radian\per\second}]$ and $[\ddqv_{\min}, \ddqv_{\max}] = [-100,100 ]~[\si{\radian\per\square\second}]$. Then, we compute the relative difference between the outputs and subsequently the mean and standard deviation. 

\section{Conclusions and Future Work}
\label{section:conclusions}

This paper presents a recursive formulation of the equations of motion for holonomically constrained serial soft robotic systems whose kinematics can described or approximated by a finite number of configuration variables. \review{The approach is independent of the body kinematics, domain and type. However, since the kinematics is treated as a black box function, recursivity can be developed only between the bodies of the system and not within the material domain.}

By considering a generic kinematic model for each soft body, the equations of motion are derived using the weak form of dynamics and the Kane method. It is then proven that these equations admit a recursive expression, enabling the development of algorithms solving the inverse dynamics problem. The procedure has linear complexity in the number of bodies, or equivalently in the number of degrees of freedom, making it optimal. Additionally, it is demonstrated that it is possible to simultaneously evaluate the mass matrix and inverse dynamics without affecting computational complexity. 

The versatility of the method is shown through simulations of a new reduced-order model of a trimmed helicoid robot and a hybrid arm consisting of rigid and soft bodies with different kinematic models. Its scalability is also illustrated in a comparison with the Euler-Lagrange method, which is the only approach that allows the computation of the inverse dynamics for the considered class of robots.

Future work will focus on the experimental validation of the method for model-based control. Additionally, we aim to investigate the modifications needed for the algorithm to handle non-holonomic constraints. Another research direction is to explore the recursive form of the dynamics to optimally solve the forward dynamics problem.

\section*{Acknowledgments}

This work was partly supported by PNRR MUR project
PE0000013-FAIR.

\appendix
\section{Appendix}

\subsection{Derivation of the differential kinematics}\label{appendix:differential kinematics}
\review{Here, we present in detail the first- and second-order differential kinematics, which are used in the recursive formulation of the equations of motion. The derivation process parallels the steps undertaken for a rigid system. However, additional terms emerge due to body deformability. Specifically, the time derivatives of $\pv_{i}$ and $\pv_{\comi}$ involve the derivatives of ${}^{i}\rv_{i}$ and ${}^{i}\pv_{\comi}$, respectively~\footnote{\color{red} In a rigid system, ${}^{i}\rv_{i}$ and ${}^{i}\pv_{\comi}$ are constant.}. Similar derivations can be found in studies on flexible link systems, such as~\cite{book1984recursive, shabana1990dynamics}. However, these works employ~\eqref{eq:r_i flexible}, which simplifies the computations. 
}

Time differentiating~\eqref{eq:com position} and~\eqref{eq:point position} yields
\begin{gather}
    \label{eq:v_com_i}
	\vv_{\comi} = \dt{\pv_{\comi}} = \vv_{i} + \omegav_{i} \times \Rotm[i]{}^{i}\pv_{\comi} + \Rotm[i] {}^{i}\dpv_{\comi},\\
	\label{eq:v_p_i}
        \dpv_{i} = \dt{\pv_{i}} = \vv_{\comi} + \omegav_{i} \times \Rotm[i]{}^{i}\rv_{i} + \Rotm[i] {}^{i}\drv_{i},
\end{gather}
where $\vv_{i}$ and $\omegav_i$, which satisfies $ \tilde{\omegav}_{i} = \Dot{\Rm}_{i}\Rotm[i]^{T}$, are the linear and angular velocity of $\{ S_{i}\}$ in $\{ S_{0} \}$, respectively. Moreover, recall from Sec.~\ref{section:kinematics}, that ${}^{i}\dpv_{\comi}$ and ${}^{i}\drv_{i}$ denote the time derivatives of ${}^{i}\pv_{\comi}$ and ${}^{i}\pv_{i}$.
Observing that $\tv_{i} = \tv_{i-1} + \Rm_{i-1}{}^{i-1}\tv_{i}$, it is possible to verify that $\vv_{i}$ and $\omegav_{i}$ admit the following recursive expressions
\begin{gather}
	\label{eq:v_i_recurisve}
	\vv_{i} = \vv_{i-1} + \omegav_{i-1} \times \Rm_{i-1} {}^{i-1}\tv_{i} + \Rm_{i-1} {}^{i-1}\vv_{i-1, i},\\
	\label{eq:omega_i_recurisve}
	\omegav_{i} = \omegav_{i-1} + \Rm_{i-1}{}^{i-1}\omegav_{i-1, i},
\end{gather}
being $^{i-1}\vv_{i-1, i} = {}^{i-1}\dtv_{i}$ and $^{i-1}\omegav_{i-1, i}$, with $^{i-1}\tilde{\omegav}_{i-1, i} = ^{i-1}\dRm_{i}{}^{i-1}\Rotm[i]^{T}$, the relative linear and angular velocity of $\{ S_{i} \}$ as seen from $\{ S_{i-1} \}$. \review{As anticipated}, the above formulas also appear in procedures that compute the FD and ID for rigid systems~\citep{featherstone2014rigid}. However, the right-hand side of~\eqref{eq:v_com_i} and~\eqref{eq:v_p_i} contains the additional terms $\Rotm[i] {}^{i}\dpv_{\comi}$ and $\Rotm[i] {}^{i}\drv_{i}$, which account for the relative motions of the center of mass and of the body points with respect to the reference configuration because of $\calB_{i}$ deformability. A similar decomposition has been exploited in formulating the dynamics of flexible links robots~\citep{deluca2016robots} and reflects an intrinsic coordinate free property of any continuum~\citep{stramigioli2024principal}.


The velocity vectors in the body frame $\{ S_{i} \}$ are defined as
\begin{equation*}
	^{i}\vv_{i} = \Rotm[i]^{T}\vv_{i},\,\,
	^{i}\omegav_{i} = \Rotm[i]^{T}\omegav_{i},\,\,
	^{i}\vv_{\comi} = \Rotm[i]^{T}\vv_{\comi}.
\end{equation*}
Making use of~\eqref{eq:v_i_recurisve} and~\eqref{eq:omega_i_recurisve}, one obtains the following recursive expressions
\begin{gather}
    \label{eq:v_i local}
	^{i}\vv_{i} = {}^{i-1}\Rotm[i]^{T}( {}^{i-1}\vv_{i-1} + {}^{i-1}\omegav_{i-1} \times {}^{i-1}\tv_{i} + {}^{i-1}\vv_{i-1, i} ),\\
    \label{eq:omega_i local}
	^{i}\omegav_{i} = {}^{i-1}\Rotm[i]^{T} ( {}^{i-1}\omegav_{i-1} + {}^{i-1}\omegav_{i-1, i} ),\\
    \label{eq:v_comi local}
	^{i}\vv_{\comi} = {}^{i}\vv_{i} + {}^{i}\omegav_{i} \times {}^{i}\pv_{\comi} + {}^{i}\dpv_{\comi},
\end{gather}
with ${}^{0}\vv_{0} = \zerov_{3}~[\si{\meter \per \second}]$ and ${}^{0}\omegav_{0} = \zerov_{3}~[\si{\radian \per \second}]$. 

Since the formulation of the dynamics requires the accelerations, we time differentiate also~\eqref{eq:v_com_i}, ~\eqref{eq:v_i_recurisve} and~\eqref{eq:omega_i_recurisve}, obtaining
\begin{gather*}
        \begin{split}
	\av_{\comi} = \dt{\vv_{\comi}} &= \av_{i} + \domegav_{i} \times \Rotm[i]{}^{i}\pv_{\comi} + \omegav_{i}\\
		& \quad\times \left( \omegav_{i} \times \Rotm[i] {}^{i}\pv_{\comi} + \Rotm[i] {}^{i}\dpv_{\comi} \right)\\
		&\quad+\omegav_{i} \times \Rotm[i] {}^{i}\dpv_{\comi} + \Rotm[i] {}^{i}\ddpv_{\comi},
	\end{split}\\
	\begin{split}
		\av_{i} = \dt{\vv_{i}} &= \av_{i-1} + \domegav_{i-1} \times \Rm_{i-1}{}^{i-1}\tv_{i} + \omegav_{i-1}\\
	& \quad \times \left( \omegav_{i-1} \times \Rm_{i-1}{}^{i-1}\tv_{i} + \Rm_{i-1} {}^{i-1}\vv_{i-1,i} \right)\\
		&\quad+ \omegav_{i-1} \times \Rm_{i-1} {}^{i-1}\vv_{i-1,i} + \Rm_{i-1} {}^{i-1}\dvv_{i-1,i},
	\end{split}
\end{gather*}
and
\begin{equation*}
\begin{split}
    \domegav_{i} = \dt{\omegav_{i}} &= \domegav_{i-1} + \omegav_{i-1} \times \Rm_{i-1} {}^{i-1}\omegav_{i-1, i}\\
    &\quad\quad+ \Rm_{i-1} {}^{i-1}\domegav_{i-1,i}.
\end{split}
\end{equation*}
From~\eqref{eq:v_p_i} and the above equations, it follows
\begin{equation}\label{eq:point acceleration}
	\begin{split}
		\ddpv_{i} &= \av_{\comi} + \domegav_{i} \times \Rotm[i]{}^{i}\rv_{i} + \omegav_{i} \times (\omegav_{i} \times \Rotm[i]{}^{i}\rv_{i})\\
		&\quad+ 2\omegav_{i} \times \Rotm[i] {}^{i}\drv_{i} + \Rotm[i] {}^{i}\ddrv_{i}.
	\end{split}
\end{equation}
After rotating once again the above vectors in the body frame, i.e., 
\begin{equation*}
	^{i}\av_{i} = \Rotm[i]^{T}\vv_{i},\,\,
	^{i}\domegav_{i} = \Rotm[i]^{T}\omegav_{i},\,\,
	^{i}\av_{\comi} = \Rotm[i]^{T}\vv_{\comi},
\end{equation*}
some computations lead to
\begin{gather}
    \label{eq:^{i}av_i}
	\begin{split}
		^{i}\av_{i} &= {}^{i-1}\Rotm[i]^{T} \left[ {}^{i-1}\av_{i-1} + {}^{i-1}\domegav_{i-1} \times {}^{i-1}\tv_{i} \right.\\
		&\quad\left.+ {}^{i-1}\omegav_{i-1} \times \left( {}^{i-1}\omegav_{i-1} \times {}^{i-1}\tv_{i} + {}^{i-1}\vv_{i-1,i} \right) \right.\\
		&\quad\left. + {}^{i-1}\omegav_{i-1} \times {}^{i-1}\vv_{i-1,i} + {}^{i-1}\dvv_{i-1,i} \right],
	\end{split}\\
    \label{eq:^{i}omegav_i}
    \begin{split}
        ^{i}\domegav_{i} &= {}^{i-1}\Rotm[i]^{T}\left( {}^{i-1}\domegav_{i-1} + {}^{i-1}\omegav_{i-1} \times {}^{i-1}\omegav_{i-1,i}\right.\\
        &\quad\quad\quad\left.+ {}^{i-1}\domegav_{i-1,i} \right),
    \end{split}\\
    \label{eq:^{i}acom_i}
    \begin{split}
		{}^{i}\av_{\comi} &= {}^{i}\av_{i} + {}^{i}\domegav_{i} \times {}^{i}\pv_{\comi}\\
		&\quad+ {}^{i}\omegav_{i} \times \left( {}^{i}\omegav_{i} \times {}^{i}\pv_{\comi} + {}^{i}\dpv_{\comi} \right)\\
		&\quad +{}^{i}\omegav_{i} \times {}^{i}\dpv_{\comi} + {}^{i}\ddpv_{\comi},
	\end{split}
\end{gather}
and
\begin{equation}\label{eq:ddpv_i local}
\begin{split}
    {}^{i}\ddpv_{i} &=  {}^{i}\av_{\comi} + {}^{i}\domegav_{i} \times {}^{i}\rv_{i} + {}^{i}\omegav_{i} \times ({}^{i}\omegav_{i} \times {}^{i}\rv_{i})\\
    &\quad+ 2{}^{i}\omegav_{i} \times {}^{i}\drv_{i} + {}^{i}\ddrv_{i}.
\end{split}
\end{equation}

\subsection{Proof of Proposition~\ref{proposition:equivalence}}\label{appendix:EL equivalence}
To show the result it suffices to prove the equivalence between the Kane and EL equations for a generic configuration variable $q_{k}$. To this end, consider the kinetic energy of the system
\begin{equation}
	\mathcal{K} =  \sum_{i = 1}^{N} \mathcal{K}_i = \sum_{i = 1}^{N} \int_{V_{i}} \frac{1}{2}\, \dpv_{i}^{T} \dpv_{i} \drm m_{i},
\end{equation}
where $\mathcal{K}_i$ is the kinetic energy of $\calB_i$.
By the Reynolds transport theorem, it follows that
\begin{equation}\label{eq:reynold_theorem}
	\begin{split}
		\dt{}\left( \int_{V_{i}} \parv{\pv_{i}}{q_{k}}^{T} \dpv_{i}\drm m_{i} \right) &= \int_{V_{i}} \dt{}\left( \parv{\pv_{i}}{q_{k}}^{T} \dpv_i\drm m_{i} \right)\\
        &=\int_{V_{i}} \dt{}\left( \parv{\pv_{i}}{q_{k}}^{T} \dpv_i \right) \drm m_{i},
	\end{split}
\end{equation}
where in the last equality $\dot{\rho}_{i} = 0$ has been used. Expanding the derivative and rearranging the terms leads to
\begin{equation}\label{eq:appendix:generalized inertial force}
    \begin{split}
		\int_{V_{i}} \parv{\pv_{i}}{q_{k}}^{T} \ddpv_{i} \drm m_{i} &= \dt{}\left( \int_{V_{i}} \parv{\pv_{i}}{q_{k}}^{T} \dpv_{i}\drm m_{i} \right)\\
		&\quad- \int_{V_{i}} \dt{}\left( \parv{\pv_{i}}{q_{k}}^{T} \right) \dpv_{i} \drm m_{i}.\\
    \end{split}
\end{equation}
Now, note that 
\begin{equation*}
    \int_{V_{i}} \parv{\pv_{i}}{q_{k}}^{T} \dpv_{i}\drm m_{i} = \parv{\int_{V_{i}} \frac{1}{2}\,\dpv_{i}^{T} \dpv_{i}\drm m_{i}}{\Dot{q}_{k}} = \parv{\mathcal{K}_i}{\Dot{q}_{k}},
\end{equation*}
and
\begin{equation*}
	\begin{split}
		\int_{V_{i}} \parv{\dpv_{i}}{q_{k}}^{T} \dpv_{i} \drm m_{i} &= \parv{\int_{V_{i}} \frac{1}{2}\, \dpv_{i}^{T} \dpv_{i} \drm m_i  }{q_{k}} = \parv{\mathcal{K}_i}{q_{k}},
	\end{split}
\end{equation*}
which substituted in~\eqref{eq:appendix:generalized inertial force} yields
\begin{equation*}
	\begin{split}
		\int_{V_{i}} \parv{\pv_{i}}{q_{k}}^{T} \ddpv_{i} \drm m_{i} &= \dt{}\left( \parv{\mathcal{K}_i}{\Dot{q}_{k}} \right) - \parv{\mathcal{K}_i}{q_{k}}.
	\end{split}
\end{equation*}
From the linearity of the differentiation operation it follows
\begin{equation*}
\begin{split}
    &\sum_{i = 1}^{N}\int_{V_{i}} \parv{\pv_{i}}{q_{k}}^{T} \ddpv_{i} \drm m_{i} = \sum_{i = 1}^{N}\dt{}\left( \parv{\mathcal{K}_i}{\Dot{q}_{k}} \right) - \parv{\mathcal{K}_i}{q_{k}}\\
    &= \dt{}\left( \parv{ \sum_{i = 1}^{N} \mathcal{K}_i}{\Dot{q}_{k}} \right)- \parv{ \sum_{i = 1}^{N} \mathcal{K}_i}{q_{k}}\\
    &= \dt{}\left( \parv{\mathcal{K}}{\Dot{q}_{k}} \right) - \parv{\mathcal{K}}{q_{k}}
\end{split}
\end{equation*}
Replacing the above equation in~\eqref{eq:kane_deformable_system} finally gives
\begin{equation*}
    Q_k = \dt{}\left( \parv{\mathcal{K}}{\Dot{q}_{k}} \right) - \parv{\mathcal{K}}{q_{k}}.
\end{equation*}
The result follows by recalling that $Q_{k} = -Q_{k}^{*}$.
%
\subsection{Proof of Theorem~\ref{theorem:recurisive M}}\label{appendix:recurisve relations details}
%
First, note that, for all $j \geq i$, the following identities hold
\begin{equation}\label{appendix:recursive 1}
	\begin{split}
		\parv{{}^{j+1}\vv_{j+1}}{\dqv_{i}} &= \parv{{}^{j}\vv_{j}}{\dqv_{i}} \Rotm[j+1][j]\\ 
		&\quad+ \parv{{}^{j}\omegav_{j}}{\dqv_{i}} {}^{j}\tv_{j+1} \times \Rotm[j+1][j],
	\end{split}
\end{equation}
\begin{equation}\label{appendix:recursive 2}
    \begin{split}
        \parv{{}^{j+1}\omegav_{j+1}}{\dqv_{i}} &= \parv{{}^{j}\omegav_{j}}{\dqv_{i}} 
            \Rotm[j+1][j].
    \end{split}
\end{equation}
The result simply follows by iteratively exploiting the above equations. In particular, consider the right-hand side of~\eqref{eq:M_{jl} recursive expression}, repeated below for the ease of readability
\begin{equation}\label{eq:M_{jl} recursive expression repeated}
    \parv{{}^{i}\vv_{i}}{\dqv_{i}} \rb{{}^{i}\Fv_{i} + {}^{i}\Fv_{i}^{*}} + \parv{{}^{i}\omegav_{i}}{\dqv_{i}} \rb{{}^{i}\Tv_{i} + {}^{i}\Tv_{i}^{*}}.
\end{equation}
Replacing~\eqref{eq:M_{jl} recursive expression update} into the previous equation gives
\begin{equation*}~\label{appendix:recursive 3}
    \begin{split}
        &\parv{{}^{i}\vv_{i}}{\dqv_{i}} \rb{{}^{i}\Fv_{i}+{}^{i}\Fv_{i}^{*}} + \parv{{}^{i}\omegav_{i}}{\dqv_{i}} \rb{{}^{i}\Tv_{i}+{}^{i}\Tv_{i}^{*}}\\
		&= \parv{{}^{i}\vv_{i}}{\dqv_{i}} \rb{{}^{i}\calFv_{i} + {}^{i}\calFv_{i}^{*} } + \parv{{}^{i}\omegav_{i}}{\dqv_{i}} \\
        &\quad\quad\quad \left( {}^{i}\calTv_{i} + {}^{i}\calTv_{i}^{*} + {}^{i}\pv_{\com_{i}} \times \rb{{}^{i}\calFv_{i} + {}^{i}\calFv_{i}^{*} } \right)\\
	&\quad+ \parv{{}^{i}\vv_{i}}{\dqv_{i}} \Rotm[i+1][i] \rb{{}^{i+1}\Fv_{i+1}+{}^{i+1}\Fv_{i+1}^{*}}\\
        &\quad+ \parv{{}^{i}\omegav_{i}}{\dqv_{i}} \Rotm[i+1][i] \rb{{}^{i+1}\Tv_{i+1}+{}^{i+1}\Tv_{i+1}^{*}}\\
		&\quad+ \parv{{}^{i}\omegav_{i}}{\dqv_{i}} {}^{i}\tv_{i+1} \times \Rotm[i+1][i]\rb{{}^{i+1}\Fv_{i+1}+{}^{i+1}\Fv_{i+1}^{*}}. 
    \end{split}
\end{equation*}
By direct inspection, one can recognize that the first two terms of the above equation correspond to the first two in the expanded right-hand side of~\eqref{eq: M_{jl} definition}. On the other hand, the remaining three elements have the same form of the right-hand side of~\eqref{appendix:recursive 1} or~\eqref{appendix:recursive 2}, which can be exploited to obtain
\begin{equation}\label{appendix:recursive 4}
	\begin{split}
		&\parv{{}^{i}\vv_{i}}{\dqv_{i}} \rb{{}^{i}\Fv_{i}+{}^{i}\Fv_{i}^{*}} + \parv{{}^{i}\omegav_{i}}{\dqv_{i}} \rb{{}^{i}\Tv_{i}+{}^{i}\Tv_{i}^{*}}\\
		&= \parv{{}^{i}\vv_{i}}{\dqv_{i}} \rb{{}^{i}\calFv_{i} + {}^{i}\calFv_{i}^{*} }\\
        &+ \parv{{}^{i}\omegav_{i}}{\dqv_{i}} \left( {}^{i}\calTv_{i} + {}^{i}\calTv_{i}^{*} + {}^{i}\pv_{\com_i} \times \rb{{}^{i}\calFv_{i} + {}^{i}\calFv_{i}^{*} } \right)\\
		&+ \parv{{}^{i+1}\vv_{i+1}}{\dqv_{i}} \rb{{}^{i+1}\Fv_{i+1} + {}^{i+1}\Fv_{i+1}^{*}}\\
        &+ \parv{{}^{i+1}\omegav_{i+1}}{\dqv_{i}} \rb{{}^{i+1}\Tv_{i+1} + {}^{i+1}\Tv_{i+1}^{*}}.
	\end{split}
\end{equation}
The last two terms of~\eqref{appendix:recursive 4} have again the same form of~\eqref{eq:M_{jl} recursive expression repeated} but with increased index. The result follows by repeating the last two steps until the $N$-th term appears. 

\review{
\subsection{Comparison with the Euler-Lagrange method}\label{appendix:EL comparison flow}
\begin{figure*}
    \centering
    \subfigure[IID algorithm]{
        \includegraphics[width=0.97\columnwidth]{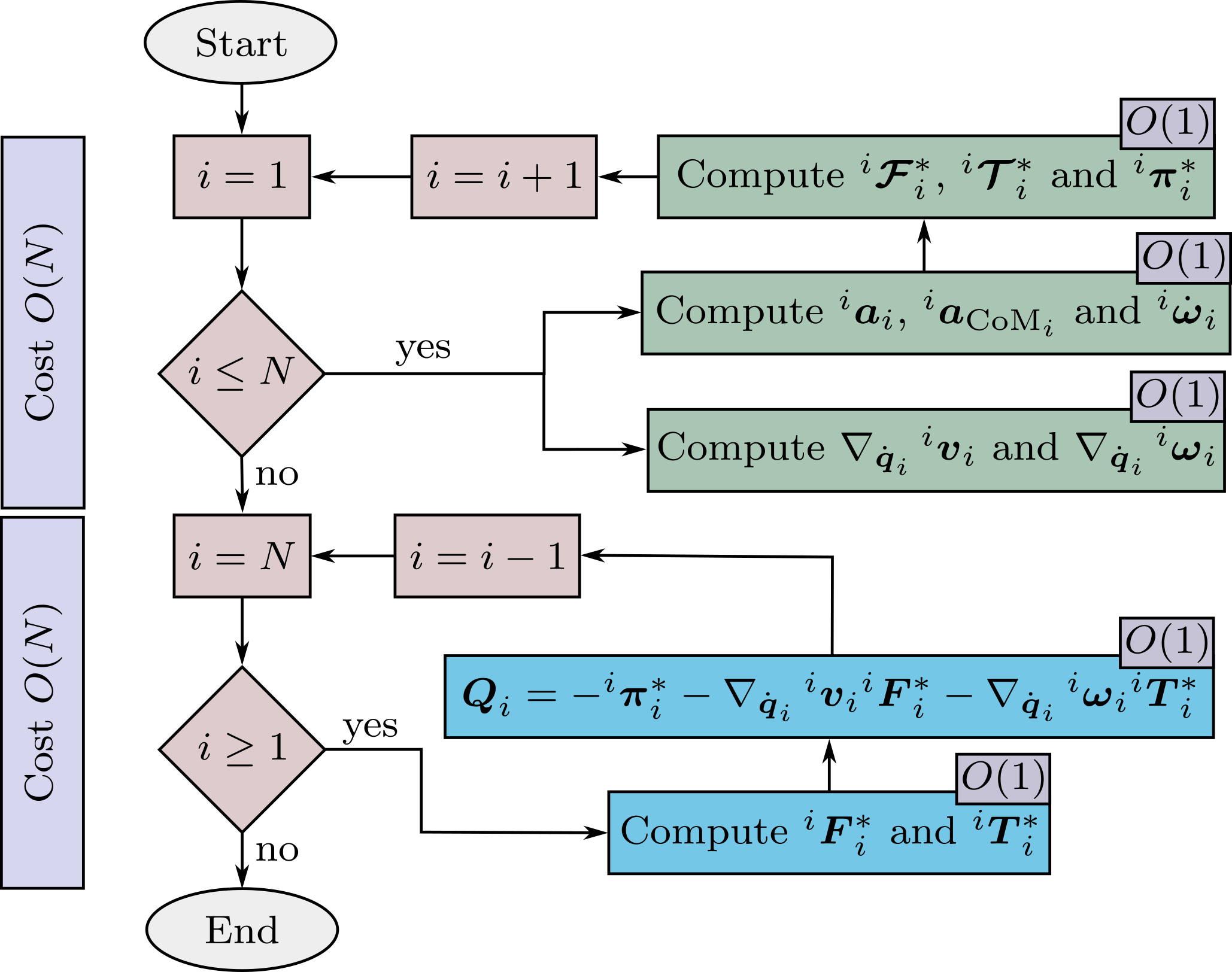}
    }
    \subfigure[EL algorithm]{
        \includegraphics[width=0.97\columnwidth]{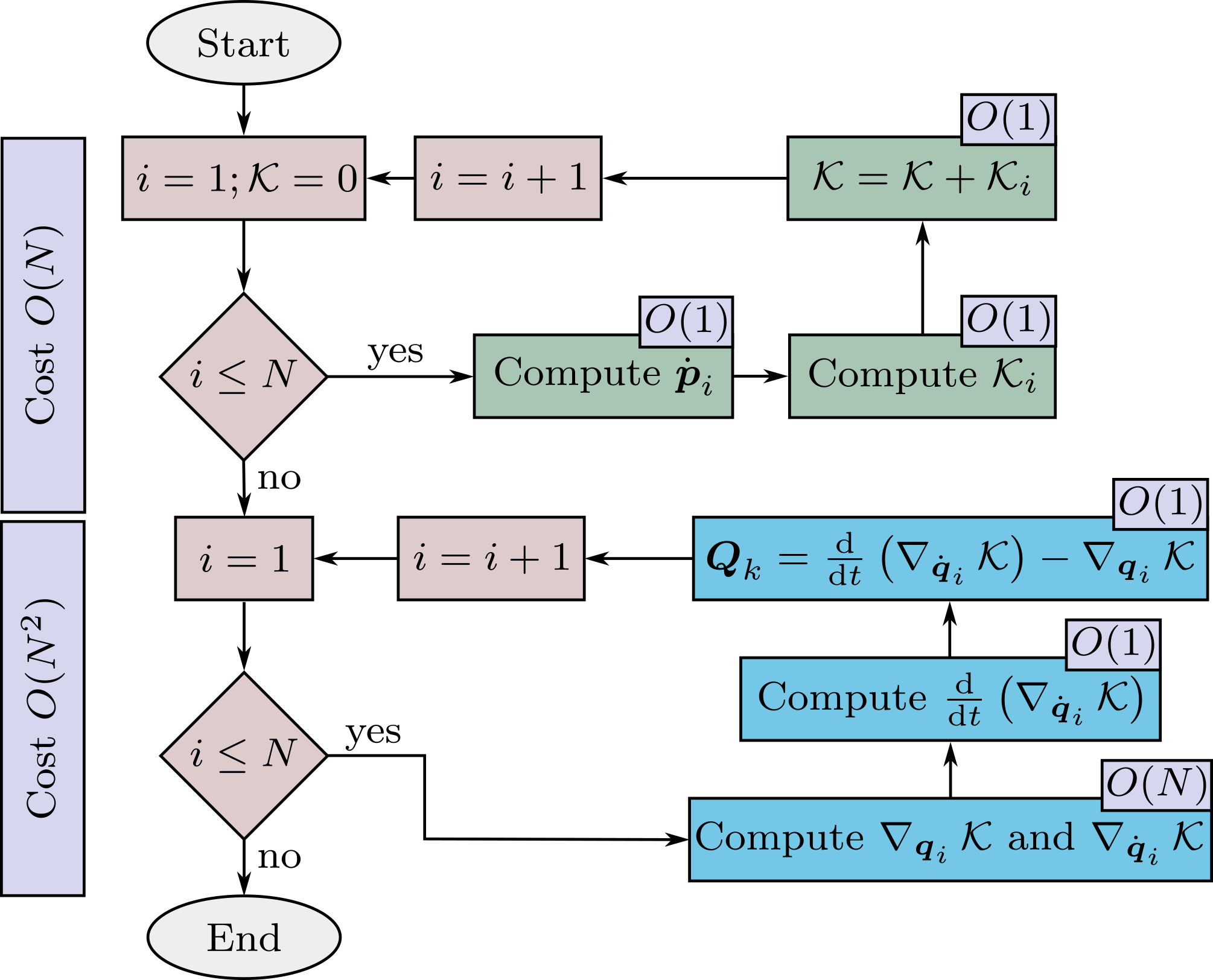}
    }
    \caption{\review{Flow diagrams representing (a)~Algorithm~\ref{alg:Kane inverse recursive general} and (b)~the EL algorithm. Both procedures can be implemented in two passes, with their primary subroutines highlighted in green and blue, respectively. The top-right corner of each subroutine indicates its associated computational complexity, assuming that the cost of kinematic evaluations is negligible. During the second pass, the first subroutine of the EL method incurs a computational cost of \( O(N) \) for every iteration index \( i \) due to the presence of a summation over all system bodies in $\mathcal{K}$. Consequently, this results in an overall complexity of \( O(N^{2}) \). In contrast, all subroutines of the Algorithm~\ref{alg:Kane inverse recursive general} maintain a constant computational complexity.}}
    \label{fig:kane_vs_el_flow}
\end{figure*}
Fig.\ref{fig:kane_vs_el_flow} illustrates the flow diagrams of Algorithm~\ref{alg:Kane inverse recursive general} and the EL approach. The latter is characterized by a computational cost that grows quadratically with the number of bodies, or equivalently, with the number of DoFs. This leads to a suboptimal implementation of the ID. Similar considerations hold when active forces are included because the evaluation of inertial forces still defines a lower bound for the computational cost.
}
\review{
\subsection{Impact evaluation of the kinematic model complexity}\label{appendix:impact evaluation}
In a deformable system, the computational complexity of solving the ID problem is influenced not only by the efficiency of evaluating the inter-body dynamics --- namely, the dynamic relationships between the bodies --- but also by the intra-body terms, i.e., those that depend on the body kinematics~\eqref{eq:body to body transformation}. Indeed, the latter may require non-negligible computational resources due to their dependence on~\eqref{eq:functional expression body kinematics}.  

To estimate how the kinematic model affects the cost of the ID algorithm, one can proceed as follows. For simplicity, consider Algorithm~\ref{alg:Kane inverse recursive general}, though similar considerations apply to all algorithms presented in this paper.

First, note that the terms appearing in the forward and backward steps of Algorithm~\ref{alg:Kane inverse recursive general} depend implicitly on the reduced-order kinematics and its Jacobian with respect to the configuration variables, as well as on the numerical method used to approximate volume integrals. Suppose that the computational complexity of the reduced-order model and its Jacobian are given by \( \alpha_{i}(n_{i}) \) and \( \beta_{i}(n_{i}) \), respectively. Additionally, let \( \gamma_{i}(n_{i}) \) denote the computational complexity of the numerical method used for computing the integrals.  
              
From Algorithm~\ref{alg:Kane inverse recursive general} and the expressions of all terms therein, it follows that the cost of both the forward and backward steps can be estimated as 
\begin{equation*}
\begin{split}
    O(\text{backward}) &= O(\text{forward})\\
    &= \sum_{i = 1}^{N} O(\gamma_{i}(\max\{\alpha_{i}(n_{i}); \beta_{i}(n_{i})\})).
\end{split}
\end{equation*}
In the above expression, we assume that algebraic computations require no resources. Consequently, we have
\begin{equation*}
\begin{split}
    O(\text{ID}) &= O(\text{backward}) + O(\text{forward})\\
    &= \sum_{i = 1}^{N} O(\gamma_{i}(\max\{\alpha_{i}(n_{i}); \beta_{i}(n_{i})\})).
\end{split}
\end{equation*}
Moreover, if we neglect the computational complexity of the kinematics, i.e., assuming \( \alpha_{i}(n_{i}) = \beta_{i}(n_{i}) = \gamma_{i}(n_{i}) = k \) for some constant integer \( k \), we recover the expected result \( O(\text{ID}) = O(N) \).

Similar considerations apply also to the EL method, for which, however, the cost is
\begin{equation*}
    \rb{\sum_{i = 1}^{N} O(\gamma_{i}(\max\{\alpha_{i}(n_{i}); \beta_{i}(n_{i})\}))}^{2}.
\end{equation*}
%
}
\subsection{Proof of $O(N)$ computation of the mass matrix}\label{appendix:mass inverse dynamics}
For the following derivations, it is convenient to split $\Mm(\qv)$ into block rows corresponding to each body, namely
\begin{equation*}
\Mm(\qv) = \begin{carray}{c}
    \Mm_{1}^{T}(\qv)\\
    \vdots\\
    \Mm_{i}^{T}(\qv)\\
    \vdots\\
    \Mm_{N}^{T}(\qv)
\end{carray},
\end{equation*}
with $\Mm_{i}^{T}(\qv)\in \R^{n_{i} \times n}$. This way,~\eqref{eq:M ID} rewrites equivalently as
\begin{equation}\label{eq:M ID j}
    \Mm_{i}^{T}(\qv) = -\jac{{\piv_{i}^{*}}^{\zerov}}{\ddqv} - \jac{\calMv_{i}^{\zerov}}{\ddqv}
\end{equation}

We begin by considering $\jac{\calMv_{i}^{\zerov}}{\ddqv}$. Recalling the backward recursive expression~\eqref{eq:M_{jl} recursive expression} one has
\begin{equation}\label{eq:jac calM j}
    \jac{ \calMv_{i}^{\zerov} }{\ddqv} = \parv{{}^{i}\vv_{i}}{\dqv_{i}} \jac{{{}^{i}\Fv_{i}^{*}}^{\zerov}}{\ddqv} + \parv{{}^{i}\omegav_{i}}{\dqv_{i}} \jac{{{}^{i}\Tv_{i}^{*}}^{\zerov}}{\ddqv},
\end{equation}
with
\begin{equation*}\label{eq:F zero velocity}
    \jac{{{}^{i}\Fv_{i}^{*}}^{\zerov}}{\ddqv} = \jac{{{}^{i}\calFv_{i}^{*}}^{\zerov}}{\ddqv} + \Rotm[i+1][i] \jac{{{}^{i+1}\Fv_{i+1}^{*}}^{\zerov}}{\ddqv},
\end{equation*}
and
\begin{equation*}\label{eq:T zero velocity}
\begin{split}
    \jac{{{}^{i}\Tv_{i}^{*}}^{\zerov}}{\ddqv} &= \jac{{{}^{i}\calTv_{i}^{*}}^{\zerov}}{\ddqv} + {}^{i}\tilde{\pv}_{\com_{i}} \jac{{{}^{i}\calFv_{i}^{*}}^{\zerov}}{\ddqv} + \Rotm[i+1][i] \jac{{{}^{i+1}\Tv_{i+1}^{*}}^{\zerov}}{\ddqv} \\
    &\quad+ {}^{i}\tilde{\tv}_{i+1} \Rotm[i+1][i] \jac{{{}^{i+1}\Fv_{i+1}^{*}}^{\zerov}}{\ddqv},
\end{split}
\end{equation*}
where the fact that the velocities are zero has been used. To compute the above expressions, one has to evaluate $\jac{{{}^{i}\calFv_{i}^{*}}^{\zerov}}{\ddqv}$ and $\jac{{{}^{i}\calTv_{i}^{*}}^{\zerov}}{\ddqv}$. Recalling \eqref{eq:inertia force} and~\eqref{eq:int r x ddpi}, it follows
\begin{equation}\label{eq:cal F zero velocity}
    \jac{{{}^{i}\calFv_{i}^{*}}^{\zerov}}{\ddqv} = -m_{i} \jac{{}^{i}\av_{\com_{i}}^{\zerov}}{\ddqv},
\end{equation}
and
\begin{equation}\label{eq:cal T zero velocity}
    \begin{split}
    \jac{{{}^{i}\calTv_{i}^{*}}^{\zerov}}{\ddqv} &= -{}^{i}\IIm_{i} \jac{^{i}\domegav_{i}^{\zerov}}{\ddqv} - \int_{V_{i}} {}^{i}\tilde{\rv}_{i} \jac{{}^{i}\ddrv_{i}^{\zerov}}{\ddqv}\drm m_{i}
    \end{split}
\end{equation}
where 
\begin{equation*}
    \jac{{}^{i}\ddrv_{i}^{\zerov}}{\ddqv} = \begin{carray}{ccccc}
        \zerov_{3 \times (n_{i}-1)} & \jac{{}^{i}\rv_{i}}{\qv_{i}} & \zerov_{3 \times (n-2n_{i} + 1)} 
    \end{carray}.
\end{equation*}
To obtain~\eqref{eq:cal F zero velocity} and~\eqref{eq:cal T zero velocity} the Jacobians of the second-order differential kinematics are needed. When $\dqv = \zerov_{n}$,~\eqref{eq:^{i}av_i}--\eqref{eq:^{i}acom_i} reduce to
\begin{gather*}
    \label{eq:^{i}av_i_ zero velocity}
	\begin{split}
		^{i}\av_{i}^{\zerov} &= {}^{i-1}\Rotm[i]^{T} \left( {}^{i-1}\av_{i-1}^{\zerov} + {}^{i-1}\domegav_{i-1}^{\zerov} \times {}^{i-1}\tv_{i} + {}^{i-1}\dvv_{i-1,i}^{\zerov} \right)\\
	\end{split}\\
    \label{eq:^{i}omegav_i zero velocity}
    \begin{split}
        ^{i}\domegav_{i}^{\zerov} &= {}^{i-1}\Rotm[i]^{T}\left( {}^{i-1}\domegav_{i-1}^{\zerov} + {}^{i-1}\domegav_{i-1,i}^{\zerov}\right),
    \end{split}
\end{gather*}
and
\begin{equation*}
    \label{eq:^{i}acom_i zero velocity}
    \begin{split}
		{}^{i}\av_{\com_{i}}^{\zerov} &= {}^{i}\av_{i}^{\zerov}+ {}^{i}\domegav_{i}^{\zerov} \times {}^{i}\pv_{\com_{i}} + {}^{i}\ddpv_{\com_{i}}^{\zerov}.
    \end{split}
\end{equation*}
The above terms depend linearly on $\ddqv$ and their Jacobian is
\begin{gather*}
    \label{eq:jacobian ^{i}av_i_ zero velocity}
	\begin{split}
		\jac{^{i}\av_{i}^{\zerov}}{\ddqv} &= {}^{i-1}\Rotm[i]^{T} \left( \jac{{}^{i-1}\av_{i-1}^{\zerov}}{\ddqv} - {}^{i-1}\tilde{\tv}_{i} \jac{{}^{i-1}\domegav_{i-1}^{\zerov}}{\ddqv} \right.\\
        &\left. \quad\quad\quad\quad + \jac{{}^{i-1}\dvv_{i-1,i}^{\zerov}}{\ddqv} \right), 
	\end{split}\\
    \label{eq:jacobian ^{i}omegav_i zero velocity}
    \begin{split}
        \jac{^{i}\domegav_{i}^{\zerov}}{\ddqv} &= {}^{i-1}\Rotm[i]^{T}\left( \jac{{}^{i-1}\domegav_{i-1}^{\zerov}}{\ddqv} + \jac{{}^{i-1}\domegav_{i-1,i}^{\zerov}}{\ddqv}\right),
    \end{split}
\end{gather*}
and
\begin{equation*}
    \label{eq:jacobian ^{i}acom_i zero velocity}
    \begin{split}
		\jac{{}^{i}\av_{\com_{i}}^{\zerov}}{\ddqv} &= \jac{{}^{i}\av_{i}^{\zerov}}{\ddqv} -{}^{i}\tilde{\pv}_{\com_{i}} \jac{{}^{i}\domegav_{i}^{\zerov}}{\ddqv} + \jac{{}^{i}\ddpv_{\com_{i}}^{\zerov}}{\ddqv}.
    \end{split}
\end{equation*}
These Jacobians can be computed recursively from the robot base by setting
\begin{equation*}
    \jac{{}^{0}\av_{0}^{\zerov}}{\ddqv} = \zerov_{3 \times n}, \,\, \jac{{}^{0}\omegav_{0}^{\zerov}}{\ddqv} = \zerov_{3 \times n},
\end{equation*}
and observing that
\begin{equation*}
    \jac{{}^{i-1}\dvv_{i-1,i}^{\zerov}}{\ddqv} = \begin{carray}{ccccc}
        \zerov_{3 \times (n_{i}-1)} & \jac{{}^{i-1}\tv_{i}}{\qv_{i}} & \zerov_{3 \times (n-2n_{i} + 1)} 
    \end{carray},
\end{equation*}
and
\begin{equation*}
    \jac{{}^{i}\ddpv_{\com_{i}}^{\zerov}}{\ddqv} = \begin{carray}{ccccc}
        \zerov_{3 \times (n_{i}-1)} & \jac{{}^{i}\pv_{\com_{i}}}{\qv_{i}} & \zerov_{3 \times (n-2n_{i} + 1)}
    \end{carray}.
\end{equation*}
Note that $\jac{{}^{i-1}\tv_{i}}{\qv_{i}}$ and $\jac{{}^{i}\pv_{\com_{i}}}{\qv_{i}}$ must also be computed for the IID and thus are readily available. 

In analogy with the computation of $\calMv_{i}$, the computation of $\jac{\calMv_{i}^{\zerov}}{\ddqv}$ can be done recursively into two stages. First, one computes $\jac{^{i}\av_{i}^{\zerov}}{\ddqv}, \jac{^{i}\domegav_{i}^{\zerov}}{\ddqv}, \jac{{}^{i}\av_{\com_{i}}^{\zerov}}{\ddqv}, \jac{{{}^{i}\calFv_{i}^{*}}^{\zerov}}{\ddqv}$ and $\jac{{{}^{i}\calTv_{i}^{*}}^{\zerov}}{\ddqv}$ forward in the chain. Then, $\jac{\calMv_{i}^{\zerov}}{\ddqv}$ is obtained backward using $\jac{{{}^{i}\calFv_{i}^{*}}^{\zerov}}{\ddqv}$ and $\jac{{{}^{i}\calTv_{i}^{*}}^{\zerov}}{\ddqv}$. 

We now move to the computation of $\jac{{\piv_{i}^{*}}^{\zerov}}{\ddqv}$, From~\eqref{eq:int dri ddpi}, we have
\begin{equation}\label{eq:jacobian pi star zero velocity} 
    \begin{split}
	\jac{{\piv_{i}^{*}}^{\zerov}}{\ddqv} &= -\parv{}{\dqv_{i}} \left( \int_{V_{i}} {}^{i}\rv_{i} \times {}^{i}\drv_{i} \drm m_{i} \right) \jac{{}^{i}\domegav_{i}^{\zerov}}{\ddqv}\\ 
	&\quad - \int_{V_{i}} \parv{{}^{i}\drv_{i}}{\dqv_{i}} \jac{{}^{i}\ddrv_{i}^{\zerov}}{\ddqv} \drm m_{i} + \parv{{}^{i}\dpv_{\com_{i}}}{\dqv_{i}} \jac{{}^{i}\calFv^{*}_{i}}{\ddqv}^{\zerov}. 
	\end{split}.
\end{equation}
Again, observe that all the terms appearing in the above equation are computed during the forward step of the IID, so implying that $\jac{{\piv_{i}^{*}}^{\zerov}}{\ddqv}$ can be evaluated in a forward pass.
%
\begin{algorithm}[t]
\caption{\small Mass Inertial Inverse Dynamics: $[\Qm, \Mm] \hspace{-2pt} = \hspace{-2pt} \text{MIID}(\qv, \dqv, \ddqv$)}\label{alg:Kane mass inverse recursive general}
\begin{algorithmic}
\Require $\qv, \dqv, \ddqv$
\For{$i = 1 \rightarrow N$}\Comment{Forward step}
    \State Compute ${}^{i}\av_{i}$, ${}^{i}\av_{\com_{i}} $ and ${}^{j}\domegav_{i}$
    \State Compute $\jac{^{i}\av_{i}^{\zerov}}{\ddqv}$, $ \jac{{}^{i}\av_{\com_{i}}^{\zerov}}{\ddqv}$ and $ \jac{^{i}\domegav_{i}^{\zerov}}{\ddqv}$
    \State Compute ${}^{i}\calFv_{i}^{*}$, ${}^{i}\calTv_{i}^{*}$ and ${}^{i}\piv_{i}^{*}$ 
    \State Compute $\jac{{{}^{i}\calFv_{i}^{*}}^{\zerov}}{\ddqv}$, $\jac{{{}^{i}\calTv_{i}^{*}}^{\zerov}}{\ddqv}$ and $\jac{{{}^{i}\piv_{i}^{*}}^{\zerov}}{\ddqv}$
\EndFor
\For{$i = N \rightarrow 1$}\Comment{Backward step}
    \State Compute $\Qm_{i}$ as given in~\eqref{eq:Q_{j}^{*} M expression} by using~\eqref{eq:pi expression final}--\eqref{eq:M_{jl} recursive expression update}
    \State Compute $\Mm_{i}^{T}$ as given in~\eqref{eq:M ID j} by using~\eqref{eq:jac calM j}--\eqref{eq:jacobian pi star zero velocity}
\EndFor
\end{algorithmic}
\end{algorithm}

Leveraging the above results, we can introduce a modified version of Algorithm~\ref{alg:Kane inverse recursive general} that evaluates $\Qm$ and $\Mm$. The pseudocode for this new procedure, called Mass Generalized Inverse Dynamics (MIID), is provided in Algorithm~\ref{alg:Kane mass inverse recursive general}.
\begin{corollary}
    The computational complexity of Algorithm~\ref{alg:Kane mass inverse recursive general} is $O(N)$.
\end{corollary}
\bibliographystyle{sageH}
\bibliography{references.bib}
\end{document}


\title{Recursive Model-agnostic Inverse Dynamics of Serial Soft-Rigid Robots: Supplementary Material}

\author{Pietro Pustina\affilnum{1, 2}, Cosimo Della Santina\affilnum{2, 3} and Alessandro De Luca\affilnum{1}}

\affiliation{\affilnum{1} Department of Computer, Control and Management Engineering, Sapienza University of Rome, Rome, IT.\\
\affilnum{2}Department of Cognitive Robotics, Delft University of Technology, Delft, NL.\\
\affilnum{3}Institute of Robotics and Mechatronics, German Aerospace Center (DLR), Oberpfaffenhofen, GER.
}

\corrauth{Pietro Pustina, pietro.pustina@uniroma1.it}

\maketitle

\section{Extra Numerical Results}\label{section:simulations}
This document presents additional simulation results with the inverse dynamics algorithms presented in~\cite{pustina2023univeral}. 

\textbf{Note:} All the numerical results are obtained with the same inverse dynamics algorithm. The MATLAB code used for the simulations can be accessed at the following link:~\url{https://github.com/piepustina/Jelly}.

\subsection{Simulation 4: PCC with variable radius}
Consider a planar soft robot actuated with three pairs of antagonistic pneumatic chambers. In the following, we show that the generic kinematics function used in~\cite{pustina2023univeral} allows us to easily model the effects of the chambers on the dynamics by introducing additional DoF for the radius change. To the best of our knowledge, this is the first time that such class of models is presented and simulated, and thus constitutes an additional contribution of our work.

Specifically, we model the kinematics by using a combination of strain and radial configuration variables. The strain configuration encodes the pose of the robot backbone, while the radial variables model the shape alterations along the robot section. The continuum is discretized into three bodies with cylindrical shape in the stress free configuration, having each rest length $L_{0_{i}} = 0.2~[\si{\meter}]$, radius $R_{0_{i}} = 0.02~[\si{\meter}]$ and mass density $\rho_{i} = 1062~[\si{\kilogram \per \cubic\meter}]; i = 1, 2$. The material visco-elastic parameters are $C_{i} = 0.501 [\si{\mega \pascal}]$ and $\eta_{i} = 0.1~[\si{\second}]$. The base is rotated so that in the straight configuration the robot is aligned with the gravitational field and points downwards. Each pneumatic chamber has a uniform distance $d_{c, i} = 0.004~[\si{\meter}]$ from the body walls.

Each body $\calB_{i}$ has in total four DoF. The first two model the curvature and elongation strains under the PCC hypothesis, i.e., 
\begin{equation*}
    \begin{carray}{c}
        \kappa_{i}(\qv_{i})\\\delta L_{i}(\qv_{i})
    \end{carray} = \frac{1}{L_{0_{i}}} \begin{carray}{cccc}
        1 & 0 & 0 & 0\\ 0 & 1 & 0 & 0
    \end{carray}\qv_{i} + \begin{carray}{c}
        0 \\ 1
    \end{carray}.
\end{equation*}
The remaining two configuration variables describe the change in radius $R_{i}$, namely
\begin{equation*}
    R_{i}(s_{i}, \phi_{i,}, \qv_{i}) = R_{0_{i}} + \delta R_{i}(s_{i}, \phi_{i}, \qv_{i}),
\end{equation*}
where
\begin{equation*}\small
    \delta R_{i} = \left\{ \begin{array}{ll}
        \multirow{2}{*}{$\begin{carray}{cccc}
            0 & 0 & b(s_{i}, \phi_{i}) & 0 
        \end{carray}\qv_{i},$} & s_{i} \in [0, L_{0, i}], \\
        & \phi_{i} \in [0, \pi);\\[4pt]
        \multirow{2}{*}{$\begin{carray}{cccc}
            0 & 0 & 0 & b(s_{i}, \phi_{i}-\pi) 
        \end{carray}\qv_{i},$} & \phi_{i} \in [0, L_{0, i}], \\
        & \phi_{i} \in [\pi, 2\pi).\\
    \end{array} \right.,
\end{equation*}
models the radius change from $R_{0_{i}}$. The angular variable $\phi_{i}$ is used to locate for each cross section the pneumatic chamber. To model the functional dependence of $\delta R_{i}$ on the material coordinates, in this case $s_{i}$, $\phi_{i}$ and the radius $R_{0_{i}}$ in the stress free-configuration, we use the multivariate bump function 
\begin{equation*}
    b(s_{i}, \phi_{i}) = e^{-\displaystyle\frac{L_{0_{i}}}{\frac{L_{0_{i}}^{2}}{4} - (s_{i} - \frac{L_{0_{i}}}{2})^{2}}}e^{-\displaystyle\frac{1}{\frac{\pi^{2}}{4} - (\phi_{i} - \frac{\pi}{2})^{2}}}.
\end{equation*}
This choice is motivated by empirical observation of pneumatically actuated soft robots, whose chamber deformation is usually larger at the middle of body. 

\begin{figure}[ht!]
    \centering
    \subfigure[{Strain configuration variables}]{
        \adjustbox{gstore width=\figureWidth, gstore        height=\figureHeight}{
            \includegraphics[width = \columnwidth]
            {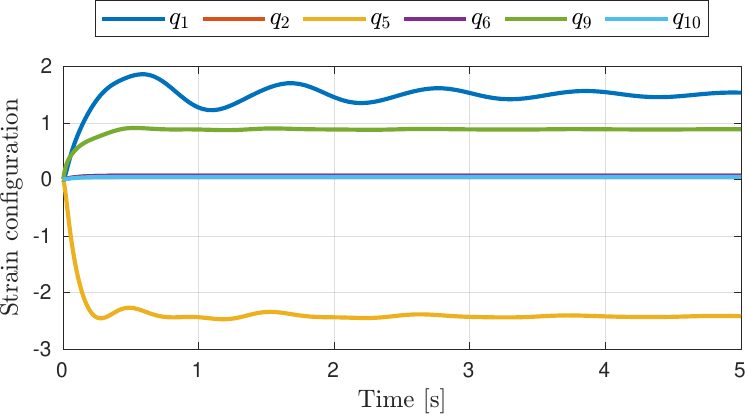}
        \label{fig:simulation 1: strain configuration variables}
        }
    }
    \subfigure[{Radius configuration variables}]{
        \adjustbox{gstore width=\figureWidth, gstore        height=\figureHeight}{
            \includegraphics[width = \columnwidth]
            {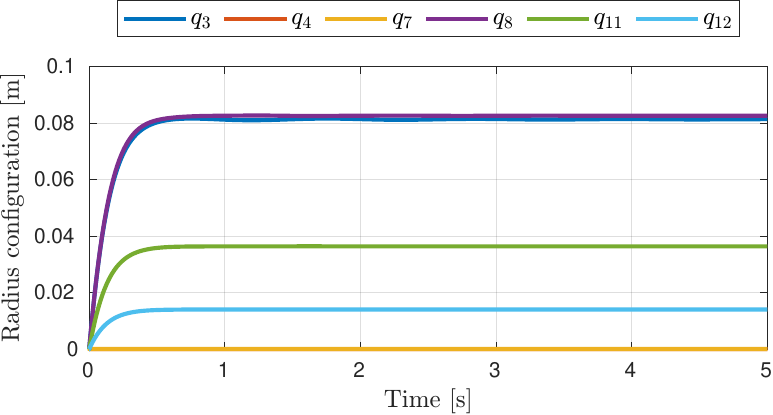}
        \label{fig:simulation 1: radius configuration variables}
        }
    }
    \subfigure[{Stroboscopic plot}]{
        \includegraphics[width = 0.6\columnwidth]
        {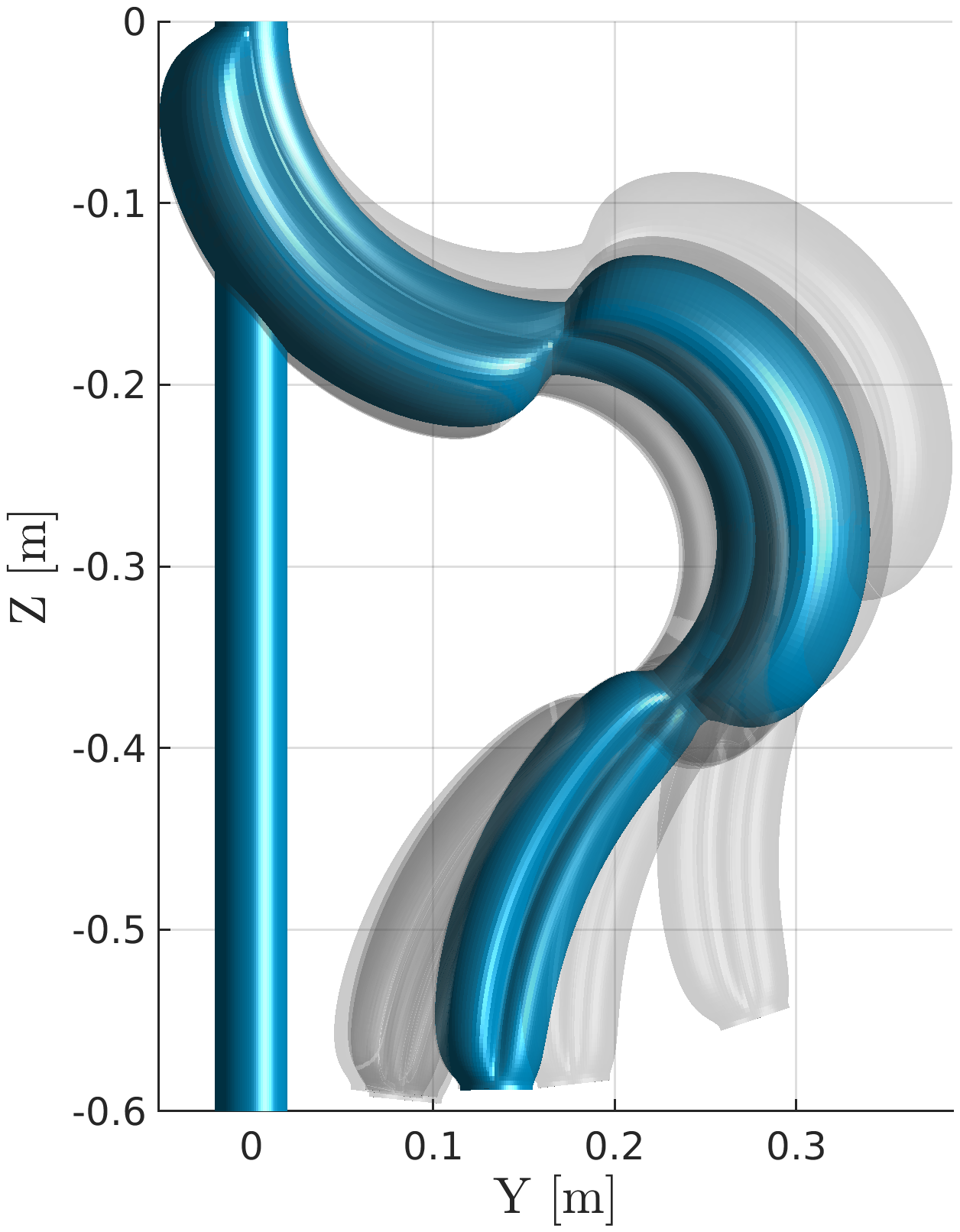}
        \label{fig:simulation 1: strobo plot}
    }
    \caption{\small Simulation 4: PCC with variable radius. Time evolution of the (a)~strain (in $[\si{\radian}]$ and $[\si{\meter}]$) and (b)~radial configuration variables, and (c)~stroboscopic plot of motion. The robot consists of three bodies with variable radius and is actuated through six pressure chambers. In~(c), the initial and final configurations are depicted in blue, while the transient motion in light gray.}
    \label{fig:simulation 1}
\end{figure}

The actuation generalized force $\nuv_{i}(\qv_{i}, \uv_{i})$ of $\calB_{i}$ is modeled using the principle of virtual works leading to
\begin{equation}\label{eq:sim1:actuation volume}
    \nuv_{i}(\qv_{i}, \uv_{i}) = \parv{\Vm_{i}}{\qv_{i}}(\qv_{i}) \uv_{i}, 
\end{equation}
where the vector $\Vm_{i}(\qv_{i}) = \rb{V_{i, 1}(\qv_{i}) \,\,\,\, V_{i, 2}(\qv_{i})}^{T}$ collects the volume of the two pneumatic chambers in the current configuration and $\uv_{i} = \rb{u_{i, 1} \,\,\,\, u_{i, 2} }^{T}$ is the difference in chamber pressure with respect to the atmospheric pressure. The robot starts at rest from the stress free configuration $\qv_{0} = \zerov_{12}$ and the dynamics is excited with a constant input equal to
\begin{equation*}
    \uv = \begin{carray}{cccccc}
        2 & 0 & 0 & 2 & 1 & 0.4
    \end{carray}^{T}~[\si{\mega \pascal}].
\end{equation*}
The volume integrals appearing in the ID algorithms and~\eqref{eq:sim1:actuation volume} are approximated using a Gaussian quadrature rule in spherical coordinates. Fig.~\ref{fig:simulation 1} reports the time evolution of the configuration variables, divided into strain and radial variables, and a stroboscopic plot of the robot motion in its workspace. 

\subsection{Simulation 5: PGC in free evolution}
\begin{figure}[ht!]
    \centering
    \subfigure[{Configuration variables}]{
        \adjustbox{gstore width=\figureWidth, gstore        height=\figureHeight}{
            \includegraphics[width = 1\columnwidth]
            {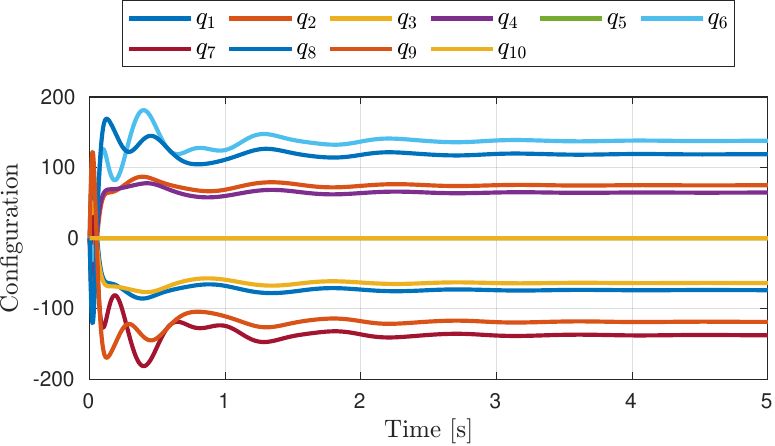}
        \label{fig:simulation 2: configuration variables}
        }
    }
    \subfigure[{Stroboscopic plot}]{
        \includegraphics[width = 0.7\columnwidth]
        {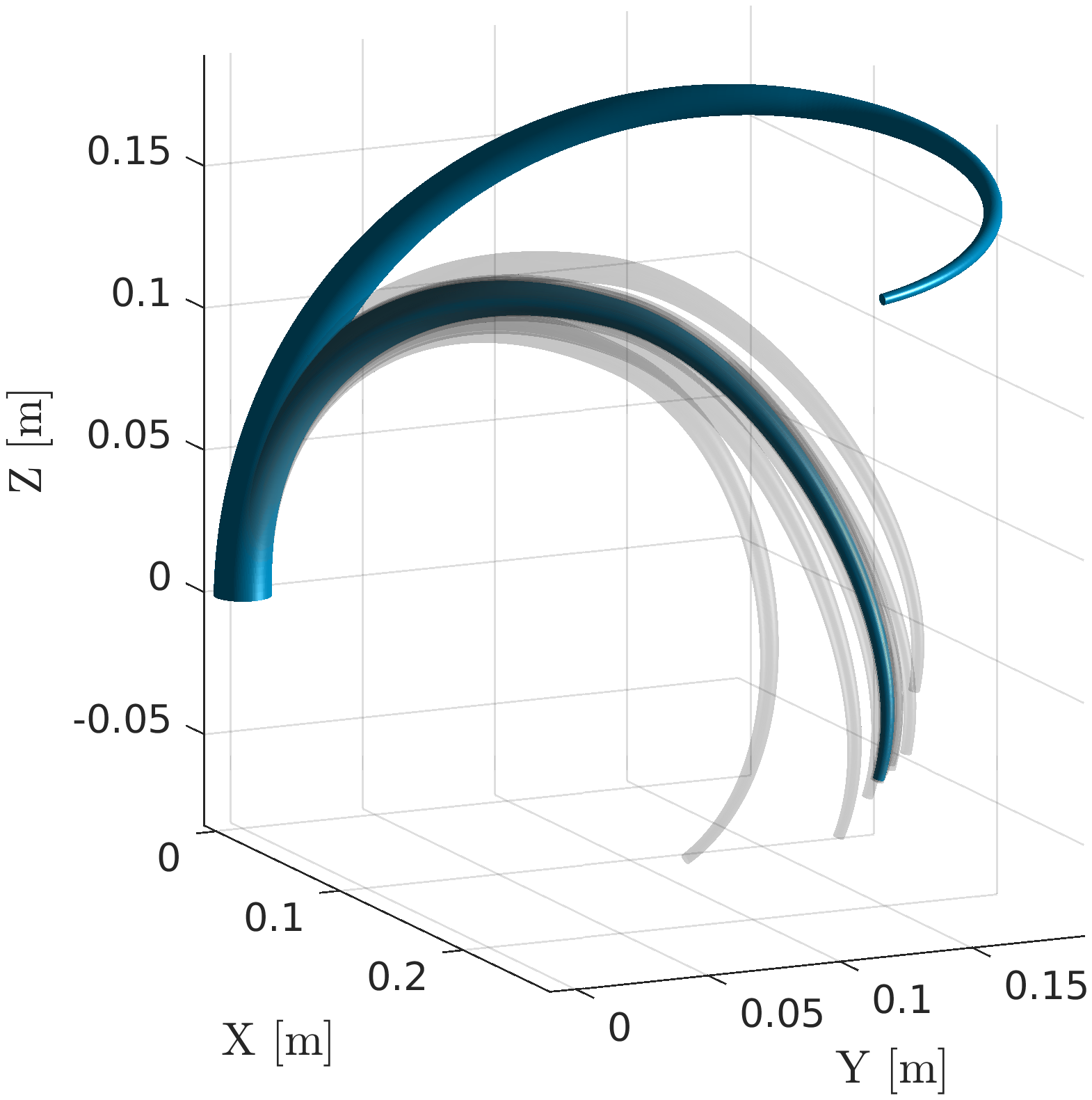}
        \label{fig:simulation 2: strobo plot}
    }
    \caption{\small Simulation 5: PGC in free evolution. Time evolution of the (a)~configuration variables (in $[\si{\radian}]$ and $[\si{\meter}]$) and (b)~stroboscopic plot of a slender continuum soft robot modeled with a Gaussian model of the strain. In~(b), the initial and final configurations are depicted in blue, while intermediate configurations in light gray.}
    \label{fig:simulation 2}
\end{figure}

Consider a tentacle-like soft robotic arm of two bodies with conical shape. The first body has base and tip radius equal to $R_{\mathrm{base}_{1}} = 0.01~[\si{\meter}]$ and $R_{\mathrm{tip}_{1}} = 0.005~ [\si{\meter}]$, respectively, while the second $R_{\mathrm{base}_{2}} = 0.005~[\si{\meter}]$ and $R_{\mathrm{tip}_{2}} = 0.002~[\si{\meter}]$. The rest length and mass density are $L_{0_{i}} = 0.2~[\si{\meter}]$ and $\rho_{i} = \SI{1070}{[\kilogram\per\cubic\meter]}$, respectively. The elastic parameter is $C_{i} = \SI{0.111}{[\giga\pascal]}$, while the material viscosity $\eta_{i} = \SI{0.01}{[\second]}; i = 1, 2$. The kinematics is computed using the Geometric Variable Strain approach of~\cite{renda2020geometric}. Inspired by~\cite{wang2022control}, we model the strain using a Piecewise Gaussian Curvature (PGC) basis with elongation, defined as follows
\begin{equation*}
\begin{carray}{c}
    \kappa_{x,i}\\
    \kappa_{y,i}\\
    \delta L_{{i}}
\end{carray} = \Phim_{i}(s_{i})\qv_{i} + \begin{carray}{c}
    0\\
    0\\
    1
\end{carray},
\end{equation*}
where
\begin{equation*}\small
     \Phim_{i}(s_{i}) = \frac{1}{L_{0_{i}}} \begin{carray}{ccccc}
        0 & 0 & -1 & -e^{-(s_{i}-L_{0_{i}}/2)^2} & 0 \\
        1 & e^{-(s_{i}-L_{0_{i}}/2)^2} & 0 & 0 & 0\\
        0 & 0 & 0 & 0 & 1
    \end{carray},
\end{equation*}
$s_{i} \in [0, L_{0_{i}}]$ and $\qv_{i} \in \R^{5}$. The robot starts at rest from the initial configuration
\begin{equation*}
    \qv_{0} = \begin{carray}{cccccccccc}
        0.3 & 1 & 0 & 1 & 0.1 & -1 & 0 & 0.1 & 2 & 0
    \end{carray}^{T},
\end{equation*}
and the base is rotated so that in the stress-free configuration the arm is aligned with the gravitational field with the tip pointing upwards. 
The results of the simulation are shown in Fig.~\ref{fig:simulation 2}. After a transient of approximately $5~[\si{\second}]$, the arm reaches a steady-state position consistent with the direction of the gravitational force. 
\subsection{Simulation 6: PCC in free evolution}\label{simulation:3}
\begin{figure}[ht!]
    \centering
    \subfigure[{Configuration variables}]{
        \adjustbox{gstore width=\figureWidth, gstore height=\figureHeight}{
            \includegraphics[width = \columnwidth]
            {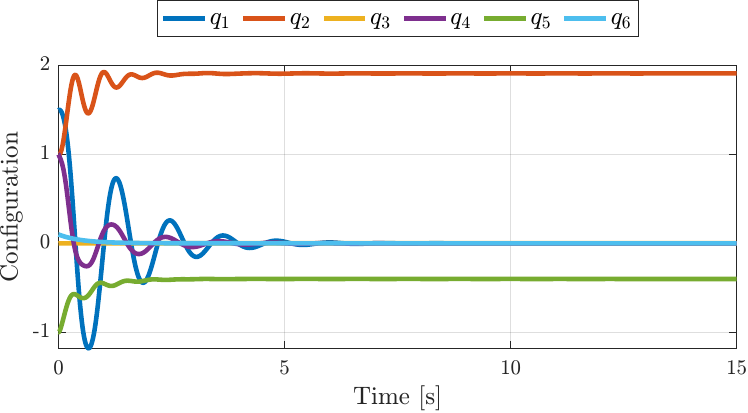}
        \label{fig:simulation 3: configuration variables}
        }
    }
    \subfigure[{Stroboscopic plot}]{
        \includegraphics[width = 0.75\columnwidth]
        {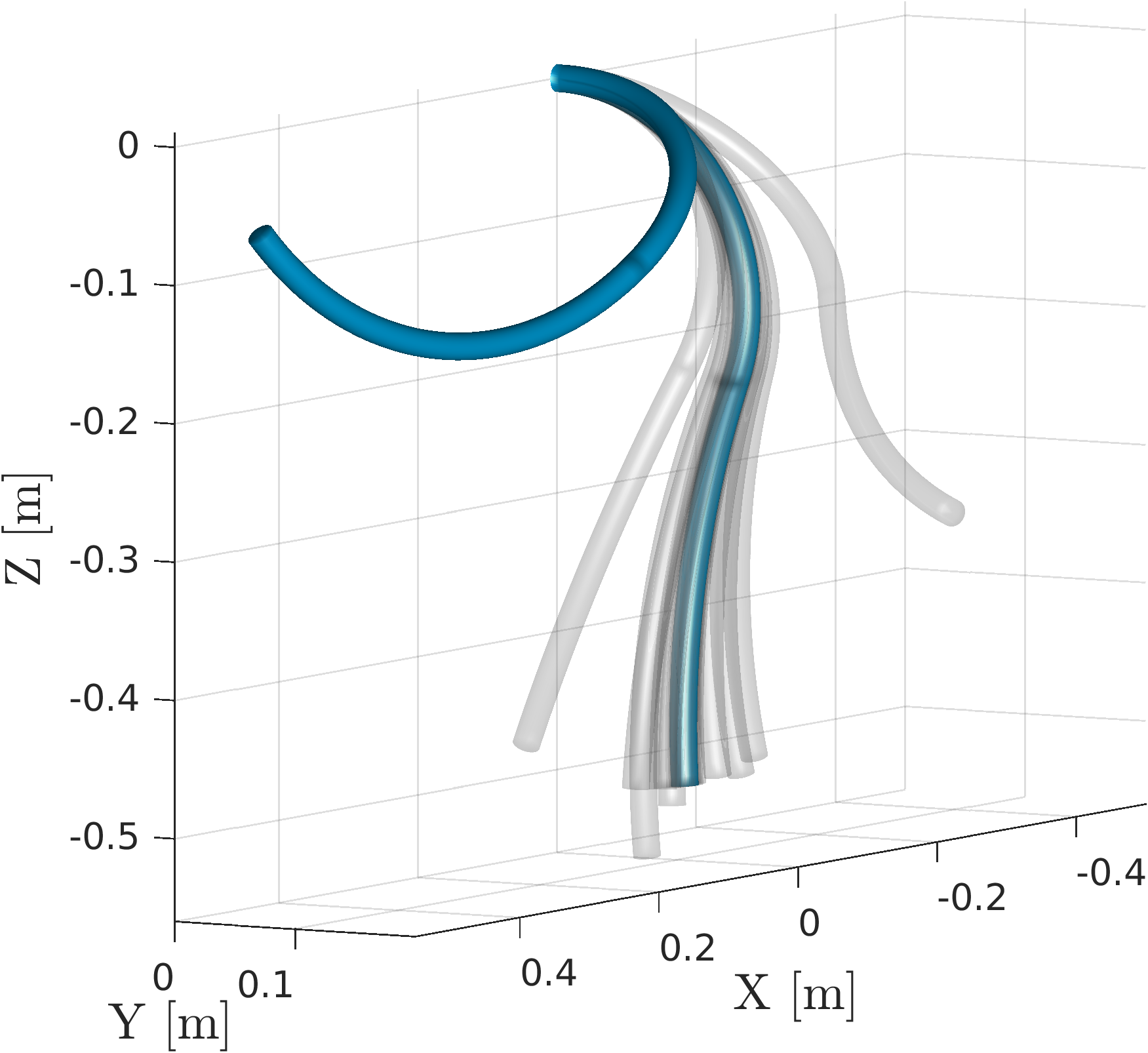}
        \label{fig:simulation 3: strobo plot}
    }
    \caption{\small Simulation 6: PCC in free evolution. Time evolution of the (a)~configuration variables (in $[\si{\radian}]$ and $[\si{\meter}]$) and (b)~stroboscopic plot of the robot motion. The arm consists of two bodies modeled under the PCC hypothesis and moves in free evolution under gravity.}
    \label{fig:simulation 3}
\end{figure}

Consider a soft robot moving in free evolution under gravity. The arm consists of two slender silicone bodies with a cylindrical shape. The base is rotated so that in the reference (straight) configuration the robot extends along the $Y$ axis, with the gravitational force acting in the $Z$ direction. We assume that a PCC model can approximate the motion of each body. Thus, the configuration space of $\calB_{i}$ has dimension three and $\qv_{\calB_{i}} = \qv_{i} = \rb{ \kappa_{x, i} \,\, \kappa_{y, i} \,\, \delta L_{i} }^{T} \in \R^{3}$; $i =1, 2,$ where $\kappa_{x, i}$ and $\kappa_{y, i}$ represent the curvature along $x$ and $y$, respectively, and $ \delta L_{i}$ is the rod elongation. The mass density is $\rho_{i} = \SI{1070}{[\kilogram\per\cubic\meter]}$ and uniformly distributed along the body, whose radius is $R_{i} = \SI{0.01}{[\meter]}$ and rest length $L_{0_{i}} = \SI{0.3}{[\meter]}$. The stress parameters are $C_{i} = \SI{0.555}{[\giga\pascal]}$ and $\eta_{i} = \SI{0.333}{[\second]}$. The robot starts at rest from the initial configuration
\begin{equation*}
    \qv_{0} = \begin{carray}{cccccc}
        1.5 & 1 & 0 & 1 & -1 & 0.1
    \end{carray}^{T}.
\end{equation*}
Fig.~\ref{fig:simulation 3} reports the time evolution of the configuration variables and a stroboscopic motion plot. 
\subsection{Simulation 7: PAC in free evolution}\label{simulation:4}
\begin{figure}[ht!]
    \centering
    \subfigure[{Configuration variables}]{
        \adjustbox{gstore width=\figureWidth, gstore        height=\figureHeight}{
            \includegraphics[width = \columnwidth]
            {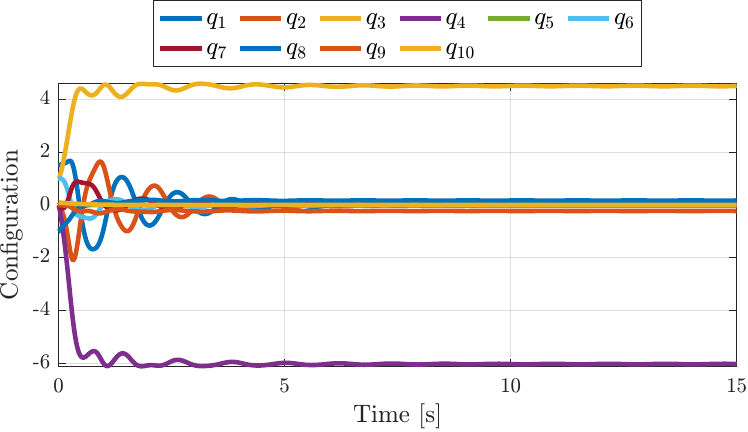}
        \label{fig:simulation 4: configuration variables}
        }
    }
    \subfigure[{Stroboscopic plot}]{
        \includegraphics[width = 0.75\columnwidth]
        {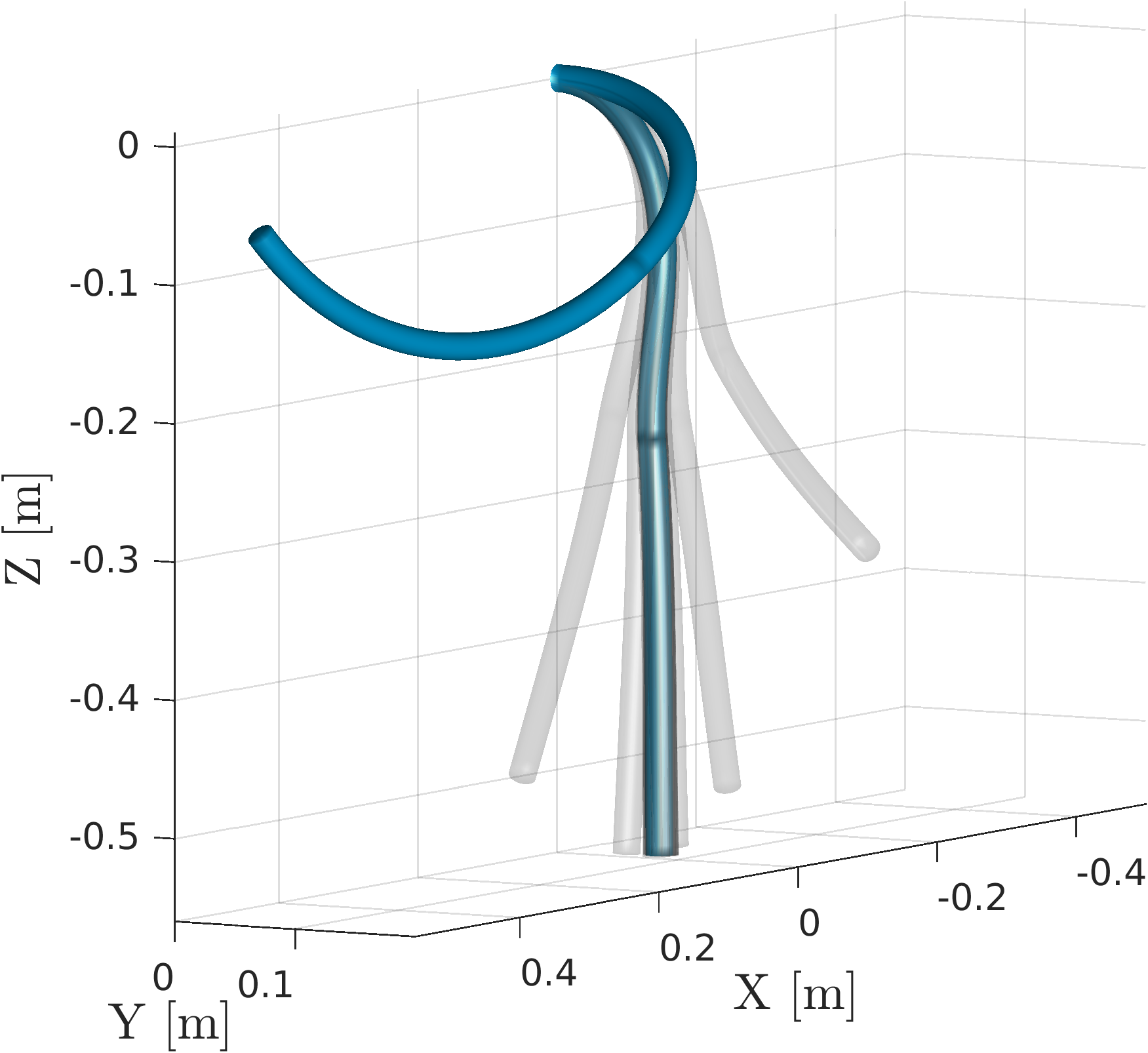}
        \label{fig:simulation 4: strobo plot}
    }
    \caption{\small Simulation 7: PAC in free evolution. Time evolution of the (a)~configuration variables (in $[\si{\radian}]$ and $[\si{\meter}]$) and (b)~stroboscopic plot of a soft robot with the same physical parameters of the robot in Simulation~6 but with an affine model for the curvature strains.}
    \label{fig:simulation 4}
\end{figure}

Consider the same soft robot of Section~\ref{simulation:3}, but with a piecewise affine curvature (PAC) model, i.e., 
\begin{equation*}
\begin{carray}{c}
    \kappa_{x,i}\\
    \kappa_{y,i}
\end{carray}
     = \frac{1}{L_{{0}_{i}}} \begin{carray}{cccc}
        0 & 0 & -1 & -\displaystyle \frac{s_{i}}{L_{0_{i}}}\\
        1 & \displaystyle \frac{s_{i}}{L_{0_{i}}} & 0 & 0
    \end{carray}\begin{carray}{c}
        q_{1,i}\\q_{2,i}\\q_{3,i}\\q_{4,i}
    \end{carray},
\end{equation*}
$s_{i} \in [0, L_{0_{i}}]$. We initialize the affine components to zero to start from the exact spatial configuration of Simulation 6. The outcomes of this simulation are depicted in Fig.~\ref{fig:simulation 4}. The steady-state configuration is different to the previous. However, as expected, the strain attains similar values at the tip of the two segments. 

\subsection{Simulation 8: PCS in free evolution}
\begin{figure}[ht!]
    \centering
    \subfigure[{Configuration variables}]{
        \adjustbox{gstore width=\figureWidth, gstore        height=\figureHeight}{
            \includegraphics[width = \columnwidth]
            {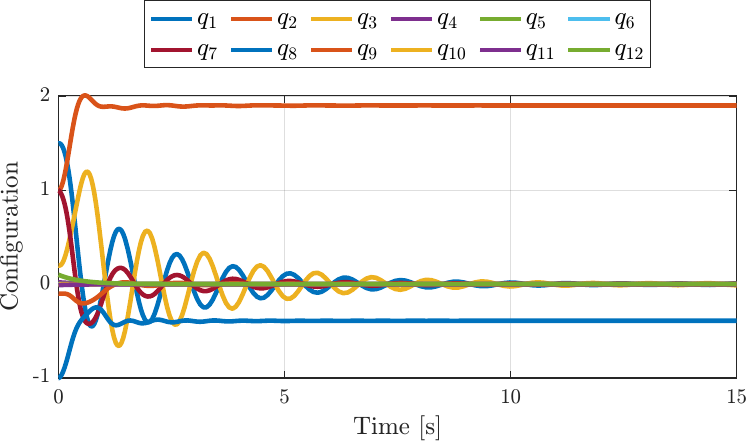}
        \label{fig:simulation 5: configuration variables}
        }
    }
    \subfigure[{Stroboscopic plot}]{
        \includegraphics[width = 0.75\columnwidth]
        {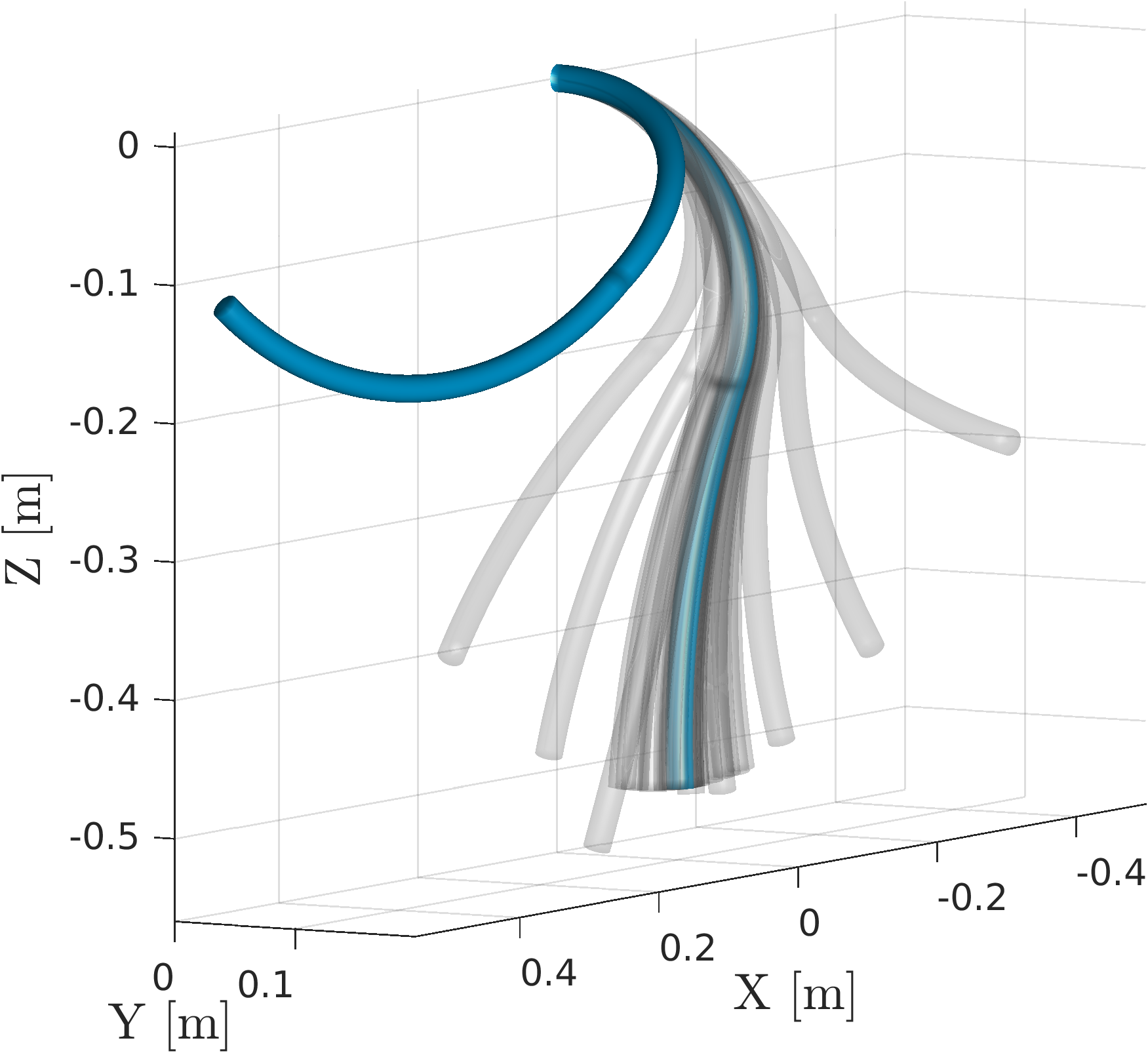}
        \label{fig:simulation 5: strobo plot}
    }
    \caption{\small Simulation 8: PCS in free evolution. Time evolution of the (a)~configuration variables (in $[\si{\radian}]$ and $[\si{\meter}]$) and (b)~stroboscopic plot for a soft robot with same physical parameters of the soft arm in Simulation~6. The configuration space is approximated adopting a PCS model of the strains.}
    \label{fig:simulation 5}
\end{figure}

We now assume that the robot takes configurations where all the strains are non-negligible and well approximated by a piecewise constant strain (PCS) model~\citep{renda2018discrete}. Consequently, the configuration of each body is a six-dimensional vector containing the strains, constant in space but variable in time. The initial configuration is 
\begin{equation*}\arraycolsep=2.8pt
    \qv_{0} = \begin{carray}{cccccccccccc}
        1.5 & 1 & 0.2 & 0.02 & 0 & 0 & 1 & -1 & -0.1 & 0 & -0.01 & 0.1
    \end{carray}^{T},
\end{equation*}
with zero velocity. In Fig.~\ref{fig:simulation 5}, we present the time evolution of the configuration variables and the robot motion. The steady-state value of the curvature variables and elongations is consistent with those observed in Simulation~6.
\subsection{Simulation 9: UR10 with PCC body}
Consider a Universal Robots UR10 collaborative manipulator carrying a passive PCC cylinder of silicone on its tip. The latter has rest length $L_0 = 0.8~[\si{\meter}]$, radius $R = \SI{0.02}{[\meter]}$, mass density $\rho = \SI{1070}{[\kilogram\per\cubic\meter]}$, elastic coefficient $C = \SI{8}{[\giga\pascal]}$ and viscous parameter $\eta = \SI{0.054}{[\second]}$. The task is to control the robot joints to the reference trajectory
\begin{equation*}
    \qv_{a_{d}}(t) = \cos(t)\vect{1}_{6}~[\si{\radian}]. 
\end{equation*}
Partial feedback linearization (PFBL) on the collocated variables is used to fulfill the task. The control law is given by 
\begin{equation}\label{simulation6: complete PFBL}
    \begin{split}
            \nuv_{a} &= (\Mm_{aa} - \Mm_{au}\Mm_{uu}^{-1}\Mm_{au}^{T})\\
            &\quad\cdot [\ddqv_{a_{d}} + \Km_{P}(\qv_{a_{d}}-\qv_{a}) + \Km_{D}(\dqv_{a_{d}}-\dqv_{a})] \\
            & -\cv_{a} - \gv_{a} - \Mm_{au}\Mm_{uu}^{-1} (\cv_{u} + \gv_{u} + \sv_{u}.
    \end{split}
\end{equation}
where the subscripts $a$ and $u$ refer to the actuated and unactuated parts of the dynamics, namely the UR10 robot and soft body, respectively. In the following, the term \textit{complete PBFL} refers to~\eqref{simulation6: complete PFBL}.
To assess how the soft body affects the motion of the UR10, a second simulation is performed with a FBL controller designed on the rigid part only, denoted as \textit{incomplete PBFL}. Both control laws are implemented using the ID algorithm described in~\cite{pustina2023univeral}. 

The control gains are $\Km_{P} = 100 \cdot \IIm_{6}~\si{[\newton\meter]}$ and $\Km_{D} = 50 \cdot \IIm_{6}\si{[\newton\meter\second]}$ and the initial configuration is
\begin{equation*}
    \qv_{0} = \begin{carray}{cccc}
        \zerov_{1 \times 6} & \pi/4 & -\pi/4 & 0
    \end{carray}^{T}.
\end{equation*}

\begin{figure*}
    \centering
    \subfigure[{Configuration variables, complete FBL}]{
        \includegraphics[width = 0.96\columnwidth]
            {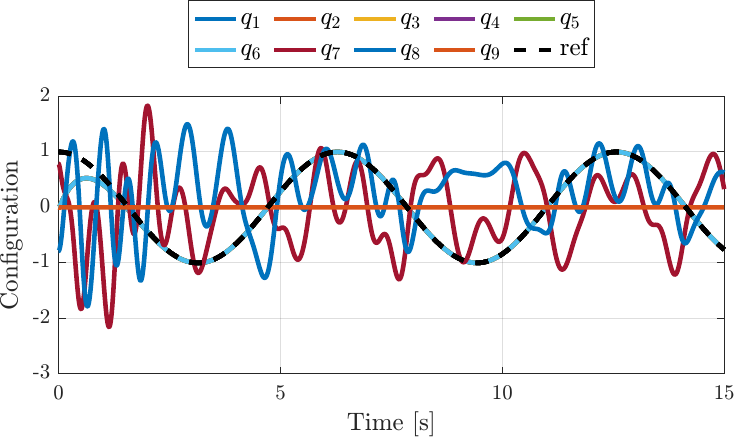}
        \label{fig:simulation 6: configuration variables}
    }
    \subfigure[{Control input, complete PFBL}]{
        \includegraphics[width = 1.0\columnwidth]{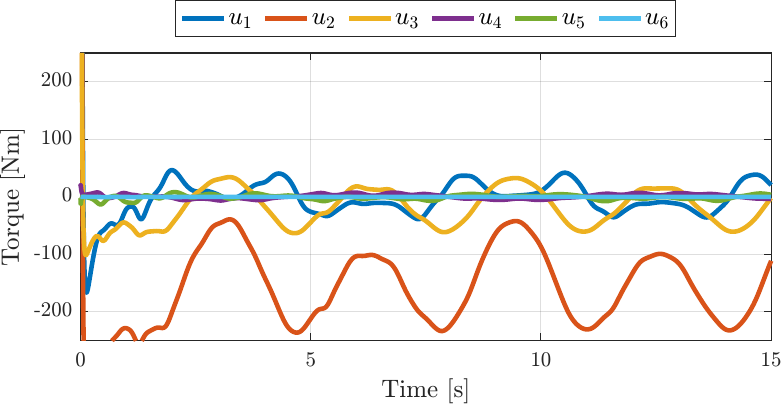}
        \label{fig:simulation 6: control action}
    }\\
    \subfigure[{Configuration variables, incomplete PFBL}]{
        \includegraphics[width = 0.96\columnwidth, trim={0 0 0 1.4cm}, clip]
            {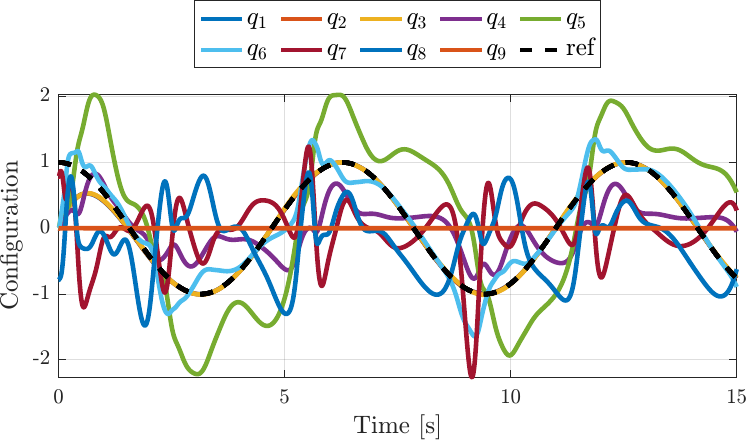}
        \label{fig:simulation 6: configuration variables no full linearization}
    }
    \subfigure[{Control input, incomplete PFBL}]{
        \includegraphics[width = 1.0\columnwidth, trim={0 0 0 0.75cm}, clip]{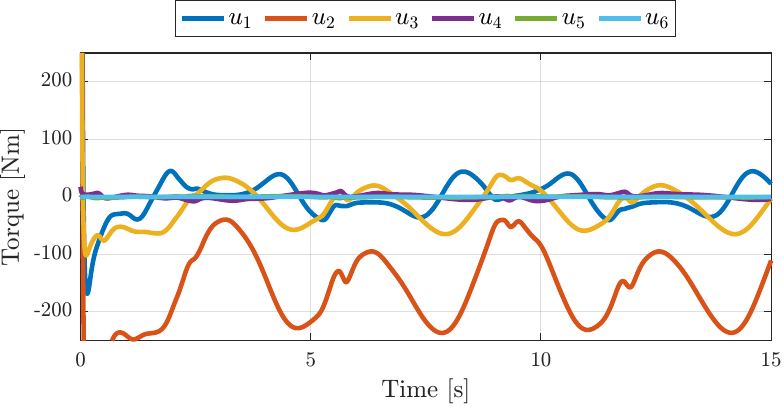}
        \label{fig:simulation 6: control action no full linearization}
    }\\
    \subfigure[{Stroboscopic plot, complete PFBL}]{
        \includegraphics[width = 0.9\columnwidth]
        {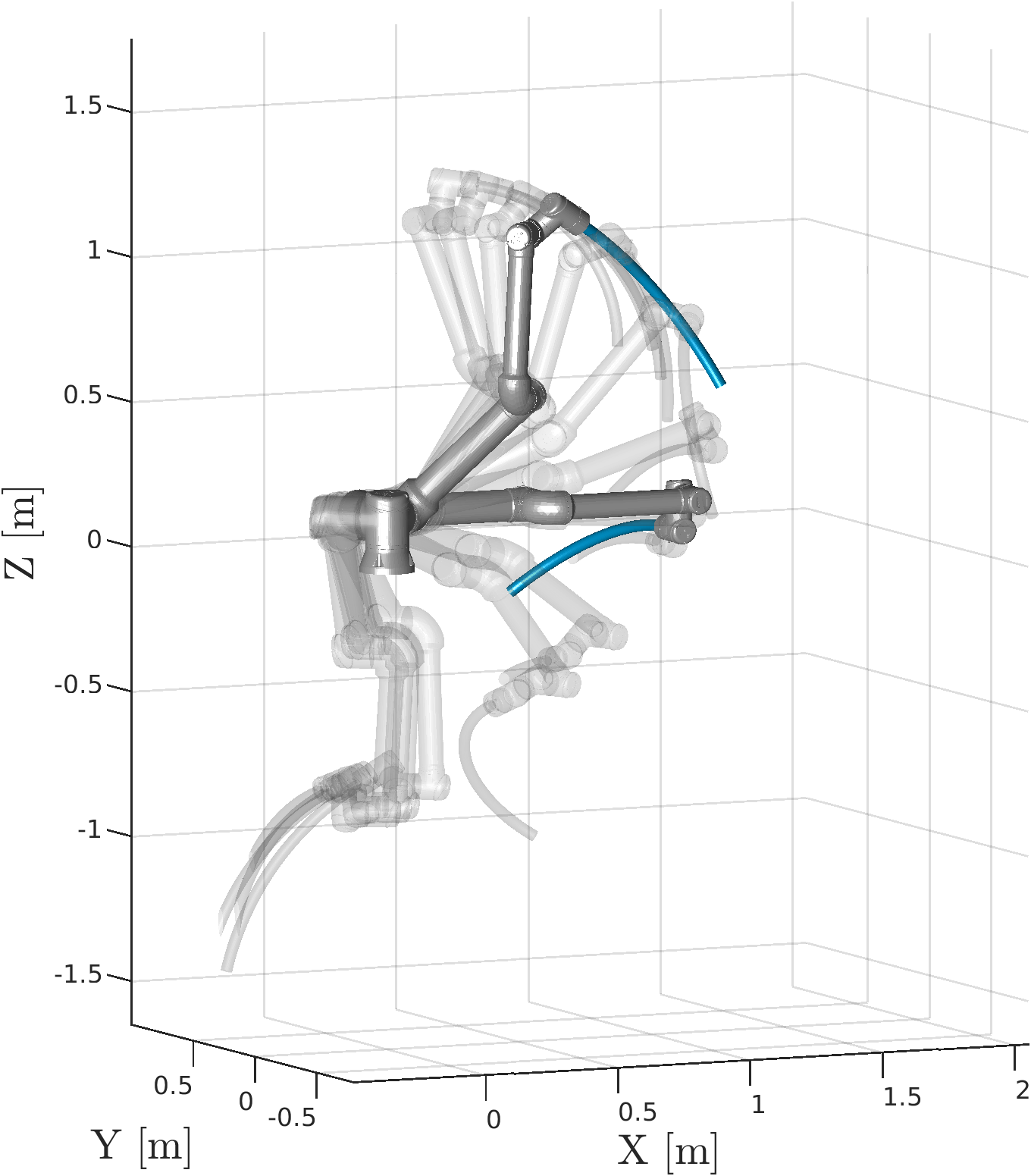}
        \label{fig:simulation 6: strobo plot}
    }
    \subfigure[{Stroboscopic plot, incomplete PFBL}]{
        \includegraphics[width = 0.9\columnwidth]
        {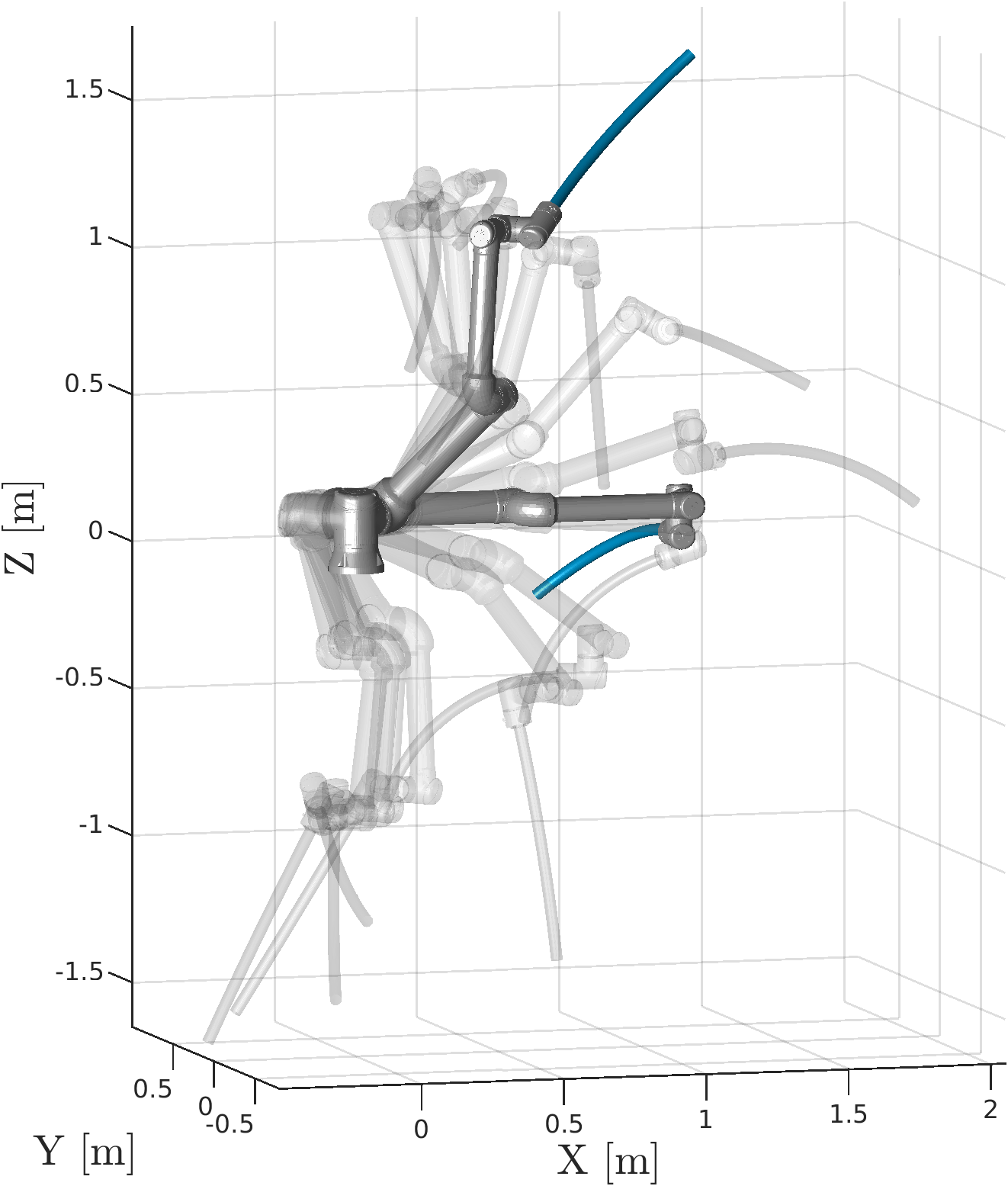}
        \label{fig:simulation 6: strobo plot no full linearization}
    }
    \caption{\small Simulation 9: UR10 with PCC body. Time evolution of the (a)~configuration variables (in $[\si{\radian}]$ and $[\si{\meter}]$) and (b)~control action under the complete PFBL controller~\eqref{simulation6: complete PFBL} for a cosinusoidal reference. Similarly, (c)--(d) show the closed-loop performance under a FBL control law designed on the rigid part only. In (e)--(f), it is possible to see the robot motion for the two controllers.}
    \label{fig:simulation 6}
\end{figure*}

Fig.~\ref{fig:simulation 6} reports the simulation results. When the complete PBFL is used, the configuration of the rigid robot converges exponentially fast to the reference, as expected. Instead, the incomplete PBFL has unsatisfactory performance, especially for the wrist joints, because the dynamic effects of the soft body are stronger at the end-effector. Nonetheless, the closed-loop is stable. In both cases, since the zero dynamics is asymptotically stable, the trajectory of the passive segment remains bounded. Initially, the motion requires a large control action because of the complete cancellation of the dynamics and the initial error. 
\subsection{Simulation 10: Regulation of PCC soft robot}
This last simulation considers the shape regulation problem for a continuum soft robot with three bodies, modeled again under the PCC hypothesis. The base is rotated so that in the straight configuration the arm is aligned with the direction of the gravitational field. The kinematic and dynamic parameters are $L_{0_{i}} = \SI{0.2}{[\meter]}$, $R_{i} = \SI{0.02}{[\meter]}$, $\rho_{i} = \SI{1070}{[\kilogram \per \cubic\meter]}$, $C_{i} = \SI{5.55}{[\giga\pascal]}$ and $\eta_{i} = 0.066~[\si{\second}]$. The robot starts at rest from the straight configuration. 

\begin{figure*}
    \centering
    \subfigure[{Configuration variables}]{
        \adjustbox{gstore width=\figureWidth, gstore        height=\figureHeight}{
            \includegraphics[width = 0.96\columnwidth]
            {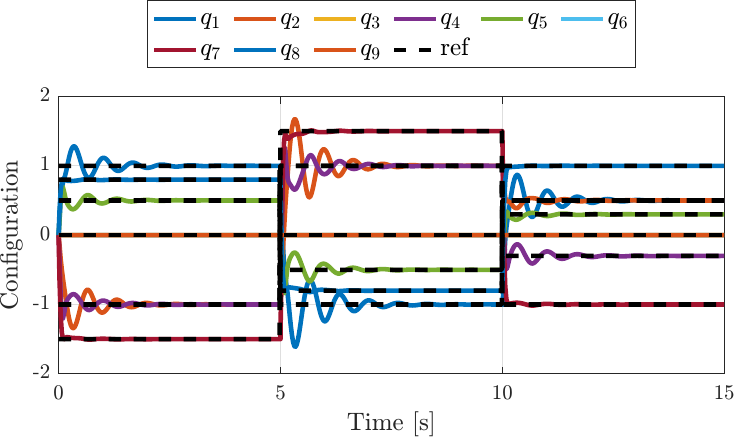}
        \label{fig:simulation 7: configuration variables}
        }
    }
    \subfigure[{Control input}]{
        \includegraphics[width = \figureWidth]{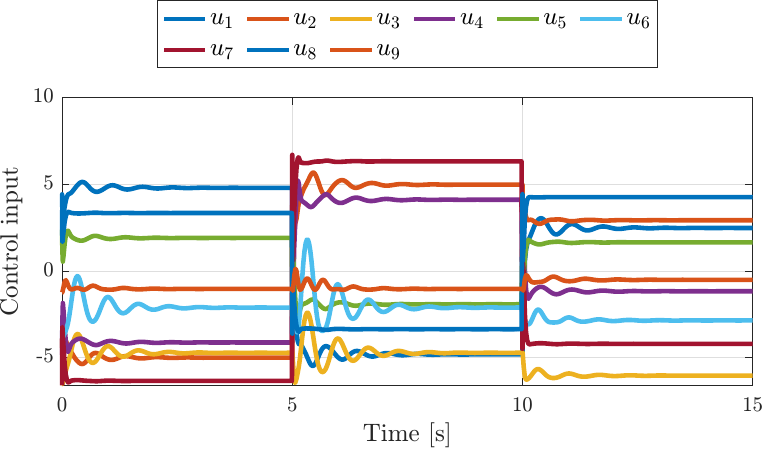}
        \label{fig:simulation 7: control action}
    }\\
    \subfigure[{Stroboscopic plot, $t \in [0;\ 5] \si{[\second]}$}]{
        \includegraphics[ height = 1.1\figureHeight]
        {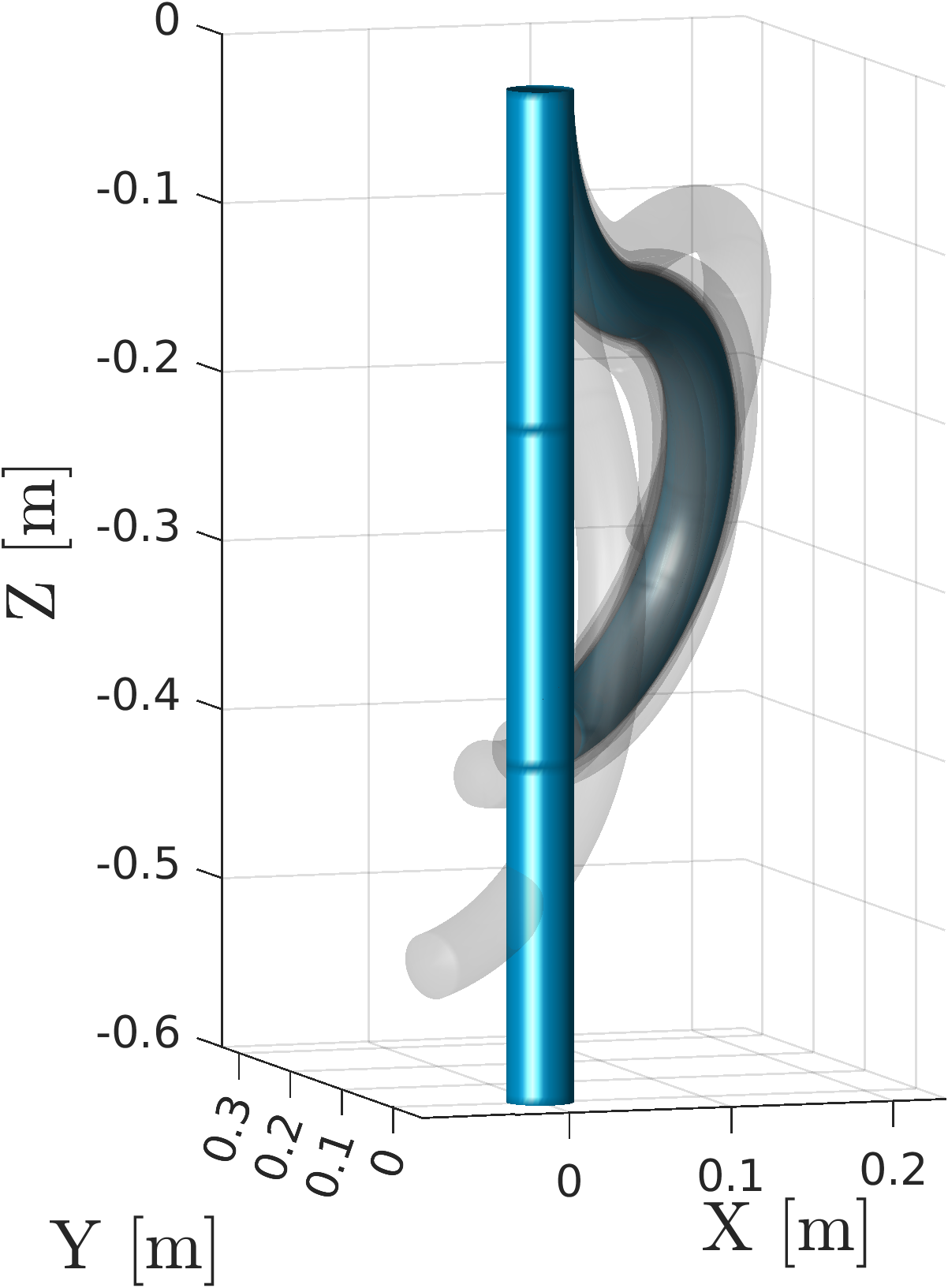}
        \label{fig:simulation 7:strobo 1}
    }
    \subfigure[{Stroboscopic plot, $t \in [5;\ 10] \si{[\second]}$}]{
        \includegraphics[ height = 1.1\figureHeight]
        {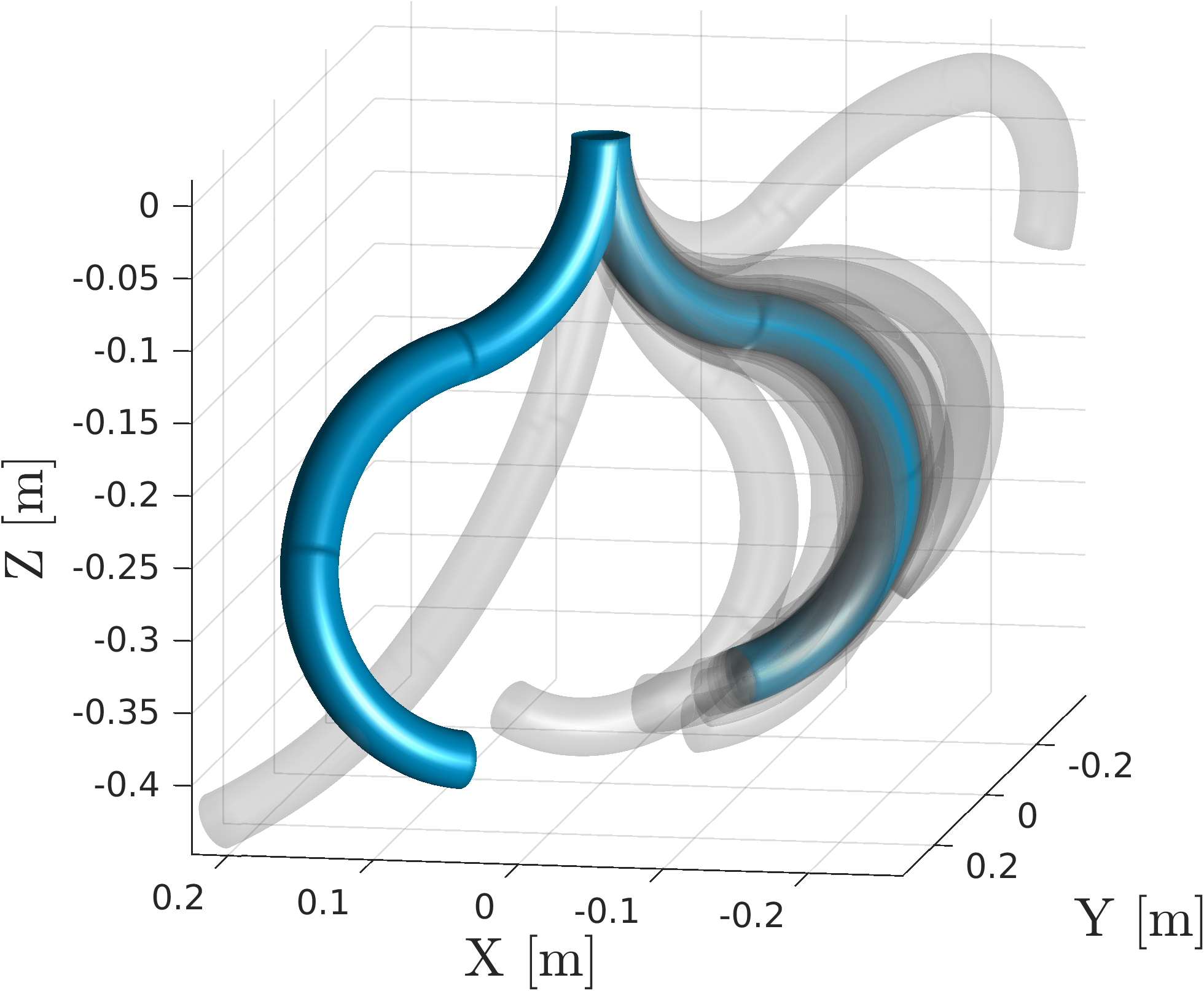}
        \label{fig:simulation 7:strobo 2}
    }
    \subfigure[{Stroboscopic plot, $t \in [10;\ 15] \si{[\second]}$}]{
        \includegraphics[ height = 1.1\figureHeight]
        {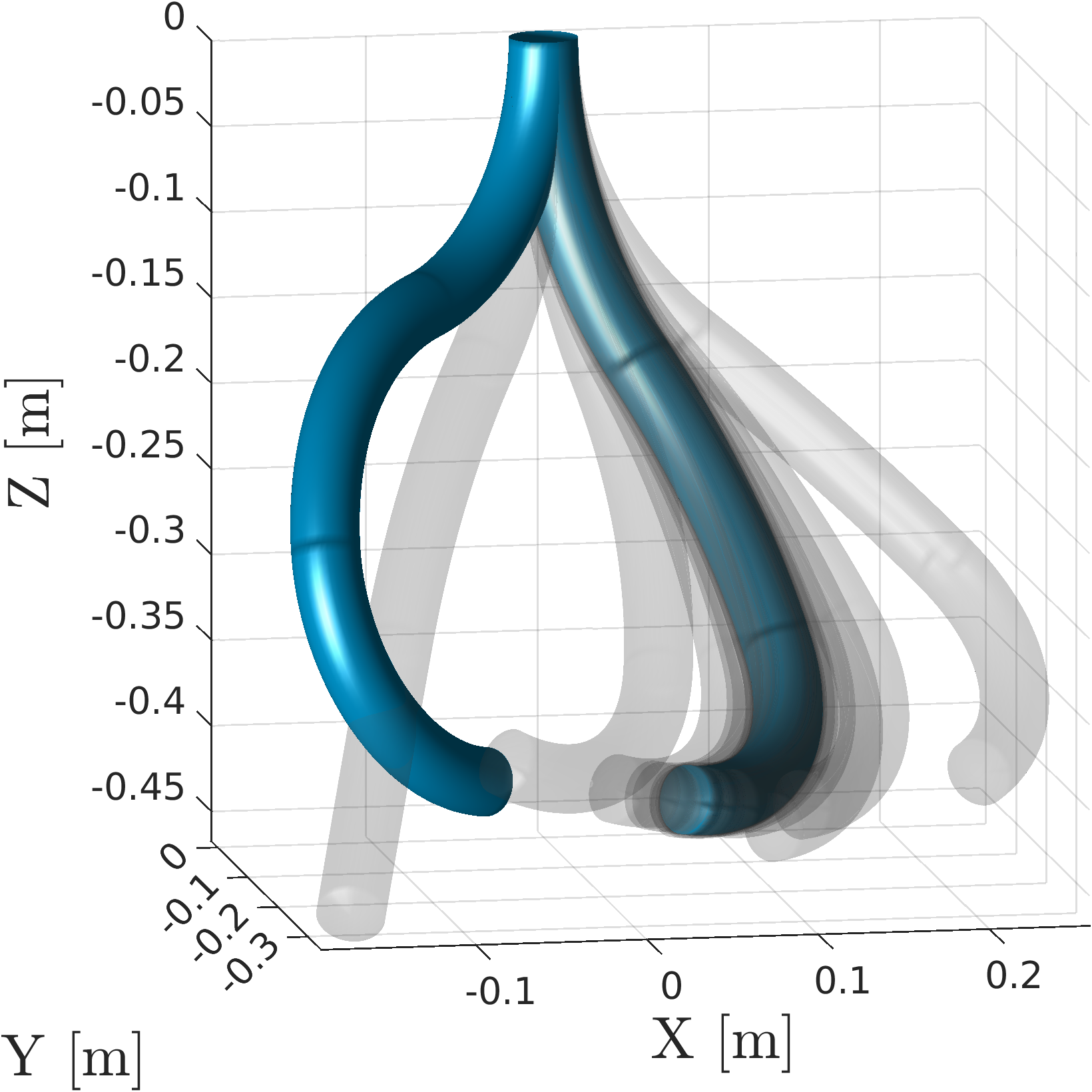}
        \label{fig:simulation 7:strobo 3}
    }
    \caption{\small Simulation 10: Regulation of PCC soft robot. Time evolution of the (a)~configuration variables (in $[\si{\radian}]$ and $[\si{\meter}]$) and (b)~control action (in $[\si{\newton \meter}]$ and $[\si{\newton}]$) under the PD+ regulator~\eqref{eq:simulation:PD+}. Figs.(c)--(e) illustrate the robot motion in its workspace for the three references.}
    \label{fig:simulation 7}
\end{figure*}

Despite oversimplifying the problem, we neglect the actuation model and assume the robot is fully actuated. Under these hypotheses, a simple PD with compensation of the potential forces fulfills the task, namely
\begin{equation}\label{eq:simulation:PD+}
    \nuv = \Km_{P}\rb{ \qv_{d} - \qv} - \Km_{D} \dqv + \gv(\qv_{d}) + \sv(\qv_{d}, \zerov_{n}), 
\end{equation}
where $\qv_{d} \in \R^{9}$ is the desired (constant) reference. The control gains are set as $\Km_{P} = 0.2~[\si{\newton\meter}]([\si{\meter}])$ and $\Km_{D} = 0.1~[\si{\newton\meter\second}]([\si{\newton\second\per\meter}])$, respectively. The reference consists of the following step commands, time spaced $\SI{5}{[\second]}$ each, 
\begin{equation*}
\begin{split}
    \qv_{d_{1}} = \begin{carray}{ccccccccc}
        1 & -1 & 0 & -1 & 0.5 & 0 & -1.5 & 0.8 & 0 
    \end{carray}^{T},\\
    \qv_{d_{2}} = \begin{carray}{ccccccccc}
        -1 & 1 & 0 & 1 & -0.5 & 0 & 1.5 & -0.8 & 0 
    \end{carray}^{T},
\end{split}
\end{equation*}
and
\begin{equation*}
    \qv_{d_{3}} = \begin{carray}{ccccccccc}
        0.5 & 0.5 & 0 & -0.3 & 0.3 & 0 & -1 & 1 & 0 
    \end{carray}^{T}. 
\end{equation*}
The above control law can be easily implemented using Algorithm~2 in~\cite{pustina2023univeral}. The only term depending on the model is the steady-state compensation of the potential forces, which can be computed with a single call to the procedure as $\gv(\qv_{d}) + \sv(\qv_{d}, \zerov_{9}) = \mathrm{ID}(\qv_{d}, \zerov_{9}, \zerov_{9})$. Fig.~\ref{fig:simulation 7} reports the closed-loop results. As expected, the configuration variables quickly converge to the reference values. However, the robot oscillates during transients, as the stroboscopic plot shows. 
\bibliographystyle{SageH}
\bibliography{references.bib}

%% file: boldfacemath.tex
\usepackage{amsmath}
\usepackage{amssymb}
\usepackage{xparse}

\newcommand{\vect}[1]{\boldsymbol{#1}}
\newcommand{\mat}[1]{\boldsymbol{#1}}

\newcommand{\diffs}[3]{\frac{\partial^2 #1}{
		\ifx#2#3 
		\partial #2^2
		\else
		\partial #2 \partial #3
		\fi
}}

\newcommand{\norm}[1]{{\left \| {#1} \right \|}}

\newcommand{\jac}[2]{\Jm_{#2}{\!#1}}

\newcommand{\R}{\mathbb{R}}

\newenvironment{carray}
{ \left( \begin{array}}
	{ \end{array} \right) }

\newcommand{\zerov}{\vect{0}}

\newcommand{\av}{\vect{a}}
\newcommand{\bv}{\vect{b}}
\newcommand{\cv}{\vect{c}}

\newcommand{\dv}{\vect{d}}

\newcommand{\fv}{\vect{f}}
\newcommand{\gv}{\vect{g}}

\newcommand{\hv}{\vect{h}}

\newcommand{\kv}{\vect{k}}
\newcommand{\lv}{\vect{l}}

\newcommand{\nv}{\vect{n}}

\newcommand{\pv}{\vect{p}}

\newcommand{\dpv}{\dot{\vect{p}}}

\newcommand{\ddpv}{\ddot{\vect{p}}}
\newcommand{\qv}{{\vect{q}}}
\newcommand{\dqv}{\dot{\vect{q}}}
\newcommand{\ddqv}{\ddot{\vect{q}}}

\newcommand{\rv}{{\vect{r}}}

\newcommand{\drv}{\dot{\vect{r}}}

\newcommand{\ddrv}{\ddot{\vect{r}}}
\newcommand{\sv}{\vect{s}}

\newcommand{\uv}{\vect{u}}
\newcommand{\vv}{\vect{v}}
\newcommand{\dvv}{\dot{\vect{v}}}

\newcommand{\xv}{\vect{x}}

\newcommand{\yv}{\vect{y}}

\newcommand{\dyv}{\dot{\vect{y}}}
\newcommand{\ddyv}{\ddot{\vect{y}}}

\newcommand{\zv}{\vect{z}}
\newcommand{\dzv}{\dot{\vect{z}}}
\newcommand{\ddzv}{\ddot{\vect{z}}}

\newcommand{\tv}{\vect{t}}

\newcommand{\dtv}{\dot{\vect{t}}}

\newcommand{\alphav}{\vect{\alpha}}

\newcommand{\piv}{\vect{\pi}}

\newcommand{\tauv}{\vect{\tau}}

\newcommand{\thetav}{\vect{\theta}}

\newcommand{\dthetav}{\dot{\vect{\theta}}}

\newcommand{\nuv}{\vect{\nu}}
\newcommand{\omegav}{\vect{\omega}}
\newcommand{\domegav}{\dot{\omegav}}

\newcommand{\Fv}{\vect{F}}

\newcommand{\Tv}{\vect{T}}

\newcommand{\Phim}{\vect{\Phi}}

\newcommand{\IIm}{\mat{I}}

\newcommand{\Am}{\mat{A}}
\newcommand{\Bm}{\mat{B}}

\newcommand{\Fm}{\mat{F}}

\newcommand{\Jm}{\mat{J}}

\newcommand{\Km}{\mat{K}}

\newcommand{\Mm}{\mat{M}}

\newcommand{\Qm}{\mat{Q}}
\newcommand{\Rm}{\mat{R}}
\newcommand{\dRm}{\dot{\Rm}}
\newcommand{\Sm}{\mat{S}}
\newcommand{\Tm}{\mat{T}}

\newcommand{\calJ}{\mathcal{J}}
\newcommand{\calB}{\mathcal{B}}

\newcommand{\calFv}{\vect{\mathcal{F}}}
\newcommand{\calTv}{\vect{\mathcal{T}}}
\newcommand{\calMv}{\vect{\mathcal{M}}}

\newcommand{\com}{\mathrm{CoM}}
\newcommand{\comi}{\mathrm{CoM}_{i}}

\newcommand{\pvone}[1]{ \nabla_{#1}\hspace{-0pt} }
\newcommand{\pvtwo}[2]{ \nabla_{#2}\hspace{-0pt}#1 }
\newcommand{\parv}[2]{\expandafter\ifx\expandafter\relax
	\detokenize{#1}\relax\pvone{#2}\else\pvtwo{#1}{#2}\fi}
\newcommand{\dt}[1]{\frac{\mathrm{d}#1}{\mathrm{d}t}}
\NewDocumentCommand{\Rotm}{ O{} O{} O{} }{{}^{#2}\Rm_{#1}#3}
\NewDocumentCommand{\Transm}{ O{} O{} O{} }{{}^{#2}\Tm_{#1}#3}
\newcommand{\Skew}[1]{\widetilde{#1}}

\newcommand{\drm}{\mathrm{d}}

\newcommand{\rb}[1]{\left( #1 \right)}

\newenvironment{sequation*}
    {\begin{equation*}\small
    }
    { 
    \end{equation*}
    }

%% file: main.bbl
\begin{thebibliography}{94}
\providecommand{\natexlab}[1]{#1}
\providecommand{\url}[1]{\texttt{#1}}
\providecommand{\urlprefix}{URL }
\expandafter\ifx\csname urlstyle\endcsname\relax
  \providecommand{\doi}[1]{DOI:\discretionary{}{}{}#1}\else
  \providecommand{\doi}{DOI:\discretionary{}{}{}\begingroup \urlstyle{rm}\Url}\fi

\bibitem[{Abraham et~al.(2016)Abraham, Marsden and Tudor}]{marsden2016manifolds}
Abraham R, Marsden JE and Tudor R (2016) \emph{Manifolds, Tensor Analysis, and Applications}.
\newblock Springer New York.

\bibitem[{Amirouche and Xie(1993)}]{amirouche1993explicit}
Amirouche FML and Xie M (1993) An explicit matrix formulation of the dynamical equations for flexible multibody systems: {A} recursive approach.
\newblock \emph{Computers \& Structures} 46(2): 311--321.

\bibitem[{Armanini et~al.(2023)Armanini, Boyer, Mathew, Duriez and Renda}]{armanini2023soft}
Armanini C, Boyer F, Mathew AT, Duriez C and Renda F (2023) Soft robots modeling: {A} structured overview.
\newblock \emph{IEEE Trans.\ on Robotics} 39(3): 1728--1748.

\bibitem[{Banerjee(2022)}]{banerjee2022flexible}
Banerjee A (2022) \emph{Flexible Multibody Dynamics: {E}fficient Formulations with Applications}.
\newblock 2nd edition. CRC Press.

\bibitem[{Banerjee and Dickens(1990)}]{banerjee1990dynamics}
Banerjee AK and Dickens JM (1990) Dynamics of an arbitrary flexible body in large rotation and translation.
\newblock \emph{J.\ of Guidance, Control, and Dynamics} 13(2): 221--227.

\bibitem[{Bevilacqua et~al.(2022)Bevilacqua, Soleti, Naso, Rizzello and Motzki}]{bevilacqua2022bio}
Bevilacqua D, Soleti G, Naso D, Rizzello G and Motzki P (2022) Bio-inspired flapping wing antagonist actuation with {SMA} wires.
\newblock In: \emph{Proc.\ Int.\ Conf.\ and Exhibition on New Actuator Systems and Applications}. pp. 1--4.

\bibitem[{Book(1984)}]{book1984recursive}
Book WJ (1984) Recursive {L}agrangian dynamics of flexible manipulator arms.
\newblock \emph{Int.\ J.\ of Robotics Research} 3(3): 87--101.

\bibitem[{Boyer et~al.(2024)Boyer, Gotelli, Tempel, Lebastard, Renda and Briot}]{boyer2024implicit}
Boyer F, Gotelli A, Tempel P, Lebastard V, Renda F and Briot S (2024) Implicit time-integration simulation of robots with rigid bodies and cosserat rods based on a newton–euler recursive algorithm.
\newblock \emph{IEEE Trans.\ on Robotics} 40: 677--696.

\bibitem[{Boyer and Khalil(1998)}]{boyer1998efficient}
Boyer F and Khalil W (1998) An efficient calculation of flexible manipulator inverse dynamics.
\newblock \emph{Int.\ J.\ of Robotics Research} 17(3): 282--293.

\bibitem[{Boyer et~al.(2020)Boyer, Lebastard, Candelier and Renda}]{boyer2020dynamics}
Boyer F, Lebastard V, Candelier F and Renda F (2020) Dynamics of continuum and soft robots: {A} strain parameterization based approach.
\newblock \emph{IEEE Trans.\ on Robotics} 37(3): 847--863.

\bibitem[{Bruder et~al.(2021)Bruder, Fu, Gillespie, Remy and Vasudevan}]{bruder2021data}
Bruder D, Fu X, Gillespie RB, Remy CD and Vasudevan R (2021) Data-driven control of soft robots using {K}oopman operator theory.
\newblock \emph{IEEE Trans. on Robotics} 37(3): 948--961.

\bibitem[{Buondonno and De~Luca(2016)}]{buondonno2016efficient}
Buondonno G and De~Luca A (2016) Efficient computation of inverse dynamics and feedback linearization for {VSA}-based robots.
\newblock \emph{IEEE Robotics and Automation Lett.} 1(2): 908--915.

\bibitem[{Caasenbrood et~al.(2023)Caasenbrood, Pogromsky and Nijmeijer}]{caasenbrood2023control}
Caasenbrood B, Pogromsky A and Nijmeijer H (2023) Control-oriented models for hyperelastic soft robots through differential geometry of curves.
\newblock \emph{Soft Robotics} 10(1): 129--148.

\bibitem[{Caasenbrood et~al.(2024)Caasenbrood, Pogromsky and Nijmeijer}]{caasenbrood2024sorotoki}
Caasenbrood BJ, Pogromsky AY and Nijmeijer H (2024) Sorotoki: {A} {MATLAB} toolkit for design, modeling, and control of soft robots.
\newblock \emph{IEEE Access} 12: 17604--17638.

\bibitem[{Chang et~al.(2023)Chang, Halder, Shih, Naughton, Gazzola and Mehta}]{chang2023energy}
Chang HS, Halder U, Shih CH, Naughton N, Gazzola M and Mehta PG (2023) Energy-shaping control of a muscular octopus arm moving in three dimensions.
\newblock \emph{Proc. of the Royal Society A} 479(2270): 20220593.

\bibitem[{Chen and Wang(2020)}]{chen2020design}
Chen F and Wang MY (2020) Design optimization of soft robots: {A} review of the state of the art.
\newblock \emph{IEEE Robotics \& Automation Mag.} 27(4): 27--43.

\bibitem[{Chen(2001)}]{chen2001dynamic}
Chen W (2001) Dynamic modeling of multi-link flexible robotic manipulators.
\newblock \emph{Computers \& Structures} 79(2): 183--195.

\bibitem[{Coevoet et~al.(2017)Coevoet, Morales-Bieze, Largilliere, Zhang, Thieffry, Sanz-Lopez, Carrez, Marchal, Goury, Dequidt et~al.}]{coevoet2017software}
Coevoet E, Morales-Bieze T, Largilliere F, Zhang Z, Thieffry M, Sanz-Lopez M, Carrez B, Marchal D, Goury O, Dequidt J et~al. (2017) Software toolkit for modeling, simulation, and control of soft robots.
\newblock \emph{Advanced Robotics} 31(22): 1208--1224.

\bibitem[{Davis and Rabinowitz(2007)}]{davis2007methods}
Davis PJ and Rabinowitz P (2007) \emph{Methods of Numerical Integration}.
\newblock Courier Corporation.

\bibitem[{De~Luca and Book(2016)}]{deluca2016robots}
De~Luca A and Book WJ (2016) Robots with flexible elements.
\newblock In: Siciliano B and Khatib O (eds.) \emph{Springer Handbook of Robotics}, 2nd edition. Springer, pp. 243--282.

\bibitem[{Della~Santina and Albu-Schaeffer(2020)}]{della2020exciting}
Della~Santina C and Albu-Schaeffer A (2020) Exciting efficient oscillations in nonlinear mechanical systems through eigenmanifold stabilization.
\newblock \emph{IEEE Control Systems Lett.} 5(6): 1916--1921.

\bibitem[{Della~Santina et~al.(2019)Della~Santina, Bicchi and Rus}]{dellasantina2019constraints}
Della~Santina C, Bicchi A and Rus D (2019) Dynamic control of soft robots with internal constraints in the presence of obstacles.
\newblock In: \emph{Proc.\ IEEE/RSJ Int.\ Conf.\ on Intelligent Robots and Systems}. pp. 6622--6629.

\bibitem[{Della~Santina et~al.(2020{\natexlab{a}})Della~Santina, Catalano and Bicchi}]{della2020soft}
Della~Santina C, Catalano MG and Bicchi A (2020{\natexlab{a}}) Soft robots.
\newblock In: Ang M, Khatib O and Siciliano B (eds.) \emph{Encyclopedia of Robotics}. Springer, pp. 1--15.

\bibitem[{Della~Santina et~al.(2023)Della~Santina, Duriez and Rus}]{dellasantina2023survey}
Della~Santina C, Duriez C and Rus D (2023) Model-based control of soft robots: {A} survey of the state of the art and open challenges.
\newblock \emph{IEEE Control Systems Mag.} 43(3): 30--65.

\bibitem[{Della~Santina et~al.(2020{\natexlab{b}})Della~Santina, Katzschmann, Bicchi and Rus}]{della2020model}
Della~Santina C, Katzschmann RK, Bicchi A and Rus D (2020{\natexlab{b}}) Model-based dynamic feedback control of a planar soft robot: {T}rajectory tracking and interaction with the environment.
\newblock \emph{Int.\ J.\ of Robotics Research} 39(4): 490--513.

\bibitem[{Della~Santina and Rus(2019)}]{della2019control}
Della~Santina C and Rus D (2019) Control oriented modeling of soft tobots: {T}he polynomial curvature case.
\newblock \emph{IEEE Robotics and Automation Lett.} 5(2): 290--298.

\bibitem[{Dinev et~al.(2018)Dinev, Liu, Li, Thomaszewski and Kavan}]{dinev2018fepr}
Dinev D, Liu T, Li J, Thomaszewski B and Kavan L (2018) {FEPR}: {F}ast energy projection for real-time simulation of deformable objects.
\newblock \emph{ACM Trans. on Graphics} 37(4): 1--12.

\bibitem[{Dubied et~al.(2022)Dubied, Michelis, Spielberg and Katzschmann}]{dubied2022sim}
Dubied M, Michelis MY, Spielberg A and Katzschmann RK (2022) Sim-to-real for soft robots using differentiable {FEM}: {R}ecipes for meshing, damping, and actuation.
\newblock \emph{IEEE Robotics and Automation Lett.} 7(2): 5015--5022.

\bibitem[{Featherstone(2014)}]{featherstone2014rigid}
Featherstone R (2014) \emph{Rigid Body Dynamics Algorithms}.
\newblock Springer.

\bibitem[{Ferrentino et~al.(2023)Ferrentino, Roels, Brancart, Terryn, Van~Assche and Vanderborght}]{ferrentino2023finite}
Ferrentino P, Roels E, Brancart J, Terryn S, Van~Assche G and Vanderborght B (2023) Finite element analysis-based soft robotic modeling: {S}imulating a soft actuator in {SOFA}.
\newblock \emph{IEEE Robotics \& Automation Mag.} 31(3): 97--105.

\bibitem[{Ferrolho et~al.(2021)Ferrolho, Ivan, Merkt, Havoutis and Vijayakumar}]{ferrolho2021inverse}
Ferrolho H, Ivan V, Merkt W, Havoutis I and Vijayakumar S (2021) Inverse dynamics vs. forward dynamics in direct transcription formulations for trajectory optimization.
\newblock In: \emph{Proc. IEEE Int. Conf. on Robotics and Automation}. pp. 12752--12758.

\bibitem[{Gaz et~al.(2019)Gaz, Cognetti, Oliva, Robuffo~Giordano and De~Luca}]{gaz2019dynamic}
Gaz C, Cognetti M, Oliva A, Robuffo~Giordano P and De~Luca A (2019) Dynamic identification of the {F}ranka {E}mika {P}anda robot with retrieval of feasible parameters using penalty-based optimization.
\newblock \emph{IEEE Robotics and Automation Lett.} 4(4): 4147--4154.

\bibitem[{Godage et~al.(2019)Godage, Webster and Walker}]{godage2019center}
Godage IS, Webster RJ and Walker ID (2019) Center-of-gravity-based approach for modeling dynamics of multisection continuum arms.
\newblock \emph{IEEE Trans.\ on Robotics} 35(5): 1097--1108.

\bibitem[{Grassmann et~al.(2022)Grassmann, Chen, Liang and Burgner-Kahrs}]{grassmann2022dataset}
Grassmann RM, Chen RZ, Liang N and Burgner-Kahrs J (2022) A dataset and benchmark for learning the kinematics of concentric tube continuum robots.
\newblock In: \emph{Proc.\ IEEE/RSJ Int.\ Conf.\ on Intelligent Robots and Systems}. pp. 9550--9557.

\bibitem[{Grazioso et~al.(2019)Grazioso, Di~Gironimo and Siciliano}]{grazioso2019geometrically}
Grazioso S, Di~Gironimo G and Siciliano B (2019) A geometrically exact model for soft continuum robots: {T}he finite element deformation space formulation.
\newblock \emph{Soft Robotics} 6(6): 790--811.

\bibitem[{Guan et~al.(2023)Guan, Stella, Della~Santina, Leng and Hughes}]{guan2023trimmed}
Guan Q, Stella F, Della~Santina C, Leng J and Hughes J (2023) Trimmed helicoids: {A}n architectured soft structure yielding soft robots with high precision, large workspace, and compliant interactions.
\newblock \emph{npj Robotics} 1: 4.

\bibitem[{Han et~al.(2021)Han, Liu, He and Li}]{han2021distributed}
Han Z, Liu Z, He W and Li G (2021) Distributed parameter modeling and boundary control of an octopus tentacle-inspired soft robot.
\newblock \emph{IEEE Trans.\ on Control Systems Technology} 30(3): 1244--1256.

\bibitem[{Hoeijmakers et~al.(2022)Hoeijmakers, Koerkamp, de~Santana, Venner, Stramigioli, Mulder, Brentjes, Gijsman and Hartman}]{hoeijmakers2022investigation}
Hoeijmakers HWM, Koerkamp LHG, de~Santana LD, Venner CH, Stramigioli S, Mulder JL, Brentjes A, Gijsman F and Hartman SA (2022) Investigation flapping-flight aerodynamics of a robotic bird.
\newblock In: \emph{Proc.\ 33th Congr.\ Int. Council of the Aeronautical Sciences}. pp. 3326--3351.

\bibitem[{Hollerbach(1980)}]{hollerbach1980recursive}
Hollerbach JM (1980) A recursive {L}agrangian formulation of maniputator dynamics and a comparative study of dynamics formulation complexity.
\newblock \emph{IEEE Trans.\ on Systems, Man, and Cybernetics} 10(11): 730--736.

\bibitem[{Hu et~al.(2020)Hu, Wang, Cheng and Bao}]{hu2020origami}
Hu F, Wang W, Cheng J and Bao Y (2020) Origami spring--inspired metamaterials and robots: {A}n attempt at fully programmable robotics.
\newblock \emph{Science Progress} 103(3): 0036850420946162.

\bibitem[{Jensen et~al.(2022)Jensen, Johnson, Lindberg and Killpack}]{jensen2022tractable}
Jensen SW, Johnson CC, Lindberg AM and Killpack MD (2022) Tractable and intuitive dynamic model for soft robots via the recursive {N}ewton-{E}uler algorithm.
\newblock In: \emph{Proc.\ IEEE Int.\ Conf.\ on Soft Robotics}. pp. 416--422.

\bibitem[{Kane and Levinson(1983)}]{kane1983use}
Kane TR and Levinson DA (1983) The use of {K}ane's dynamical equations in robotics.
\newblock \emph{Int. J. of Robotics Research} 2(3): 3--21.

\bibitem[{Khalil et~al.(2017)Khalil, Boyer and Morsli}]{khalil2017general}
Khalil W, Boyer F and Morsli F (2017) General dynamic algorithm for floating base tree structure robots with flexible joints and links.
\newblock \emph{ASME J.\ of Mechanisms and Robotics} 9(3): 031003.

\bibitem[{Korayem et~al.(2014)Korayem, Shafei and Dehkordi}]{korayem2014systematic}
Korayem MH, Shafei AM and Dehkordi SF (2014) Systematic modeling of a chain of {N}-flexible link manipulators connected by revolute--prismatic joints using recursive {G}ibbs-{A}ppell formulation.
\newblock \emph{Archive of Applied Mechanics} 84: 187--206.

\bibitem[{Lacarbonara(2013)}]{lacarbonara2013nonlinear}
Lacarbonara W (2013) \emph{Nonlinear Structural Mechanics: Theory, Dynamical Phenomena and Modeling}.
\newblock Springer.

\bibitem[{Lee et~al.(2009)Lee, Noh, Lee and Park}]{lee2009development}
Lee S, Noh S, Lee Y and Park JH (2009) Development of bio-mimetic robot hand using parallel mechanisms.
\newblock In: \emph{Proc.\ IEEE Int.\ Conf.\ on Robotics and Biomimetics}. pp. 550--555.

\bibitem[{Li et~al.(2021)Li, Shintake and Hayashibe}]{li2021deep}
Li G, Shintake J and Hayashibe M (2021) Deep reinforcement learning framework for underwater locomotion of soft robot.
\newblock In: \emph{Proc.\ IEEE Int.\ Conf.\ on Robotics and Automation}. pp. 12033--12039.

\bibitem[{Li et~al.(2024)Li, Donato, Lomonaco and Falotico}]{li2024continual}
Li L, Donato E, Lomonaco V and Falotico E (2024) Continual policy distillation of reinforcement learning-based controllers for soft robotic in-hand manipulation.
\newblock In: \emph{Proc.\ IEEE Int. Conf.\ on Soft Robotics}. pp. 1026--1033.

\bibitem[{Li et~al.(2022)Li, Wang and Kwok}]{li2022towards}
Li Y, Wang X and Kwok KW (2022) Towards adaptive continuous control of soft robotic manipulator using reinforcement learning.
\newblock In: \emph{Proc.\ IEEE/RSJ Int.\ Conf.\ on Intelligent Robots and Systems}. pp. 7074--7081.

\bibitem[{Lilly and Orin(1990)}]{lilly1990n}
Lilly K and Orin D (1990) O({N}) recursive algorithm for the operational space inertia matrix of a robot manipulator.
\newblock \emph{IFAC Proceedings Volumes} 23(8): 275--279.

\bibitem[{Lin et~al.(2022)Lin, Jiang and Shang}]{lin2022emerging}
Lin Z, Jiang T and Shang J (2022) The emerging technology of biohybrid micro-robots: {A} review.
\newblock \emph{Bio-Design and Manufacturing} 5: 107--132.

\bibitem[{Liu et~al.(2023{\natexlab{a}})Liu, Bai, Li and Fantuzzi}]{liu2023large}
Liu TW, Bai JB, Li SL and Fantuzzi N (2023{\natexlab{a}}) Large deformation and failure analysis of the corrugated flexible composite skin for morphing wing.
\newblock \emph{Engineering Structures} 278: 115463.

\bibitem[{Liu et~al.(2023{\natexlab{b}})Liu, Li, Duan, Wu, Wang, Fan, Lin and Hu}]{liu2023hierarchical}
Liu Y, Li T, Duan J, Wu X, Wang H, Fan Q, Lin J and Hu Y (2023{\natexlab{b}}) On a hierarchical adaptive and robust inverse dynamic control strategy with experiment for robot manipulators under uncertainties.
\newblock \emph{Control Engineering Practice} 138: 105604.

\bibitem[{Longhini et~al.(2023)Longhini, Moletta, Reichlin, Welle, Held, Erickson and Kragic}]{longhini2023edo}
Longhini A, Moletta M, Reichlin A, Welle MC, Held D, Erickson Z and Kragic D (2023) {EDO}-net: {L}earning elastic properties of deformable objects from graph dynamics.
\newblock In: \emph{Proc.\ IEEE Int.\ Conf.\ on Robotics and Automation}. pp. 3875--3881.

\bibitem[{Mathew et~al.(2024)Mathew, Feliu-Talegon, Alkayas, Boyer and Renda}]{anup2024reduced}
Mathew AT, Feliu-Talegon D, Alkayas AY, Boyer F and Renda F (2024) Reduced order modeling of hybrid soft-rigid robots using global, local, and state-dependent strain parameterization.
\newblock \emph{Int.\ J.\ of Robotics Research} 44(1): 129--154.

\bibitem[{Mathew et~al.(2022)Mathew, Hmida, Armanini, Boyer and Renda}]{mathew2022sorosim}
Mathew AT, Hmida IB, Armanini C, Boyer F and Renda F (2022) {SoRoSim}: {A} {MATLAB} toolbox for hybrid rigid--soft robots based on the geometric variable-strain approach.
\newblock \emph{IEEE Robotics \& Automation Mag.} 30(3): 106--122.

\bibitem[{My et~al.(2019)My, Bien, Le and Packianather}]{my2019efficient}
My CA, Bien DX, Le CH and Packianather M (2019) An efficient finite element formulation of dynamics for a flexible robot with different type of joints.
\newblock \emph{Mechanism and Machine Theory} 134: 267--288.

\bibitem[{Ménager et~al.(2023)Ménager, Navez, Goury and Duriez}]{menager2023direct}
Ménager E, Navez T, Goury O and Duriez C (2023) Direct and inverse modeling of soft robots by learning a condensed {FEM} model.
\newblock In: \emph{Proc.\ IEEE Int.\ Conf.\ on Robotics and Automation}. pp. 530--536.

\bibitem[{Pinskier and Howard(2022)}]{pinskier2022bioinspiration}
Pinskier J and Howard D (2022) From bioinspiration to computer generation: {D}evelopments in autonomous soft robot design.
\newblock \emph{Advanced Intelligent Systems} 4(1): 2100086.

\bibitem[{Pu et~al.(1996)Pu, M{\"u}ller, Abdalla, Abdelatif, Bark and {Nour Eldin}}]{pu1996parallel}
Pu H, M{\"u}ller M, Abdalla E, Abdelatif L, Bark E and {Nour Eldin} H (1996) Parallel computation of the inertia matrix of a tree type robot using one directional recursion of {N}ewton-{E}uler formulation.
\newblock \emph{J.\ of Intelligent and Robotic Systems} 15: 33--39.

\bibitem[{Pustina et~al.(2024)Pustina, Della~Santina, Boyer, De~Luca and Renda}]{pustina2024input}
Pustina P, Della~Santina C, Boyer F, De~Luca A and Renda F (2024) Input decoupling of {L}agrangian systems via coordinate transformation: {G}eneral characterization and its application to soft robotics.
\newblock \emph{IEEE Trans.\ on Robotics} 40: 2098--2110.

\bibitem[{Pustina et~al.(2022)Pustina, Della~Santina and De~Luca}]{pustina2022feedback}
Pustina P, Della~Santina C and De~Luca A (2022) Feedback regulation of elastically decoupled underactuated soft robots.
\newblock \emph{IEEE Robotics and Automation Lett.} 7(2): 4512--4519.

\bibitem[{Qin et~al.(2024)Qin, Peng, Huang, Liu and Huang}]{qin2024modeling}
Qin L, Peng H, Huang X, Liu M and Huang W (2024) Modeling and simulation of dynamics in soft robotics: {A} review of numerical approaches.
\newblock \emph{Current Robotics Reports} 5: 1--13.

\bibitem[{Renda et~al.(2020)Renda, Armanini, Lebastard, Candelier and Boyer}]{renda2020geometric}
Renda F, Armanini C, Lebastard V, Candelier F and Boyer F (2020) A geometric variable-strain approach for static modeling of soft manipulators with tendon and fluidic actuation.
\newblock \emph{IEEE Robotics and Automation Lett.} 5(3): 4006--4013.

\bibitem[{Renda et~al.(2022)Renda, Armanini, Mathew and Boyer}]{renda2022geometrically}
Renda F, Armanini C, Mathew A and Boyer F (2022) Geometrically-exact inverse kinematic control of soft manipulators with general threadlike actuators’ routing.
\newblock \emph{IEEE Robotics and Automation Lett.} 7(3): 7311--7318.

\bibitem[{Renda et~al.(2018)Renda, Boyer, Dias and Seneviratne}]{renda2018discrete}
Renda F, Boyer F, Dias J and Seneviratne L (2018) Discrete {C}osserat approach for multisection soft manipulator dynamics.
\newblock \emph{IEEE Trans.\ on Robotics} 34(6): 1518--1533.

\bibitem[{Renda and Seneviratne(2018)}]{renda2018geometric}
Renda F and Seneviratne L (2018) A geometric and unified approach for modeling soft-rigid multi-body systems with lumped and distributed degrees of freedom.
\newblock In: \emph{Proc. IEEE Int. Conf. on Robotics and Automation}. pp. 1567--1574.

\bibitem[{Rodriguez and Kreutz-Delgado(1992)}]{rodriguez1992spatial}
Rodriguez G and Kreutz-Delgado K (1992) Spatial operator factorization and inversion of the manipulator mass matrix.
\newblock \emph{IEEE Trans. on Robotics and Automation} 8(1): 65--76.

\bibitem[{Rone and Ben-Tzvi(2013)}]{rone2013continuum}
Rone WS and Ben-Tzvi P (2013) Continuum robot dynamics utilizing the principle of virtual power.
\newblock \emph{IEEE Trans. on Robotics} 30(1): 275--287.

\bibitem[{Rus and Tolley(2015)}]{rus2015design}
Rus D and Tolley MT (2015) Design, fabrication and control of soft robots.
\newblock \emph{Nature} 521(7553): 467--475.

\bibitem[{Russo et~al.(2023)Russo, Sadati, Dong, Mohammad, Walker, Bergeles, Xu and Axinte}]{russo2023continuum}
Russo M, Sadati SMH, Dong X, Mohammad A, Walker ID, Bergeles C, Xu K and Axinte DA (2023) Continuum robots: {A}n overview.
\newblock \emph{Advanced Intelligent Systems} 5(5): 2200367.

\bibitem[{Sadati et~al.(2020)Sadati, Naghibi, Da~Cruz and Bergeles}]{sadati2020reduced}
Sadati S, Naghibi SE, Da~Cruz L and Bergeles C (2020) Reduced-order modeling and model order reduction for soft robots.
\newblock \emph{ResearchGate Preprint} \doi{10.13140/RG.2.2.24115.86568}.

\bibitem[{Sadati et~al.(2021)Sadati, Naghibi, Shiva, Michael, Renson, Howard, Rucker, Althoefer, Nanayakkara, Zschaler, Bergeles, Hauser and Walker}]{sadati2021tmtdyn}
Sadati SH, Naghibi SE, Shiva A, Michael B, Renson L, Howard M, Rucker CD, Althoefer K, Nanayakkara T, Zschaler S, Bergeles C, Hauser H and Walker ID (2021) {TMTD}yn: {A} {Matlab} package for modeling and control of hybrid rigid–continuum robots based on discretized lumped systems and reduced-order models.
\newblock \emph{Int.\ J.\ of Robotics Research} 40(1): 296--347.

\bibitem[{Sadati et~al.(2017)Sadati, Naghibi, Walker, Althoefer and Nanayakkara}]{sadati2017control}
Sadati SH, Naghibi SE, Walker ID, Althoefer K and Nanayakkara T (2017) Control space reduction and real-time accurate modeling of continuum manipulators using {Ritz} and {R}itz--{G}alerkin methods.
\newblock \emph{IEEE Robotics and Automation Lett.} 3(1): 328--335.

\bibitem[{Santiba{\~n}ez and Kelly(2001)}]{santibanez2001pd}
Santiba{\~n}ez V and Kelly R (2001) {PD} control with feedforward compensation for robot manipulators: {A}nalysis and experimentation.
\newblock \emph{Robotica} 19(1): 11--19.

\bibitem[{Saunders et~al.(2010)Saunders, Trimmer and Rife}]{saunders2010modeling}
Saunders F, Trimmer BA and Rife J (2010) Modeling locomotion of a soft-bodied arthropod using inverse dynamics.
\newblock \emph{Bioinspiration \& Biomimetics} 6(1): 016001.

\bibitem[{Sfakiotakis et~al.(2014)Sfakiotakis, Kazakidi, Chatzidaki, Evdaimon and Tsakiris}]{sfakiotakis2014multi}
Sfakiotakis M, Kazakidi A, Chatzidaki A, Evdaimon T and Tsakiris DP (2014) Multi-arm robotic swimming with octopus-inspired compliant web.
\newblock In: \emph{Proc.\ IEEE/RSJ Int.\ Conf.\ on Intelligent Robots and Systems}. pp. 302--308.

\bibitem[{Shabana(1990)}]{shabana1990dynamics}
Shabana AA (1990) Dynamics of flexible bodies using generalized {N}ewton-{E}uler equations.
\newblock \emph{ASME J.\ of Dynamic Systems, Measurement, and Control} 112(3): 496--503.

\bibitem[{Sharp et~al.(2023)Sharp, Romero, Jacobson, Vouga, Kry, Levin and Solomon}]{sharp2023data}
Sharp N, Romero C, Jacobson A, Vouga E, Kry P, Levin DI and Solomon J (2023) Data-free learning of reduced-order kinematics.
\newblock In: \emph{Proc. ACM SIGGRAPH}. pp. 1--9.

\bibitem[{Singh et~al.(1985)Singh, VanderVoort and Likins}]{singh1985dynamics}
Singh RP, VanderVoort RJ and Likins PW (1985) Dynamics of flexible bodies in tree topology --- {A} computer-oriented approach.
\newblock \emph{J. of Guidance, Control, and Dynamics} 8(5): 584--590.

\bibitem[{Spong(1994)}]{spong1994partial}
Spong MW (1994) Partial feedback linearization of underactuated mechanical systems.
\newblock In: \emph{Proc.\ IEEE Int.\ Conf.\ on Intelligent Robots and Systems}. pp. 314--321.

\bibitem[{Stramigioli(2024)}]{stramigioli2024principal}
Stramigioli S (2024) The principal bundle structure of continuum mechanics.
\newblock \emph{J.\ of Geometry and Physics} 200: 105172.

\bibitem[{Sun et~al.(2020)Sun, Yu, Chen, Bian, Ye, Sun and Zhao}]{sun2020biohybrid}
Sun L, Yu Y, Chen Z, Bian F, Ye F, Sun L and Zhao Y (2020) Biohybrid robotics with living cell actuation.
\newblock \emph{Chemical Society Reviews} 49(12): 4043--4069.

\bibitem[{Tiburzio et~al.(2024)Tiburzio, Coleman and Della~Santina}]{tiburzio2024model}
Tiburzio S, Coleman T and Della~Santina C (2024) Model-based manipulation of deformable objects with non-negligible dynamics as shape regulation.
\newblock \emph{arXiv:2402.16114} .

\bibitem[{Tonkens et~al.(2021)Tonkens, Lorenzetti and Pavone}]{tonkens2021soft}
Tonkens S, Lorenzetti J and Pavone M (2021) Soft robot optimal control via reduced order finite element models.
\newblock In: \emph{Proc.\ IEEE Int.\ Conf.\ on Robotics and Automation}. pp. 12010--12016.

\bibitem[{Wang et~al.(2021)Wang, Chan, Yuan, Wang, Xia, Yang, Ko, Wang, Sung, Chiu and Zhang}]{wang2021endoscopy}
Wang B, Chan KF, Yuan K, Wang Q, Xia X, Yang L, Ko H, Wang YXJ, Sung JJY, Chiu PWY and Zhang L (2021) Endoscopy-assisted magnetic navigation of biohybrid soft microrobots with rapid endoluminal delivery and imaging.
\newblock \emph{Science Robotics} 6(52): eabd2813.

\bibitem[{Webster~III and Jones(2010)}]{webster2010design}
Webster~III RJ and Jones BA (2010) Design and kinematic modeling of constant curvature continuum robots: {A} review.
\newblock \emph{Int.\ J.\ of Robotics Research} 29(13): 1661--1683.

\bibitem[{Whittaker(1964)}]{whittaker1964treatise}
Whittaker ET (1964) \emph{A Treatise on the Analytical Dynamics of Particles and Rigid Bodies}.
\newblock CUP Archive.

\bibitem[{Wotte et~al.(2023)Wotte, Dummer, Botteghi, Brune, Stramigioli and Califano}]{wotte2023discovering}
Wotte YP, Dummer S, Botteghi N, Brune C, Stramigioli S and Califano F (2023) Discovering efficient periodic behaviors in mechanical systems via neural approximators.
\newblock \emph{Optimal Control Applications and Methods} 44(6): 3052--3079.

\bibitem[{Xu and Chirikjian(2023)}]{xu2023model}
Xu Y and Chirikjian GS (2023) Model reduction in soft robotics using locally volume-preserving primitives.
\newblock \emph{IEEE Robotics and Automation Lett.} 8(9): 5831--5838.

\bibitem[{Yin et~al.(2021)Yin, Varava and Kragic}]{yin2021modeling}
Yin H, Varava A and Kragic D (2021) Modeling, learning, perception, and control methods for deformable object manipulation.
\newblock \emph{Science Robotics} 6(54): eabd8803.

\bibitem[{Zhang(2009)}]{zhang2009recursive}
Zhang Dg (2009) Recursive {L}agrangian dynamic modeling and simulation of multi-link spatial flexible manipulator arms.
\newblock \emph{Applied Mathematics and Mechanics} 30(10): 1283--1294.

\bibitem[{Zheng and Lin(2022)}]{zheng2022pde}
Zheng T and Lin H (2022) {PDE}-based dynamic control and estimation of soft robotic arms.
\newblock In: \emph{Proc.\ IEEE Conf.\ on Decision and Control}. pp. 2702--2707.

\bibitem[{Zhu et~al.(2022)Zhu, Cherubini, Dune, Navarro-Alarcon, Alambeigi, Berenson, Ficuciello, Harada, Kober, Li, Pan, Yuan and Gienger}]{zhu2022challenges}
Zhu J, Cherubini A, Dune C, Navarro-Alarcon D, Alambeigi F, Berenson D, Ficuciello F, Harada K, Kober J, Li X, Pan J, Yuan W and Gienger M (2022) Challenges and outlook in robotic manipulation of deformable objects.
\newblock \emph{IEEE Robotics \& Automation Mag.} 29(3): 67--77.

\end{thebibliography}
